\documentclass[10pt, journal]{IEEEtran}

\usepackage{times}
\usepackage{epsfig}
\usepackage{graphicx}
\usepackage{amsmath}
\usepackage{amssymb}
\usepackage{cite}
\usepackage{amsmath,amssymb,amsfonts}
\usepackage{textcomp}
\usepackage{subfigure}
\usepackage[linesnumbered,boxed,ruled,commentsnumbered]{algorithm2e}
\usepackage{algpseudocode}  
\usepackage{amsmath}  
\usepackage{multirow}
\usepackage{color}
\UseRawInputEncoding

\usepackage{caption}
\usepackage{url}

\IEEEoverridecommandlockouts                              

\title{DANIEL: A Fast and Robust Consensus Maximization Method for Point Cloud Registration with High Outlier Ratios}

\author{Lei Sun$^{*}$%
\thanks{$^{*}$Corresponding author.}%
\thanks{Lei Sun is with the School of Mechanical and Power Engineering, East China University of Science and Technology, Shanghai 200237, China
         {(e-mail: \tt\footnotesize leisunjames@126.com)}}%
\thanks{This work was not supported by any organization.}%
}

\begin{document}

\maketitle

\begin{abstract}

Correspondence-based point cloud registration is a cornerstone in geometric computer vision, robotics perception, photogrammetry and remote sensing, which seeks to estimate the best rigid transformation between two point clouds from the correspondences established over 3D keypoints. However, due to limited robustness and accuracy, current 3D keypoint matching techniques are very prone to yield outliers, probably even in very large numbers, making robust estimation for point cloud registration of great importance. Unfortunately, existing robust methods may suffer from high computational cost or insufficient robustness when encountering high (or even extreme) outlier ratios, hardly ideal enough for practical use. In this paper, we present a novel time-efficient RANSAC-type consensus maximization solver, named DANIEL (Double-layered sAmpliNg with consensus maximization based on stratIfied Element-wise compatibiLity), for robust registration. DANIEL is designed with two layers of random sampling, in order to find inlier subsets with the lowest computational cost possible. Specifically, we: (i) apply the rigidity constraint to prune raw outliers in the first layer of one-point sampling, (ii) introduce a series of stratified element-wise compatibility tests to conduct rapid compatibility checking between minimal models so as to realize more efficient consensus maximization in the second layer of two-point sampling, and (iii) probabilistic termination conditions are employed to ensure the timely return of the final inlier set. Based on a variety of experiments over multiple real datasets, we show that DANIEL is robust against over 99\% outliers and also significantly faster than existing state-of-the-art robust solvers (e.g. RANSAC, FGR, GORE). To be specific, with 1000 correspondences and as many as 99\% outliers, DANILE can render accurate results within 3 seconds, over 10000 times faster than RANSAC.

\end{abstract}

\begin{IEEEkeywords}
Point cloud registration, robust estimation, consensus maximization, RANSAC, compatibility
\end{IEEEkeywords}

\section{Introduction}

Point cloud registration is a fundamental task in the fields of 3D computer vision, robotics, photogrammetry and remote sensing. It aims to align two point clouds by estimating the optimal rigid transformation (including the rotation and translation) between them, and has been broadly applied in 3D reconstruction~\cite{henry2012rgb,choi2015robust,zhang2015visual}, object recognition and localization~\cite{drost2010model,zeng2017multi,wong2017segicp,marion2018label}, SLAM~\cite{zhang2014loam}, medical imaging~\cite{audette2000algorithmic}, archaeology~\cite{chase2012geospatial}, etc.

Though Iterative Closet Point (ICP)~\cite{besl1992method} is a popular registration method, its performance is still limited since it highly relies on the initial guess provided by users and is prone to converge to the local minima when the initialization is not good enough. To circumvent the need of initial guess, people tend to use 3D keypoint detecting and matching techniques (e.g. FPFH~\cite{rusu2009fast}, ISS~\cite{zhong2009intrinsic}, 3DSmoothNet~\cite{gojcic2019perfect}) to establish correspondences between the point clouds, and then estimate the transformation based on these correspondences. However, 3D keypoint matching is much less robust and accurate than 2D keypoint matching (e.g. SIFT~\cite{lowe2004distinctive}, SURF~\cite{bay2006surf}), so it may easily generate a huge number of false matches, usually called \textit{outliers}, among the putative correspondences. Consequently, robust estimation methods must be adopted to solve the correct transformation from the potentially abundant outliers.

Unfortunately, many existing robust methods have their own nonnegligible limitations. RANSAC~\cite{fischler1981random} is a well-known hypothesis-and-test consensus maximization paradigm for robust estimation, but it has exponentially growing computational cost w.r.t. the outlier ratio, thus unsuitable for handling high-outlier situations. Note that an outlier ratio of over 95\% is common after 3D keypoint matching in real scenes~\cite{bustos2017guaranteed}, so RANSAC is not a generally practical option. Branch-and-Bound (BnB)~\cite{parra2014fast,horst2013global}, as another famous robust estimator, can yield the globally optimal solution, but it also suffers from the worst-case exponential runtime w.r.t. the problem size. Other robust methods include the non-minimal global solvers such as FGR~\cite{zhou2016fast}, GNC~\cite{yang2020graduated} and ADAPT~\cite{tzoumas2019outlier} which are limited in robustness and generally cannot tolerate outlier ratios higher than 90\%, the guaranteed outlier removal method GORE~\cite{bustos2017guaranteed} which may also be too slow for use due to its probable internal use of BnB, and the certifiably optimal solver TEASER~\cite{yang2019polynomial,yang2020teaser} which is also slow without using parallelism programming. Hence, we can see that almost all of these methods have limited performance in some ways.

Therefore, our goal in this paper is to propose a new robust estimation approach, which can handle registration problems with high or extreme outliers in a time-efficient way.

\textbf{Our Contributions.} We present a specialized consensus maximization method to realize rapid robust estimation for point cloud registration with even extremely high outliers (e.g. up to 99\%). 

First, we abandon the traditional time-consuming single-layered three-point sampling framework used in RANSAC, and present a more efficient strategy by smartly decomposing the three-point layer into: (i) a one-point sampling layer that serves as a raw outlier `filter' on the basis of the 3D-geometric rigidity constraint to diminish the correspondence size, and (ii) a two-point sampling layer that performs random sampling and minimal model estimating. This strategy can significantly reduce the computational cost for obtaining an all-inlier subset from the random samples.

Moreover, for faster consensus maximization, we propose a compatibility-based consensus building strategy by introducing a novel stratified element-wise compatibility checking technique that is merely made up of very simple calculations and boolean conditions, in order to effectively cut down the time cost for the repetitive construction of consensus set as in RANSAC. 

These two contributions lead to our robust solver DANIEL (Double-layered sAmpliNg with consensus maximization based on stratIfied Element-wise compatibiLity). Comprehensive experimental evaluation on various real-world datasets shows that the proposed solver is highly robust against over 99\% outliers and is also rather fast in practice (running within 3 seconds for solving a 99\%-outlier registration problem), outperforming existing state-of-the-art robust registration solvers.

\section{Related Work}

This section provides brief reviews on addressing the point cloud registration problem with correspondences.

Before the application of robust solvers, correspondences have to be first established between the point clouds. 3D keypoint detecting and matching with feature descriptors~\cite{rusu2009fast,zhong2009intrinsic,zhong2009intrinsic} is a widely-used process for building the correspondences, based on which registration solvers are able to estimate the best transformation.

In the most ideal case, assuming that there is no outlier among these putative correspondences, closed-form estimators can efficiently solve the optimal transformation based on eigenvector~\cite{horn1987closed} or Singular Value Decomposition (SVD)~\cite{arun1987least}. More recent methods include the optimal BnB solver~\cite{olsson2008branch} and the certifiable Semi-Definite relaxation method~\cite{briales2017convex}. 

However, it is known to all that 3D keypoint matching is less robust than its 2D counterpart such as SIFT~\cite{lowe2004distinctive} or SURF~\cite{bay2006surf} and could easily generate outliers to corrupt the performance of these outlier-intolerant solvers in practical use. Even worse, the outliers may sometimes occupy a massive majority of the correspondences (as shown in~\cite{bustos2017guaranteed} and Section~\ref{Experiments}), making inliers (correct matches) fairly sparse. In this case, we need to apply robust estimation to reject outliers and find the true inliers to make reasonable estimates, where the robustness as well as efficiency of the robust methods could determine the registration results to a great extent.

We introduce several types of robust solvers as follows.

\subsection{Consensus Maximization}

Consensus maximization consists in finding a model that can maximize the number of correspondences with residual errors lower than a certain inlier threshold w.r.t. this model. RANSAC~\cite{fischler1981random} is a very common consensus maximization paradigm via the hypothesize-and-test model-fitting framework. Besides, further techniques, including local optimization~\cite{chum2003locally,lebeda2012fixing} correspondence sorting~\cite{chum2005matching}, etc, have been applied to improve the performance of RANSAC. However, the time cost of these RANSAC solvers generally increases exponentially with the outlier ratio. For instance, RANSAC should require more than 4.6$\times$10$^6$ samples to select one all-inlier subset with 0.99 confidence if the outlier ratio is 99\% in one registration problem. This would make RANSAC infeasible for use in realistic problems. 

Another typical consensus maximization approach is BnB. It addresses the optimization problem globally optimally by searching in the parameter space (e.g. $SO(3)$ w.r.t. rotation or $SE(3)$ w.r.t. transformation). Unfortunately, BnB runs in exponential time and could not scale to problems with large correspondence numbers, but thousands of correspondences is quite common in reality, which limits the practicality of BnB.

ADAPT (Adaptive Trimming)~\cite{tzoumas2019outlier} provides a non-minimal way to solve the consensus maximization problem. It can be directly in conjunction with standard non-minimal solvers to find the maximum consensus set through iterations, but it has the issue of limited robustness. In the registration problem, ADAPT could hardly tolerate more than 90\% outliers.

In essence, our solver DANIEL is also a consensus maximization method based on the RANSAC paradigm. But much differently, DANIEL employs a smarter framework by decomposing the sampling into two layers and applying the stratified compatibility checking before the consensus building, so it exceeds RANSAC in time-efficiency by tens to tens of thousands of times.

\subsection{M-Estimation}

M-estimation can actively decrease the weights of outliers by incorporating robust loss functions into the optimization process. Local M-estimators (e.g.~\cite{agarwal2013robust,kummerle2011g,sunderhauf2012towards}) can be used to minimize the object function, but the problem is that they must require the initial guess and hence is liable to converge to local solutions if the initial guess is poor. FGR~\cite{zhou2016fast} first applied Graduated Non-Convexity (GNC) in the registration problem, and more recently, GNC is incorporated with non-minimal solvers and extended to more robotics and computer vision problems~\cite{yang2020graduated}. However, these solvers generally have confined robustness against outliers. For example, FGR and GNC-TLS would both become brittle once the outlier ratio exceeds 90\%.

\subsection{Downside of RANSAC}

We now provide a more explicit analysis on the limitation of RANSAC. In traditional RANSAC or many of its variants, when we obtain one minimal subset in a certain iteration of random sampling, we must then: (i) estimate the coarse minimal model ($\boldsymbol{R}^{*},\boldsymbol{t}^{*}$), and (ii) build the consensus set with this model by computing the residual errors w.r.t. all the correspondences. This consensus maximization strategy has two apparent downsides: (i) when the outlier ratio is high, the probability to sample an all-inlier subset can be extremely low, thus making a great number of iterations necessary (exponentially growing with the outlier ratio), and (ii) it is time-consuming to build the consensus set in every iteration, especially when the problem size is large (e.g. thousands of correspondences). And these two limitations are the main issues that we aim to circumvent by using our solver.

\begin{figure*}[t]
\centering
\setlength\tabcolsep{1pt}
\addtolength{\tabcolsep}{0pt}
\begin{tabular}{ccc}

\footnotesize{(a) Correspondences with 99\% outliers}
&
\footnotesize{(b) Registration by DANIEL}
&
\footnotesize{(c) Overview of DANIEL}

\\

\begin{minipage}[t]{0.32\linewidth}
\centering
\includegraphics[width=1\linewidth]{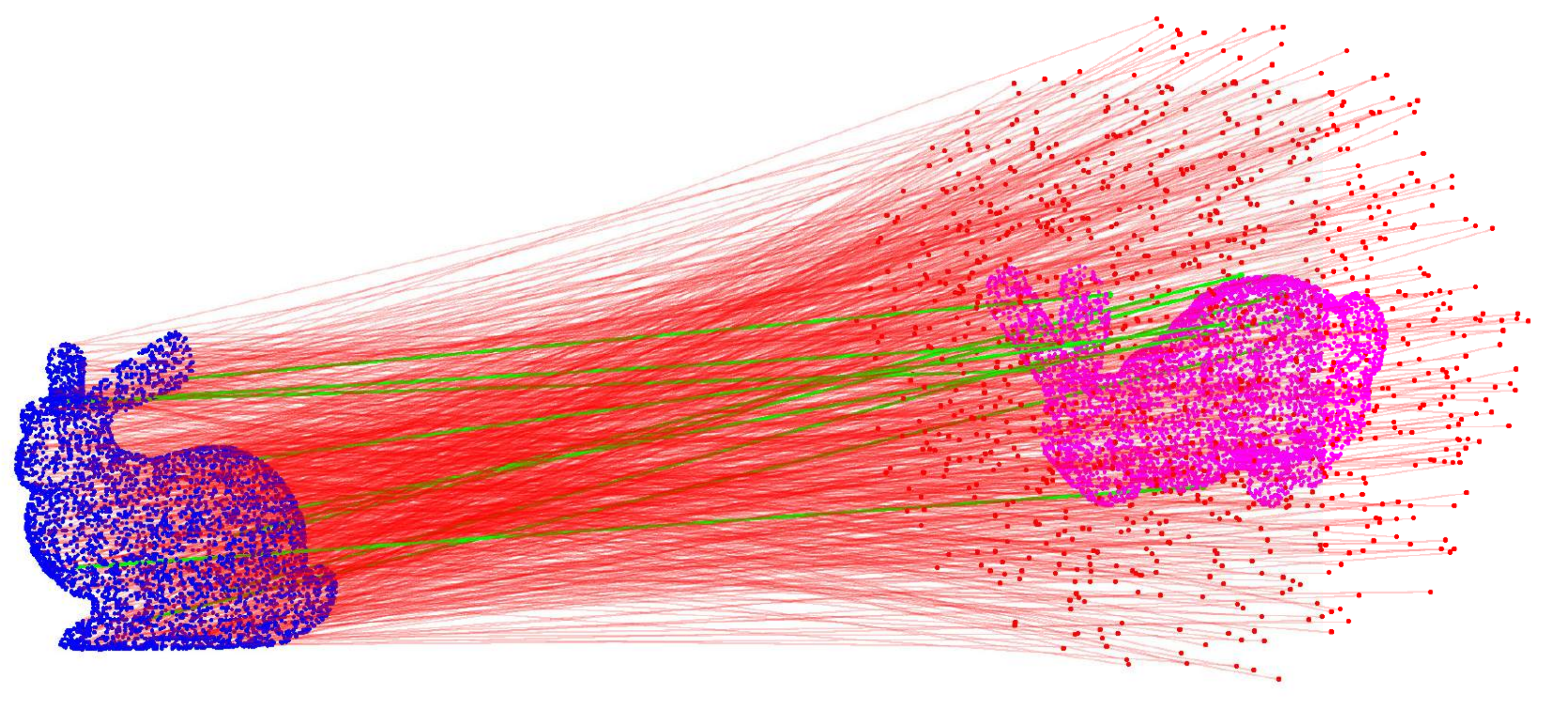}
\end{minipage}

&

\begin{minipage}[t]{0.16\linewidth}
\centering
\includegraphics[width=1\linewidth]{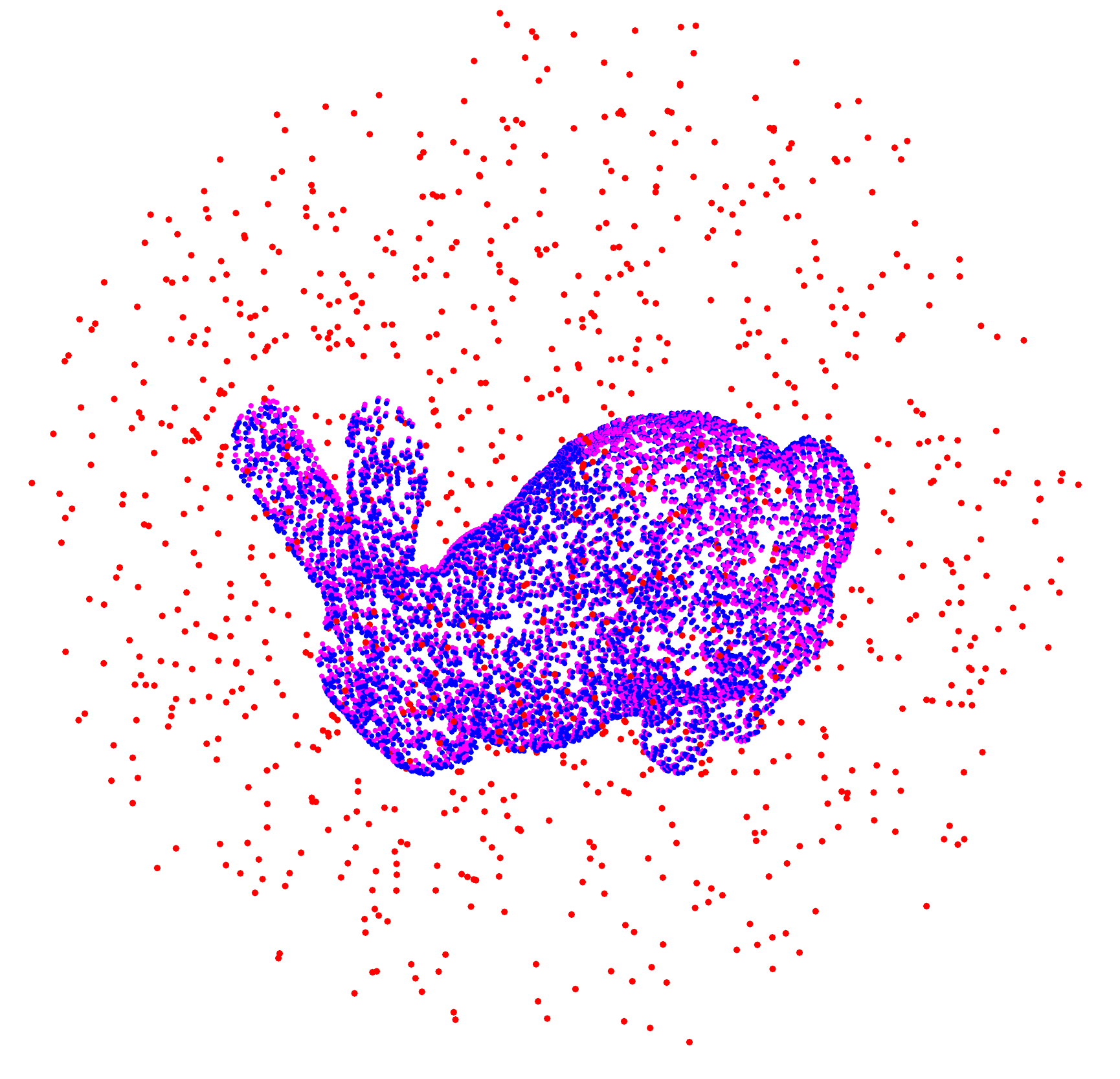}
\end{minipage}



&

\begin{minipage}[t]{0.5\linewidth}
\centering
\includegraphics[width=0.96\linewidth]{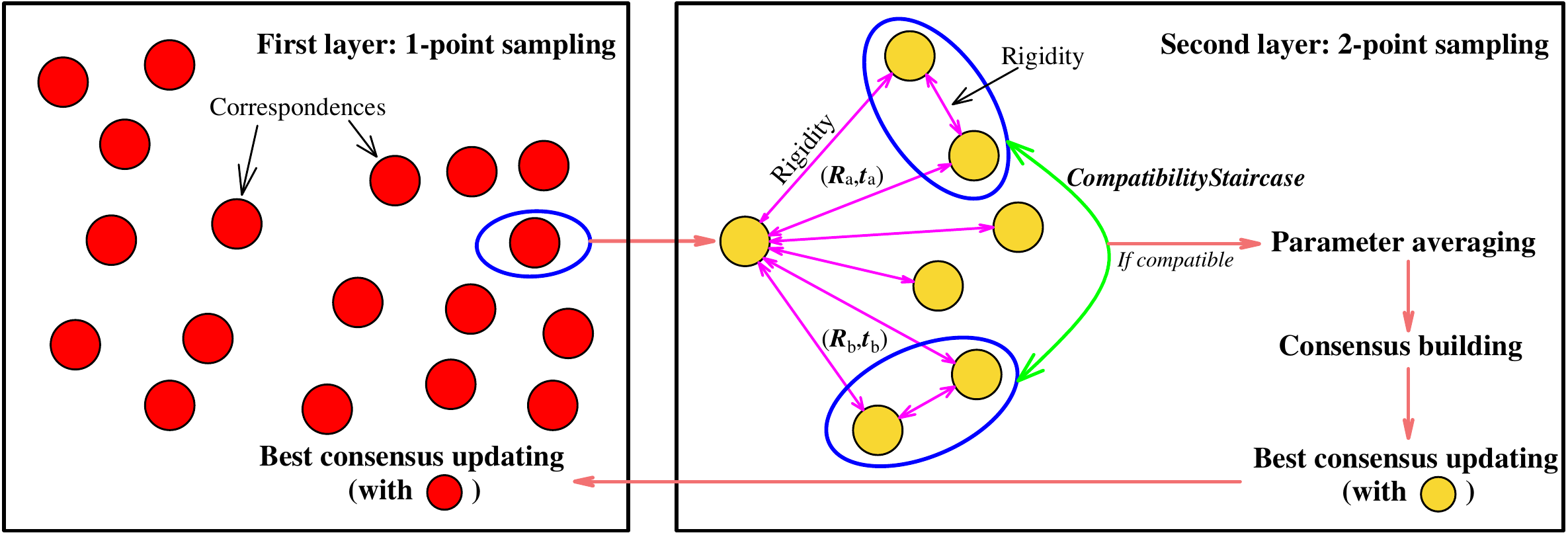}
\end{minipage}

\end{tabular}
\vspace{-1mm}
\caption{Illustration of robust point cloud registration using our solver DANIEL. (a) An example of a registration problem with $N=1000$ correspondences and only 10 inliers. (b) DANIEL can robustly estimate the best transformation in only 2.597 seconds. (c) An intuitive overview of DANILE. Please see Section~\ref{overview} for explicit descriptions.}
\label{demo-for-show}
\vspace{-3mm}
\end{figure*}

\section{Our Method: DANIEL}

In this section, we present our robust point cloud registration method DANIEL.

\subsection{Problem Formulation: Consensus Maximization}

First of all, we introduce the consensus maximization formulation for the registration problem.

Assume that we have two sets of 3D points: $\mathcal{X}=\{\boldsymbol{x}_i\}_{i=1}^{N}$ and $\mathcal{Y}=\{\boldsymbol{y}_i\}_{i=1}^{N}$ where $\boldsymbol{x}_i\leftrightarrow\boldsymbol{y}_i\,(\boldsymbol{x}_i,\boldsymbol{y}_i\in\mathbb{R}^3)$ makes up a putative correspondence. Point cloud registration aims to estimate the best transformation including rotation $\boldsymbol{R}\in SO(3)$ and translation $\boldsymbol{t}\in \mathbb{R}^{3}$ that can align point set $\mathcal{X}$ and $\mathcal{Y}$. If correspondence $\boldsymbol{x}_i\leftrightarrow\boldsymbol{y}_i$ is an inlier, then we can have:
\begin{equation}\label{prob-form1}
\boldsymbol{y}_i=\boldsymbol{R}\boldsymbol{x}_i+\boldsymbol{t}+\boldsymbol{\epsilon}_i,
\end{equation}
where $\boldsymbol{\epsilon}_i\in\mathbb{R}^3$ denotes the noise measurement, so that the following relation 
\begin{equation}
\left\|\boldsymbol{R}\boldsymbol{x}_i+\boldsymbol{t}-\boldsymbol{y}_i\right\|\leq \xi
\end{equation}
could be satisfied, where $\xi\geq\|\boldsymbol{\epsilon}\|$ denotes the inlier threshold; however, if $\boldsymbol{x}_i\leftrightarrow\boldsymbol{y}_i$ is an outlier, we can assume:
\begin{equation}
\left\|\boldsymbol{R}\boldsymbol{x}_i+\boldsymbol{t}-\boldsymbol{y}_i\right\|>\xi.
\end{equation}
Usually, we assume the noise to be isotropic Gaussian with standard deviation $\sigma$, so the inlier threshold can be typically set as: $\xi=5\sim 6\sigma$.

In this case, letting set $\mathcal{N}=\{1,2,\dots,N\}$, the registration problem can be written as the following formulation:
\begin{equation}\label{CM}
\begin{gathered}
\underset{\mathcal{M}\subset \mathcal{N}}{\max}\, |\mathcal{M}|, \\
s.t. \,\|\boldsymbol{R}_{\mathcal{M}}\boldsymbol{x}_i+\boldsymbol{t}_{\mathcal{M}}-\boldsymbol{y}_i\| \leq \xi,\,(\forall i\in\mathcal{M})
\end{gathered}
\end{equation}
where $(\boldsymbol{R}_{\mathcal{M}},\boldsymbol{t}_{\mathcal{M}})$ represents a rigid transformation and $\mathcal{M}$ is its corresponding consensus set. The goal of formulation~\eqref{CM} is to find the transformation $(\boldsymbol{R}_{\mathcal{M}},\boldsymbol{t}_{\mathcal{M}})$ that maximizes the size of consensus $\mathcal{M}$, the process of which is typically called \textit{consensus maximization}. The goal of our solver is to solve~\eqref{CM} in a computationally efficient way.

\subsection{Methodology Overview}\label{overview}

Fig.~\ref{demo-for-show} illustrates the effectiveness and the overview of our proposed solver DANIEL with descriptions given below.

In general, DANIEL maximizes the consensus set also by random sampling, but its main operation structure is significantly different from the traditional RANSAC solvers. To be specific, DANIEL consists of two random-sampling layers, where the latter is embedded into the main framework of the former. The first layer adopts one-point sampling in each iteration (in order to maximize the probability to obtain inliers) and then reduces the correspondence set size by using rigidity constraint. (See Section~\ref{first-layer} for the first layer)

In one certain iteration of the first layer and after the reduction of the correspondence set, the second layer performs continuous random sampling of two points as well as the computing of the respective minimal models. In the meantime, a new stratified element-wise compatibility checking approach is proposed to conduct rapid compatibility checking between pairs of minimal models obtained, which serves as a prerequisite for consensus building in DANIEL. Only when we can obtain two minimal subsets (models) that are compatible with each other can we average the parameters of these two models and build the consensus set. Furthermore, we can derive probabilistic termination conditions (by computing maximum iteration numbers) for both of the two layers, and the maximized consensus in the second layer will be used to further maximize the consensus in the first layer. Finally, we can return the maximum consensus set in the first layer to further obtain the ultimate full inlier set. (See Section~\ref{second-layer} for the second layer) 

\subsection{First Layer: One-Point Sampling and Inlier Candidate Searching with Rigidity Constraint}\label{first-layer}

Rigidity~\cite{michel2017global,zach2015dynamic,quan2020compatibility,yang2019polynomial} (invariance of length) is a common pairwise constraint in 3D geometry. Specifically, for any two correspondences $\boldsymbol{x}_i\leftrightarrow\boldsymbol{y}_i$ and $\boldsymbol{x}_j\leftrightarrow\boldsymbol{y}_j$ that are both inliers, we can have the rigidity constraint such that
\begin{equation}\label{rigidity}
\left|\left\|\boldsymbol{y}_i-\boldsymbol{y}_j\right\|-\left\|\boldsymbol{x}_i-\boldsymbol{x}_j\right\|\right| \leq 2\xi,
\end{equation}
which can be derived, based on triangular inequality and the property that norm is invariant to $\boldsymbol{R}$, as follows:
\begin{equation}\label{proof-of-rigidity}
\begin{gathered}
\left|{\|\boldsymbol{y}_i-\boldsymbol{y}_j\|}-{\|\boldsymbol{x}_i-\boldsymbol{x}_j\|}\right| \\
=\left|{\left\|\boldsymbol{R} (\boldsymbol{x}_i-\boldsymbol{x}_j+\boldsymbol{R}^{\top}\boldsymbol{\epsilon}_i-\boldsymbol{R}^{\top}\boldsymbol{\epsilon}_j)\right\|}-{\|\boldsymbol{x}_i-\boldsymbol{x}_j\|}\right| \\
\leq \left\| (\boldsymbol{x}_i-\boldsymbol{x}_j)-({\boldsymbol{x}_i-\boldsymbol{x}_j})+\boldsymbol{R}^{\top}\boldsymbol{\varepsilon}_i-\boldsymbol{R}^{\top}\boldsymbol{\epsilon}_j \right\| \\ 
= {2\|(\boldsymbol{\epsilon}_i-\boldsymbol{\epsilon}_j)\|}\leq 2\xi.
\end{gathered}
\end{equation}
Intuitively, this constraint indicates that the length between two points (both inliers) remains fixed before and after the rigid transformation, which underlies our strategy to reduce the correspondence set in our first sampling layer.

In the first layer of DANIEL, we perform one-point sampling, that is, to select only one random correspondence in each iteration, say correspondence $n\in\mathcal{N}$ ($\mathcal{N}=\{1,2,\dots,N\}$ denotes the full correspondence set). Then, we employ rigidity constraint to sift out all the eligible correspondences from $\mathcal{N}$ that can satisfy~\eqref{rigidity} with $n$, and then add them to set $\mathcal{C}$, called the `\textit{inlier candidate set}'. (See \textbf{lines 3-9} in Algorithm~\ref{DANIEL-algo})

The insight here lies in that if the sampled point $n$ is an inlier, then all the other inliers must fulfill condition~\eqref{rigidity} with $n$ and must therefore lie within set $\mathcal{C}$. In this way, we can establish a smaller correspondence set for our second layer. Note that this step could be rather effective especially when the outlier ratio is high, since a large portion of `raw' outliers can be swiftly removed by the rigidity constraint, exponentially increasing the probability of sampling an all-inlier subset later in the second layer.

\begin{algorithm}[t]
\caption{\textit{CompatibilityStaircase}}
\label{Staircase}
\SetKwInOut{Input}{\textbf{Input}}
\SetKwInOut{Output}{\textbf{Output}}
\Input{two minimal models: ($\boldsymbol{R}^*_{a},\boldsymbol{t}^*_{a}$) and ($\boldsymbol{R}^*_{b},\boldsymbol{t}^*_{b}$); element-wise thresholds $\theta^{r}$ and $\theta^{t}$\;}
\Output{boolean compatibility status $comp$\;}
\BlankLine
$\boldsymbol{T}_{a}\leftarrow{[{vec(\boldsymbol{R}^*_{a})}^{\top},{\boldsymbol{t}^{*}_{a}}^{\top}]}^{\top}$, $\boldsymbol{T}_{b}\leftarrow{[{vec(\boldsymbol{R}^*_{b})}^{\top},{\boldsymbol{t}^{*}_{b}}^{\top}]}^{\top}$, $\boldsymbol{\theta}\leftarrow{[\underbrace{\theta^r,\dots,\theta^r}_{9\times\theta^r}, \underbrace{\theta^t,\dots,\theta^t}_{3\times\theta^t}]}^{\top}$, and $comp\leftarrow 0$\;
\For{$rep=1:12$}{
\If{$\left|\boldsymbol{T}_{a}(rep)-\boldsymbol{T}_{b}(rep)\right|>\boldsymbol{\theta}(rep)$}{
\textbf{break}}
\If{$rep=12$}{
$comp\leftarrow 1$\;
}
}
\Return boolean compatibility status $comp$\;
\end{algorithm}

\subsection{Second Layer: Two-Point Sampling and Consensus Maximization based on Stratified Element-wise Compatibility}\label{second-layer}

In the second layer of DANIEL, with the correspondence $n$ selected in the first layer, we only need to sample two more points to construct a three-point minimal subset for the estimation of the transformation, which requires much less computational cost in comparison with the traditional three-point RANSAC since: (i) the correspondence set is reduced from the original set $\mathcal{N}$ to set $\mathcal{C}$ now, and (ii) the sampling dimension is reduced from 3 to 2 (three-point to two-point). Subsequently, rigidity constraint~\eqref{rigidity} can be applied once again for `raw' outlier removal. When we select a pair of random correspondences $\{a,b\}\subset\mathcal{C}$, knowing that both $(a,n)$ and $(b,n)$ have already satisfied rigidity in the first layer, we can further test $(a,b)$ with~\eqref{rigidity}. If satisfied, the three-point set $\{n,a,b\}$ is entirely rigid (each line between any two points has fixed length after the transformation) and can be used to solve the transformation model $(\boldsymbol{R}^*,\boldsymbol{t}^*)$ minimally using Horn's triad-based method~\cite{horn1987closed}. (See \textbf{lines 12-14} in Algorithm~\ref{DANIEL-algo})

More importantly, we depart from the traditional sampling-and-consensus-building technique as applied in RANSAC, and introduce a novel compatibility-based consensus maximization paradigm, which is partially inspired by~\cite{sun2021ransic}.

Different from the standard procedure of first making an estimate with a minimal subset and then establishing the consensus set in every single iteration in traditional RANSAC, when we compute a minimal model ($\boldsymbol{R}^*,\boldsymbol{t}^*$) with a random subset, we store its parameters and check the \textit{mutual compatibility} between this model and every single existing model already in storage. (Intuitively, compatibility checking can be operated by measuring the mathematical error between the two models and then judging whether it is small enough.) Only when we have found two models that fulfill the mutual compatibility could we build the consensus set with these two models by computing the residual errors w.r.t. all the correspondences in set $\mathcal{C}$. This strategy is inspired by the fact that models estimated with different true inliers should still be sufficiently similar (not equivalent due to the presence of noise), and it can tremendously curtail the time cost of the repeated but `redundant' consensus building in every iteration of sampling. (See \textbf{lines 15-17} in Algorithm~\ref{DANIEL-algo})

However, when the outlier ratio is high, we may have to sample a large number of random subsets and store many minimal models so as to achieve two all-inlier subsets to satisfy this compatibility condition. Consequently, we further propose a series of \textit{stratified element-wise compatibility tests} in order to ensure high efficiency in our compatibility checking process.

\subsubsection{Stratified Element-wise Model Compatibility}

Instead of the holistic checking of compatibility, our idea is that two models must be mutually compatible as long as all their respective parameters are compatible.
The parameters of the model involved in the registration problem include 9 scalars from the rotation matrix $\boldsymbol{R}\in SO(3)$ plus 3 scalars from the translation vector $\boldsymbol{t}\in\mathbb{R}^3$, which can be represented as $\boldsymbol{R}=\left[\begin{array}{ccc} r_1, &r_4, & r_7 \\  r_2, &r_5, & r_8 \\  r_3, &r_6, & r_9  \end{array}\right]$ and $\boldsymbol{t}=[t_1,t_2,t_3]^{\top}$, respectively. 

For a certain translation vector $\boldsymbol{t}^*$ estimated from a minimal subset, it can be rewritten as:
\begin{equation}
\boldsymbol{t}^*=\boldsymbol{t}_{gt}+\boldsymbol{\epsilon}^{\boldsymbol{t}},
\end{equation}
where $\boldsymbol{t}_{gt}$ represents the ground-truth translation and $\boldsymbol{\epsilon}^{\boldsymbol{t}}\in\mathbb{R}^3$ denotes the noise measurement on translation. Now assume that we have two such minimal translations, say $\boldsymbol{t}^*_{a}$ and $\boldsymbol{t}^*_{b}$. Then, we can derive the following compatibility condition if they are both inliers:
\begin{equation}\label{t-comp}
\left\|\boldsymbol{t}^*_{a}-\boldsymbol{t}^*_{b}\right\|=\left\|\boldsymbol{\epsilon}^{\boldsymbol{t}_a}-\boldsymbol{\epsilon}^{\boldsymbol{t}_b}\right\| \leq 2\xi^{\boldsymbol{t}},
\end{equation}
where $\xi^{\boldsymbol{t}}\geq\|\boldsymbol{\epsilon}^{\boldsymbol{t}}\|$ is the noise upper-bound for translation, and we usually set $\xi^{\boldsymbol{t}}=5\sigma$. Moreover, since the three elements in translation are completely independent, compatibility~\eqref{t-comp} can be easily decoupled into three errors in L1 norm: $|t^*_{a1}-t^*_{b1}|$, $|t^*_{a2}-t^*_{b2}|$ and $|t^*_{a3}-t^*_{b3}|$. We can then set a scalar threshold for each of them in order to fulfil condition~\eqref{t-comp} such that
\begin{equation}\label{t-comp-each}
\theta^{t}=\left|{t}^*_{ap}-{t}^*_{bp}\right| \leq \frac{2\mu}{\sqrt{3}}\xi^{\boldsymbol{t}},\,(\forall p\in\{1,2,3\})
\end{equation}

\clearpage
\begin{algorithm*}[h]
\caption{DANIEL}
\label{DANIEL-algo}
\SetKwInOut{Input}{\textbf{Input}}
\SetKwInOut{Output}{\textbf{Output}}
\Input{correspondences $\mathcal{P}=\{(\mathbf{x}_i,\mathbf{y}_i)\}_{i=1}^N$; noise $\sigma$; minimum inlier number $I_{min}=\max(5,0.01N)$\;}
\Output{optimal $(\boldsymbol{R}^{\star}, \boldsymbol{t}^{\star})$; inlier set $\mathcal{N}^{\star}$ \;}
\BlankLine
$\mathcal{N}\leftarrow [1,2,\dots,N]$, $samp1\leftarrow 0$, $\mathcal{N}_{best}\leftarrow\emptyset$, $bestSize_1\leftarrow0$, $maxItr_1\leftarrow 459$, and get $\theta^r$ and $\theta^t$ with~\eqref{r-comp-each} and~\eqref{t-comp-each}\;
\While{$samp_1\leq maxItr_1$}{
Select a random $n\in\mathcal{N}$, $\mathcal{C}\leftarrow \emptyset$, and $samp_1\leftarrow samp_1+1$\; 
\For{\textbf{\textup{all}} $i\in\mathcal{N}$ and $i\neq n$}{
\If{correspondence pair $(i,n)$ can satisfy condition~\eqref{rigidity}}{
$\mathcal{C}=\mathcal{C}\cup{\{i\}}$\;
}
}
\If{$|\mathcal{C}|\geq I_{min}$}{
$bestSize_2\leftarrow0$, $samp_2\leftarrow 0$ and $\mathcal{S}_{best}\leftarrow\emptyset$\;
\While{$samp_2\leq maxItr_2$}{
Select a random subset $\{a,b\}\subset\mathcal{C}$, and $samp_2\leftarrow samp_2+1$\;
\If{correspondence pair $(a,b)$ can satisfy condition~\eqref{rigidity}}{
Estimate $(\boldsymbol{R}^{*},\boldsymbol{t}^{*})$ minimally using Horn's method~\cite{horn1987closed}\;
$\mathcal{R}\leftarrow\mathcal{R}\cup\{\boldsymbol{R}^{*}\}$, and $\mathcal{T}\leftarrow\mathcal{T}\cup\{\boldsymbol{t}^{*}\}$\;
\For{$z=1:\left(|\mathcal{R}|-1\right)$}{
$comp\leftarrow$\textit{CompatibilityStaircase}($\mathcal{R}_{(z)}$, $\mathcal{T}_{(z)}$, $\boldsymbol{R}^{*}$, $\boldsymbol{t}^{*}$, $\theta^r$, $\theta^t$)\;
\If{$comp=1$}{
$\boldsymbol{R}^{\circ}\leftarrow$\textit{AverageRotPara}($\mathcal{R}_{(z)}$, $\boldsymbol{R}^{*}$), $\boldsymbol{t}^{\circ}\leftarrow$\textit{AverageTranPara}($\mathcal{T}_{(z)}$, $\boldsymbol{t}^{*}$), and $\mathcal{S}\leftarrow\emptyset$\;
\% Obtain the consensus set of the averaged model: $\boldsymbol{R}^{\circ}$ and $\boldsymbol{t}^{\circ}$ \%\\
\For{\textbf{\textup{all}} $j\in\mathcal{C}$}{
\If{$\|\boldsymbol{R}^{\circ}\boldsymbol{x}_j+\boldsymbol{t}^{\circ}-\boldsymbol{y}_j|\leq 5\sigma$}{
$\mathcal{S}\leftarrow\mathcal{S}\cup\{j\}$\;
}
}
\If{$|\mathcal{S}|\geq bestSize_2$}{
$\mathcal{S}_{best}\leftarrow\mathcal{S}$, $bestSize_2\leftarrow |\mathcal{S}|$, and update $maxItr_2$ with~\eqref{max-itr-2}\;
}
}
}
}
}
} 
\If{$bestSize_2\geq bestSize_1$}{
Solve $(\boldsymbol{R}^{\dagger}, \boldsymbol{t}^{\dagger})$ non-minimally with $\mathcal{S}_{best}$ using SVD~\cite{arun1987least}, and $\mathcal{N}_{best}\leftarrow\emptyset$\;
\% Obtain the consensus set of current $\boldsymbol{R}^{\dagger}$ and $\boldsymbol{t}^{\dagger}$ \%\\
\For{\textbf{\textup{all}} $j\in\mathcal{N}$}{
\If{$\|\boldsymbol{R}^{\dagger}\boldsymbol{x}_j+\boldsymbol{t}^{\dagger}-\boldsymbol{y}_j\|\leq 5\sigma$}{
$\mathcal{N}_{best}\leftarrow\mathcal{N}_{best}\cup\{j\}$\;
}
}
$bestSize_1\leftarrow |\mathcal{N}_{best}|$, and update $maxItr_1$ with~\eqref{max-itr-1}\;
}
}
Solve $(\boldsymbol{R}^{\star}, \boldsymbol{t}^{\star})$ non-minimally with $\mathcal{N}_{best}$ using SVD, and $\mathcal{N}^{\star}\leftarrow\emptyset$\;
\% Obtain the consensus set with current $\boldsymbol{R}^{\star}$ and $\boldsymbol{t}^{\star}$ \%\\
\For{\textbf{\textup{all}} $k\in\mathcal{N}$}{
\If{$\|\boldsymbol{R}^{\star}\boldsymbol{x}_k+\boldsymbol{t}^{\star}-\boldsymbol{y}_k\|\leq 5\sigma$}{
$\mathcal{N}^{\star}\leftarrow\mathcal{N}^{\star}\cup\{j\}$\;
}
}
Solve $(\boldsymbol{R}^{\star}, \boldsymbol{t}^{\star})$ non-minimally with $\mathcal{N}^{\star}$ using SVD\;
\Return $(\boldsymbol{R}^{\star}, \boldsymbol{t}^{\star})$ and $\mathcal{N}^{\star}$\;
\end{algorithm*}
\clearpage

\noindent where $\mu \geq1$. Note that if $\mu=1$,  \eqref{t-comp-each} $\Rightarrow$ \eqref{t-comp}. But we want a slightly more lenient threshold, so $\mu$ can be set to slightly larger than 1. In practice, we can empirically set $\mu=1.2$.

For rotation, the widely-used geodesic distance~\cite{hartley2013rotation} is represented by angle and requires relatively complex computation including $\arccos$ and $trace$, so it is difficult to be rewritten as element-wise. But according to~\cite{hartley2013rotation}, the chordal distance, having natural relation to the geodesic distance, enables us to derive a series of element-wise compatibility conditions.

Assume that we have two rotations $\boldsymbol{R}^*_a$ and $\boldsymbol{R}^*_b$ estimated from two minimal subsets. $d_{\text{chord}}$ denotes the chordal distance between them that can be written as:
\begin{equation}\label{chordal}
d_{\text{chord}}(\boldsymbol{R}^*_a,\boldsymbol{R}^*_b)=\left\|\boldsymbol{R}^*_a-\boldsymbol{R}^*_b\right\|_{F}=2\sqrt{2}\sin (\frac{d_{\text{geo}}(\boldsymbol{R}^*_a,\boldsymbol{R}^*_b)}{2}),
\end{equation}
where $\|\cdot\|_F$ denotes the Frobenius norm and $d_\text{geo}(\cdot,\cdot)$ means the geodesic distance~\cite{hartley2013rotation} such that
\begin{equation}\label{geodesic}
d_{\text{geo}}(\boldsymbol{R}^*_a,\boldsymbol{R}^*_b)=\left|\arccos\left(\frac{trace({\boldsymbol{R}_a^*}^{\top}\boldsymbol{R}^*_b)-1}{2}\right)\right|.
\end{equation}
Chordal distance~\eqref{chordal} intuitively represents the square root of the sum of sqaures of all the 9 parameters in matrix $\boldsymbol{R}^*_a-\boldsymbol{R}^*_b$. Now our goal is to set a proper threshold for each of them as in~\eqref{t-comp-each}. Similarly, a minimal rotation $\boldsymbol{R}^*$ can be described by:
\begin{equation}
\boldsymbol{R}^*=\boldsymbol{R}_{gt}\cdot Exp\left([\boldsymbol{\epsilon}^{\boldsymbol{R}}]_{\times}\right),
\end{equation}
where $\boldsymbol{R}_{gt}$ is the ground-truth rotation, $\boldsymbol{\epsilon}^{\boldsymbol{R}}\in\mathbb{R}^3$ is the noise measurement on rotation, $[\,\cdot\,]_{\times}$ denotes the skew-symmetric matrix for the size-3 vector, and $Exp(\,\cdot\,)$ is the exponential map of matrix. Then we let $\boldsymbol{\epsilon}^{\boldsymbol{R}}=\mathit{S}\cdot\boldsymbol{e}$ where $\boldsymbol{e}$ is a random vector of unit length and $\mathit{S}$ denotes its scale, and according to~\cite{barfoot2017state}, the geodesic error between $\boldsymbol{R}^*$ and $\boldsymbol{I}_3$ can be written as:
\begin{equation}
d_{\text{geo}}(\boldsymbol{R}^*,\boldsymbol{I}_3)=\mathit{S},
\end{equation}
and based on the triangular inequality of geodesic distances, we can then have that
\begin{equation}
d_{\text{geo}}(\boldsymbol{R}^*_a,\boldsymbol{R}^*_b) \leq d_{\text{geo}}(\boldsymbol{R}^*_a,\boldsymbol{I}_3)+d_{\text{geo}}(\boldsymbol{R}_b^*,\boldsymbol{I}_3)=2\mathit{S},
\end{equation}
so that the chordal distance between $\boldsymbol{R}^*_a$ and $\boldsymbol{R}^*_b$ satisfies:
\begin{equation}
d_{\text{chord}}(\boldsymbol{R}^*_a,\boldsymbol{R}^*_b)\leq 2\sqrt{2}\sin \left( \mathit{S} \right).
\end{equation}

Afterwards, the threshold for the L1-norm distance of each of the 9 element in rotation should be:
\begin{equation}\label{r-comp-each}
\theta^{r}=\left|r^*_{aq}-r^*_{bq}\right|\leq \mu\cdot\frac{2\sqrt{2}\sin \left( \mathit{S} \right)}{3},\,(\forall q\in\{1,2,\dots,9\})
\end{equation}
where $\mu\geq1$ and we can choose $\mu=1.2$ for practical use, similar to~\eqref{t-comp-each}. Note that in practice, if the maximum distances (diameters) of the point cloud in the 3 axes (X,Y and Z) are $D_X$, $D_Y$ and $D_Z$ and their mean value is $\widetilde{D}$, we can provide an empirical choice for $\mathit{S}$ such that
\begin{equation}
\mathit{S}=10\frac{\sigma}{\widetilde{D}}.
\end{equation}

Up to now, we can render our fast compatibility checking method that we name \textit{CompatibilityStaircase} because it is operated like a staircase, only when the first condition is satisfied could we proceed to the second condition, as demonstrated in Algorithm~\ref{Staircase}. (Note that we use $\boldsymbol{V}(y)$ to denote the $y_{th}$ entry of the vector $\boldsymbol{V}$.)

\textbf{Description of Algorithm~\ref{Staircase}:} With two minimal models ($\boldsymbol{R}^*_{a},\boldsymbol{t}^*_{a}$) and ($\boldsymbol{R}^*_{b},\boldsymbol{t}^*_{b}$) estimated from two minimal subsets, we vectorize $\boldsymbol{R}^*_{a}$ (and $\boldsymbol{R}^*_{b}$) in column-wise and stack it with $\boldsymbol{t}^*_{a}$ (and $\boldsymbol{t}^*_{b}$) to form the size-12 vector $\boldsymbol{T}_{a}$ (and $\boldsymbol{T}_{b}$) (\textbf{line 1}). Subsequently, we check the element-wise compatibility for the 12 pairs of elements based on condition~\eqref{t-comp-each} and~\eqref{r-comp-each} in sequence. If the $i_{th}$ element pair is not mutually compatible, then the rest $12-i$ element pairs are no longer required to be checked. This operation could be easily realized by \textbf{lines 2-9}, where `$comp=1$' defines that the two models are compatible with each other whereas `$comp=1$' symbolizes their mutual incompatibility. Note that the operations mainly are subtracting, getting absolute values, and boolean conditions, which are all rather fast to implement.

\subsubsection{Parameter Averaging for Consensus Building}

Once two mutually compatible models (subsets) are found, implying that two different subsets are approximately pointing to the same model, it is reasonable and necessary to establish and evaluate the consensus set over these two models. Note that in traditional RANSAC, minimal models may be easily affected by noise, hence probably deviating from the ground-truth model to a great extent. But fortunately, since we have two models now, averaging them could reduce the influence of noise. Here, we prefer parameter averaging to applying the non-minimal solvers (e.g. SVD~\cite{arun1987least}) because: (i) the former is more time-efficient, and (ii) the advantage of non-minimal solvers can be hardly shown here because the correspondence number is relatively small (no greater than 6). 

For translation, we directly adopt the mean vector:
\begin{equation}\label{t-average}
\boldsymbol{t}^{\circ}=\frac{\boldsymbol{t}_a+\boldsymbol{t}_b}{2},
\end{equation}
while for rotation, we adopt the geodesic L2-mean~\cite{hartley2013rotation} such that
\begin{equation}\label{R-average}
\boldsymbol{R}^{\circ}=\boldsymbol{R}_a\cdot Exp\left(\frac{log(\boldsymbol{R}_a^{\top}\boldsymbol{R}_b)}{2}\right),
\end{equation}
where $log(\,\cdot\,)$ denotes the logarithm of matrix. Here, we introduce functions: \textit{AverageRotPara}($\boldsymbol{R}_a,\boldsymbol{R}_b$), \textit{AverageTranPara}($\boldsymbol{t}_a,\boldsymbol{t}_b$), to represent the parameter averaging operations in~\eqref{R-average} and~\eqref{t-average}, respectively.

Subsequently, we can compute the residual errors of the correspondences in set $\mathcal{C}$ using $\boldsymbol{R}^{\circ}$ and $\boldsymbol{t}^{\circ}$ and then build the consensus set. (See \textbf{lines 18-29} in Algorithm~\ref{DANIEL-algo})

\begin{figure*}[t]
\centering

\begin{tabular}{cc}
\begin{minipage}[t]{0.46\linewidth}
\centering
\includegraphics[width=1\linewidth]{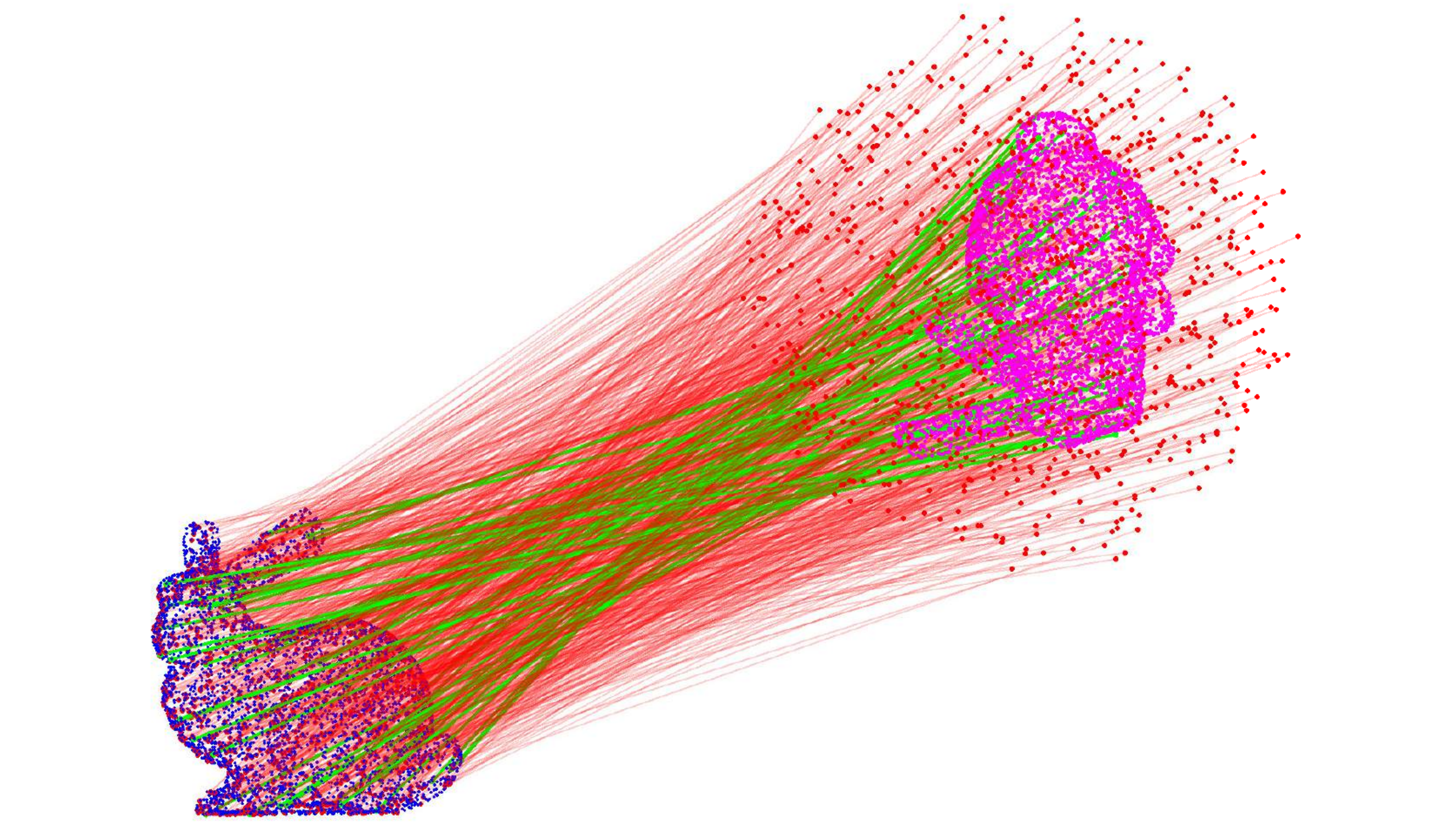}
\end{minipage}
&
\begin{minipage}[t]{0.46\linewidth}
\centering
\includegraphics[width=1\linewidth]{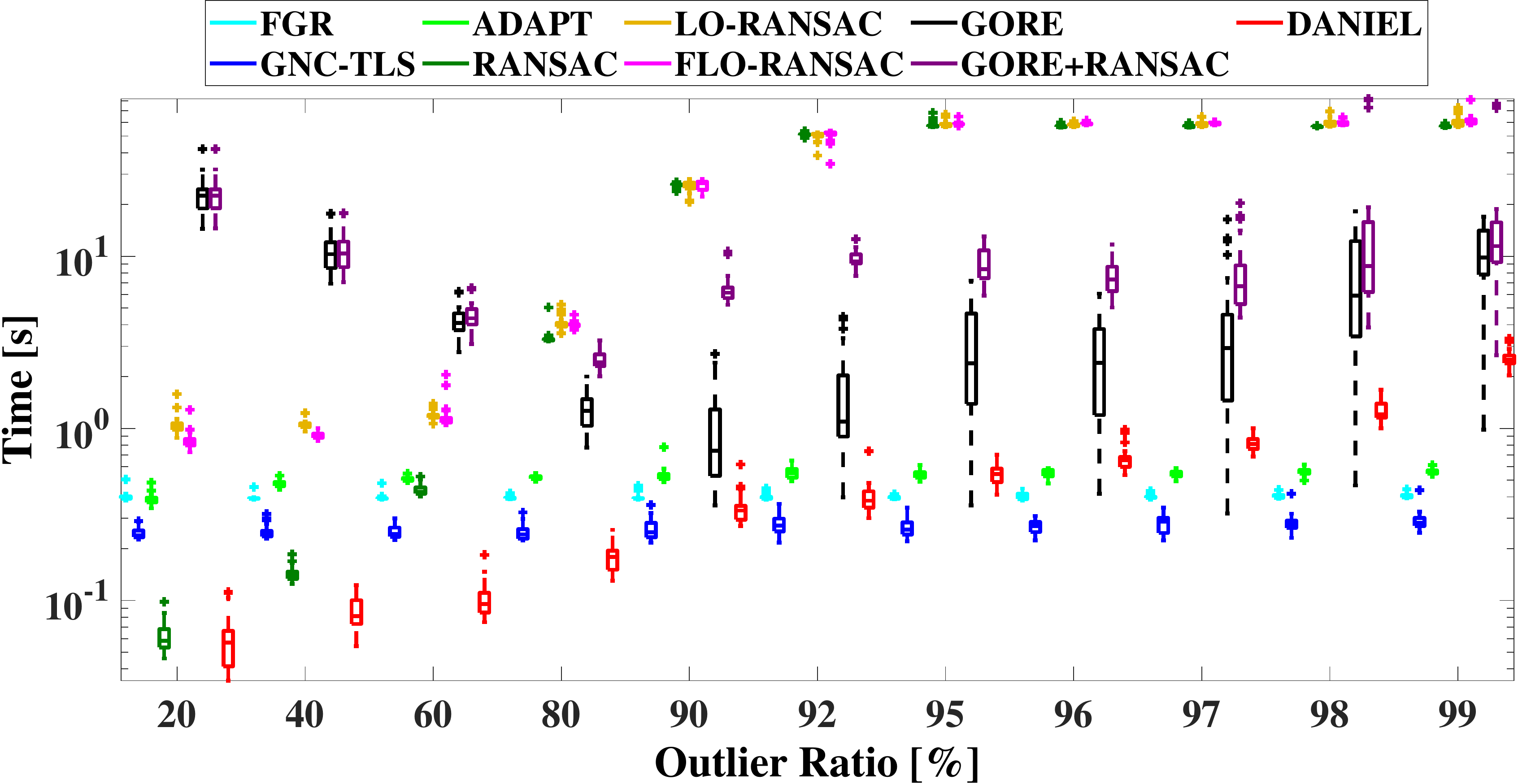}
\end{minipage}
\\
\begin{minipage}[t]{0.46\linewidth}
\centering
\includegraphics[width=1\linewidth]{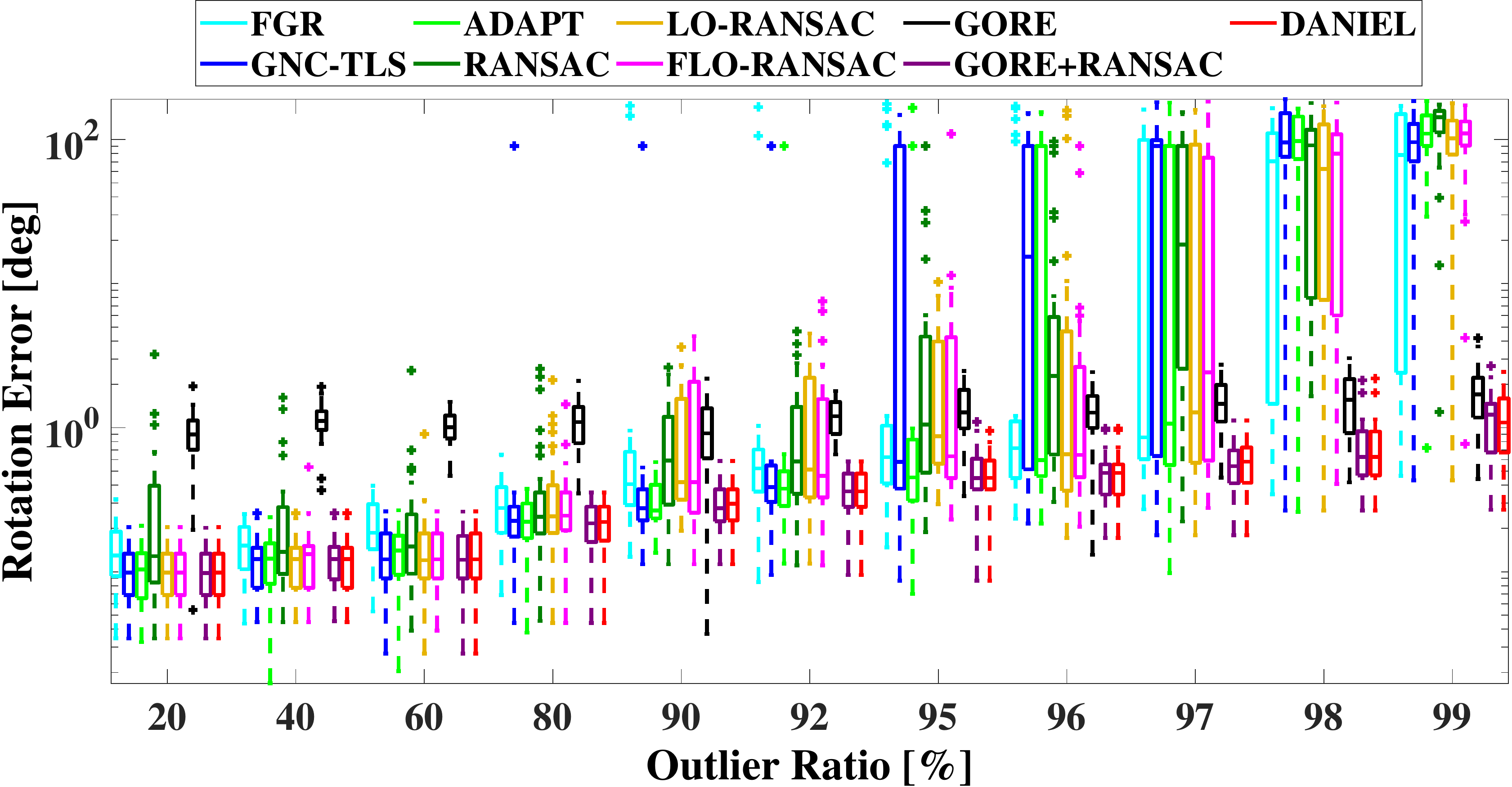}
\end{minipage}
&
\begin{minipage}[t]{0.46\linewidth}
\centering
\includegraphics[width=1\linewidth]{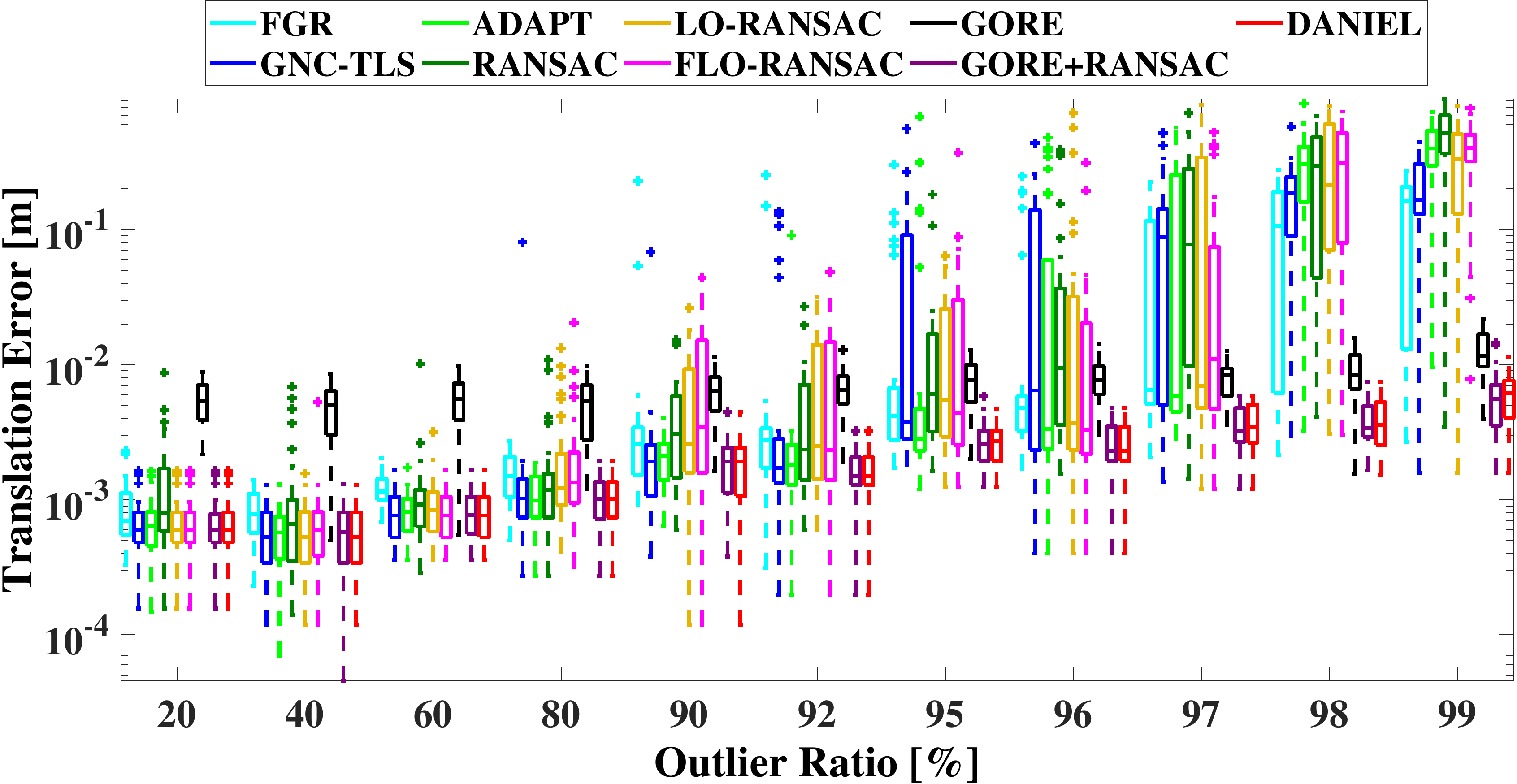}
\end{minipage} 
\end{tabular}

\caption{Standard benchmarking results in boxplot. We show the rotation and translation estimation accuracy as well as the runtime of different solvers over the `\textit{bunny}' dataset~\cite{curless1996volumetric} w.r.t. increasing outlier ratios from 20\% up to 99\%. The top-left image exemplifies the correspondences with 95\% outliers in our experimental environment, where green lines denote the inliers while red lines denote the outliers.}
\label{Benchmarking}
\vspace{-3mm}
\end{figure*}

\subsection{Probabilistic Computation of Maximum Iteration Numbers}\label{probability}

There exist two maximum iteration numbers involved in DANIEL, one for the one-point sampling in the first layer and the other for the two-point sampling in the second layer. 

In the first layer, according to~\cite{fischler1981random,chum2003locally}, given the size of the minimal subset $A_1=1$ (since we use one-point sampling), the so-far-the-best consensus set $\mathcal{N}_{best}$, and the probability of sampling at least one all-inlier subset $P_1=0.99$ (sufficiently close to 1), we can compute the expected maximum number of (random sampling) iteration such that
\begin{equation}\label{max-itr-1}
maxItr_1\geq\frac{\log{(1-P_1)}}{\log{\left(1-{\left(\frac{|\mathcal{N}_{best}|}{N}\right)}^{A_1}\right)}}.
\end{equation}
\noindent{At} the beginning of the algorithm, we are required to preset a minimum inlier number $I_{min}$ (usually we set $I_{min}=0.01N$ in practice), so replacing $|\mathcal{N}_{best}|$ with $I_{min}$ in~\eqref{max-itr-1}, we can have $maxItr_1=459$. During the sampling process, whenever a new consensus set is obtained, we update $maxItr_1$ with~\eqref{max-itr-1}. (See \textbf{line 42} in Algorithm~\ref{DANIEL-algo})

In the second layer, we set $A_2=2$ (two-point sampling) and require to sample at least two all-inlier minimal subsets, so we can set $P_2=0.995$, and then compute $maxItr_2$ as
\begin{equation}\label{max-itr-2}
maxItr_2\geq2\cdot\frac{\log{(1-P_2)}}{\log{\left(1-{\left(\frac{I_{min}}{|\mathcal{C}|}\right)}^{A_2}\right)}}.
\end{equation}
In this case, the probability of sampling two all-inlier subsets to fulfill the mutual compatibility should be $0.995^2\approx0.99$. Similarly, during the sampling process, when we construct a new consensus set after the compatibility of two subsets, we can update $maxItr_2$ with~\eqref{max-itr-2}. (See \textbf{lines 27} in Algorithm~\ref{DANIEL-algo})

\begin{figure*}[t]
\centering
\setlength\tabcolsep{0pt}
\addtolength{\tabcolsep}{0pt}
\begin{tabular}{c|cccc|ccc}

\quad & \,\, & \footnotesize{FPFH Correspondences} & \footnotesize{Registration by DANIEL} & \quad & \,\, & \footnotesize{FPFH Correspondences} & \footnotesize{Registration by DANIEL} \\
\hline 
&&&&&&&
\\
\rotatebox{90}{\scriptsize{bunny, $N=1401$, 92.78\%}}\,
& &
\begin{minipage}[t]{0.2\linewidth}
\centering
\includegraphics[width=1\linewidth]{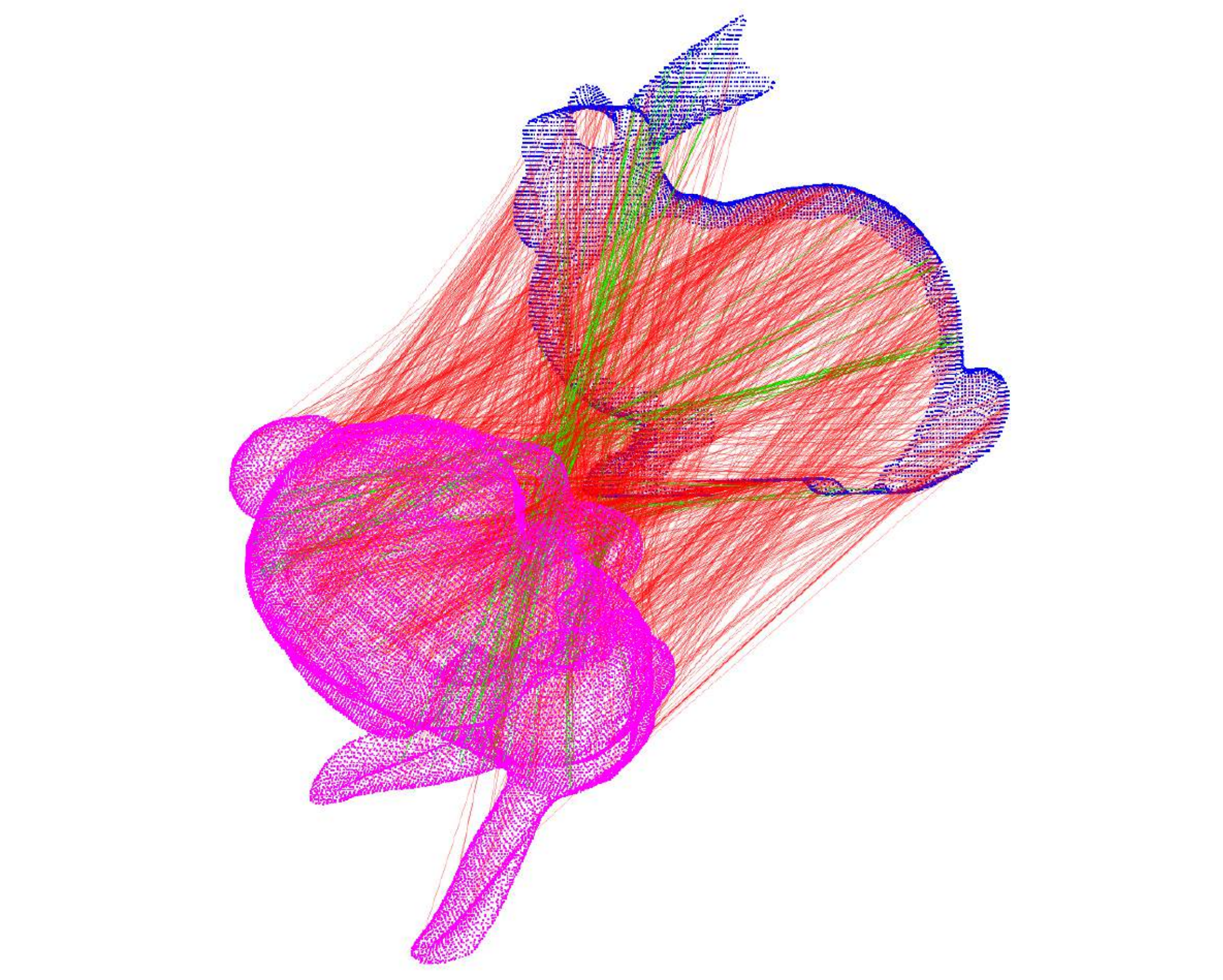}
\end{minipage}
& 
\begin{minipage}[t]{0.2\linewidth}
\centering
\includegraphics[width=1\linewidth]{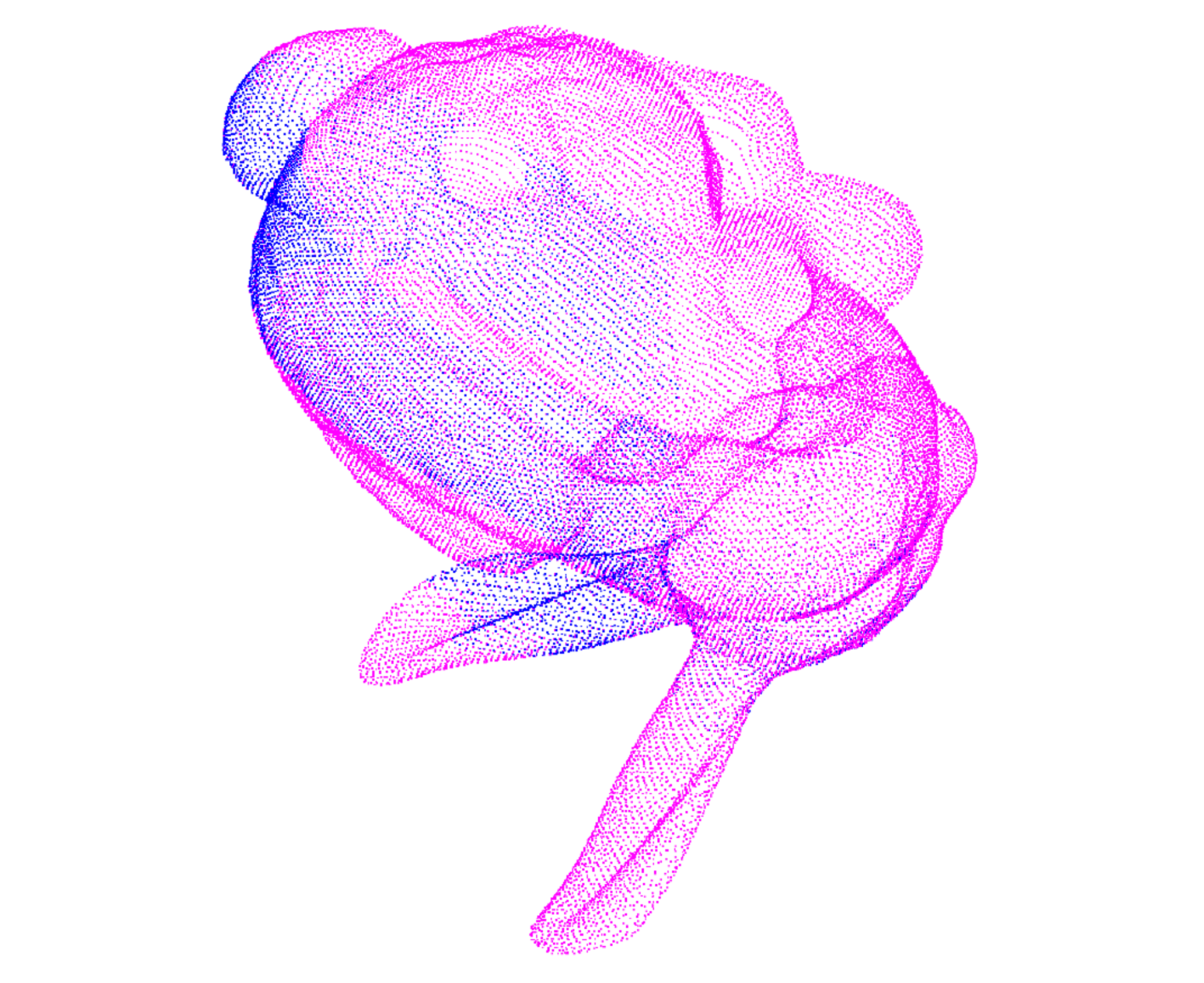}
\end{minipage}
&
\rotatebox{90}{\scriptsize{armadillo, $N=876$, 89.81\%}}\,
& &
\begin{minipage}[t]{0.19\linewidth}
\centering
\includegraphics[width=1\linewidth]{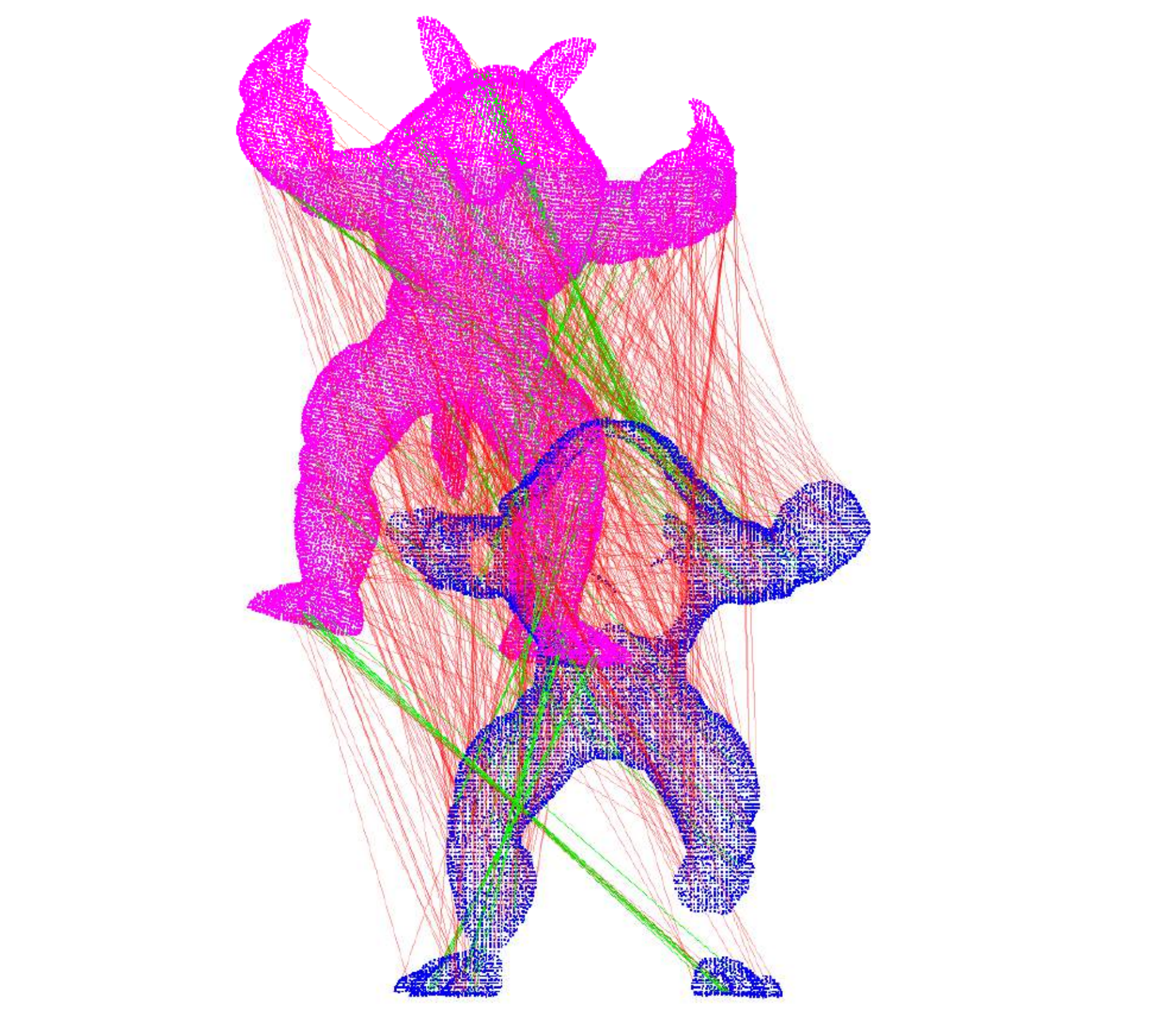}
\end{minipage}
&
\begin{minipage}[t]{0.18\linewidth}
\centering
\includegraphics[width=1\linewidth]{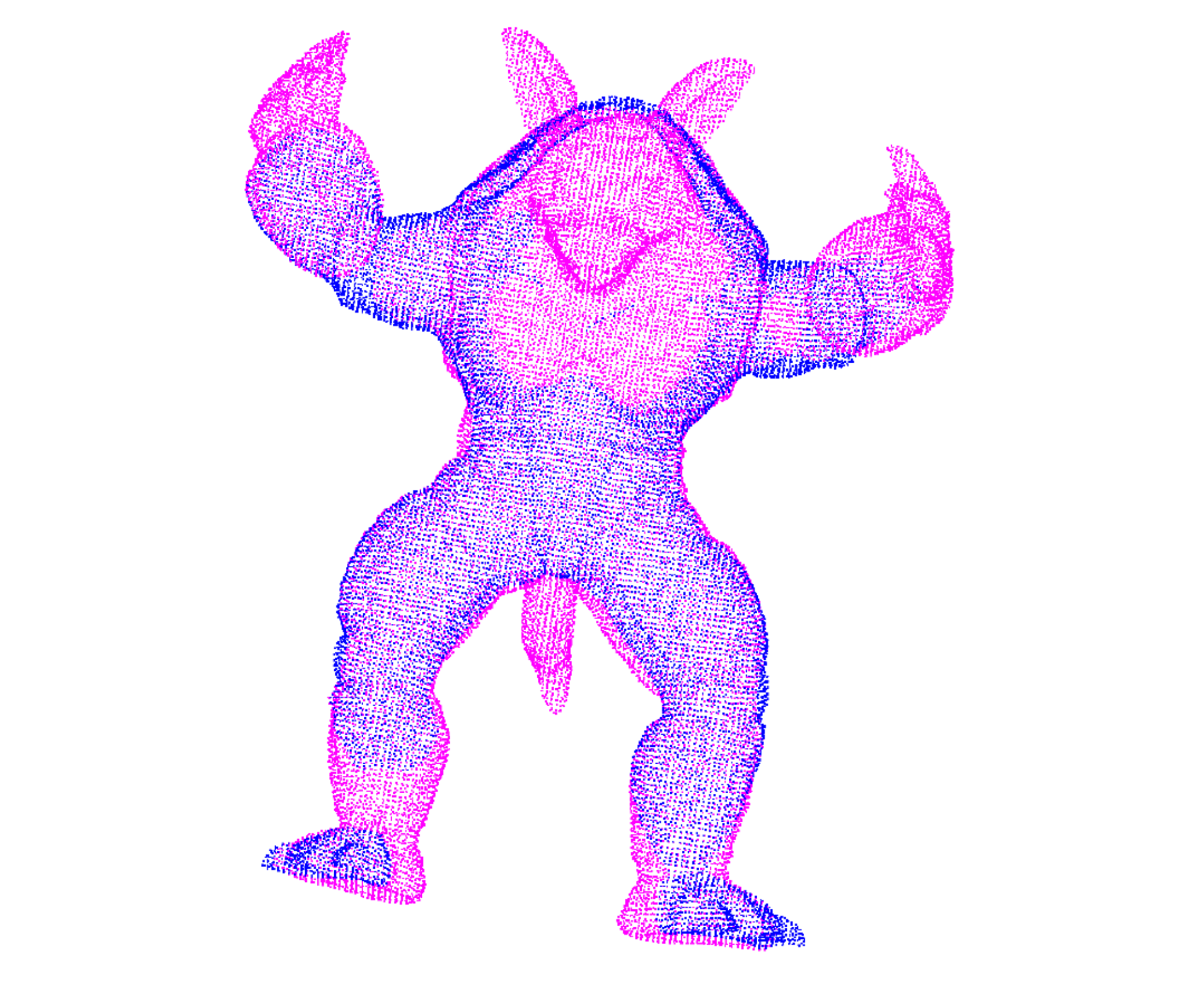}
\end{minipage}
\\
\rotatebox{90}{\scriptsize{dragon, $N=1208$, 89.97\%\quad}}\,
& &
\begin{minipage}[t]{0.2\linewidth}
\centering
\includegraphics[width=1\linewidth]{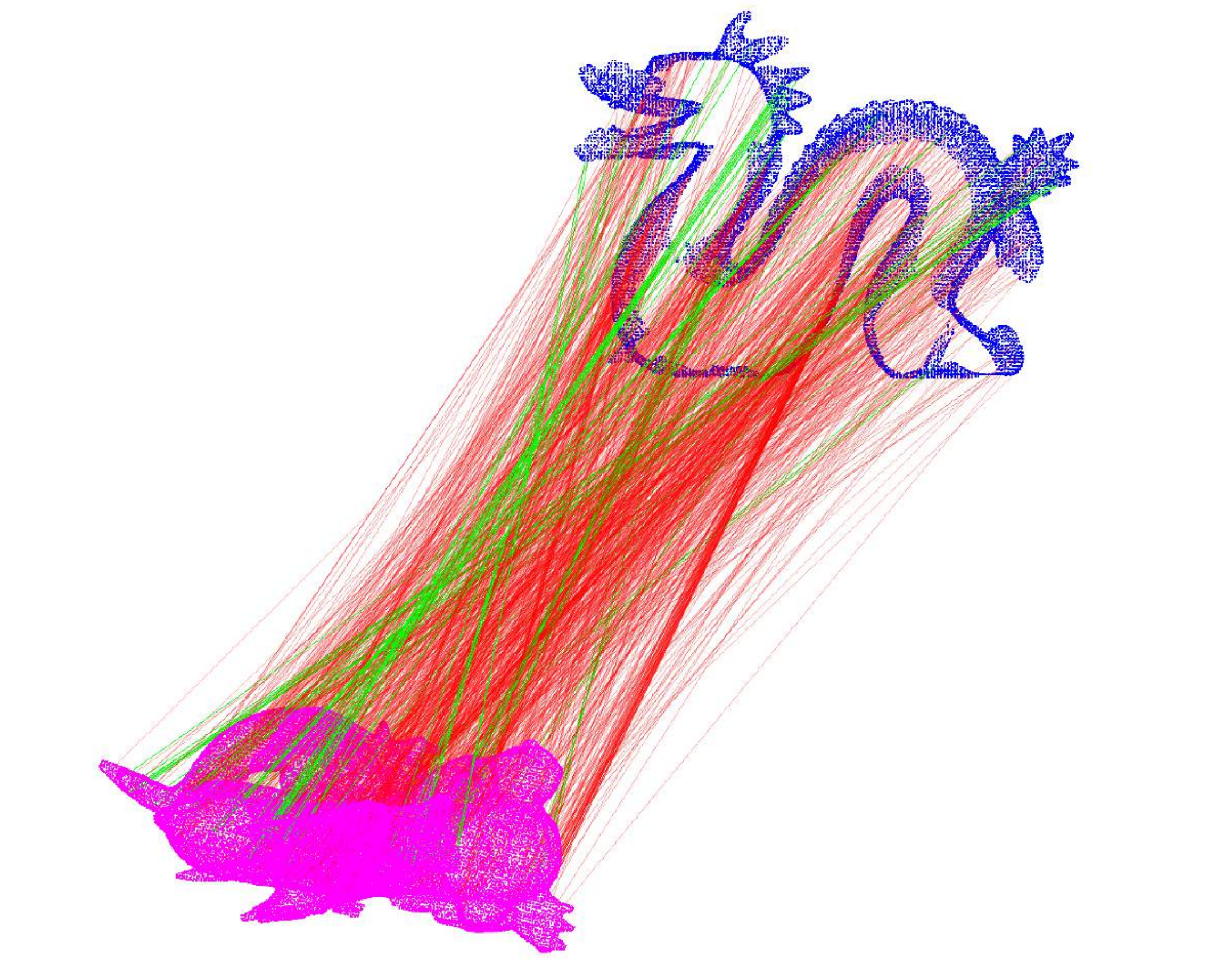}
\end{minipage}
&
\begin{minipage}[t]{0.21\linewidth}
\centering
\includegraphics[width=1\linewidth]{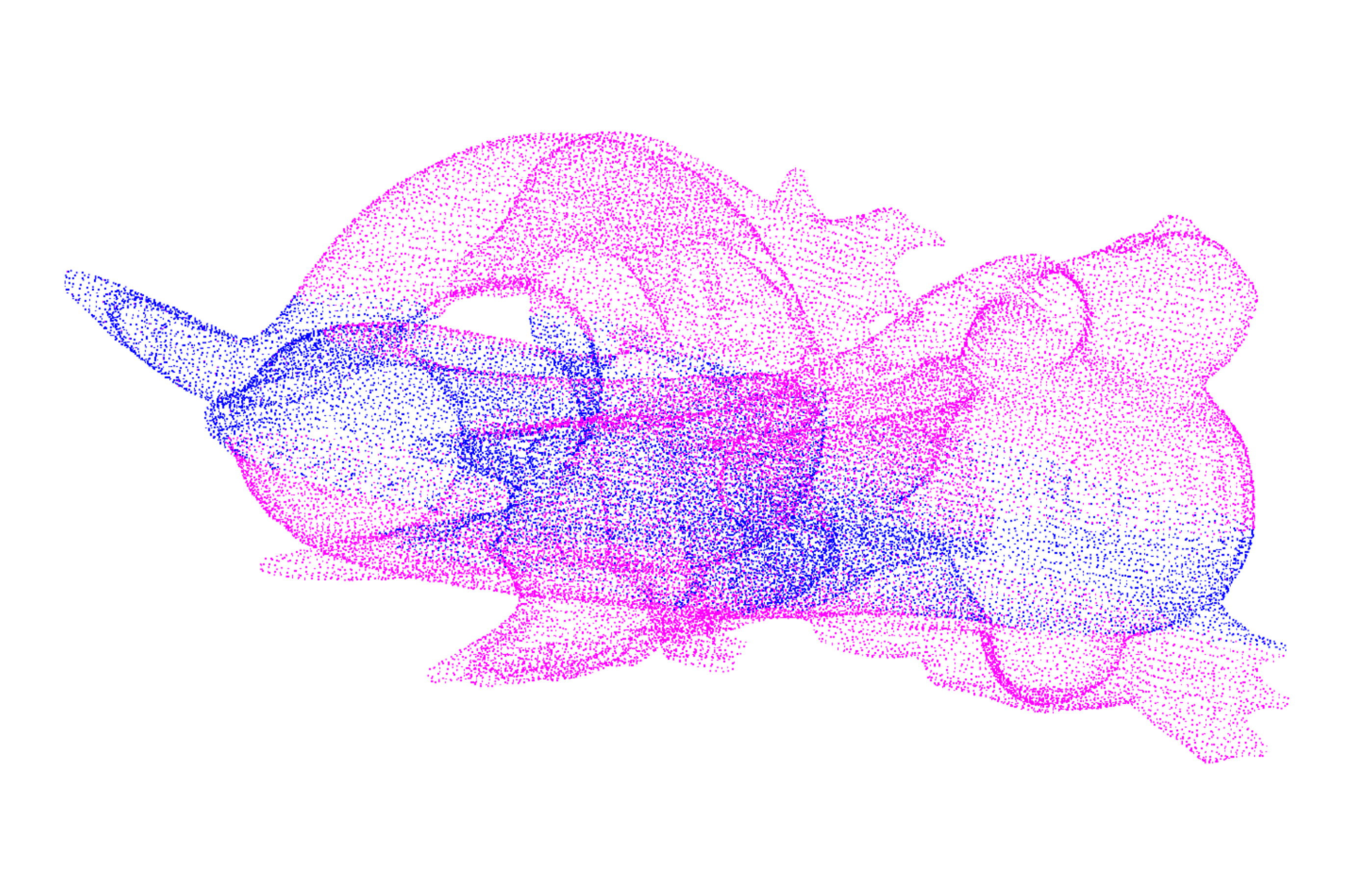}
\end{minipage}
&
\rotatebox{90}{\scriptsize{cheff, $N=1500$, 93.09\%}}\,
& &
\begin{minipage}[t]{0.2\linewidth}
\centering
\includegraphics[width=1\linewidth]{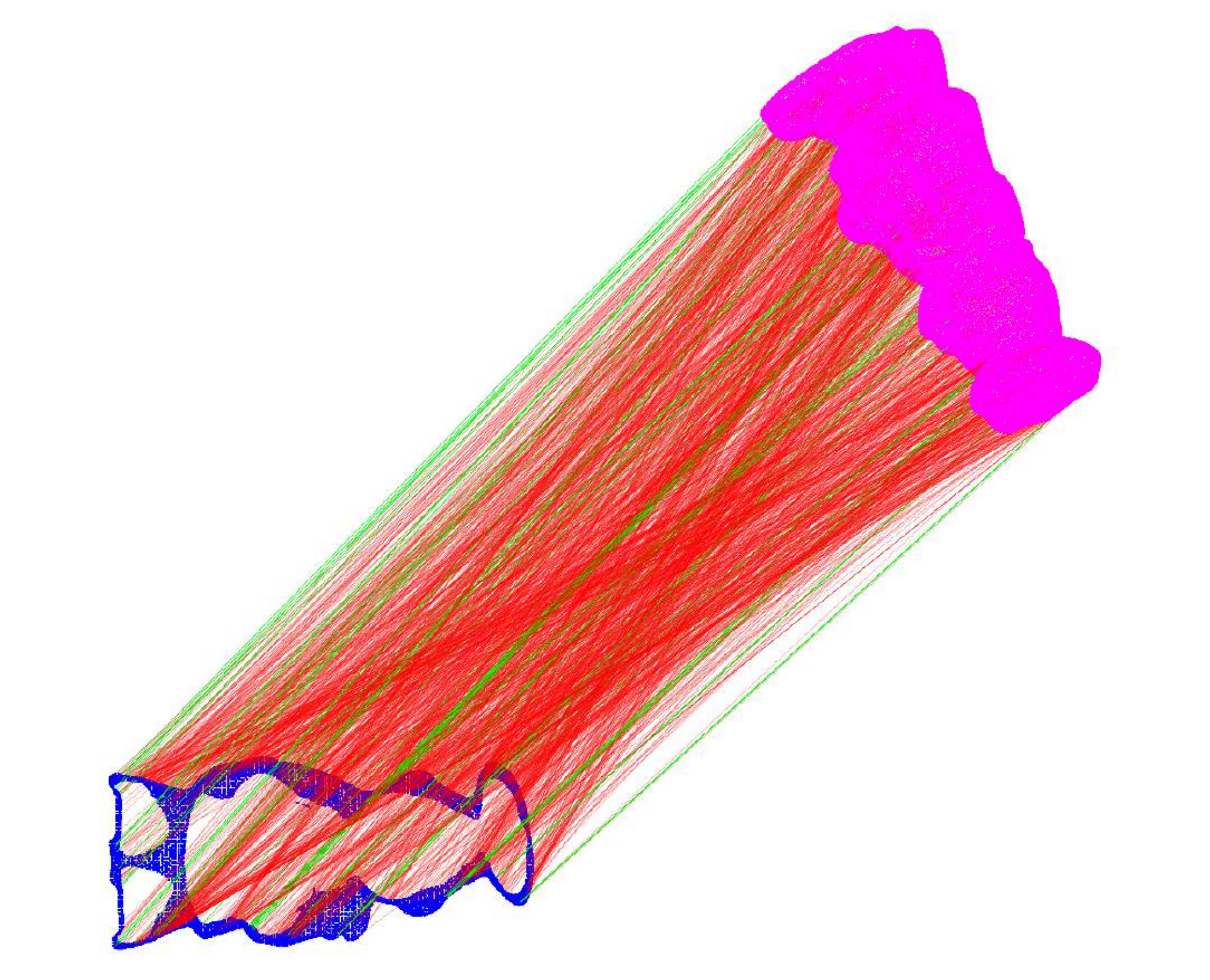}
\end{minipage}
&
\begin{minipage}[t]{0.205\linewidth}
\centering
\includegraphics[width=1\linewidth]{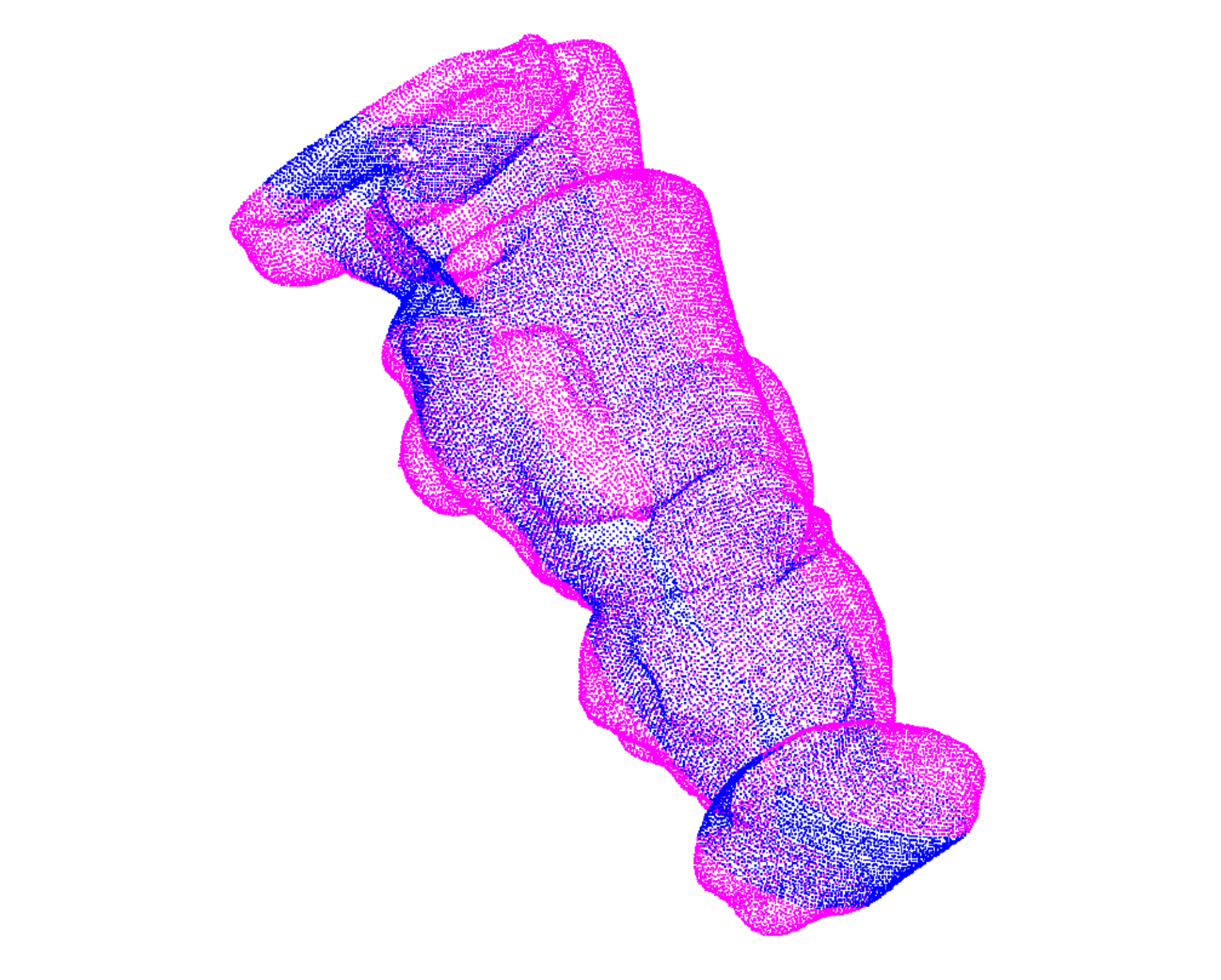}
\end{minipage}
\\
\rotatebox{90}{\scriptsize{chicken, $N=1500$, 91.73\%\quad}}\,
& &
\begin{minipage}[t]{0.2\linewidth}
\centering
\includegraphics[width=1\linewidth]{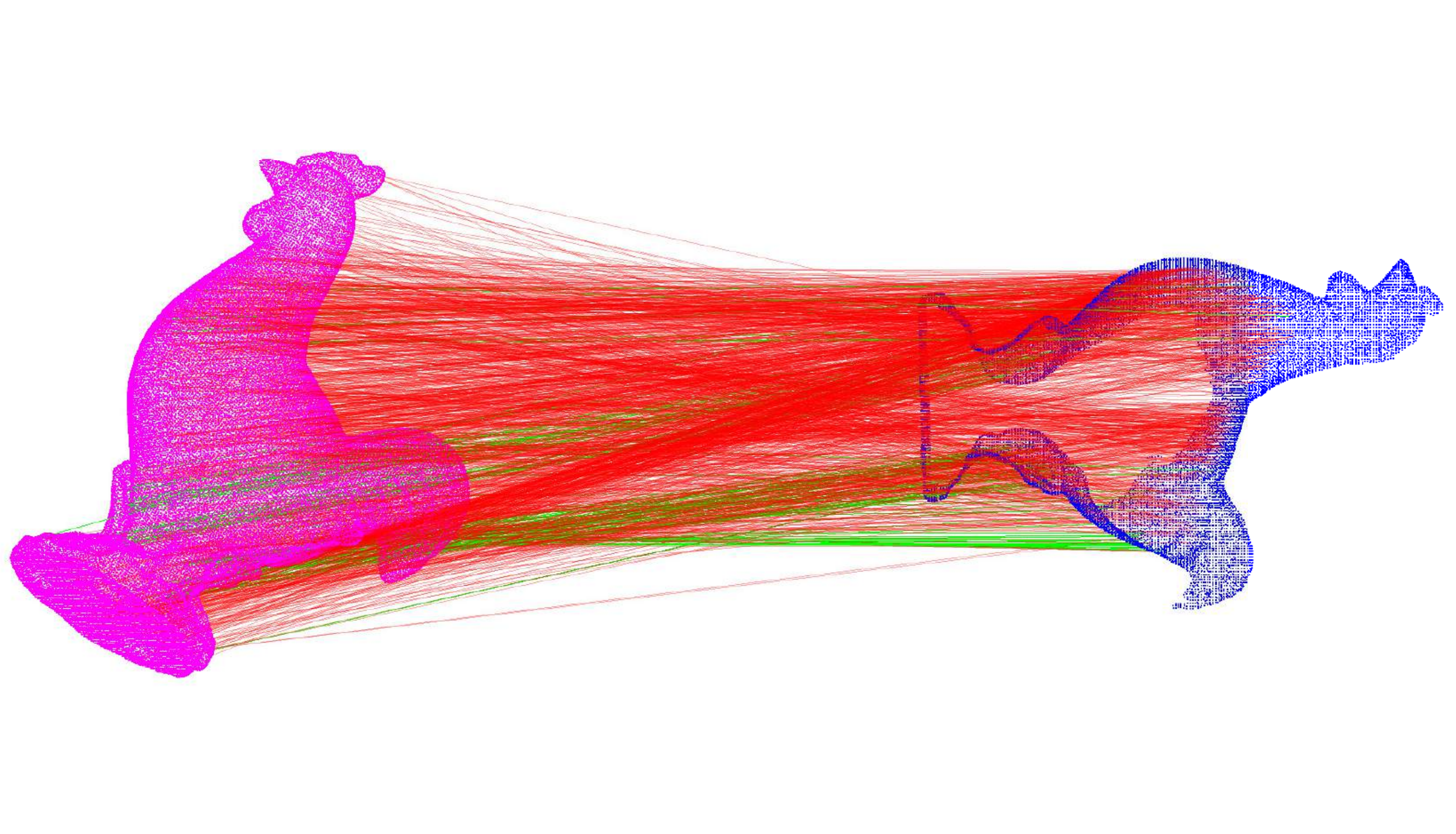}
\end{minipage}
&
\begin{minipage}[t]{0.205\linewidth}
\centering
\includegraphics[width=1\linewidth]{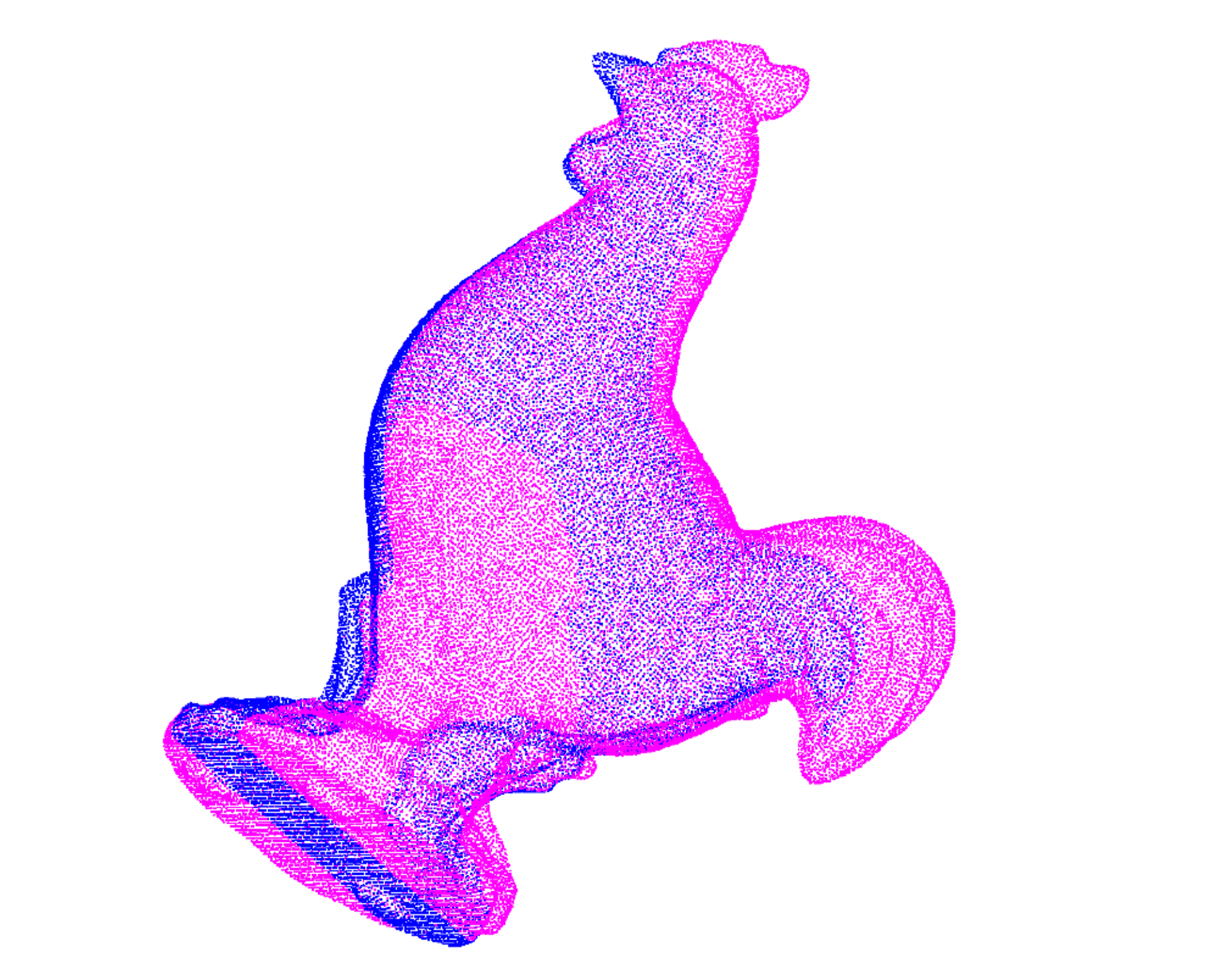}
\end{minipage}
&
\rotatebox{90}{\scriptsize{rhino, $N=1500$, 93.88\%}}\,
& &
\begin{minipage}[t]{0.2\linewidth}
\centering
\includegraphics[width=1\linewidth]{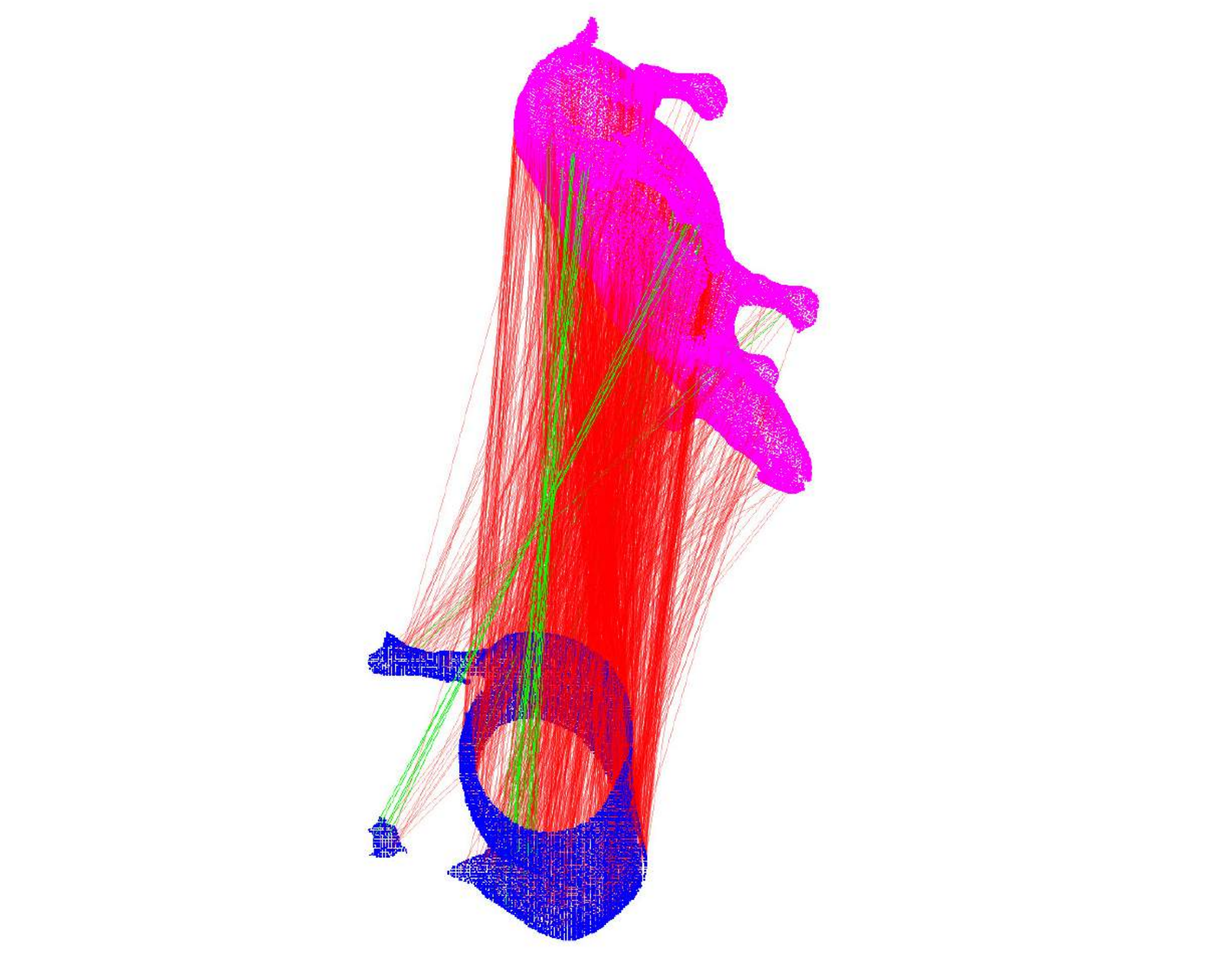}
\end{minipage}
&
\begin{minipage}[t]{0.21\linewidth}
\centering
\includegraphics[width=1\linewidth]{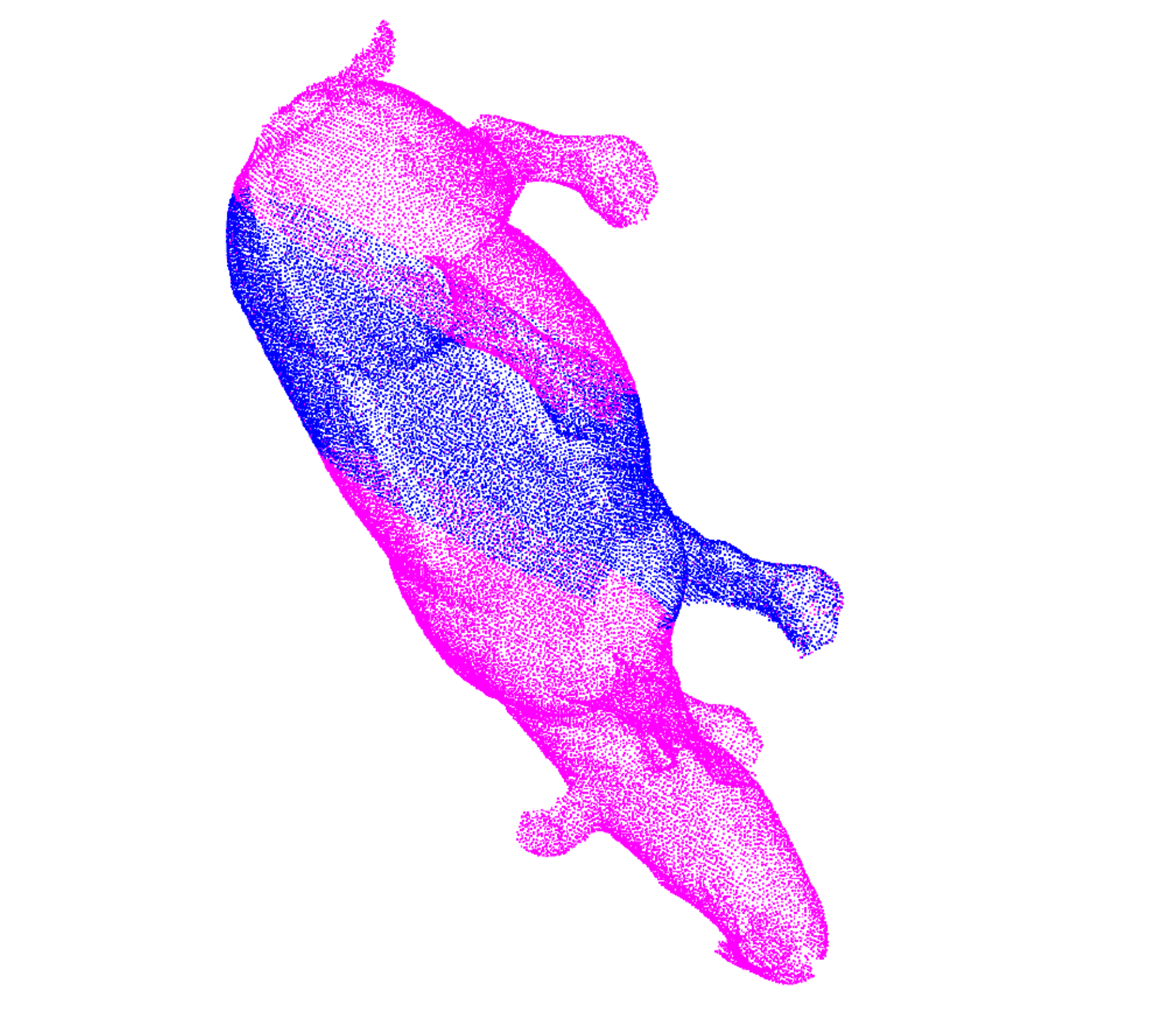}
\end{minipage}
\\
\rotatebox{90}{\scriptsize{parasauro, $N=1500$, 92.98\% \quad}}\,
& &
\begin{minipage}[t]{0.2\linewidth}
\centering
\includegraphics[width=1\linewidth]{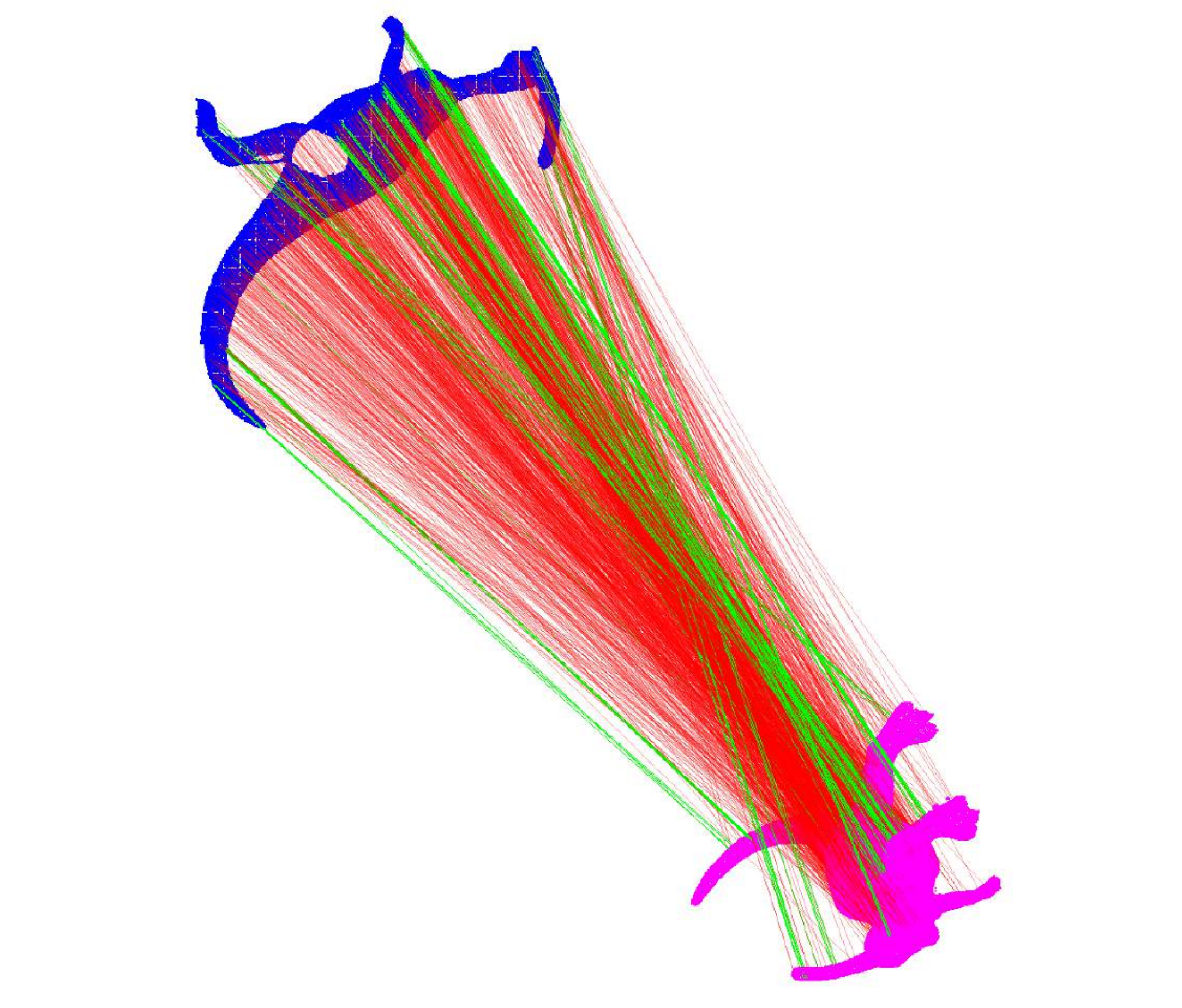}
\end{minipage}
&
\begin{minipage}[t]{0.21\linewidth}
\centering
\includegraphics[width=1\linewidth]{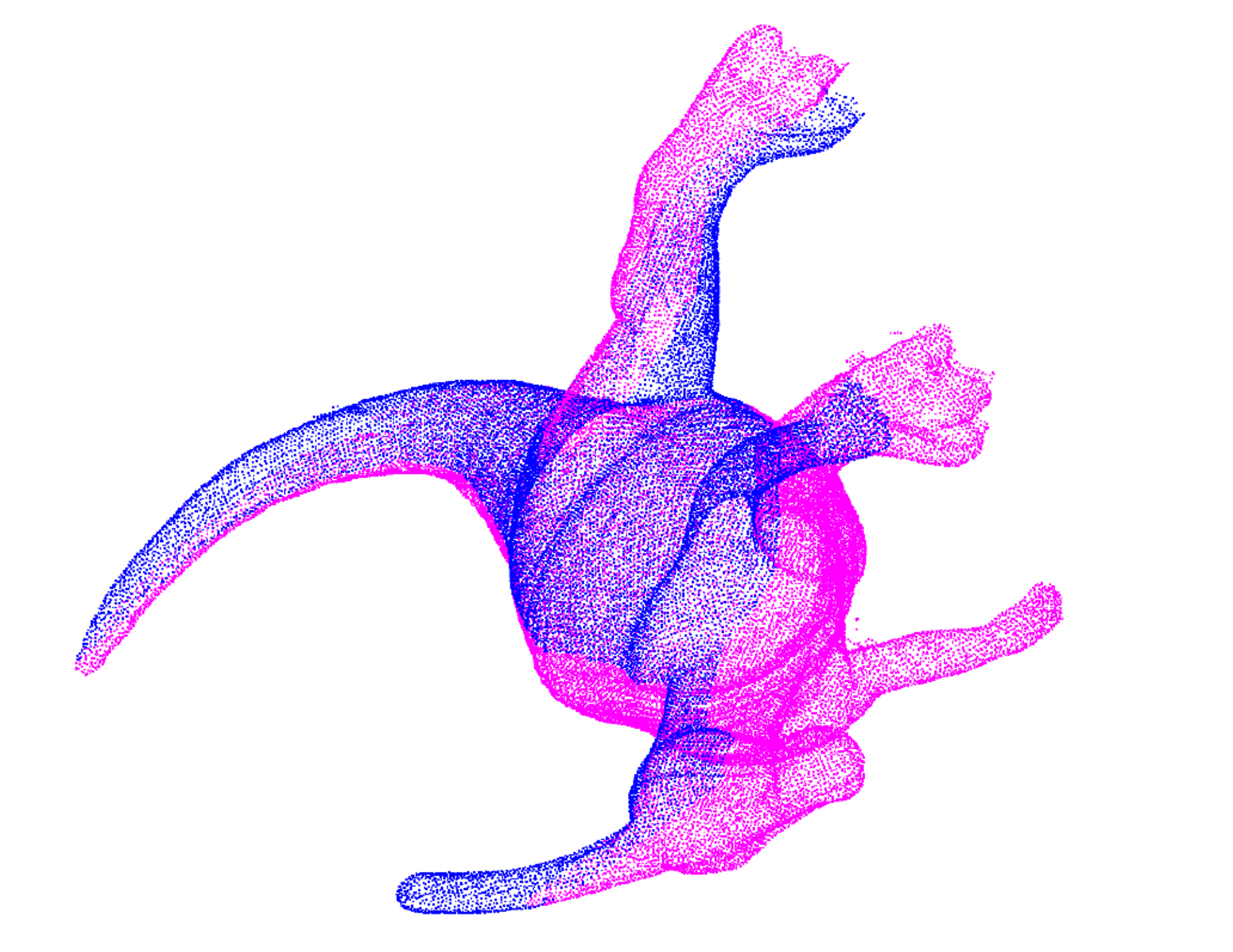}
\end{minipage}
&
\rotatebox{90}{\scriptsize{T-rex, $N=1500$, 91.95\%}}\,
& &
\begin{minipage}[t]{0.2\linewidth}
\centering
\includegraphics[width=1\linewidth]{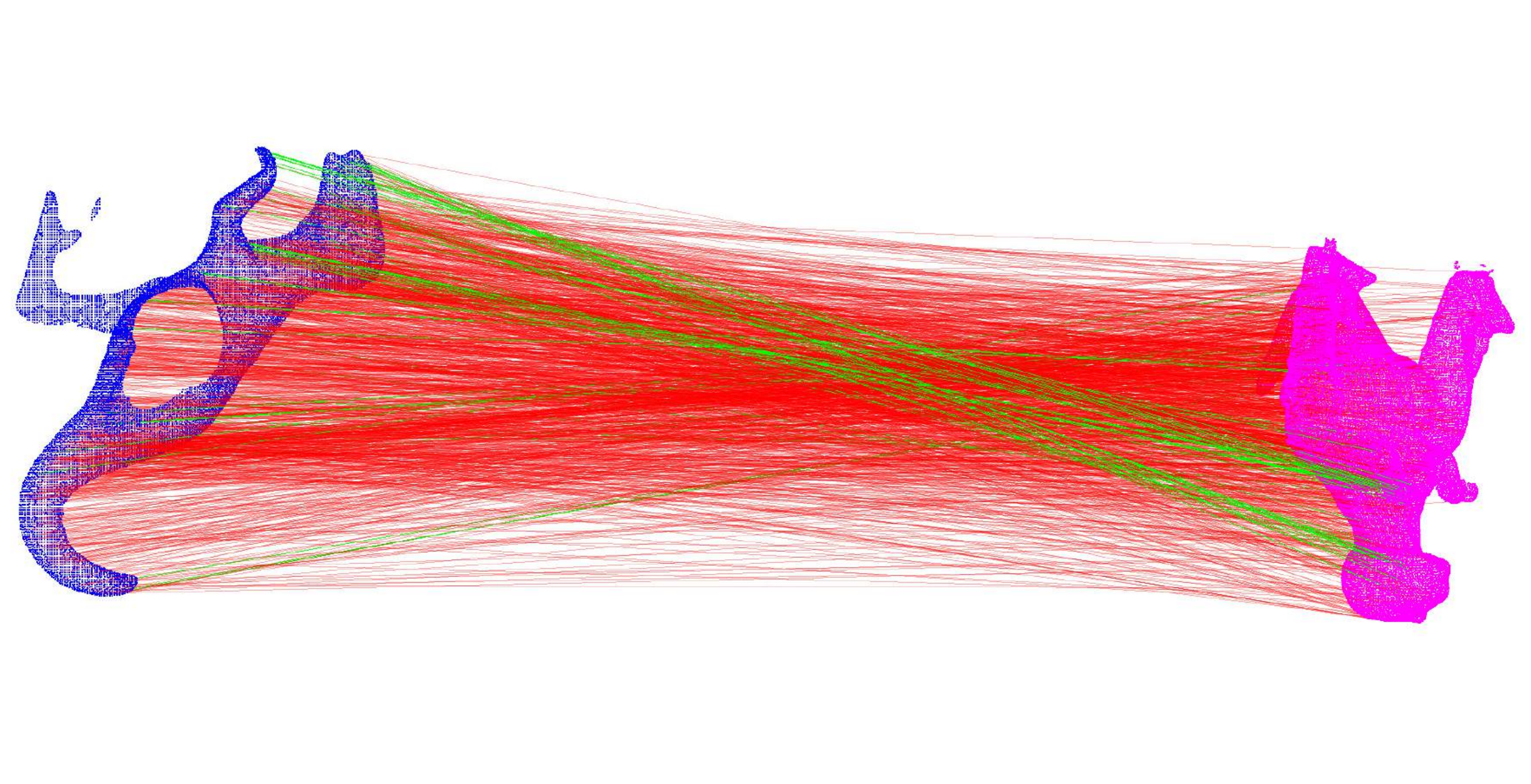}
\end{minipage}
&
\begin{minipage}[t]{0.21\linewidth}
\centering
\includegraphics[width=1\linewidth]{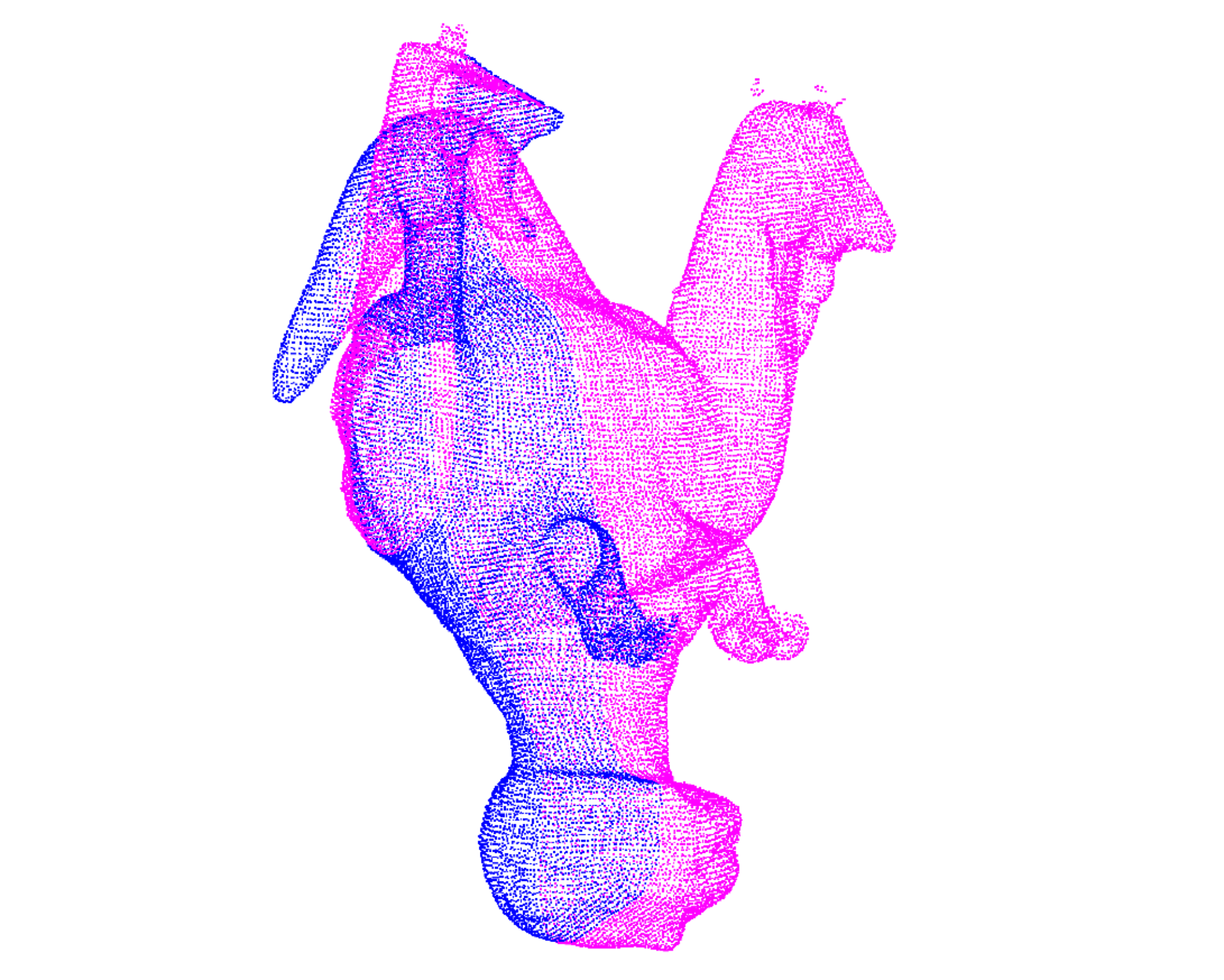}
\end{minipage}

\end{tabular}

\caption{Qualitative registration results (boxplot) on multiple real point cloud datasets using our solver DANIEL. The first column shows the correspondences matched by FPFH~\cite{rusu2009fast} where green lines denote the inliers while red lines denote the outliers. The second column displays the registration (projection) result using the transformation estimated by DANIEL.}
\label{qualit-partial}
\vspace{-3mm}
\end{figure*}

\subsection{Main Algorithm}

The pseudocode of the main algorithm of DANIEL is provided in Algorithm~\ref{DANIEL-algo}. 

\textbf{Description of Algorithm~\ref{DANIEL-algo}:} After the initialization setup, we start the first layer of one-point random sampling with rigidity examining, as described in Section~\ref{first-layer}. If the size of set $\mathcal{C}$ exceeds $I_{min}$, meaning that $n$ is likely to be an inlier, we then move on to the second layer of two-point random sampling, as described in Section~\ref{second-layer}. This layer has two sequential steps: (i) continuous sampling and model estimating with mutual compatibility checking, and (ii) parameter averaging to build the consensus set after a pair of compatible models are detected. In both sampling layers, we update the best consensus set (replacing the smaller consensus set with the larger one) as well as the maximum iteration numbers (according to Section~\ref{probability}) iteratively. Note that the consensus updating and maximizing procedure in the second layer is conducted within set $\mathcal{C}$, while in the first layer it uses the complete correspondence set $\mathcal{N}$. Eventually, when the first layer converges\footnote[1]{Here, convergence means that the actual iteration number reaches the current maximum iteration number computed with the best consensus set so far using formulation~\eqref{max-itr-1} or~\eqref{max-itr-2}. It is the stop condition of random sampling.}, the best consensus set can be applied to obtain the final inlier set and to compute the optimal transformation for this registration problem.

\textbf{`1+2$<$3'? Why is DANIEL Faster?} Our DANIEL uses a one-point sampling layer plus an inner two-point sampling layer, while traditional RANSAC adopts one single three-point sampling layer. Though it seems that 1+2 should be equal to 3, DANIEL can be in fact up to over 10 thousand times faster than RANSAC. The main reasons are given as follows.

First, in the RANSAC paradigm, the theoretical maximum iteration number of finding the best consensus set should be
\begin{equation}\label{max-itr}
maxItr^{\star}=\frac{\log{(1-0.99)}}{\log{\left(1-{\left(\frac{N_{inlier}}{N}\right)}^{A}\right)}},
\end{equation}
where $N_{inlier}$ is the true inlier number and $A$ is the subset size. The function of the first layer is two-fold: (i) to reduce $N$, which exponentially decrease $maxItr^{\star}$, and (ii) to lower $A$ (from 3 to 2), which also exponentially decrease $maxItr^{\star}$. And when we reach the second layer, where both $N$ and $A$ are reduced, the maximum iteration number has enormously declined. Besides, the first layer only involves the computation of norms and the judging of boolean conditions, so its own computational cost is fairly low. Using such a `cheap' operation in exchange for the exponential decrease of  iteration number is completely rewarding, and that is why DANIEL is much faster than RANSAC.

Second, another reason lies in our stratified compatibility checking method. Note that in each iteration, we replace the traditional repeated consensus building operation with the compatibility checking, where the latter also only consists of norm computing and boolean conditions, so the efficiency can be greatly enhanced. Generally, the larger the problem size $N$ is, the more apparent the speed superiority of DANIEL will be. In experiments, we further compare the runtime of DANIEL against that of RANSAC (Fig.~\ref{RANSAC-vs-DANIEL}).

\begin{figure*}[t]
\centering
\setlength\tabcolsep{1pt}
\addtolength{\tabcolsep}{0pt}
\begin{tabular}{ccc}

\begin{minipage}[t]{0.32\linewidth}
\centering
\includegraphics[width=1\linewidth]{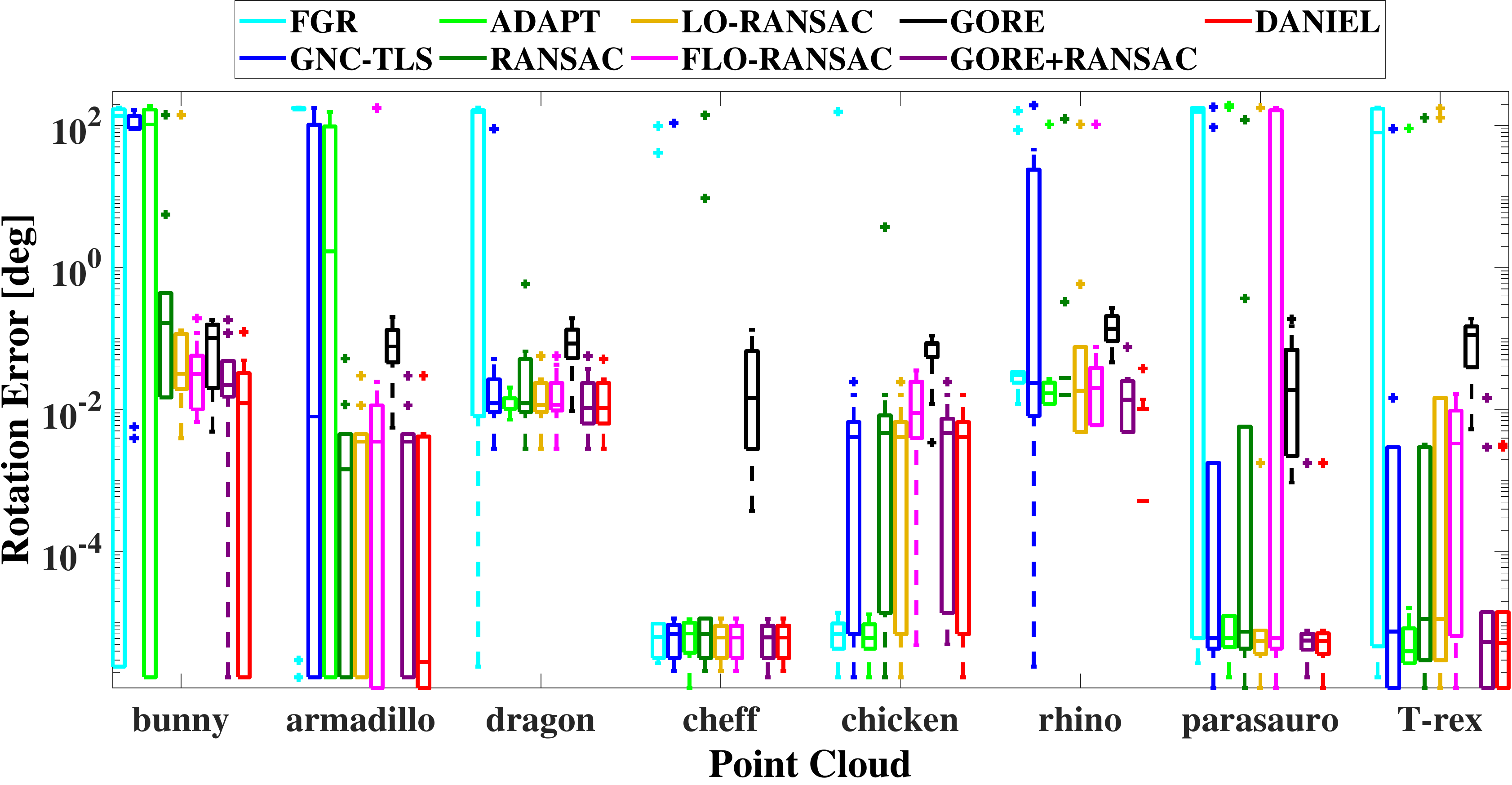}
\end{minipage}

&

\begin{minipage}[t]{0.32\linewidth}
\centering
\includegraphics[width=1\linewidth]{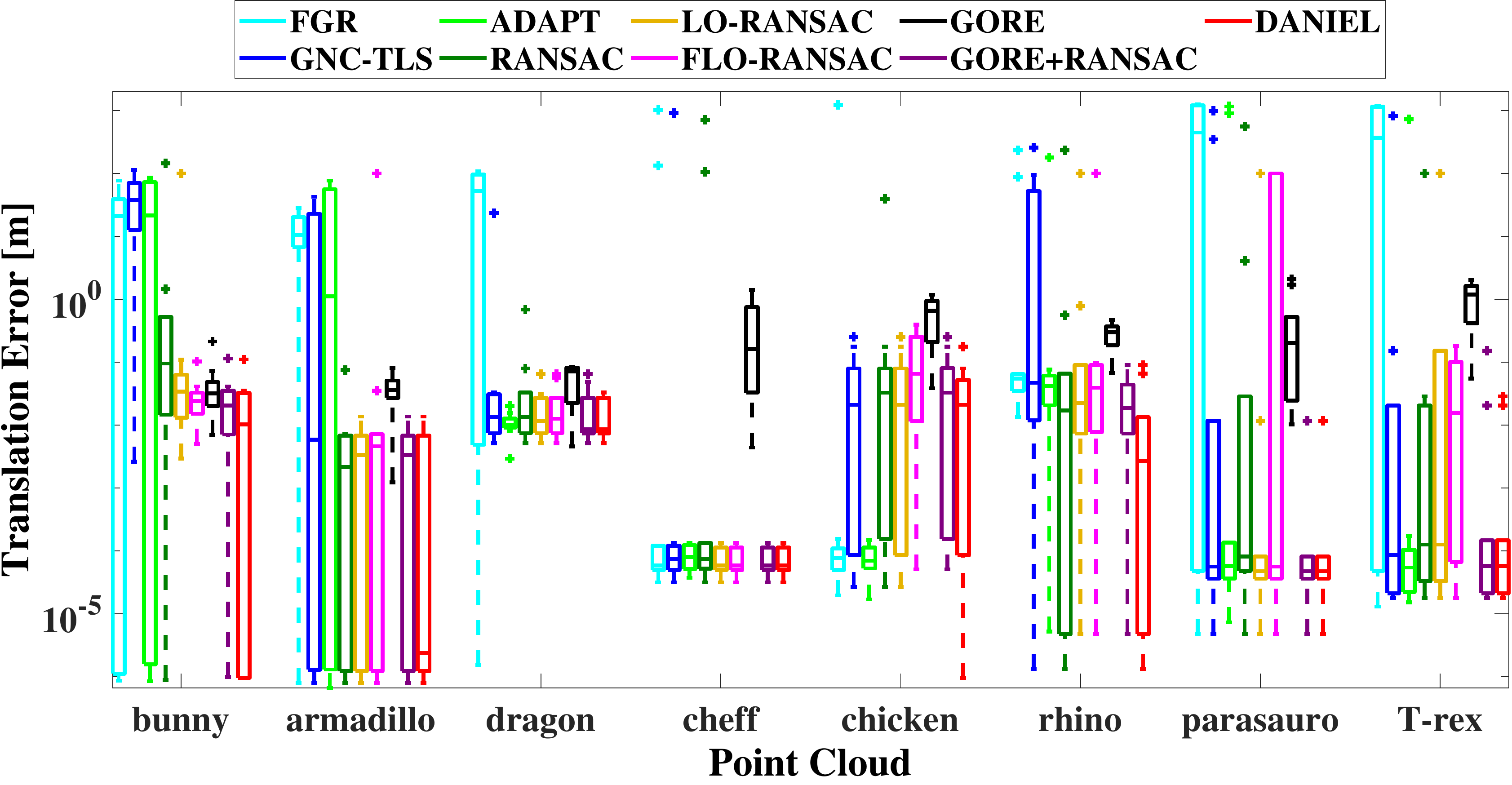}
\end{minipage}

&

\begin{minipage}[t]{0.32\linewidth}
\centering
\includegraphics[width=1\linewidth]{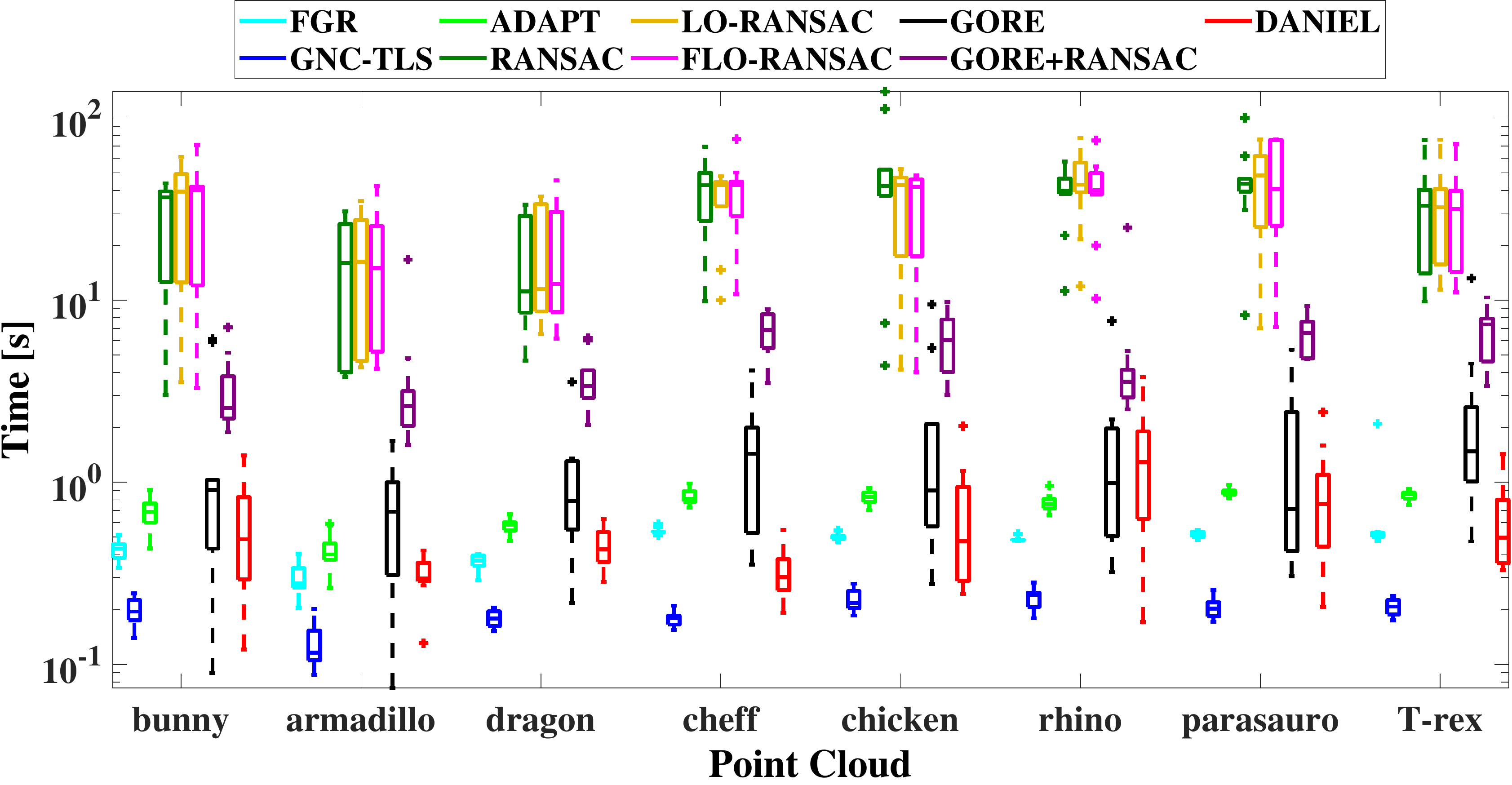}
\end{minipage}

\end{tabular}

\caption{Quantitative registration results (boxplot) on multiple real point cloud datasets. The performances (estimation errors and runtime) of different solvers over different point clouds tested are shown.}
\label{Partial-reg}
\vspace{-3mm}
\end{figure*}

\section{Experiments and Applications}\label{Experiments}

In this section, we conduct multiple comprehensive experiments on the basis of various realistic datasets in order to fully evaluate the performance of DANIEL, in comparison with other state-of-the-art competitors. All the experiments are implemented in Matlab over a laptop with a 2.8Ghz CPU and a 16GB RAM without using any parallelism programming.

\subsection{Standard Benchmarking on Semi-Synthetic Data}\label{sec-bench}

We first evaluate our DANIEL in the standard benchmarking experiment (the point cloud used is realistic while the outliers are artificially generated, thus called \textit{semi-synthetic}), in comparison with the existing state-of-the-art solvers including: (i) three non-minimal solvers: FGR~\cite{zhou2016fast}, GNC-TLS~\cite{yang2020graduated} and ADAPT~\cite{tzoumas2019outlier}, (ii) three RANSAC solvers: RANSAC~\cite{fischler1981random}, LO-RANSAC~\cite{chum2003locally} and FLO-RANSAC~\cite{lebeda2012fixing} (also called LO$^+$-RANSAC), and (iii) the guaranteed outlier removal solver GORE~\cite{bustos2017guaranteed} as well as its enhanced version GORE+RANSAC\footnote[2]{Further using RANSAC to find the best consensus set after the guaranteed outlier removal of GORE.}. Note that the RANSAC solvers are all set with 10000 maximum iterations and 0.99 confidence, and the local optimization is set with 10 iterations. For all solvers, the inlier threshold is set to $\xi=6\sigma$. For quantitative evaluation of estimation errors, we use the geodesic error~\eqref{geodesic} to represent rotation error in degrees such that
\begin{equation}\label{geodesic}
E_{\text{rot}}(\boldsymbol{R}_{gt}, \boldsymbol{R}^{\star})=\left|\arccos\left(\frac{trace({\boldsymbol{R}_{gt}}^{\top}\boldsymbol{R}^{\star})-1}{2}\right)\right|\cdot\frac{180}{\pi}^{\circ},
\end{equation}
and use L2-norm to represent translation error in meters such that
\begin{equation}\label{geodesic}
E_{\text{tran}}(\boldsymbol{t}_{gt}, \boldsymbol{t}^{\star})=\left\|\boldsymbol{t}_{gt}-\boldsymbol{t}^{\star}\right\|\, m,
\end{equation}
where $\boldsymbol{R}_{gt}$ and $\boldsymbol{t}_{gt}$ denotes the ground-truth transformation.

We adopt the `\textit{bunny}' point cloud dataset from the Stanford 3D Scanning Repository~\cite{curless1996volumetric}. This point cloud is downsampled to $N=1000$ and resized to be placed a $[-0.5,0.5]^3m$ box as our initial point set $\mathcal{X}=\{\boldsymbol{x}_i\}_{i=1}^{N}$. Then we transform $\mathcal{X}$ with a random transformation such that $\boldsymbol{R}\in SO(3)$ and $\boldsymbol{t}\in\mathbb{R}^3$ ($||\boldsymbol{t}||\leq 3$), and add random noise with $\sigma=0.01m$ to the transformed point set $\mathcal{Y}=\{\boldsymbol{y}_i\}_{i=1}^{N}$. To create outliers simulating cluttered scenes, we artificially corrupt (replace) 20\% to 99\% of the points in $\mathcal{Y}$ with completely random points inside a 3D sphere of radius 1. The boxplot benchmarking results are shown in  Fig.~\ref{Benchmarking}, in which 50 Monte Carlo runs are implemented to obtain the result data.

From Fig.~\ref{Benchmarking}, we can clearly observe that non-minimal solvers (FGR, GNC-TLS and ADAPT) fail at 90-92\% outliers, and the RANSAC solvers (RANSAC, LO-RANSAC and FLO-RANSAC) with 10000 maximum iterations break at over 95\% outliers, meanwhile requiring very long computational time with high outlier ratios, whereas GORE, GORE+RANSAC and our DANIEL are robust against as many as 99\% outliers. Furthermore, DANIEL is the fastest solver at all outlier ratios (from 20\% to 99\%) with (at least one of) the highest estimation accuracy, superior to all the other tested competitors in overall performance.

\begin{figure}[h]
\centering
\begin{minipage}[t]{1\linewidth}
\centering
\includegraphics[width=0.87\linewidth]{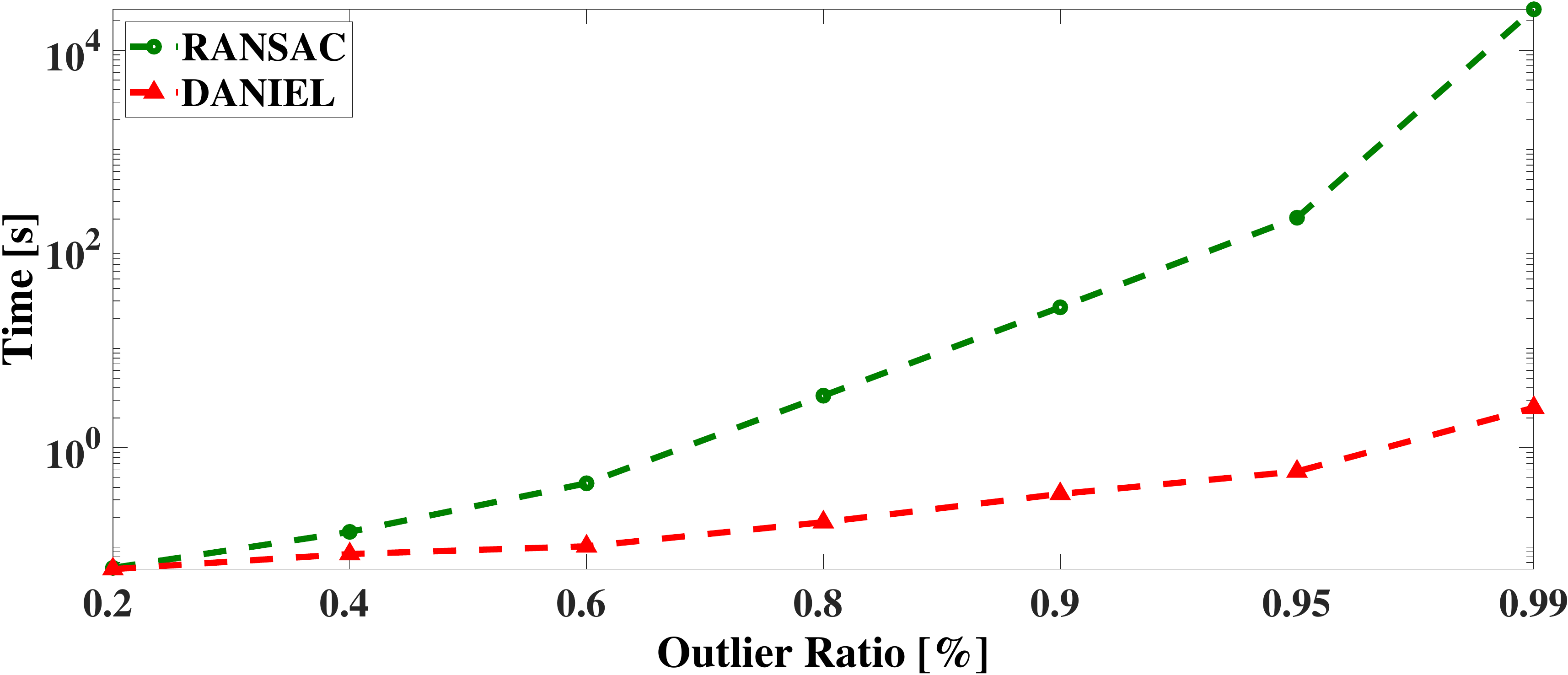}
\end{minipage}
\caption{Comparison of mean runtime between DANIEL and RANSAC w.r.t. increasing outlier ratios.}
\label{RANSAC-vs-DANIEL}
\vspace{-1mm}
\end{figure}

\subsection{Runtime Comparison against RANSAC}

Since that RANSAC has been a de-facto standard among all the robust solvers, we exclusively compare the runtime of our DANIEL with that of RANSAC, in order to more clearly observe DANIEL's performance in efficiency. Note that in this experiment, no maximum iteration number is imposed to RANSAC and we simply wait for it until convergence.

We also adopt the experimental setup in Section~\ref{sec-bench} and demonstrate the mean runtime (based on 50 data points) of these two solvers w.r.t. increasing outlier ratios from 20\% up to 95\%. But when the outlier ratio is 99\%, RANSAC would take over 7 hours per run, making it difficult to measure its mean runtime. Fortunately, we can reasonably deduce its runtime at 99\% by computing the theoretical maximum iteration based on~\eqref{max-itr}. Note that in each iteration of RANSAC, the operations are almost constant and fixed, that is, minimal sampling, model estimating and consensus building, so its runtime should be mainly dependent on its iteration number. For example, in our experiment, RANSAC's mean runtime at 95\% is found to be 7.97 times greater than that at 90\%, which fits well with the fact that the maximum iteration number required for 95\% is 8 times greater than that required for 90\% according to~\eqref{max-itr}. As a result, since that the maximum iteration number of RANSAC at 99\% outliers should be 125 greater than that at 95\% which is about 206.37 seconds in average, the runtime of RANSAC at 99\% can be computed as 206.37$\times$125$\approx$25,796 seconds.

Thus, according to Fig.~\ref{RANSAC-vs-DANIEL}, we can see that DANIEL is 75 times faster than RANSAC at 90\%, 357 times faster at 95\%, and more than 10000 times faster at 99\%, so that the efficiency superiority of DANIEL is extremely obvious.

\begin{figure*}[t]
\centering
\setlength\tabcolsep{0.1pt}
\addtolength{\tabcolsep}{0pt}
\begin{tabular}{c|cc|ccccc}

\quad &\,&\,\footnotesize{Correspondences}\, &\,&  \footnotesize{GNC-TLS} & \footnotesize{FLO-RANSAC} & \footnotesize{GORE+RANSAC} & \footnotesize{DANIEL} \\

\hline

&&\footnotesize{$N=534$, 97.94\%}& & \,\footnotesize{$133.649^{\circ},2.206m,0.065s$}\, &\,\footnotesize{$90.351^{\circ},2.177m,15.391s$}\, &\,\footnotesize{$\textbf{0.807}^{\circ},\textbf{0.024}m,0.998s$}\,&\,\footnotesize{$\textbf{0.807}^{\circ},\textbf{0.024}m,\textbf{0.259}s$}\,

\\

\rotatebox{90}{\,\,\footnotesize{\textit{Scene-01, mug}}\,}\,

& &

\begin{minipage}[t]{0.19\linewidth}
\centering
\includegraphics[width=1\linewidth]{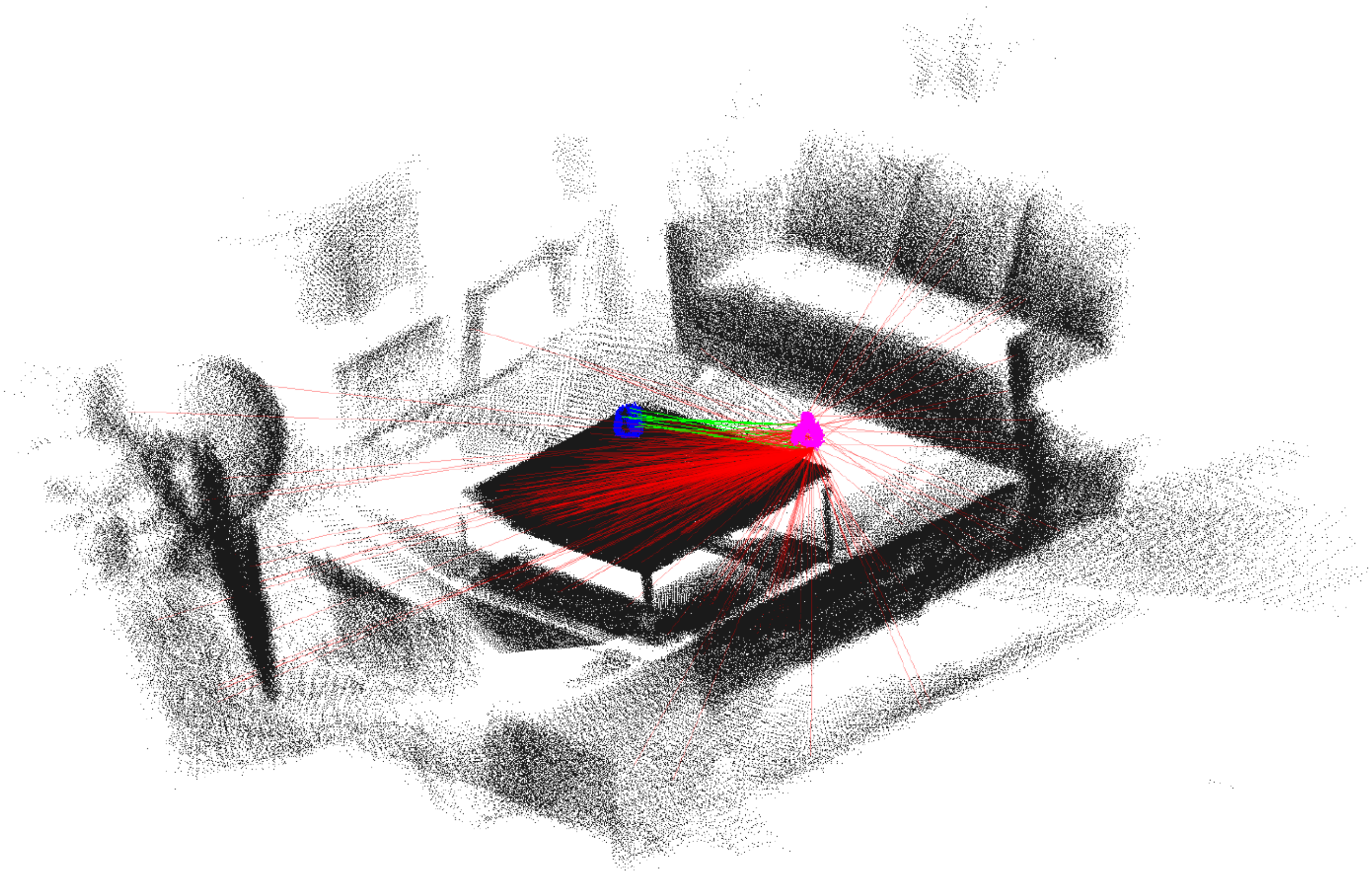}
\end{minipage}

& &

\begin{minipage}[t]{0.19\linewidth}
\centering
\includegraphics[width=1\linewidth]{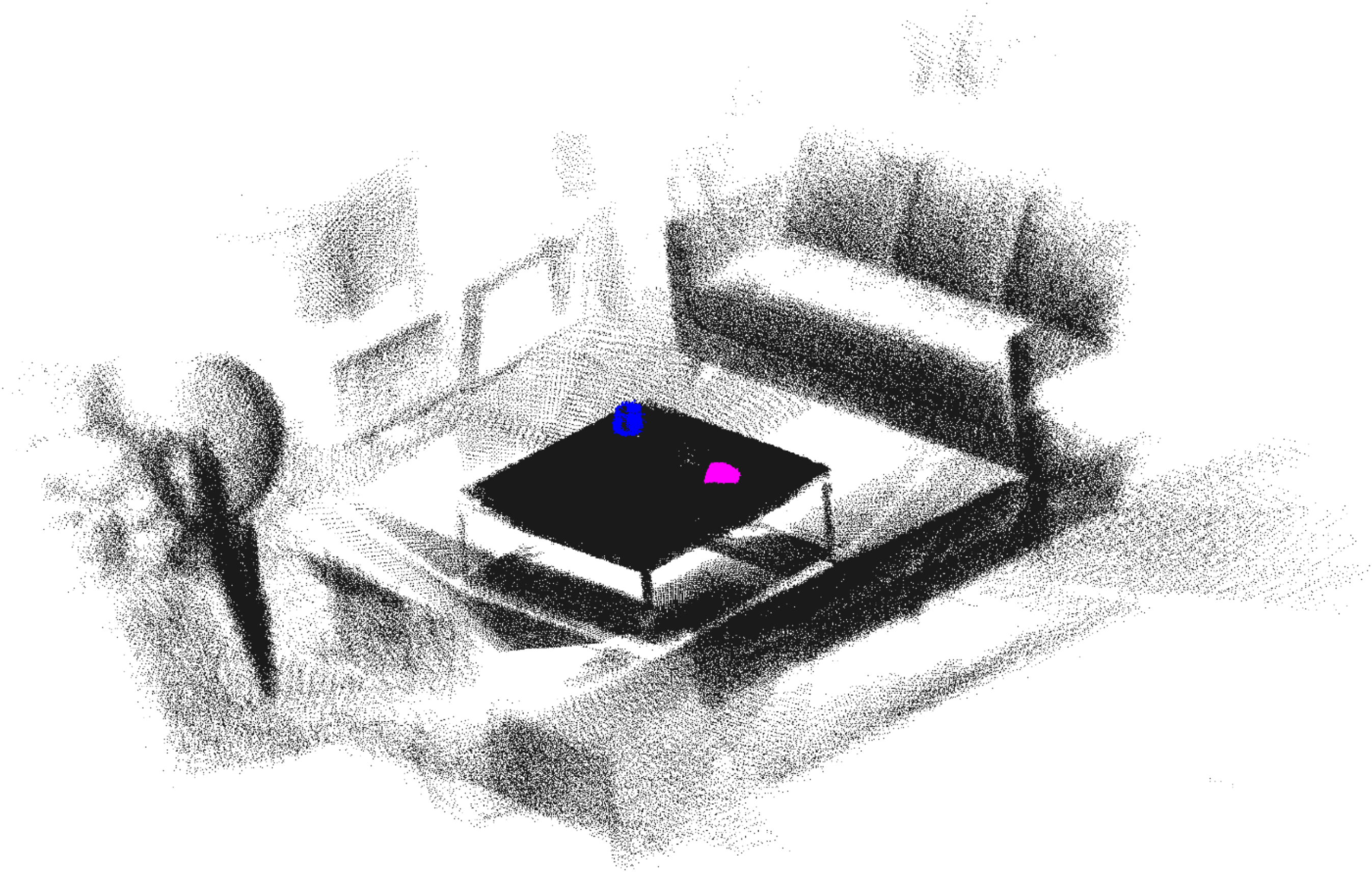}
\end{minipage}

&

\begin{minipage}[t]{0.19\linewidth}
\centering
\includegraphics[width=1\linewidth]{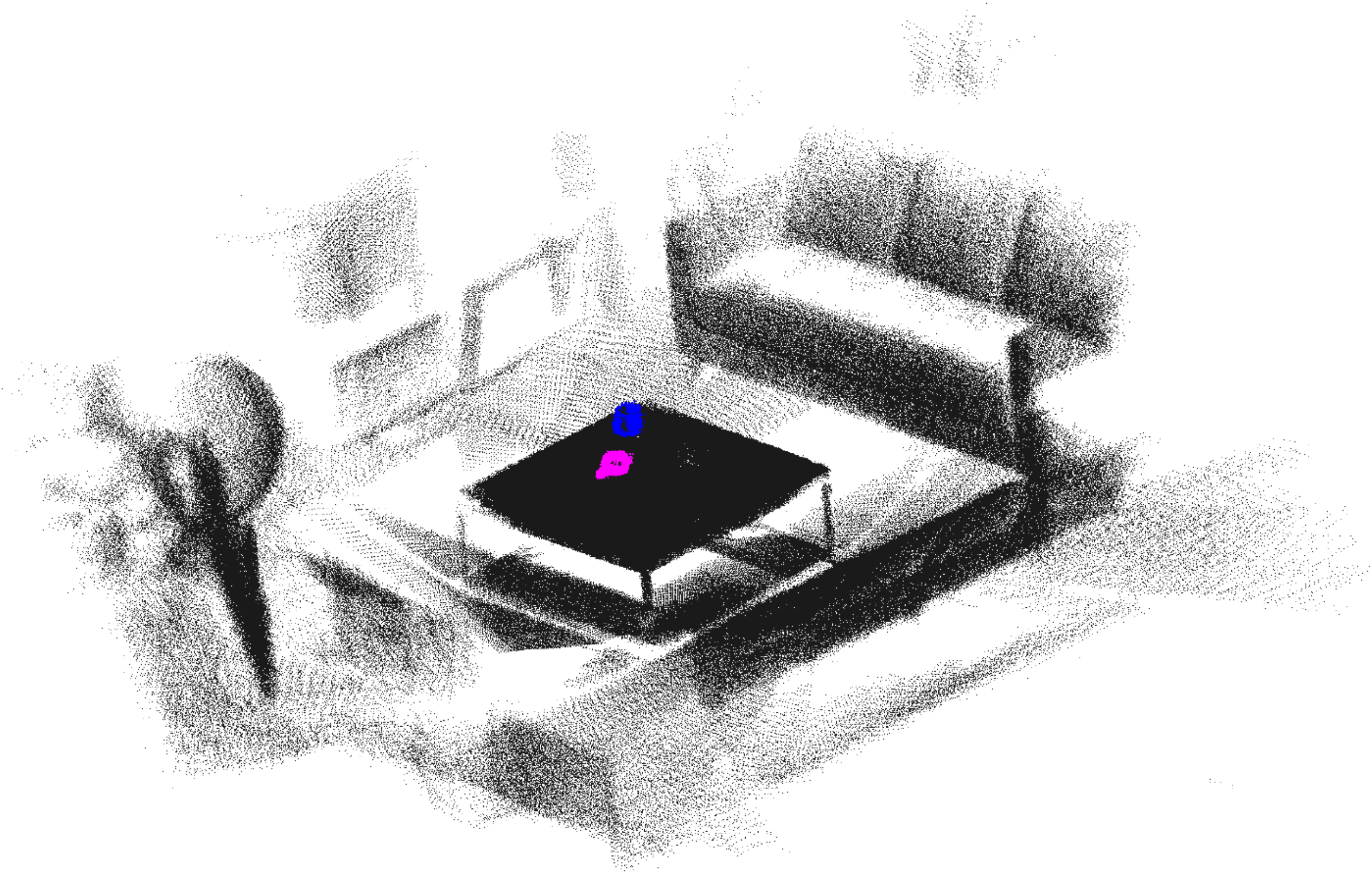}
\end{minipage}

&

\begin{minipage}[t]{0.19\linewidth}
\centering
\includegraphics[width=1\linewidth]{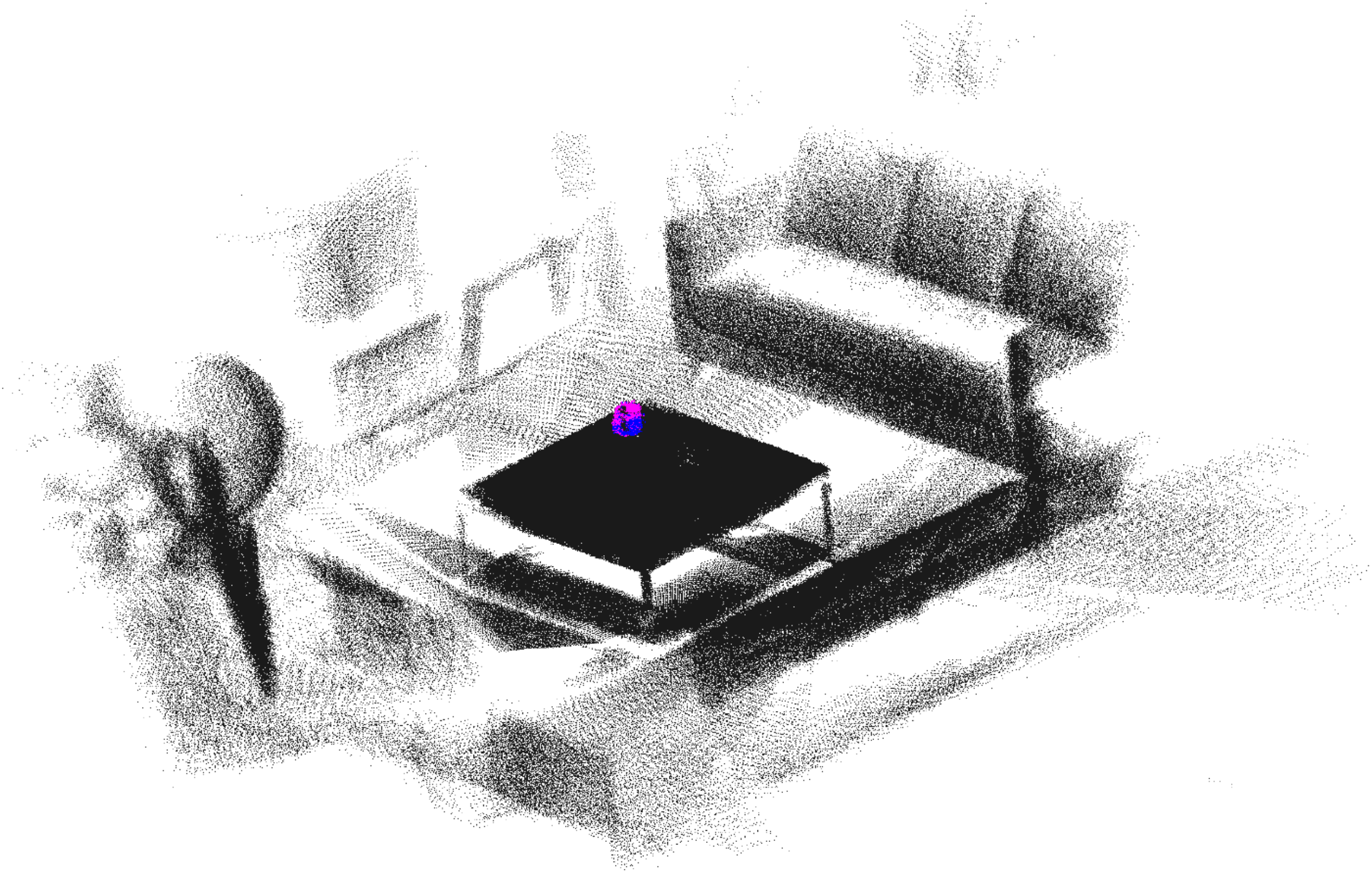}
\end{minipage}

&

\begin{minipage}[t]{0.19\linewidth}
\centering
\includegraphics[width=1\linewidth]{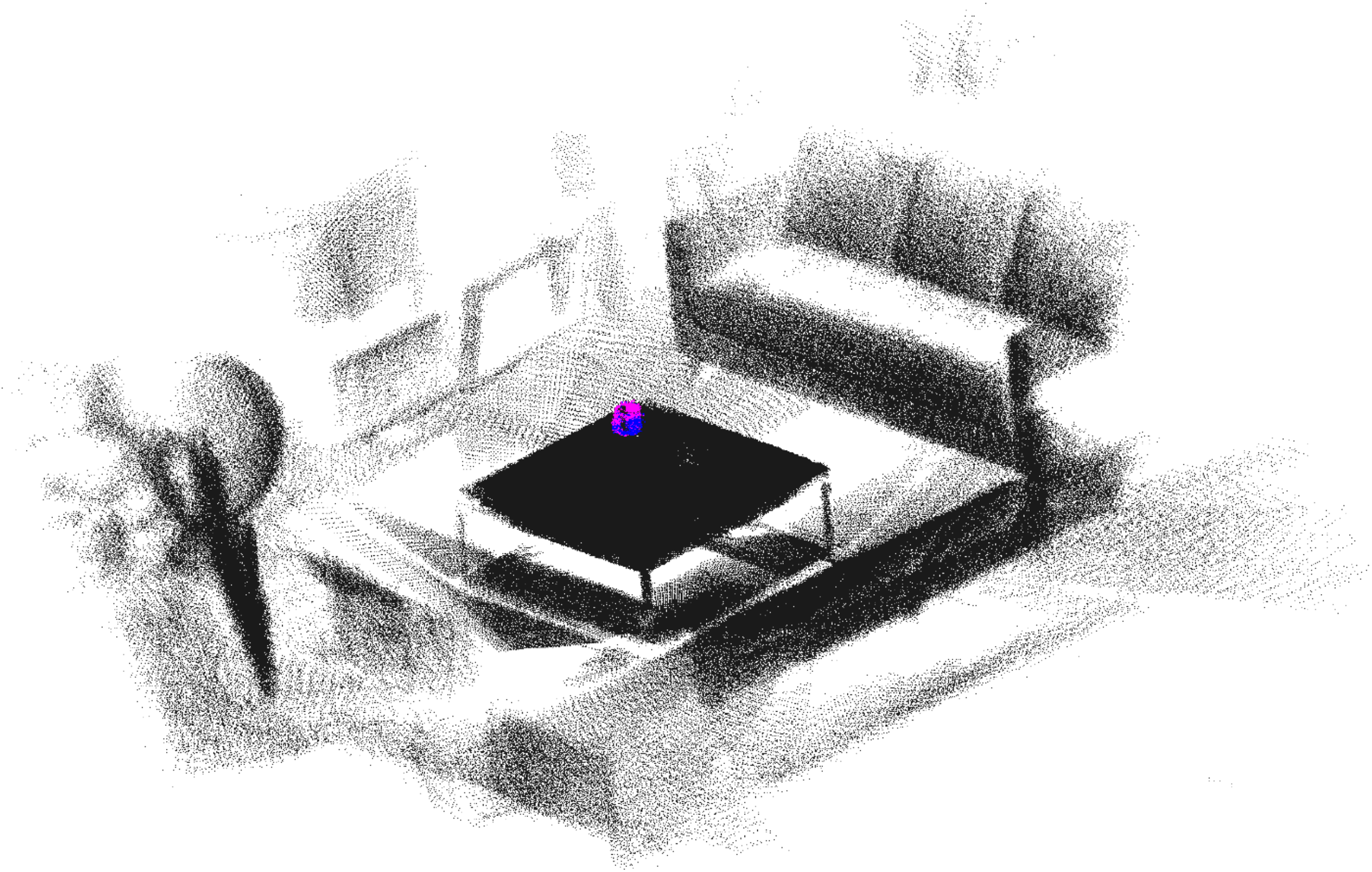}
\end{minipage}

\\

&&\footnotesize{$N=458$, 95.41\%}& & \,\footnotesize{$101.597^{\circ},1.632m,0.103s$}\, &\,\footnotesize{$171.168^{\circ},2.401m,18.079s$}\, &\,\footnotesize{$\textbf{0.187}^{\circ},\textbf{0.003}m,0.827s$}\,&\,\footnotesize{$\textbf{0.187}^{\circ},\textbf{0.003}m,\textbf{0.291}s$}\,

\\

\rotatebox{90}{\,\,\footnotesize{\textit{Scene-02, box}}\,}\,

& &

\begin{minipage}[t]{0.19\linewidth}
\centering
\includegraphics[width=1\linewidth]{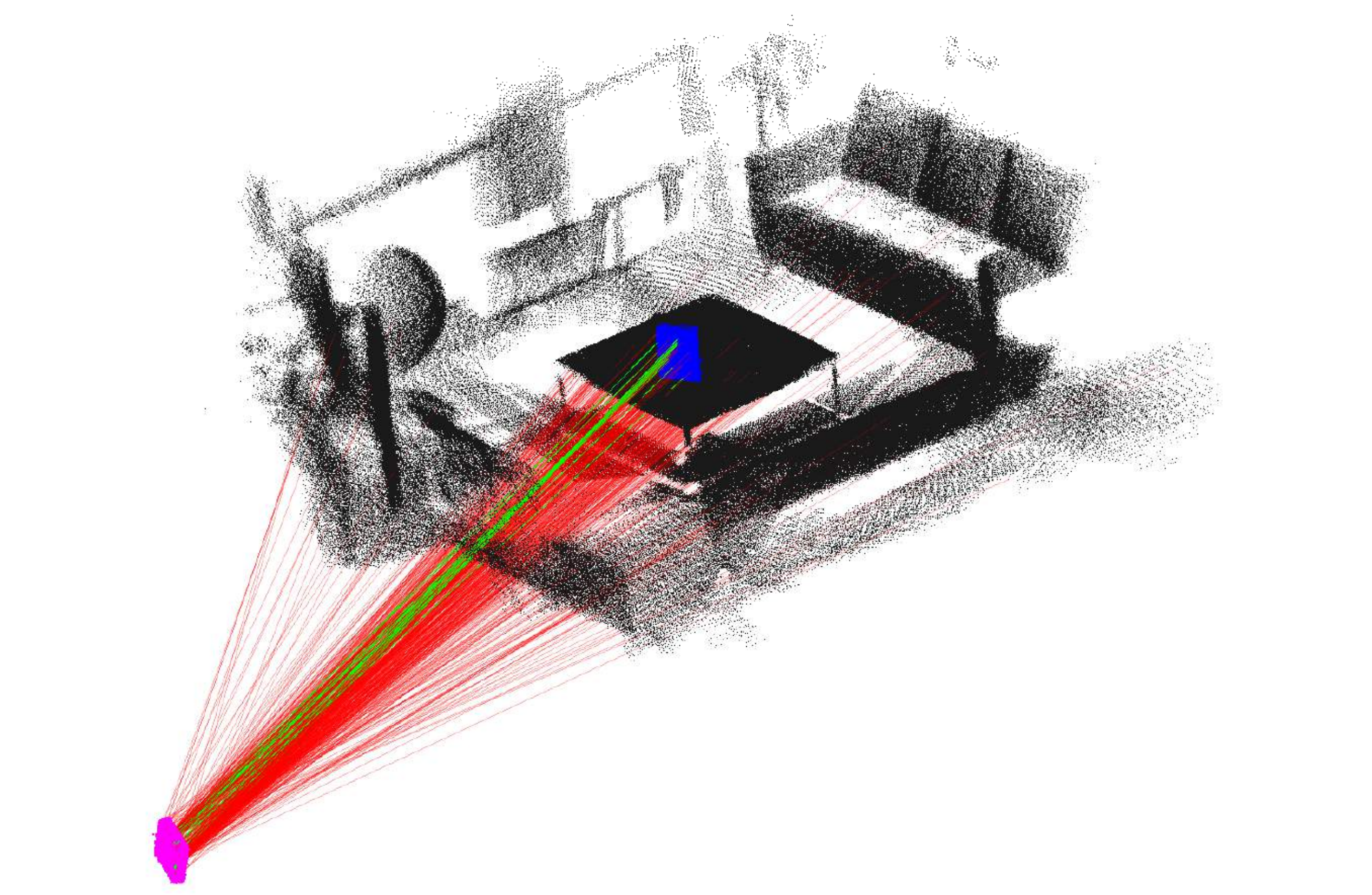}
\end{minipage}

& &

\begin{minipage}[t]{0.19\linewidth}
\centering
\includegraphics[width=1\linewidth]{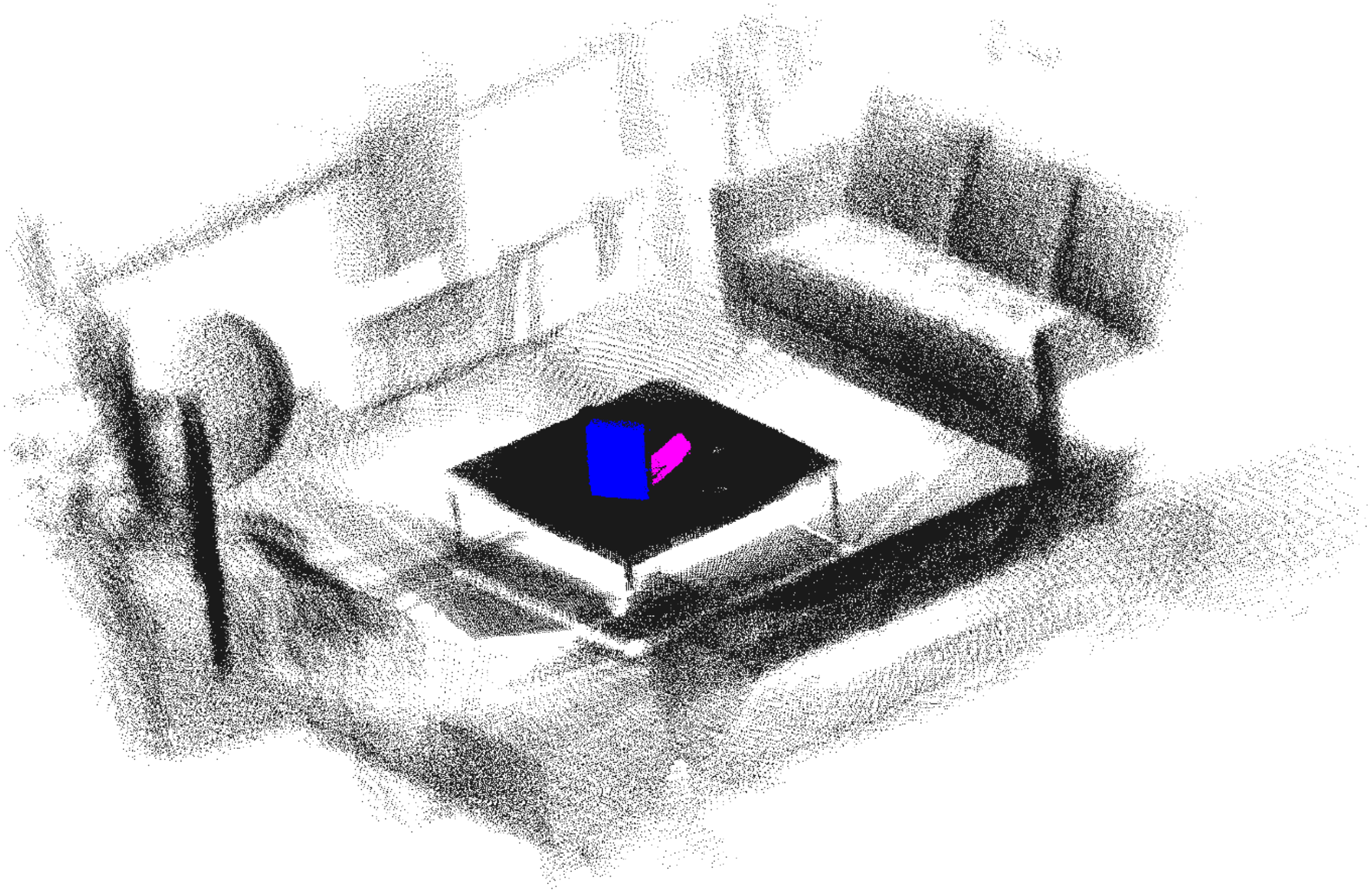}
\end{minipage}

&

\begin{minipage}[t]{0.19\linewidth}
\centering
\includegraphics[width=1\linewidth]{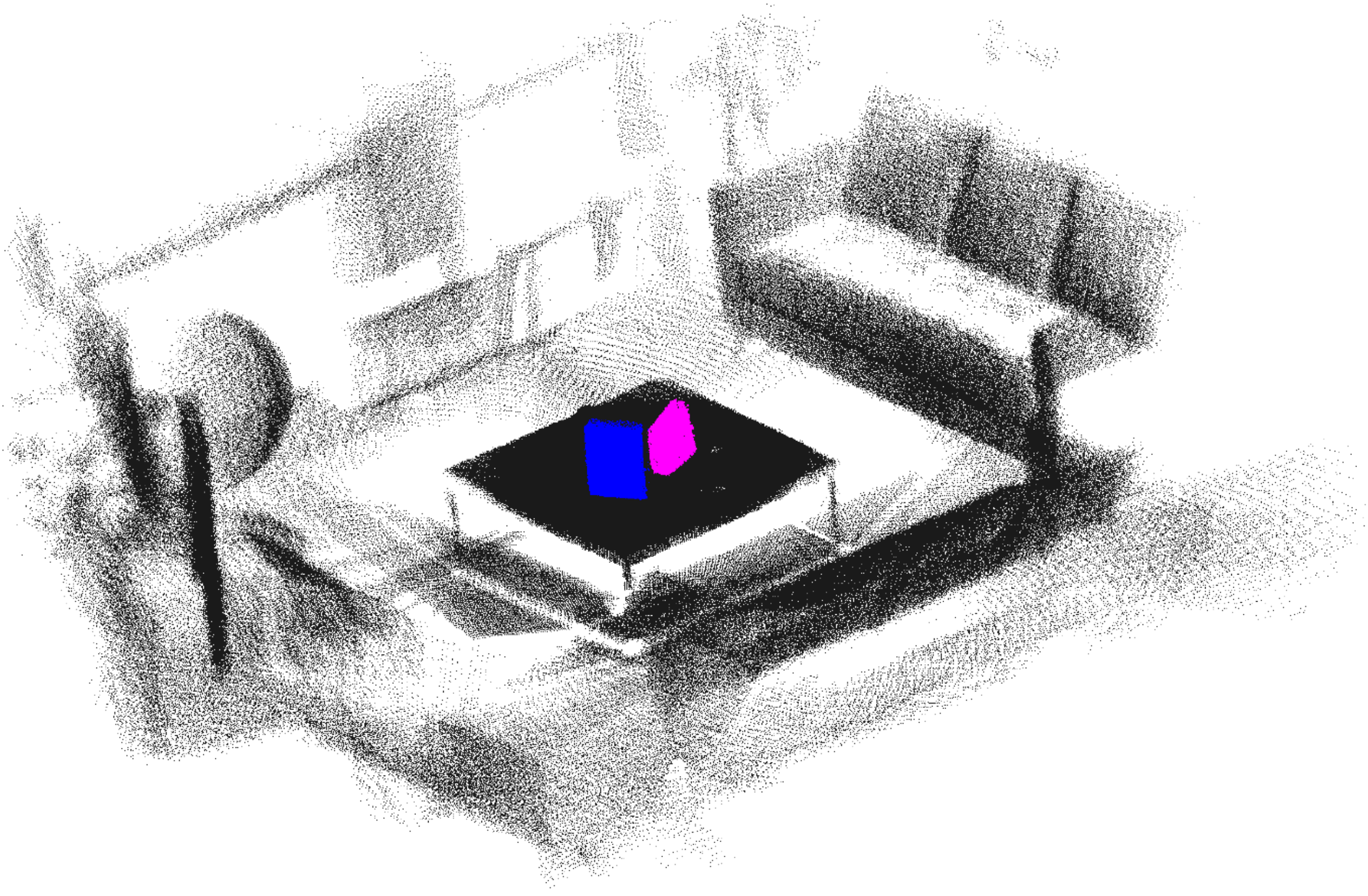}
\end{minipage}

&

\begin{minipage}[t]{0.19\linewidth}
\centering
\includegraphics[width=1\linewidth]{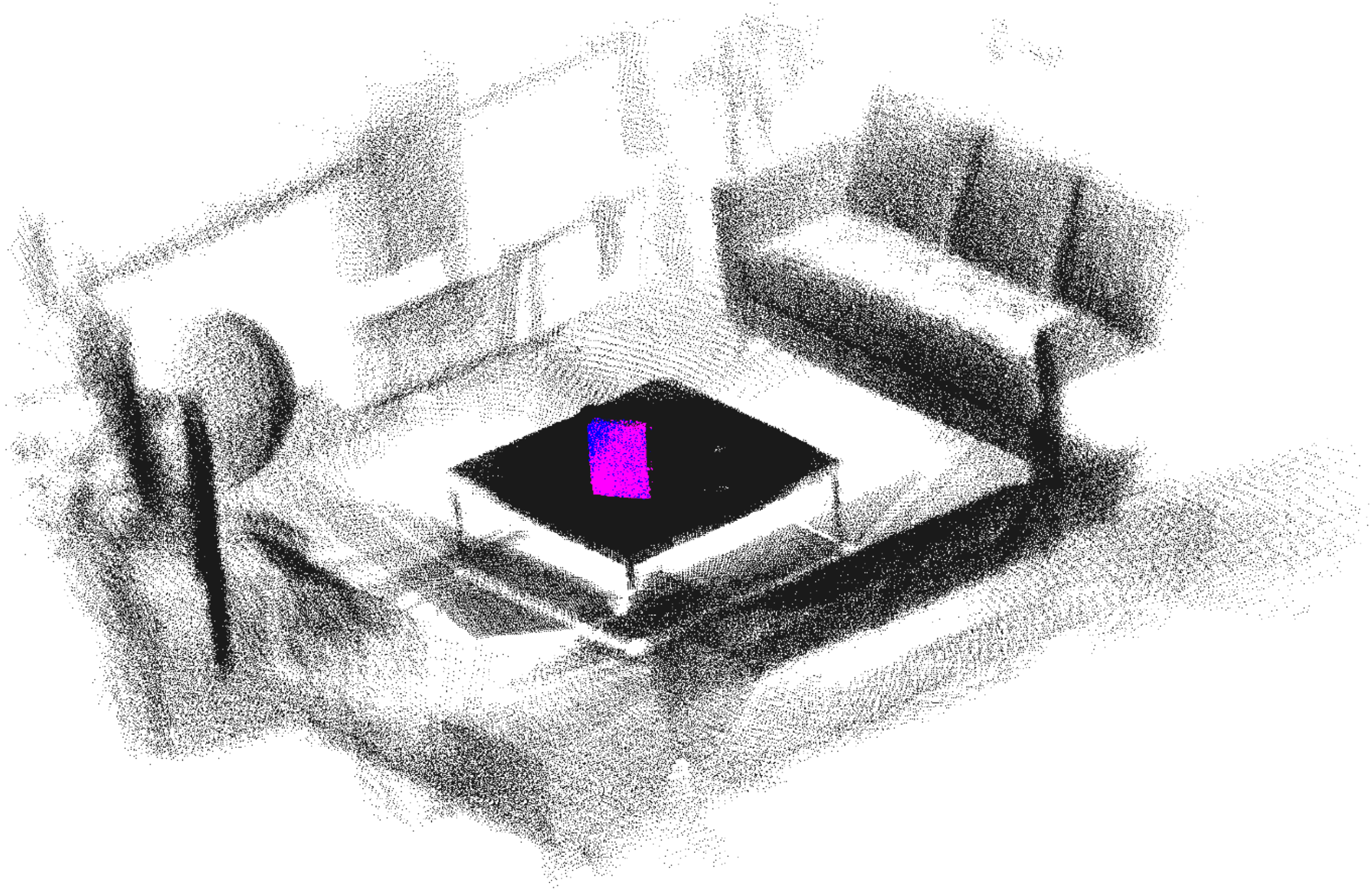}
\end{minipage}

&

\begin{minipage}[t]{0.19\linewidth}
\centering
\includegraphics[width=1\linewidth]{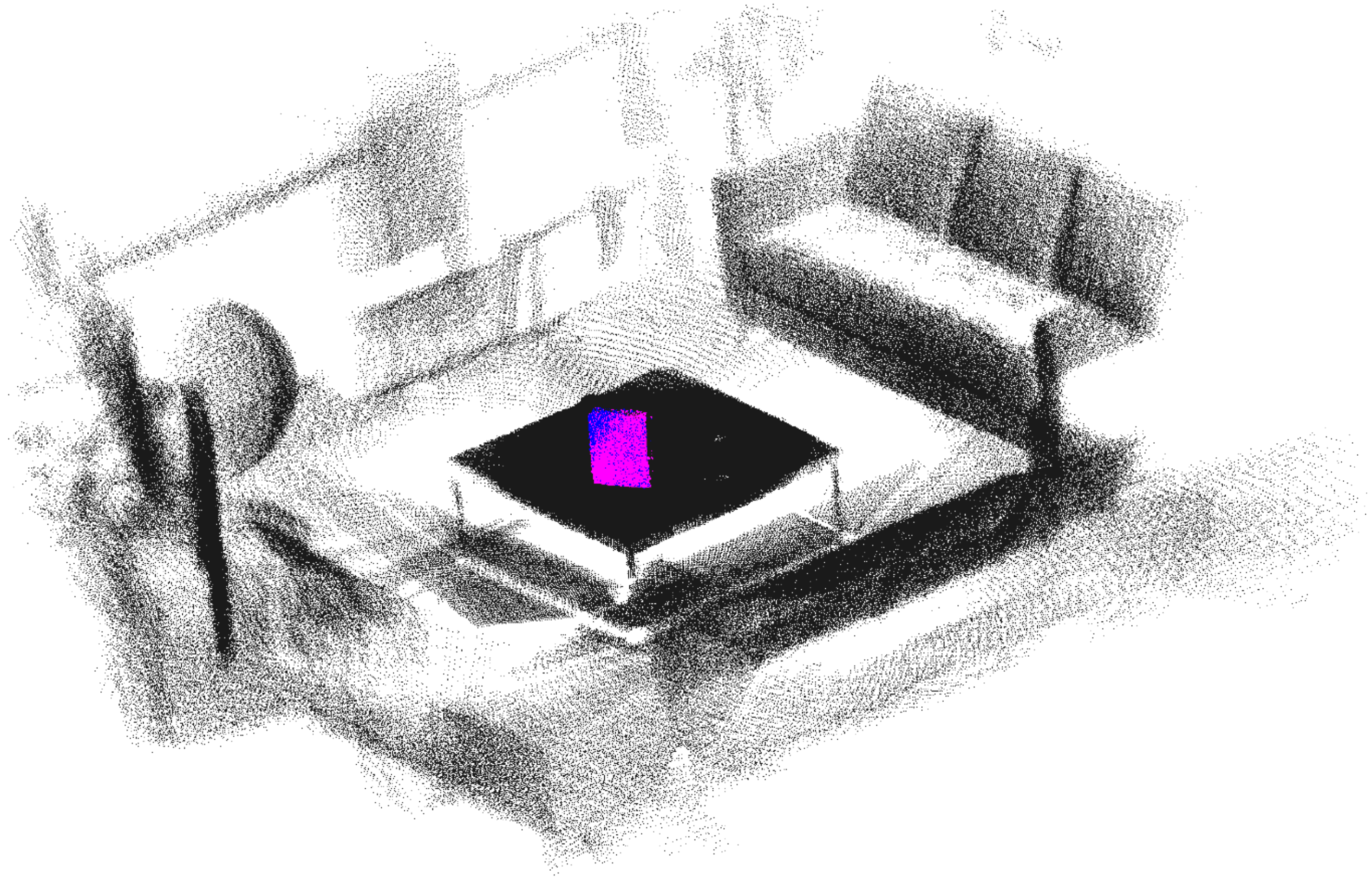}
\end{minipage}

\\

&&\footnotesize{$N=689$, 97.68\%}& & \,\footnotesize{$92.381^{\circ},0.591m,0.055s$}\, &\,\footnotesize{$110.968^{\circ},2.199m,15.839s$}\, &\,\footnotesize{${0.198}^{\circ},\textbf{0.003}m,1.329s$}\,&\,\footnotesize{$\textbf{0.139}^{\circ},\textbf{0.003}m,\textbf{0.103}s$}\,

\\

\rotatebox{90}{\,\,\footnotesize{\textit{Scene-03, cap}}\,}\,

& &

\begin{minipage}[t]{0.19\linewidth}
\centering
\includegraphics[width=1\linewidth]{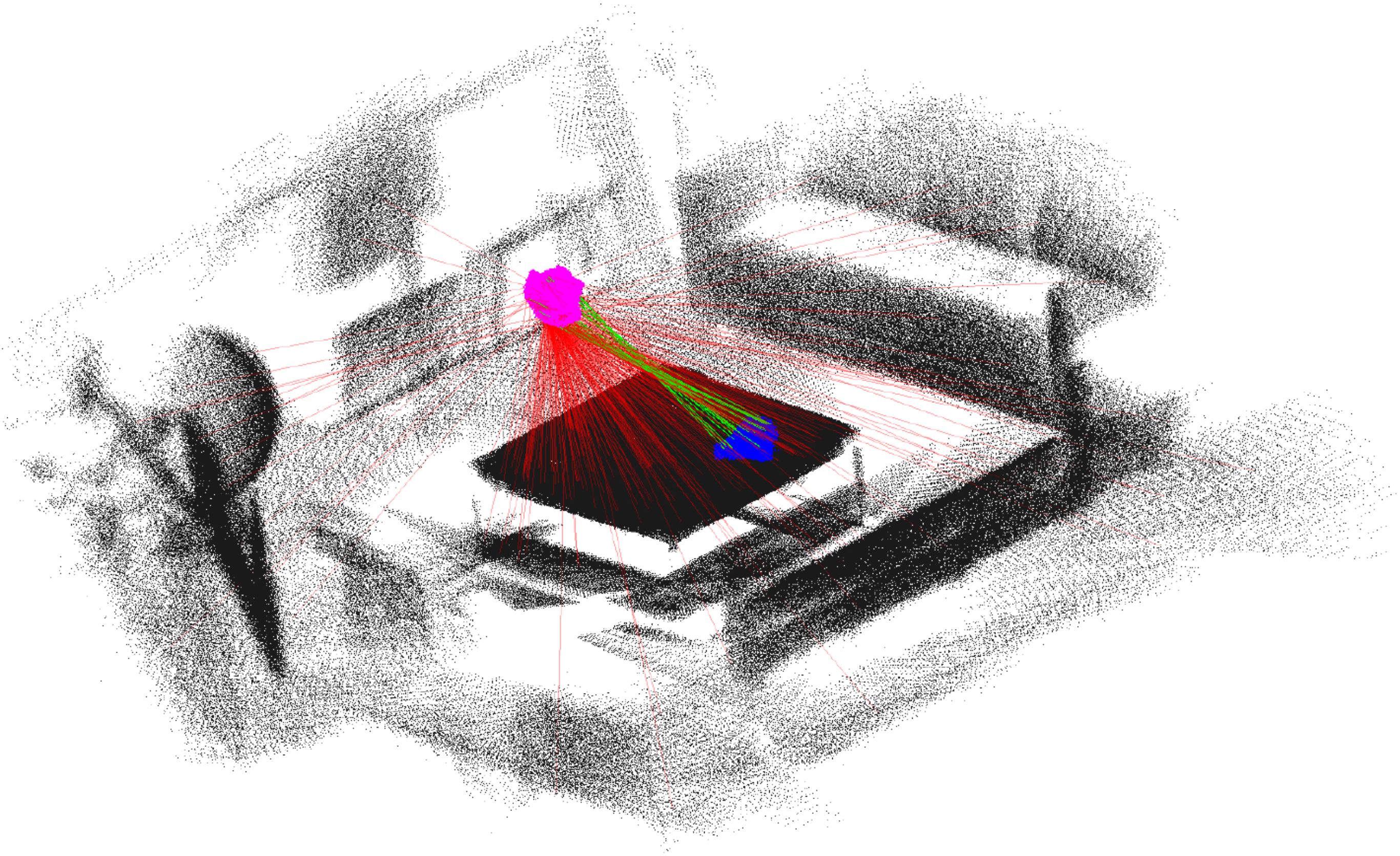}
\end{minipage}

& &

\begin{minipage}[t]{0.19\linewidth}
\centering
\includegraphics[width=1\linewidth]{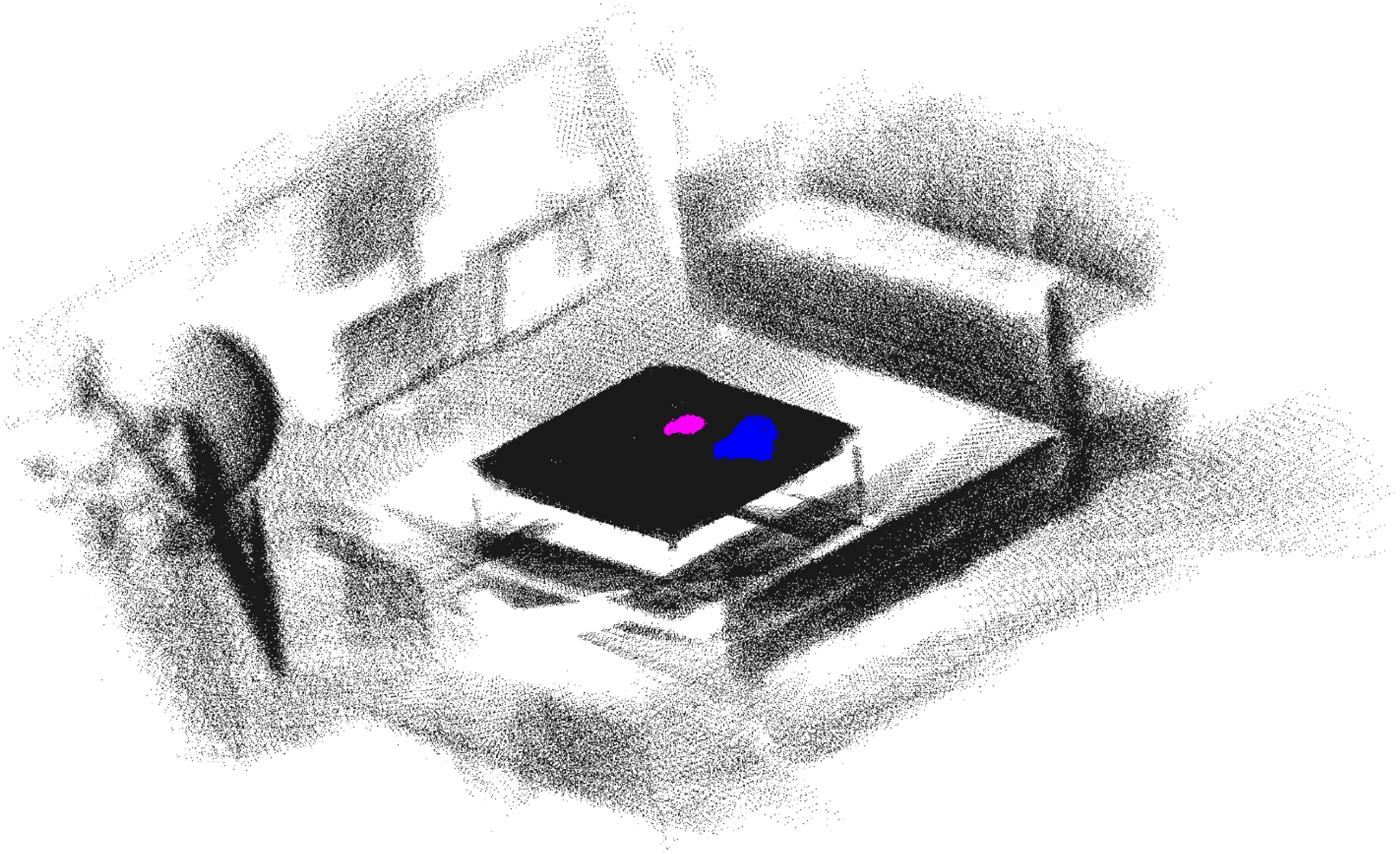}
\end{minipage}

&

\begin{minipage}[t]{0.19\linewidth}
\centering
\includegraphics[width=1\linewidth]{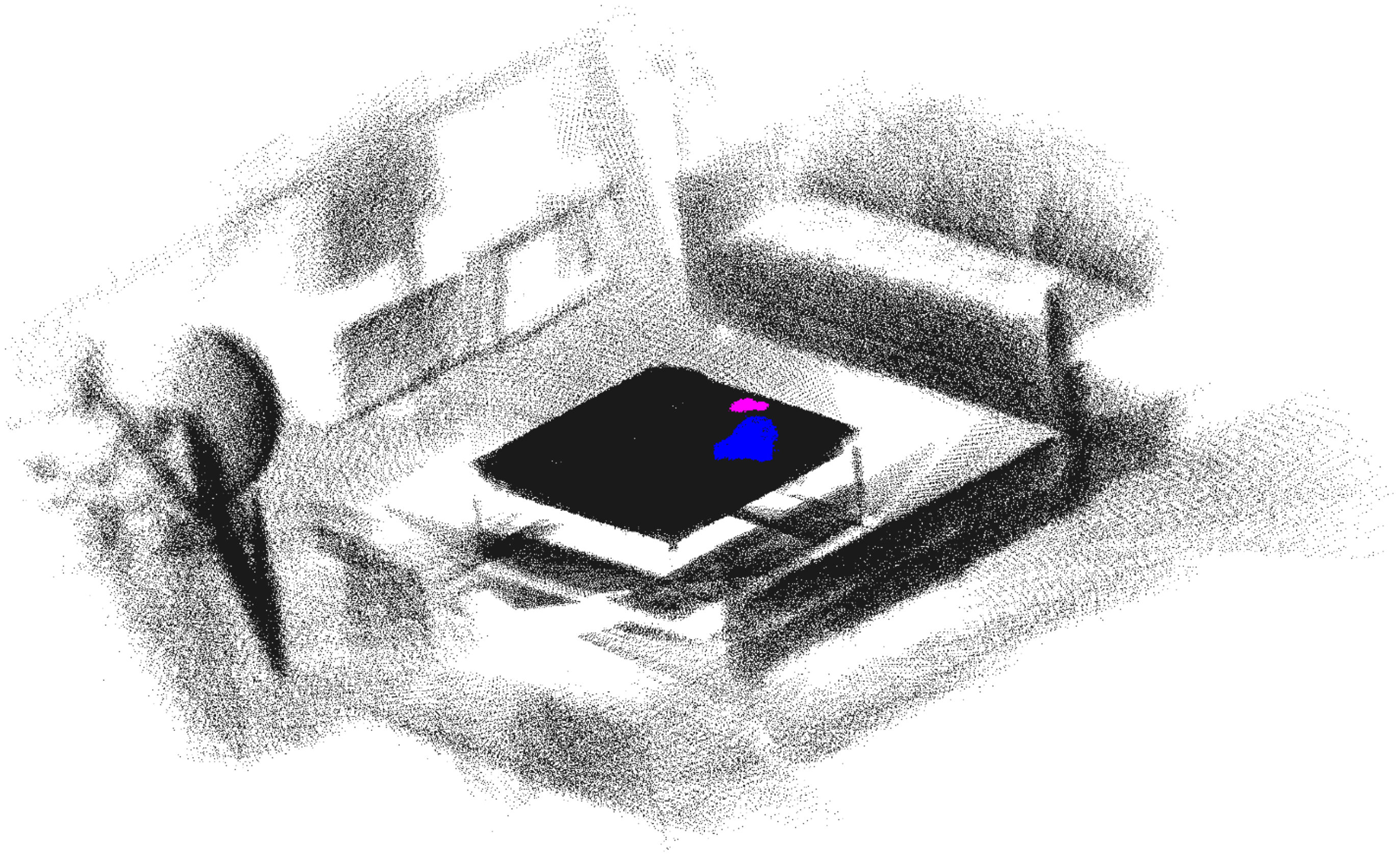}
\end{minipage}

&

\begin{minipage}[t]{0.19\linewidth}
\centering
\includegraphics[width=1\linewidth]{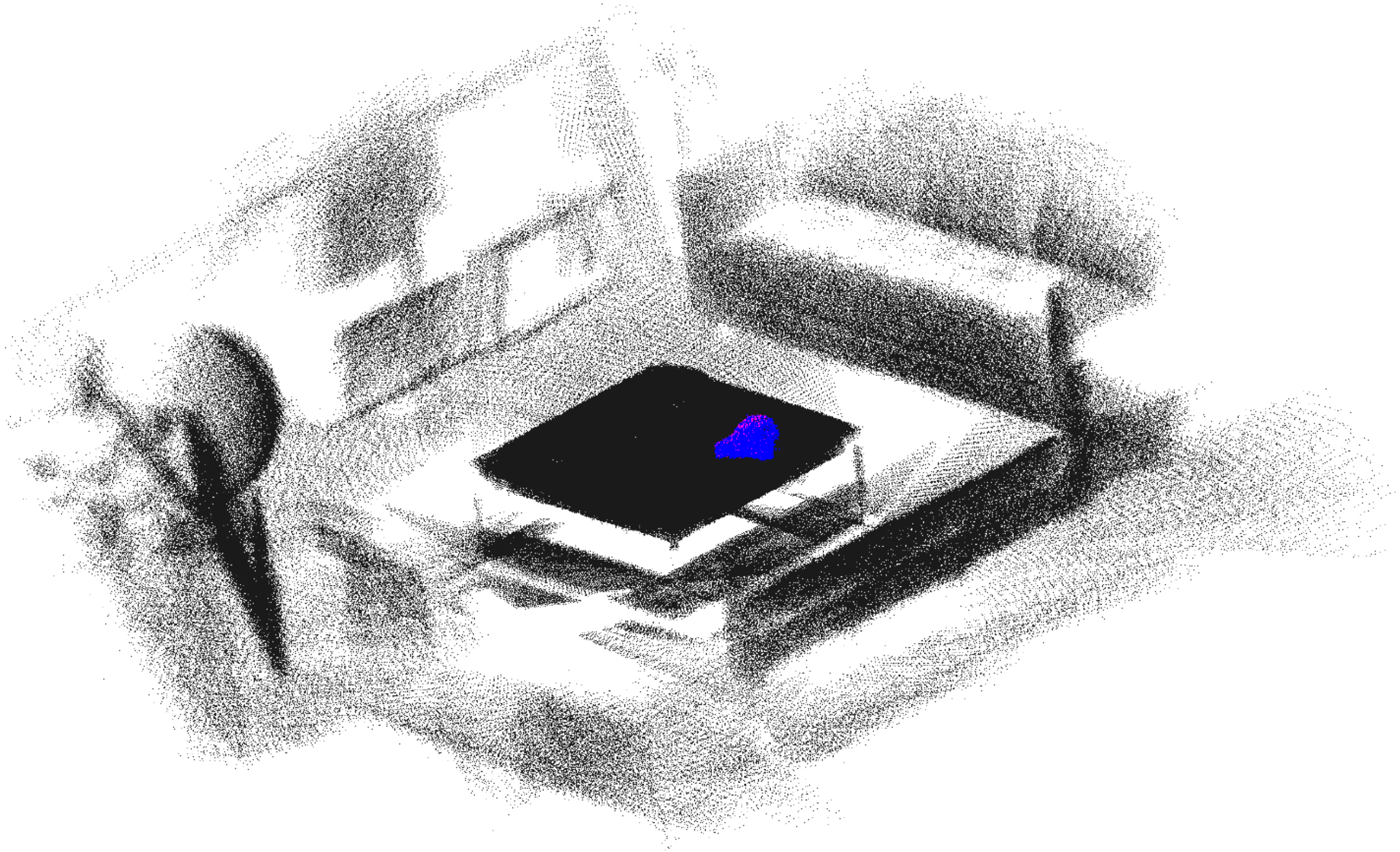}
\end{minipage}

&

\begin{minipage}[t]{0.19\linewidth}
\centering
\includegraphics[width=1\linewidth]{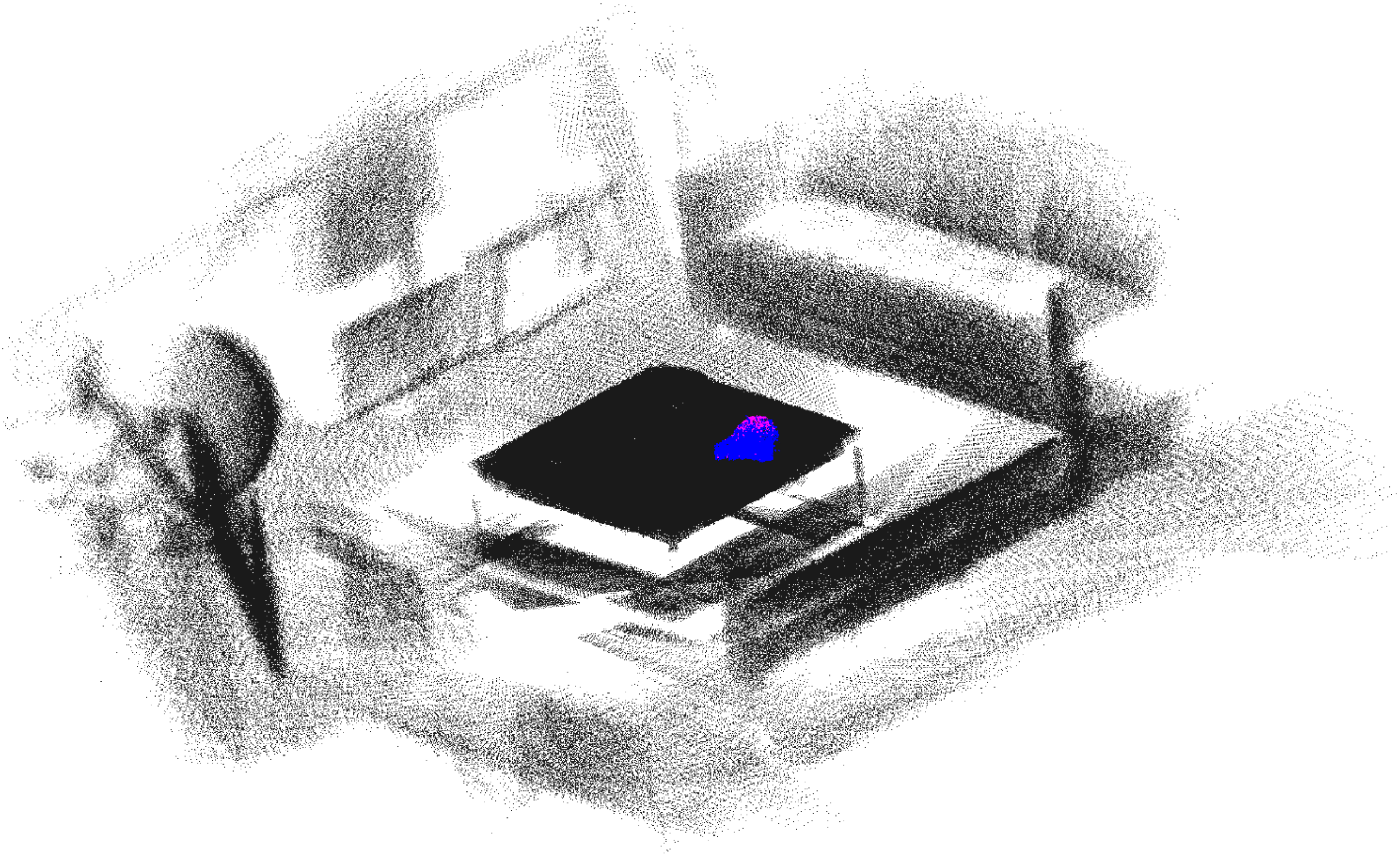}
\end{minipage}

\\

&&\footnotesize{$N=304$, 97.04\%}& & \,\footnotesize{$150.055^{\circ},1.566m,0.041s$}\, &\,\footnotesize{$33.426^{\circ},0.749m,12.562s$}\, &\,\footnotesize{$\textbf{1.969}^{\circ},\textbf{0.024}m,0.763s$}\,&\,\footnotesize{$\textbf{1.969}^{\circ},\textbf{0.024}m,\textbf{0.118}s$}\,

\\

\rotatebox{90}{\,\,\footnotesize{\textit{Scene-04, mug}}\,}\,

& &

\begin{minipage}[t]{0.19\linewidth}
\centering
\includegraphics[width=1\linewidth]{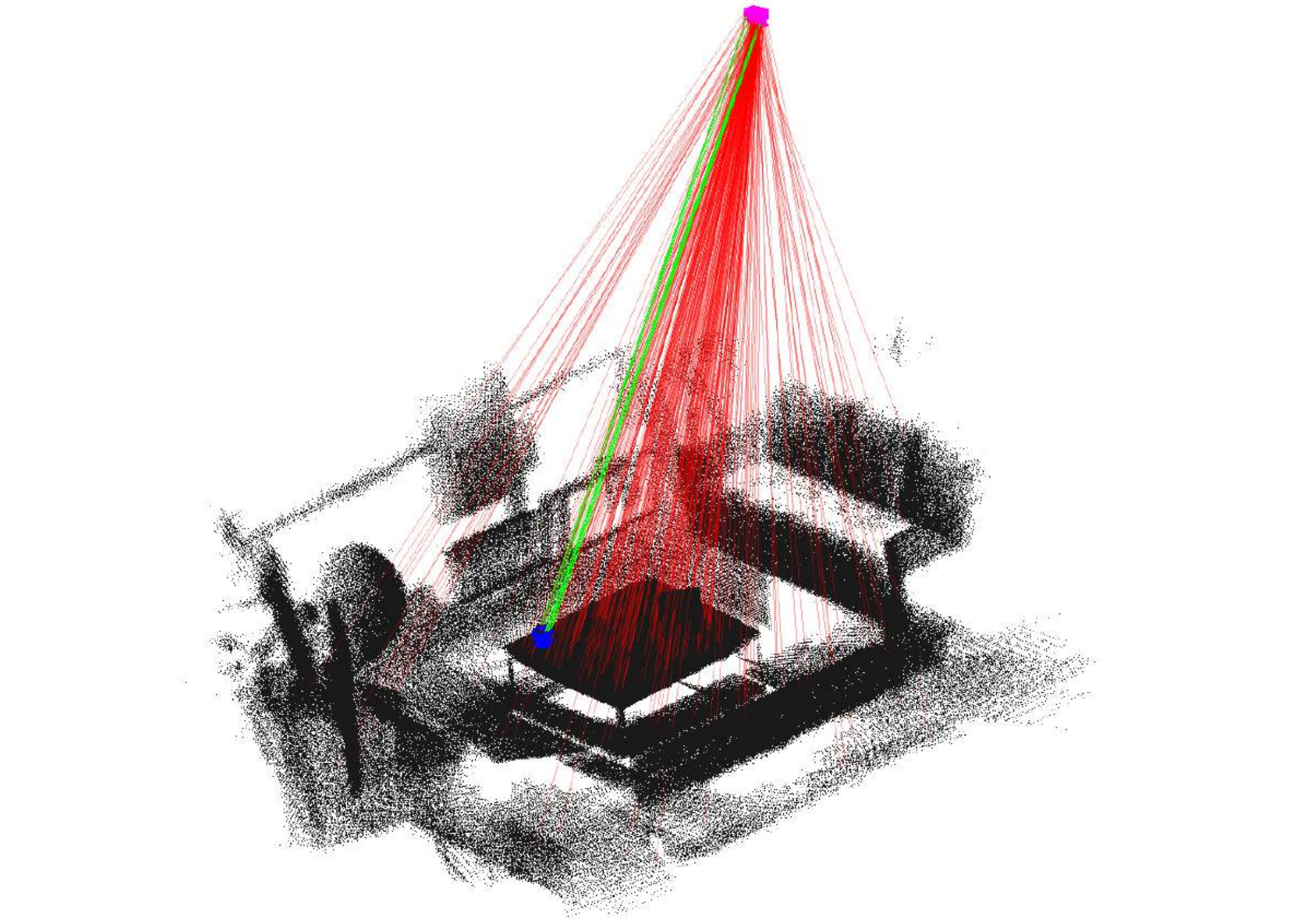}
\end{minipage}

& &

\begin{minipage}[t]{0.19\linewidth}
\centering
\includegraphics[width=1\linewidth]{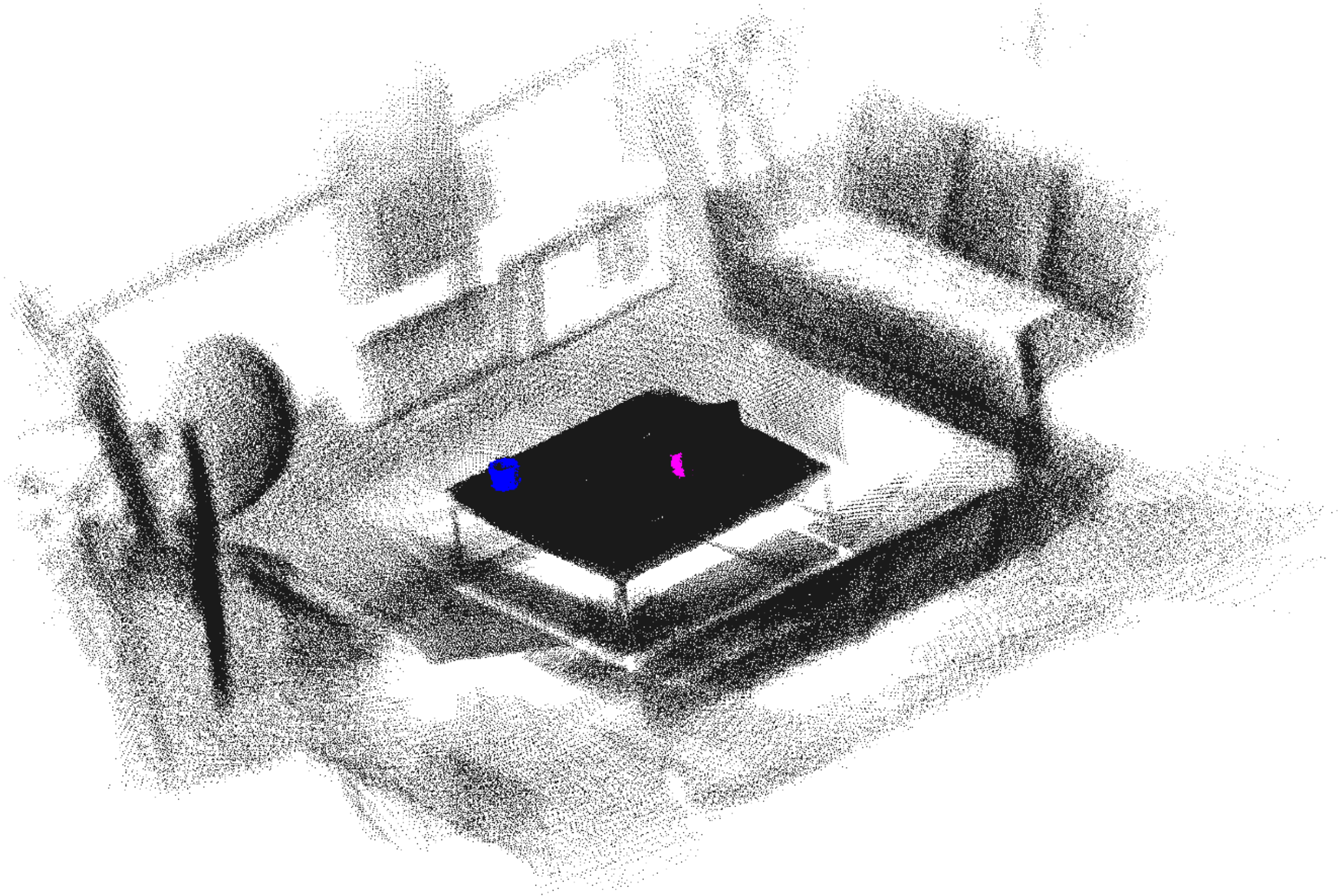}
\end{minipage}

&

\begin{minipage}[t]{0.19\linewidth}
\centering
\includegraphics[width=1\linewidth]{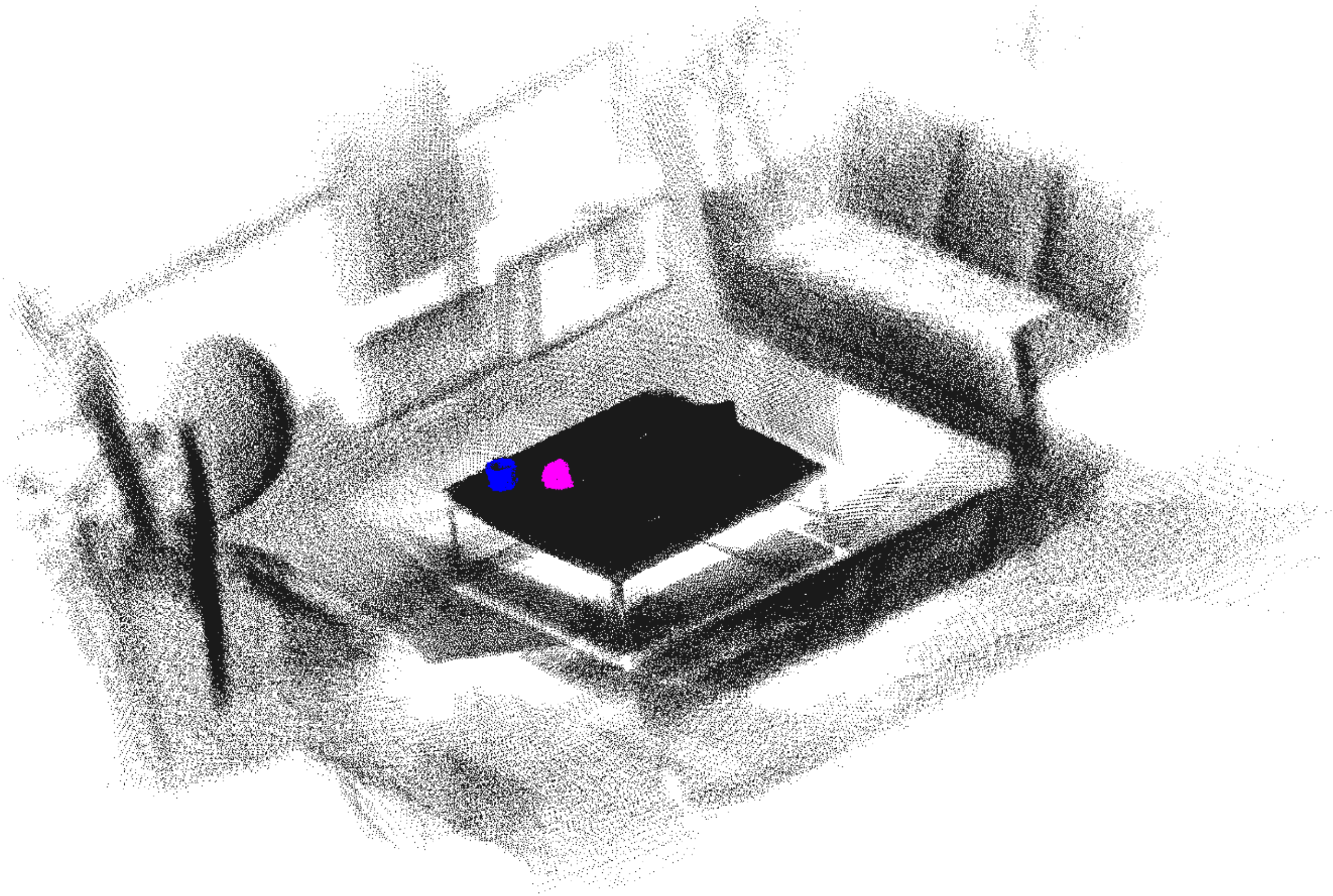}
\end{minipage}

&

\begin{minipage}[t]{0.19\linewidth}
\centering
\includegraphics[width=1\linewidth]{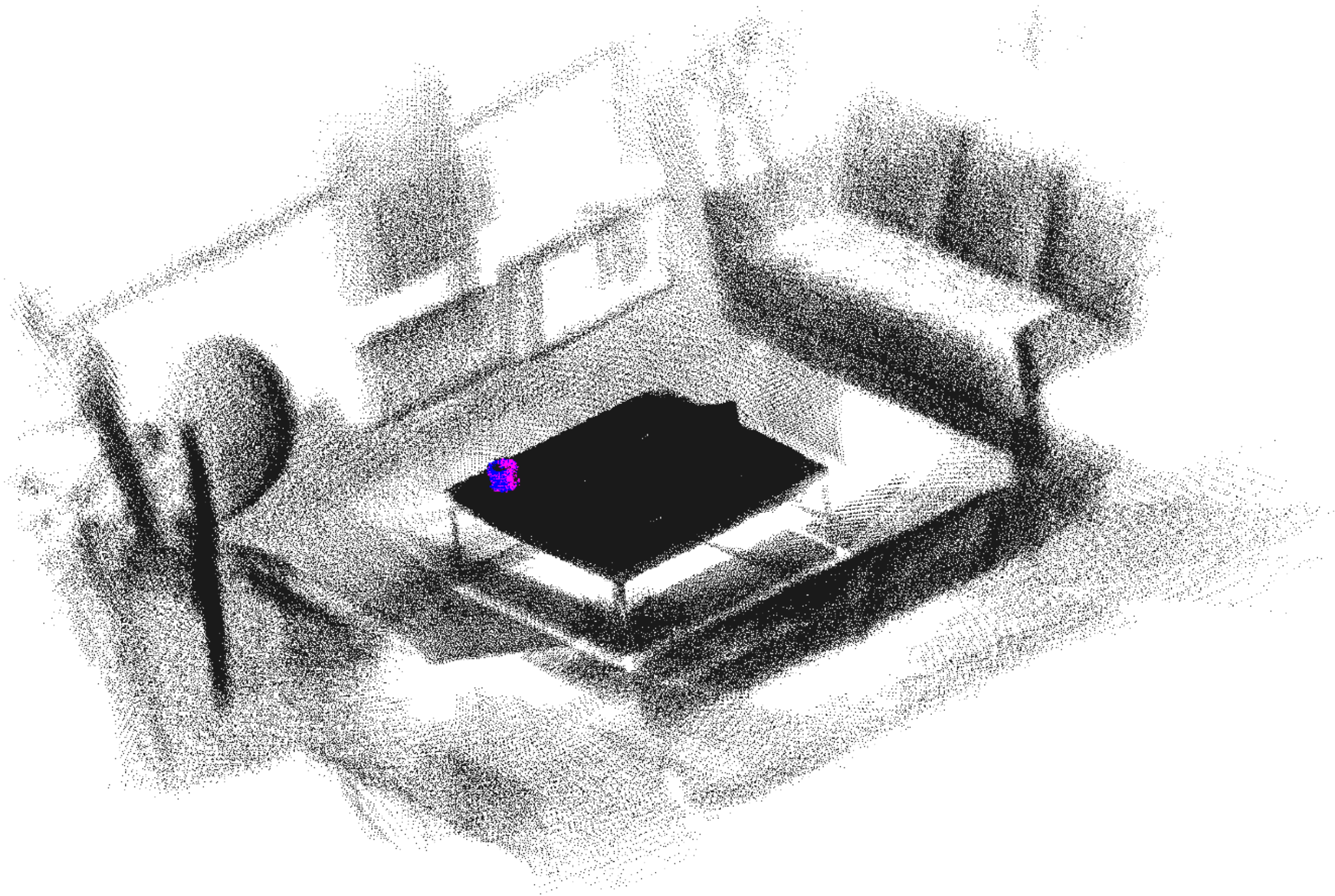}
\end{minipage}

&

\begin{minipage}[t]{0.19\linewidth}
\centering
\includegraphics[width=1\linewidth]{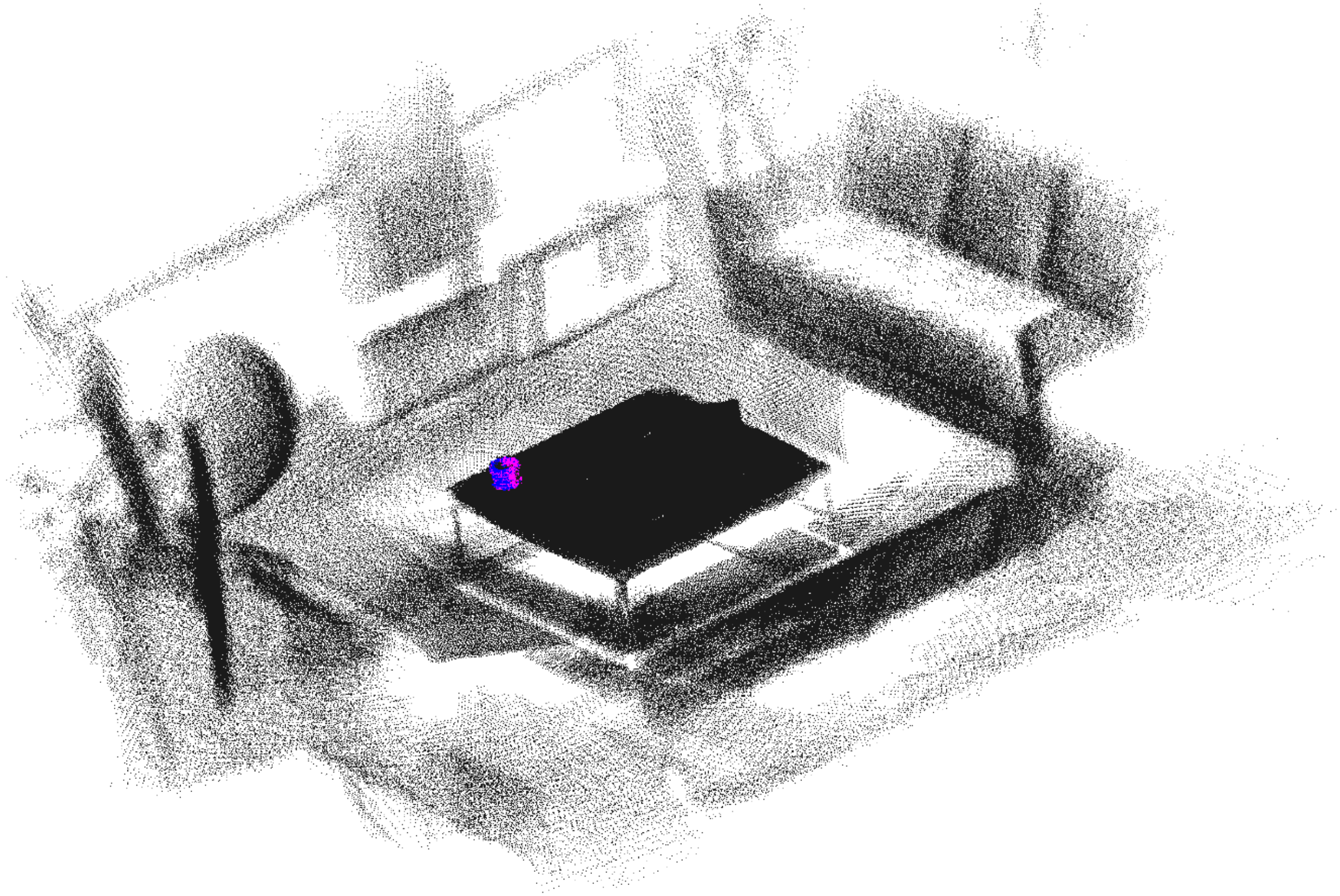}
\end{minipage}

\\

&&\footnotesize{$N=773$, 97.41\%}& & \,\footnotesize{$138.748^{\circ},2.567m,0.132s$}\, &\,\footnotesize{$152.764^{\circ},2.566m,29.398s$}\, &\,\footnotesize{$\textbf{0.427}^{\circ},\textbf{0.011}m,1.321s$}\,&\,\footnotesize{$\textbf{0.427}^{\circ},\textbf{0.011}m,\textbf{0.310}s$}\,

\\

\rotatebox{90}{\,\,\footnotesize{\textit{Scene-05, box}}\,}\,

& &

\begin{minipage}[t]{0.19\linewidth}
\centering
\includegraphics[width=1\linewidth]{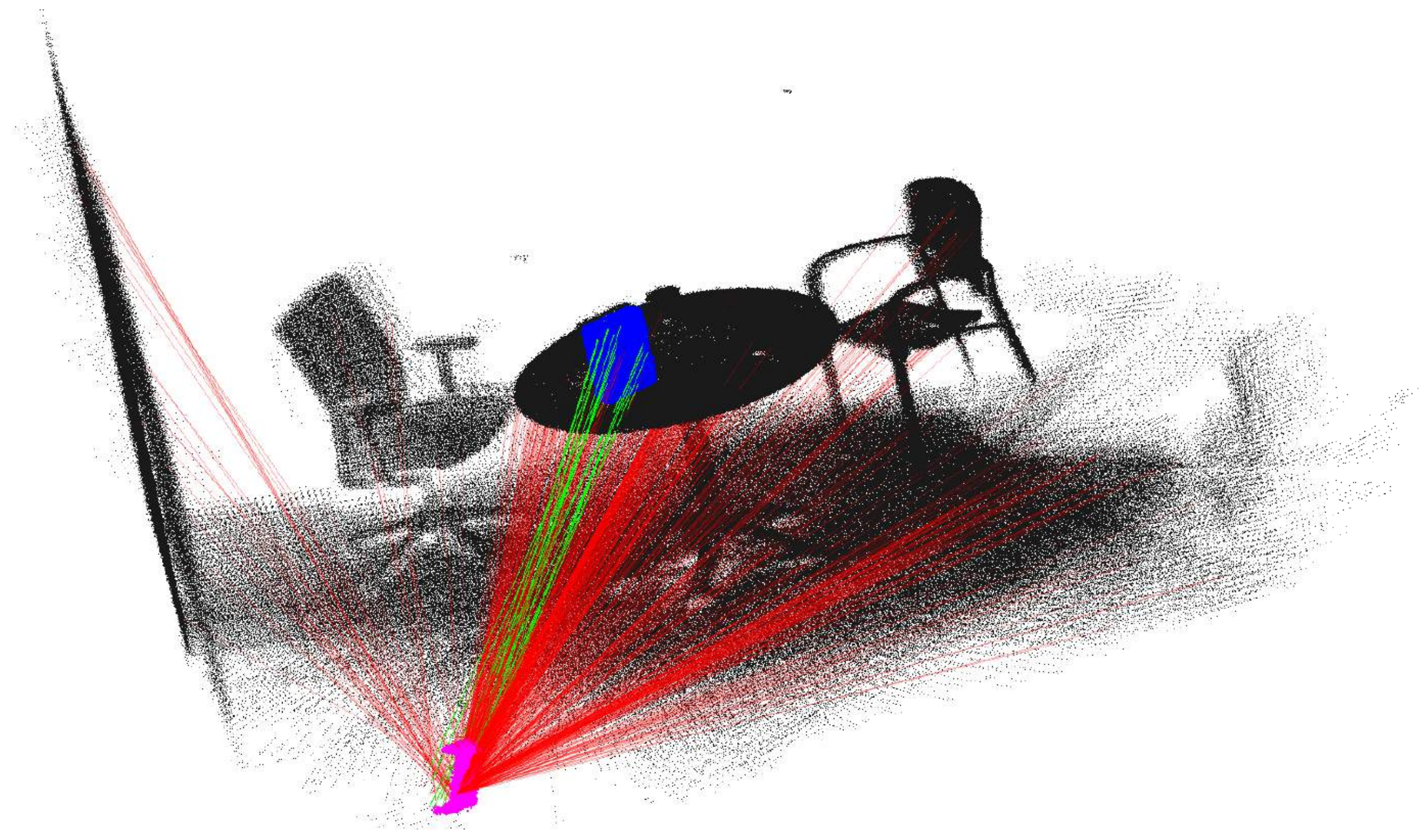}
\end{minipage}

& &

\begin{minipage}[t]{0.19\linewidth}
\centering
\includegraphics[width=1\linewidth]{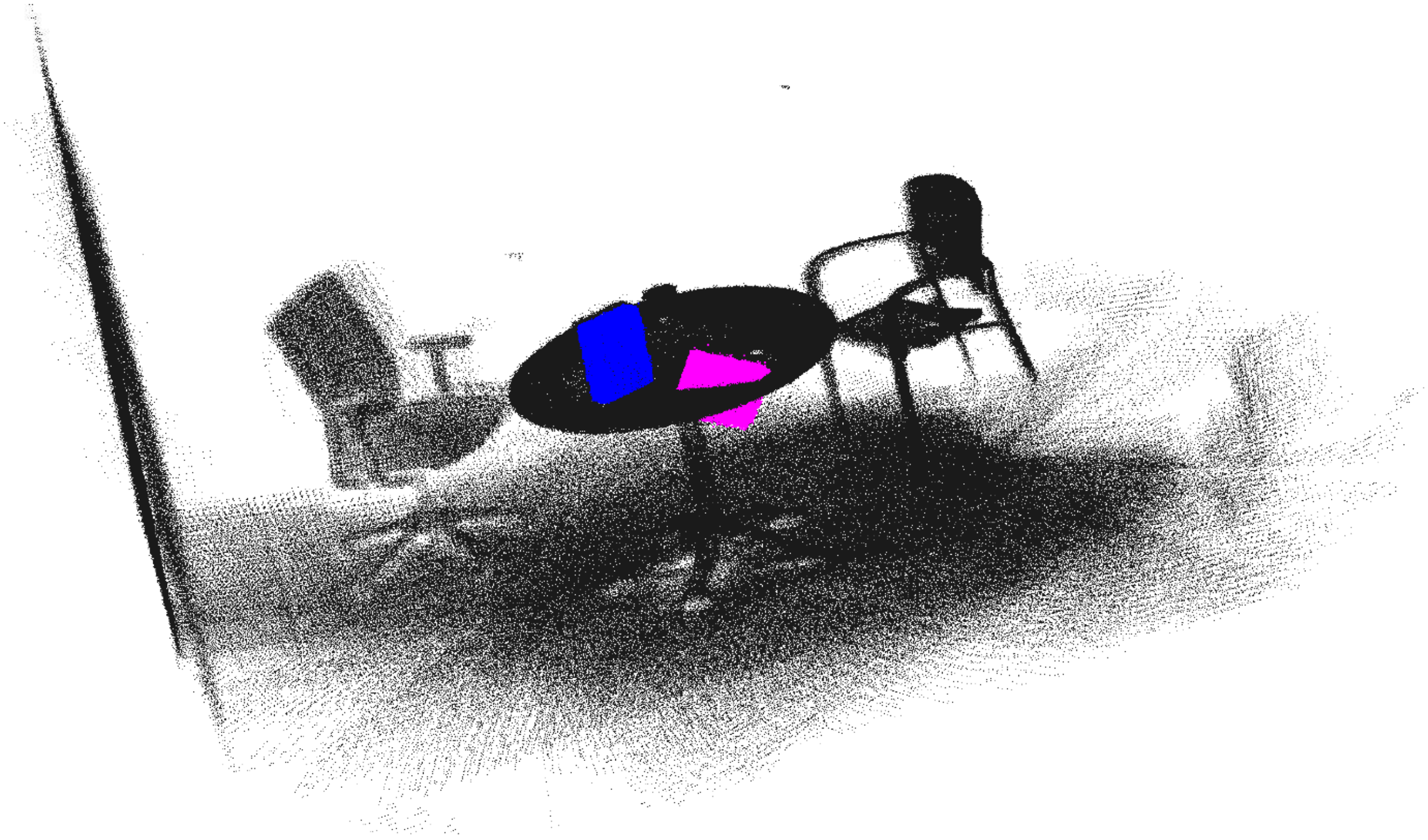}
\end{minipage}

&

\begin{minipage}[t]{0.19\linewidth}
\centering
\includegraphics[width=1\linewidth]{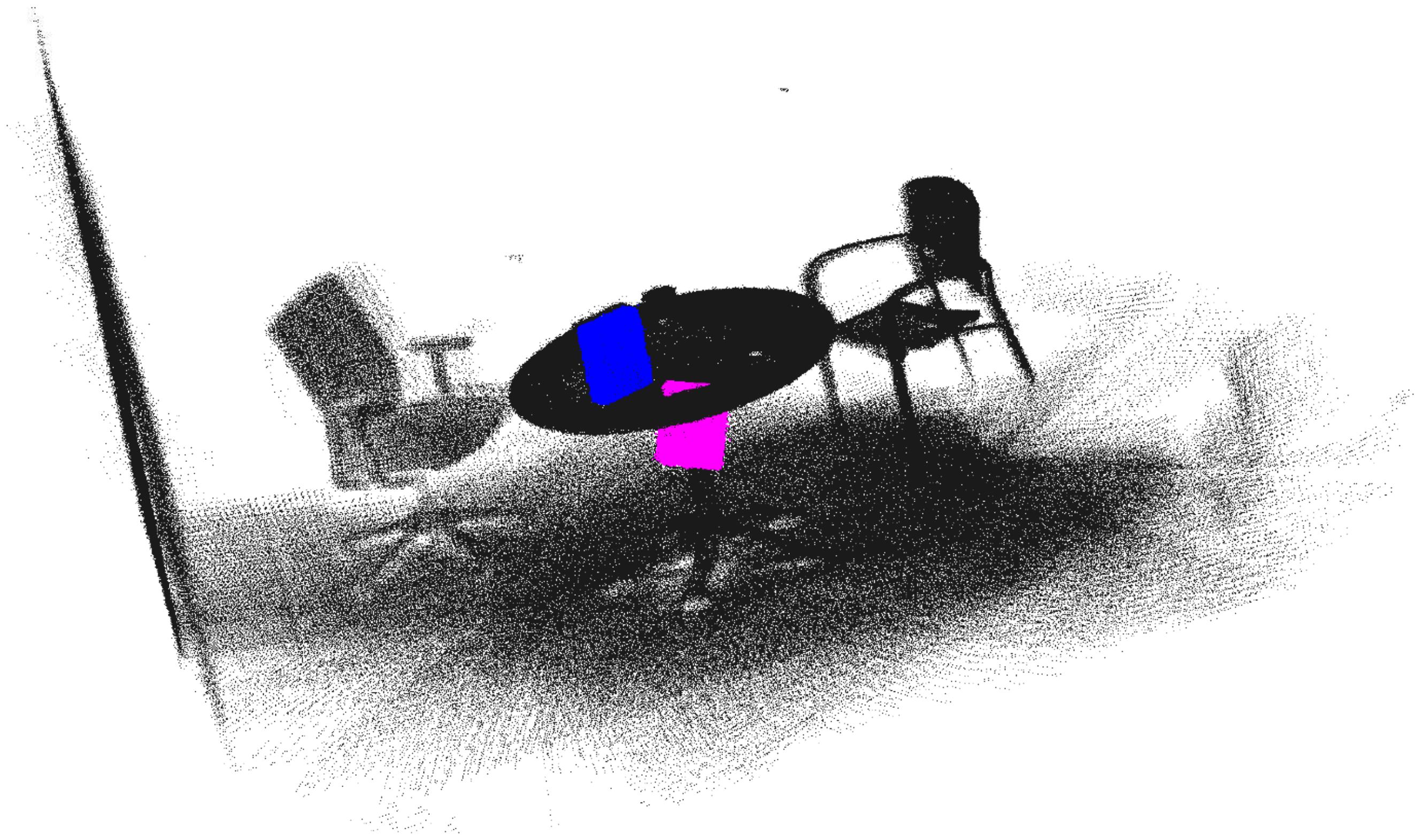}
\end{minipage}

&

\begin{minipage}[t]{0.19\linewidth}
\centering
\includegraphics[width=1\linewidth]{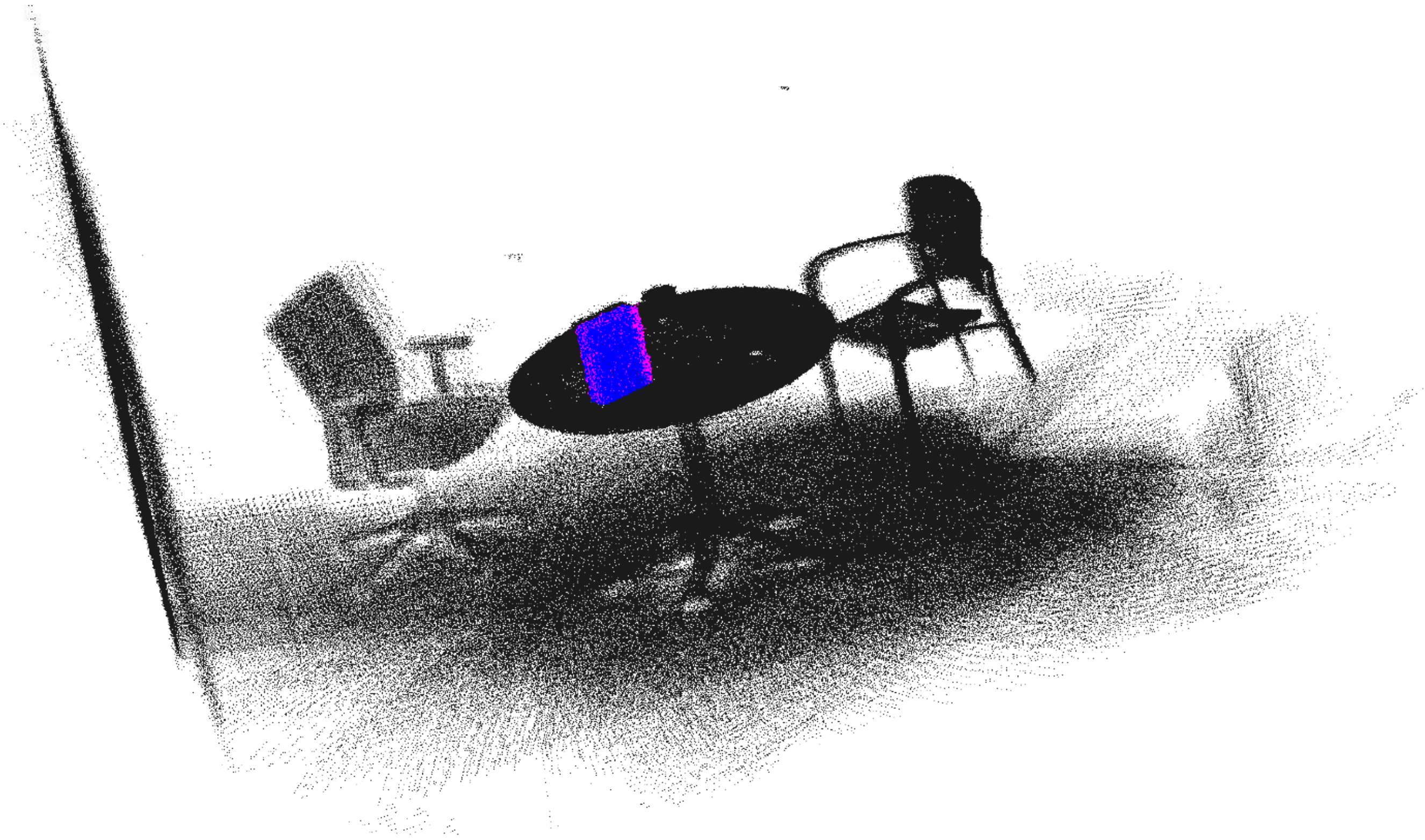}
\end{minipage}

&

\begin{minipage}[t]{0.19\linewidth}
\centering
\includegraphics[width=1\linewidth]{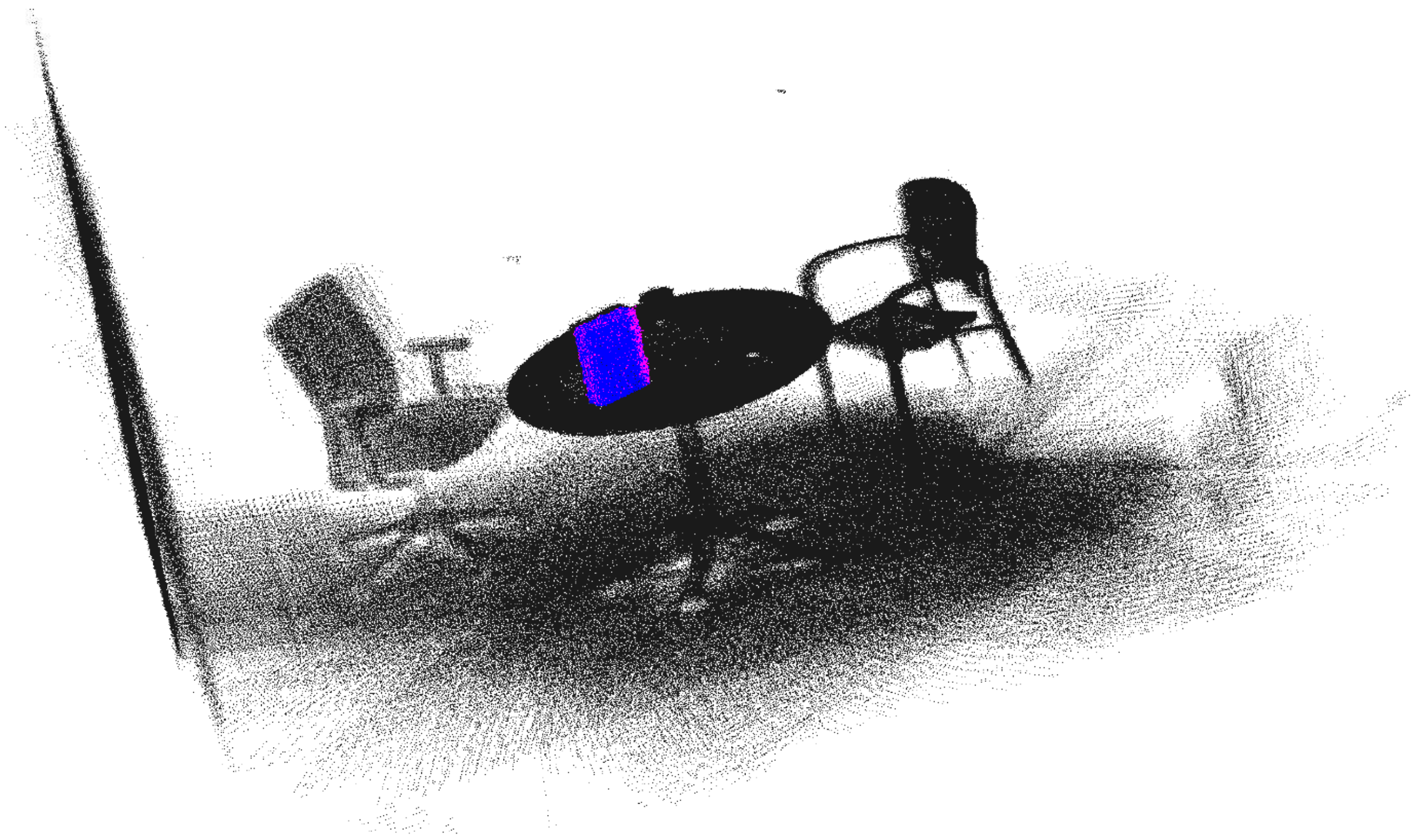}
\end{minipage}

\end{tabular}

\caption{Object localization and pose estimation results on the RGB-D scenes dataset~\cite{lai2011large}. The left-most column shows the raw FPFH correspondences where information regarding the correspondence number and outlier ratio is also provided, and the rest columns show the registration results (reprojecting the object back to the scene with the transformation estimated) using GNC-TLS, FLO-RANSAC, GORE+RANSAC and DANIEL, respectively. On top of the results of each solver, from left to right, we display the rotation error (in degrees), translation error (in meters) and runtime (in seconds), respectively. Best results are shown in \textbf{bold} font.}
\label{obj-local1}
\vspace{-3mm}
\end{figure*}

\subsection{Reslistic Registration on Real Data}

We conduct realistic point cloud registration experiments for further comparative evaluation over more datasets, including the `\textit{bunny}', `\textit{armadillo}' and `\textit{dragon}' from the Stanford 3D Scanning Repository~\cite{curless1996volumetric}, as well as the `\textit{cheff}', `\textit{chicken}', `\textit{rhino}', `\textit{parasaurolophus}' and `\textit{T-rex}' from Mian's dataset~\cite{mian2006three,mian2010repeatability} (8 point cloud datasets in total). For each point cloud above, we first resize it to fit in a $[-50,50]^3m$ box, then manually divide its full scan into 10 random partial scans where each scan shares 30\%-35\% overlapping with the original full scan, generate random transformations ($\boldsymbol{R}\in SO(3)$ and $\boldsymbol{t}\in\mathbb{R}^3$ where $||\boldsymbol{t}||\leq 100$) to transform these partial scans, and finally use FPFH~\cite{rusu2009fast} (Matlab function: \textit{extractFPFHFeatures}) to establish correspondences between each of the transformed partial scans and the initial full scan. When $N>1500$, we only preserve the first 1500 correspondences. Owing to serious partiality, the FPFH correspondences often contain huge amounts of outliers. Subsequently, these correspondences are fed to our DANIEL as well as the other solvers included in Section~\ref{sec-bench} to robustly solve the registration problems (10 for each point cloud, hence 80 problems in total), where the noise is set to $\sigma=0.1m$.

The qualitative registration examples (one example for each point cloud) are shown in Fig.~\ref{qualit-partial}, where the average correspondence number and outlier ratio of each point cloud are labeled on the left side of the examples. In addition, the quantitative results in boxplot are displayed in Fig.~\ref{Partial-reg}. We can find that our DANIEL (i) is the most robust and accurate solver, never yielding any single wrong estimate throughout the tests, and (ii) is fast in practice. Note that though the non-minimal solvers seem to show similar or even better (e.g. GNC-TLS) efficiency compared to DANIEL, they are prone to generate wrong results, unable to keep stably robust in all tests. Besides, the RANSAC solvers require relatively long runtime, and also occasionally return failure cases, while single-handed GORE has poor accuracy despite its promising runtime. GORE+RANSAC is the only solver that can have comparable robustness and accuracy with our DANIEL, but it is apparently slower than DANIEL. So overall, DANIEL manifests the best performance.

\begin{figure*}[t]
\centering
\setlength\tabcolsep{0.1pt}
\addtolength{\tabcolsep}{0pt}
\begin{tabular}{c|cc|ccccc}

\quad &\,&\,\footnotesize{Correspondences}\, &\,&  \footnotesize{GNC-TLS} & \footnotesize{FLO-RANSAC} & \footnotesize{GORE+RANSAC} & \footnotesize{DANIEL} \\

\hline

&&\footnotesize{$N=186$, 95.16\%}& & \,\footnotesize{$131.278^{\circ},2.784m,0.032s$}\, &\,\footnotesize{$\textbf{0.261}^{\circ},\textbf{0.005}m,6.024s$}\, &\,\footnotesize{$\textbf{0.261}^{\circ},\textbf{0.005}m,0.604s$}\,&\,\footnotesize{$\textbf{0.261}^{\circ},\textbf{0.005}m,\textbf{0.057}s$}\,

\\

\rotatebox{90}{\,\,\footnotesize{\textit{Scene-06, can}}\,}\,

& &

\begin{minipage}[t]{0.19\linewidth}
\centering
\includegraphics[width=1\linewidth]{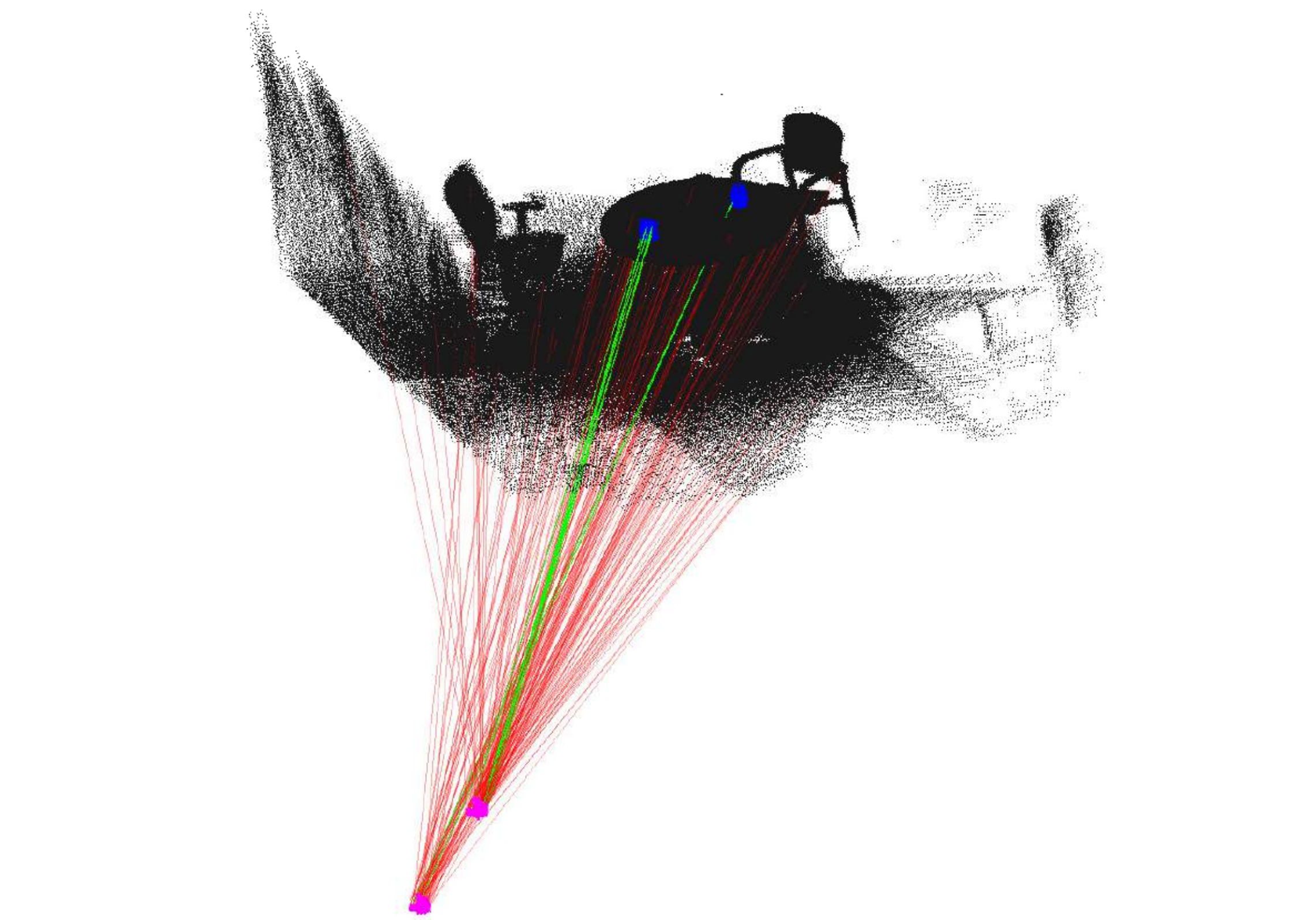}
\end{minipage}

& &

\begin{minipage}[t]{0.19\linewidth}
\centering
\includegraphics[width=1\linewidth]{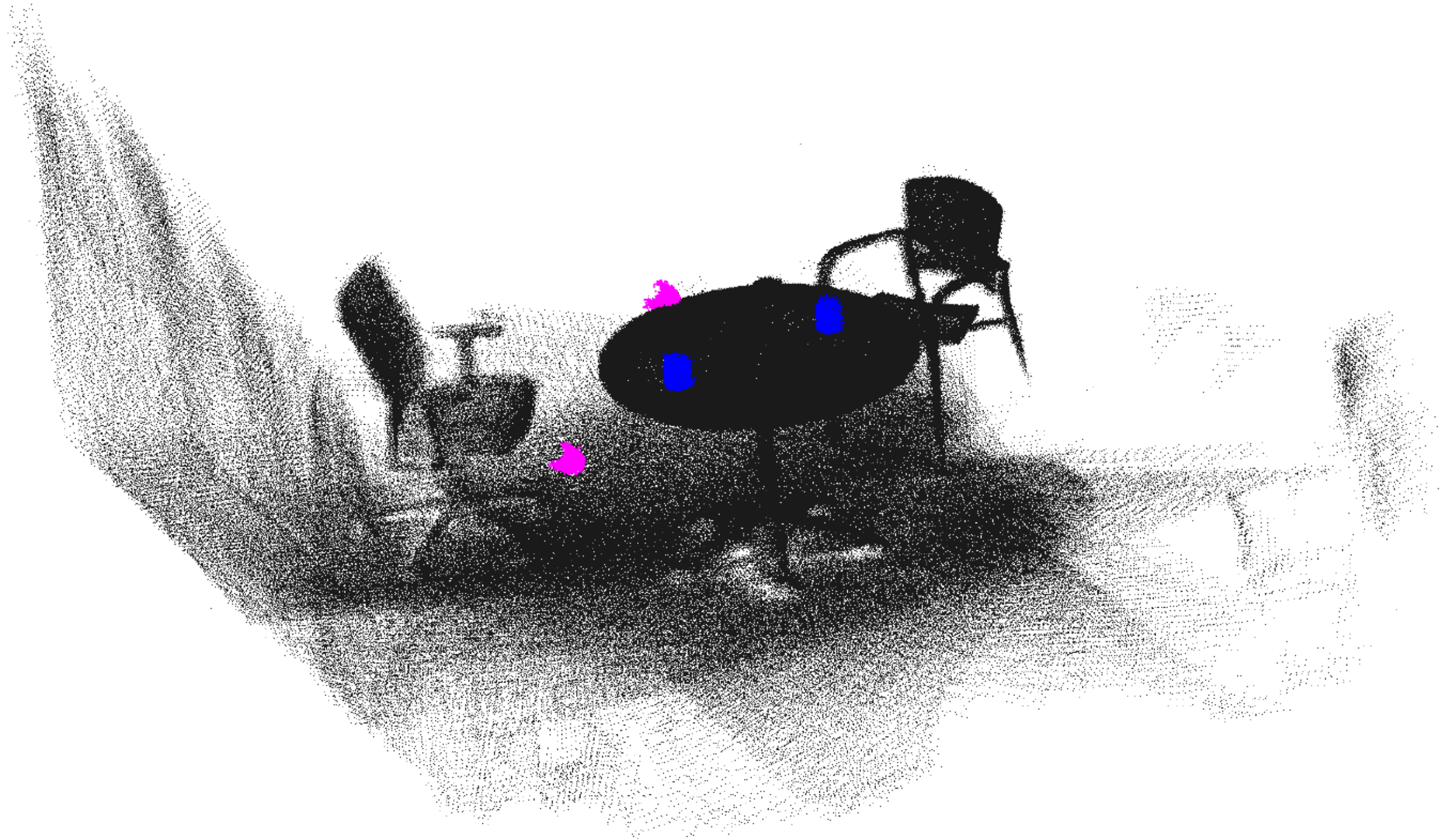}
\end{minipage}

&

\begin{minipage}[t]{0.19\linewidth}
\centering
\includegraphics[width=1\linewidth]{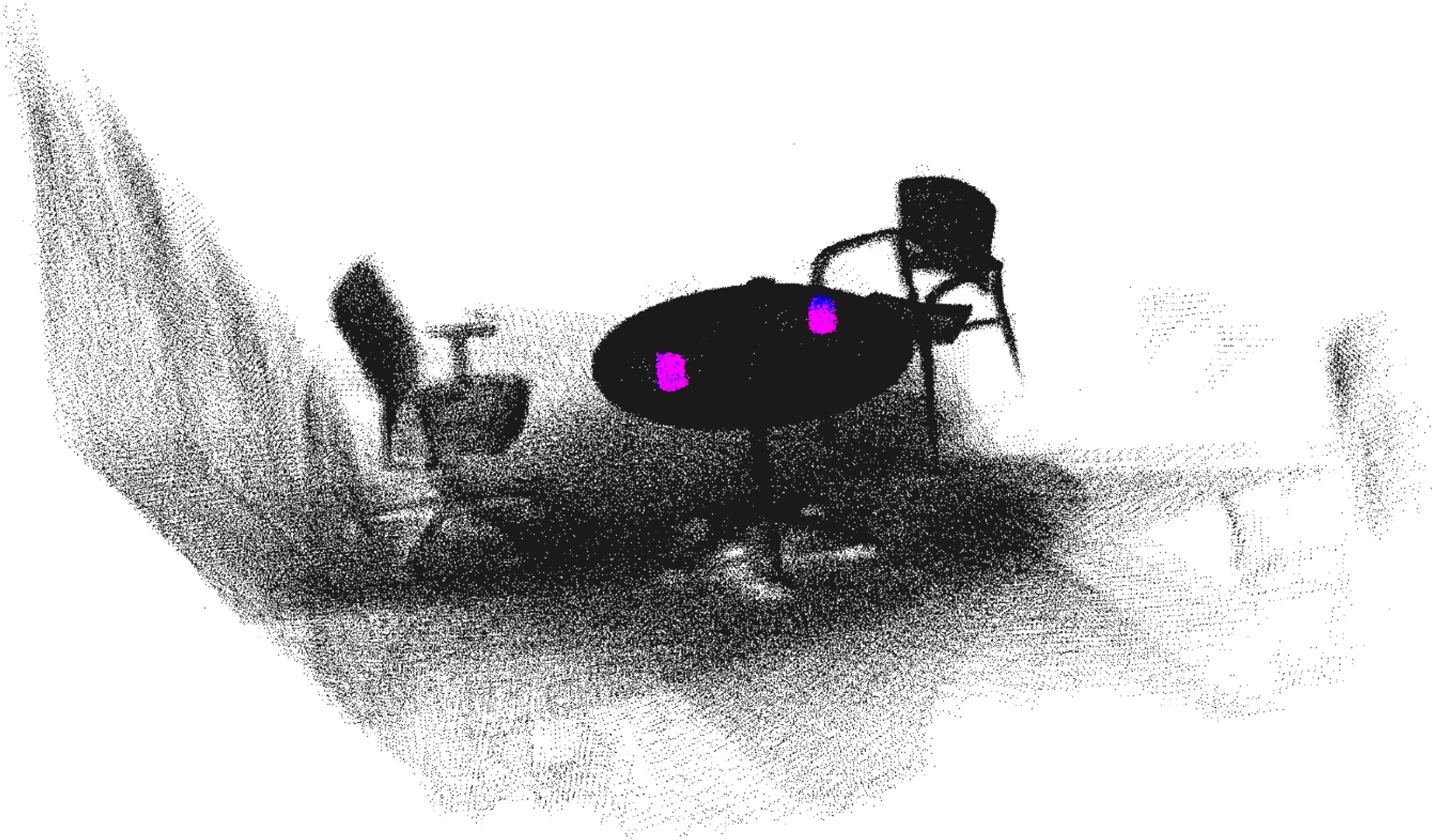}
\end{minipage}

&

\begin{minipage}[t]{0.19\linewidth}
\centering
\includegraphics[width=1\linewidth]{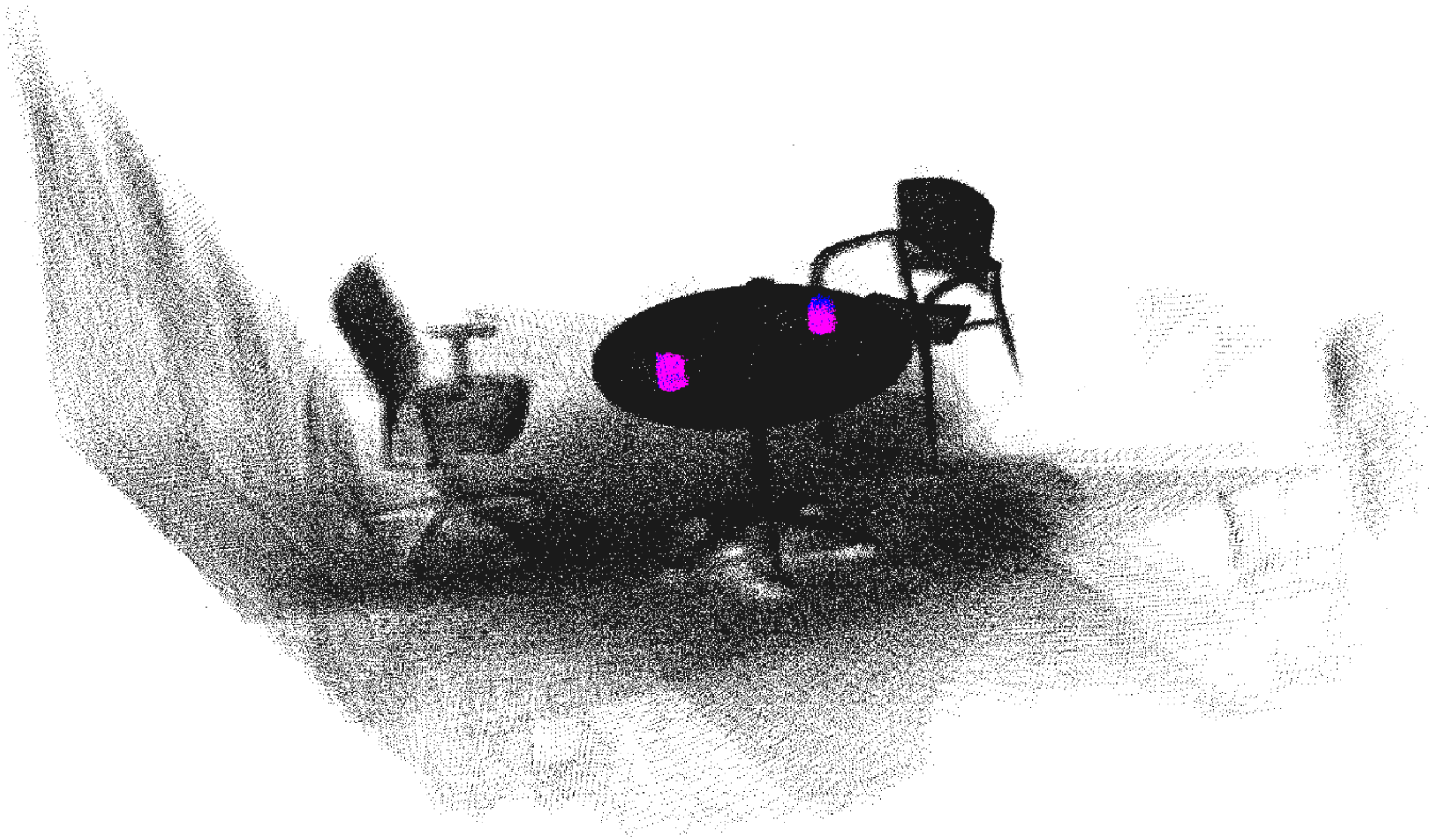}
\end{minipage}

&

\begin{minipage}[t]{0.19\linewidth}
\centering
\includegraphics[width=1\linewidth]{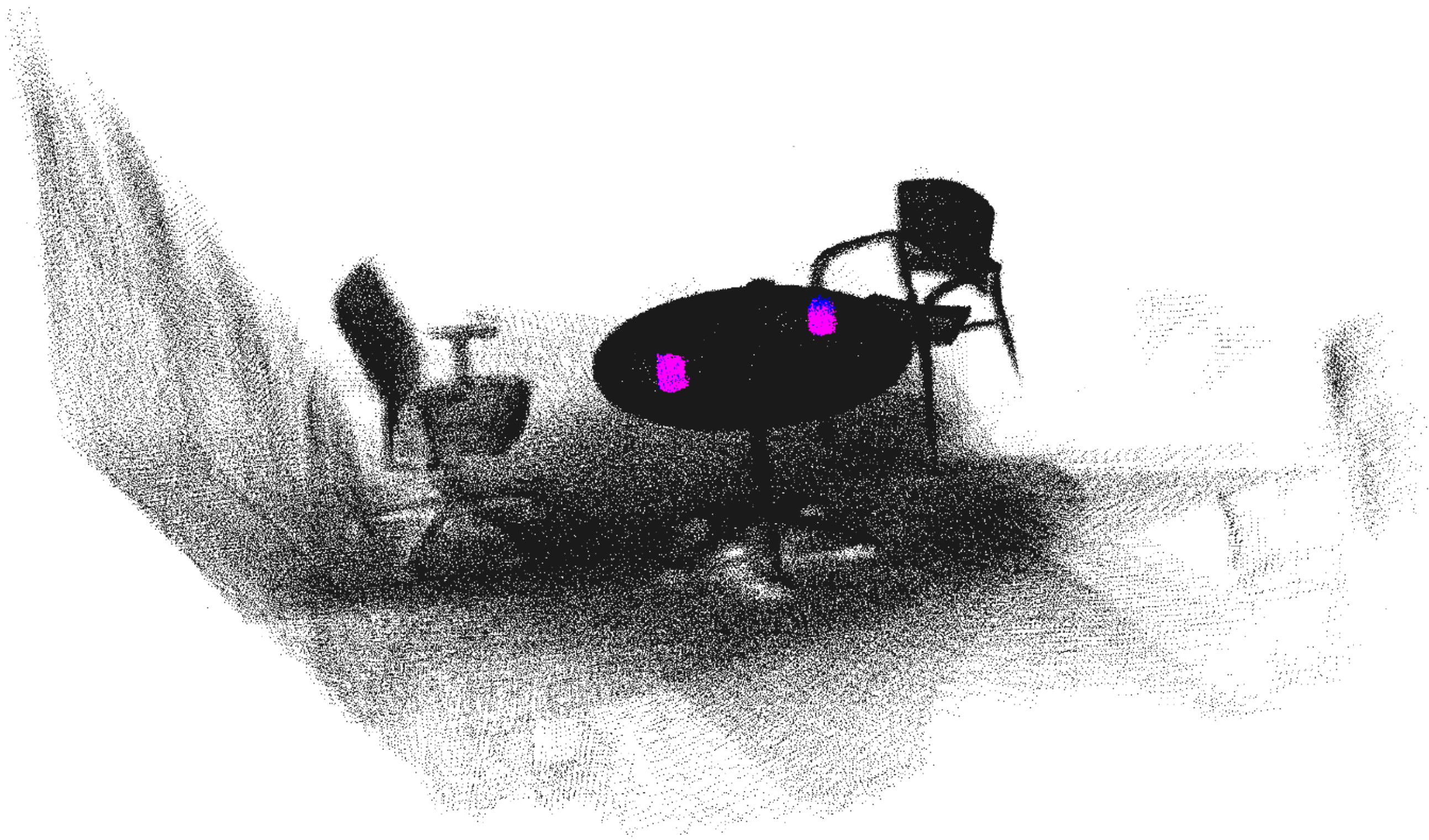}
\end{minipage}

\\

&&\footnotesize{$N=214$, 95.33\%}& & \,\footnotesize{$104.921^{\circ},1.537m,0.031s$}\, &\,\footnotesize{$\textbf{0.544}^{\circ},\textbf{0.012}m,6.814s$}\, &\,\footnotesize{$\textbf{0.544}^{\circ},\textbf{0.012}m,0.744s$}\,&\,\footnotesize{$\textbf{0.544}^{\circ},\textbf{0.012}m,\textbf{0.059}s$}\,

\\

\rotatebox{90}{\,\,\footnotesize{\textit{Scene-07, cap}}\,}\,

& &

\begin{minipage}[t]{0.19\linewidth}
\centering
\includegraphics[width=1\linewidth]{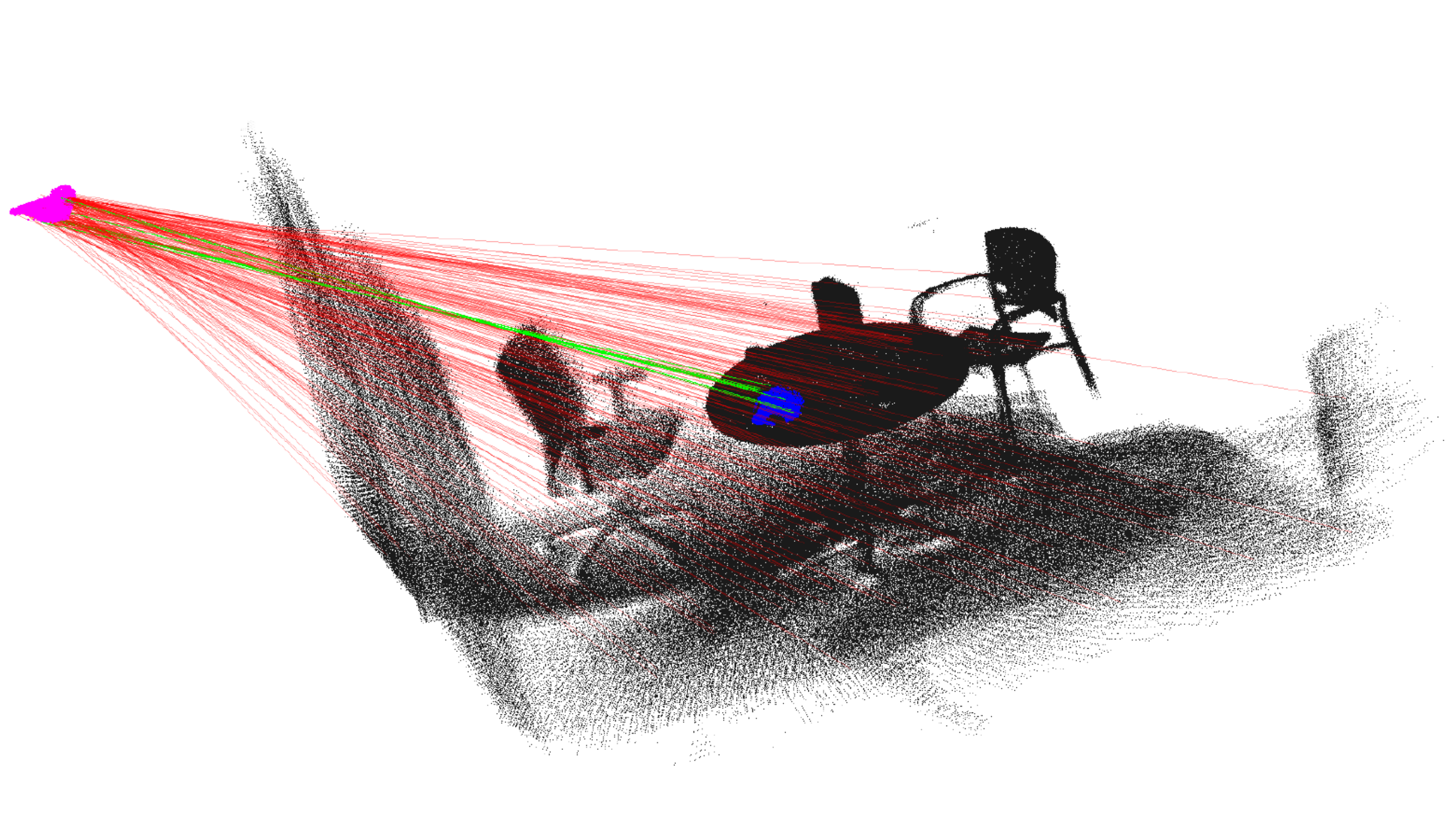}
\end{minipage}

& &

\begin{minipage}[t]{0.19\linewidth}
\centering
\includegraphics[width=1\linewidth]{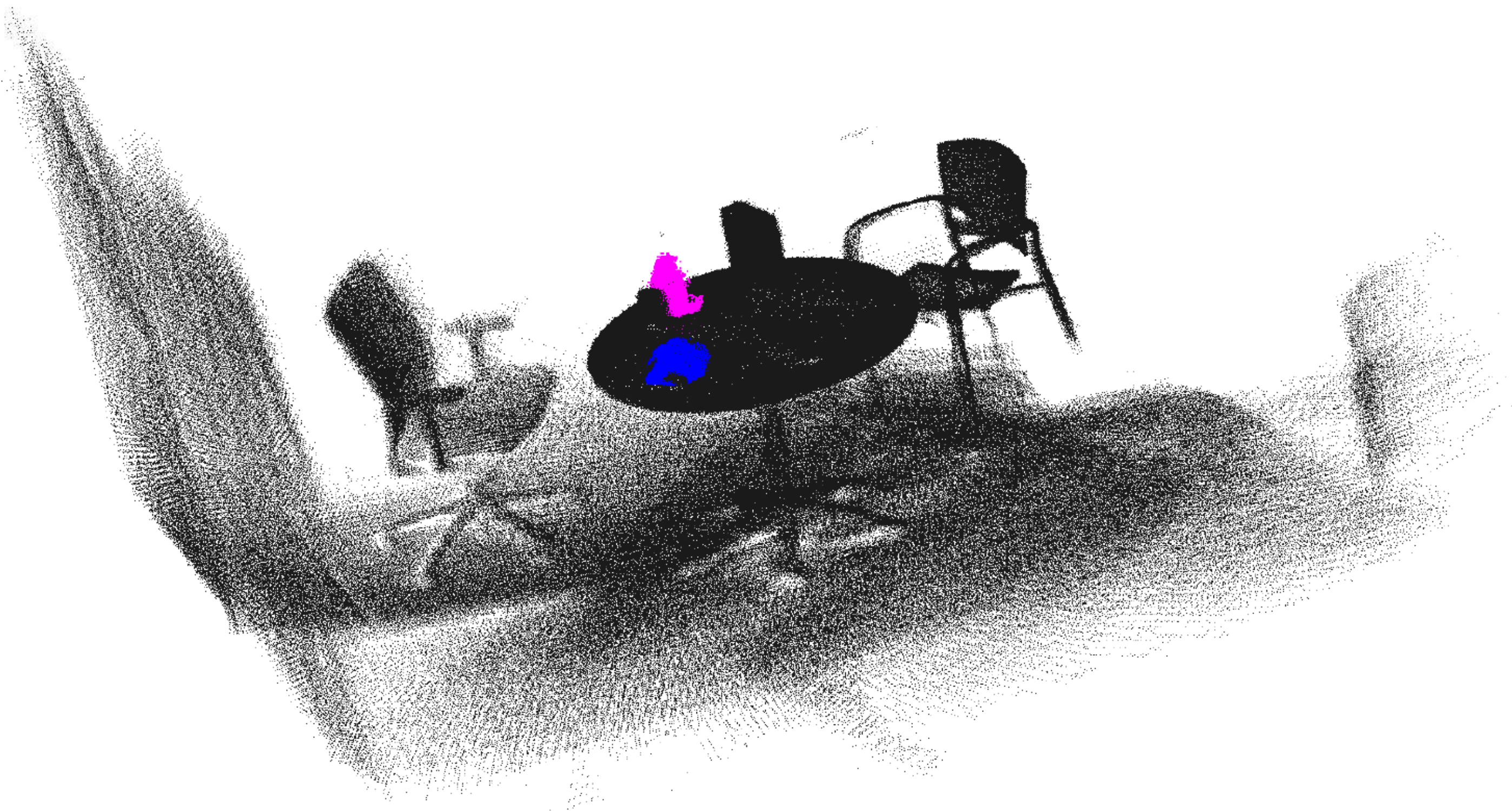}
\end{minipage}

&

\begin{minipage}[t]{0.19\linewidth}
\centering
\includegraphics[width=1\linewidth]{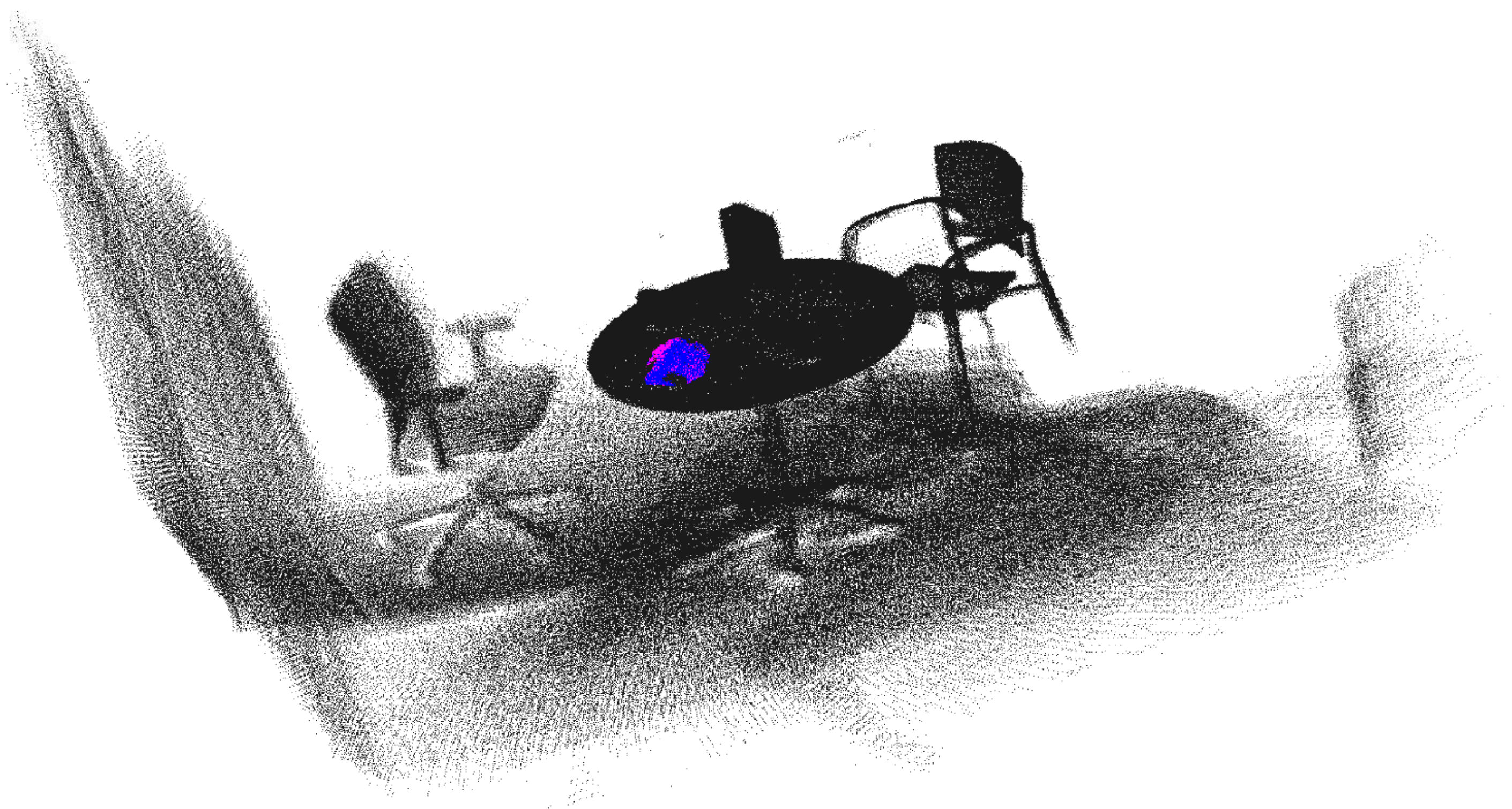}
\end{minipage}

&

\begin{minipage}[t]{0.19\linewidth}
\centering
\includegraphics[width=1\linewidth]{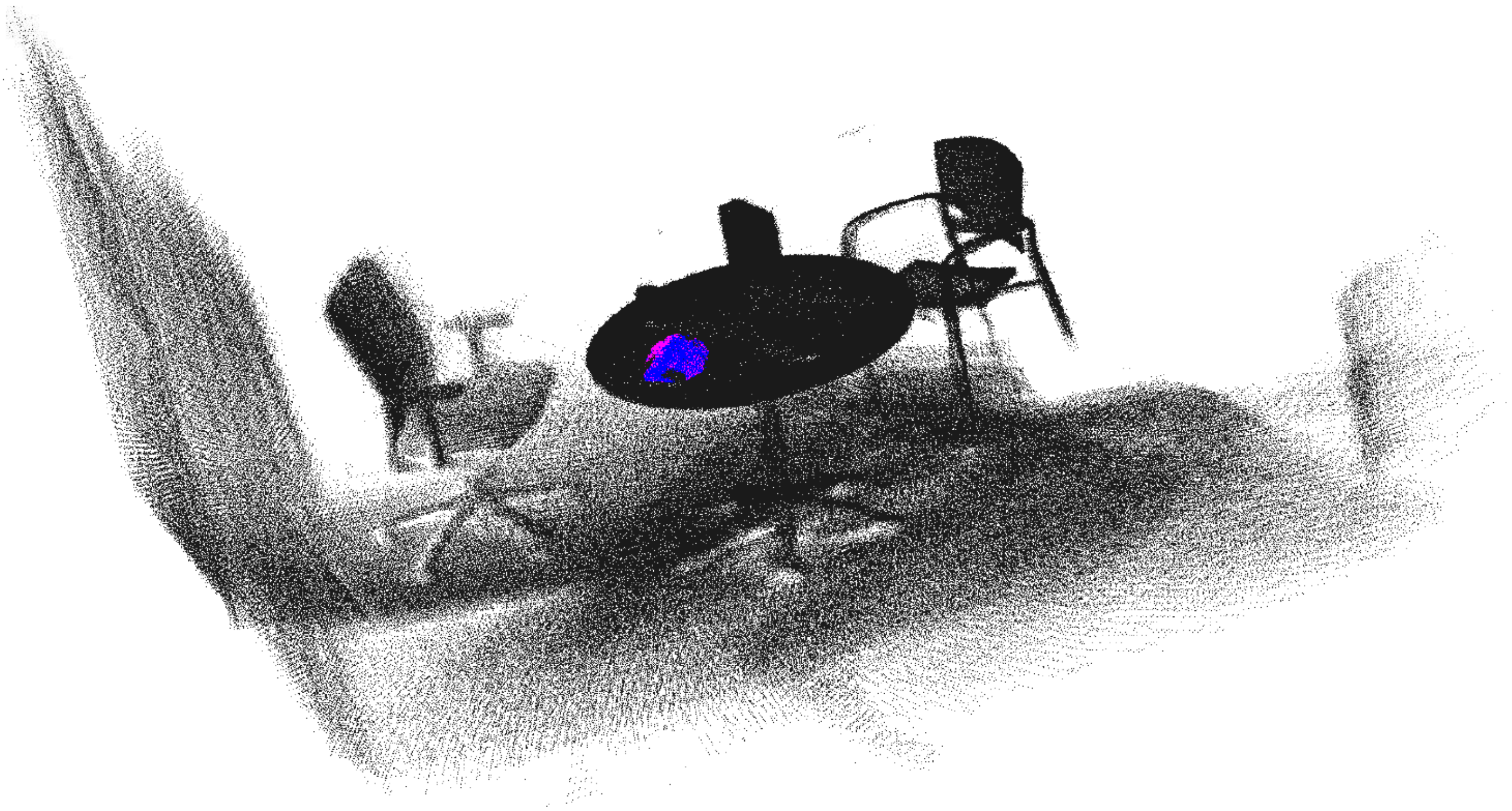}
\end{minipage}

&

\begin{minipage}[t]{0.19\linewidth}
\centering
\includegraphics[width=1\linewidth]{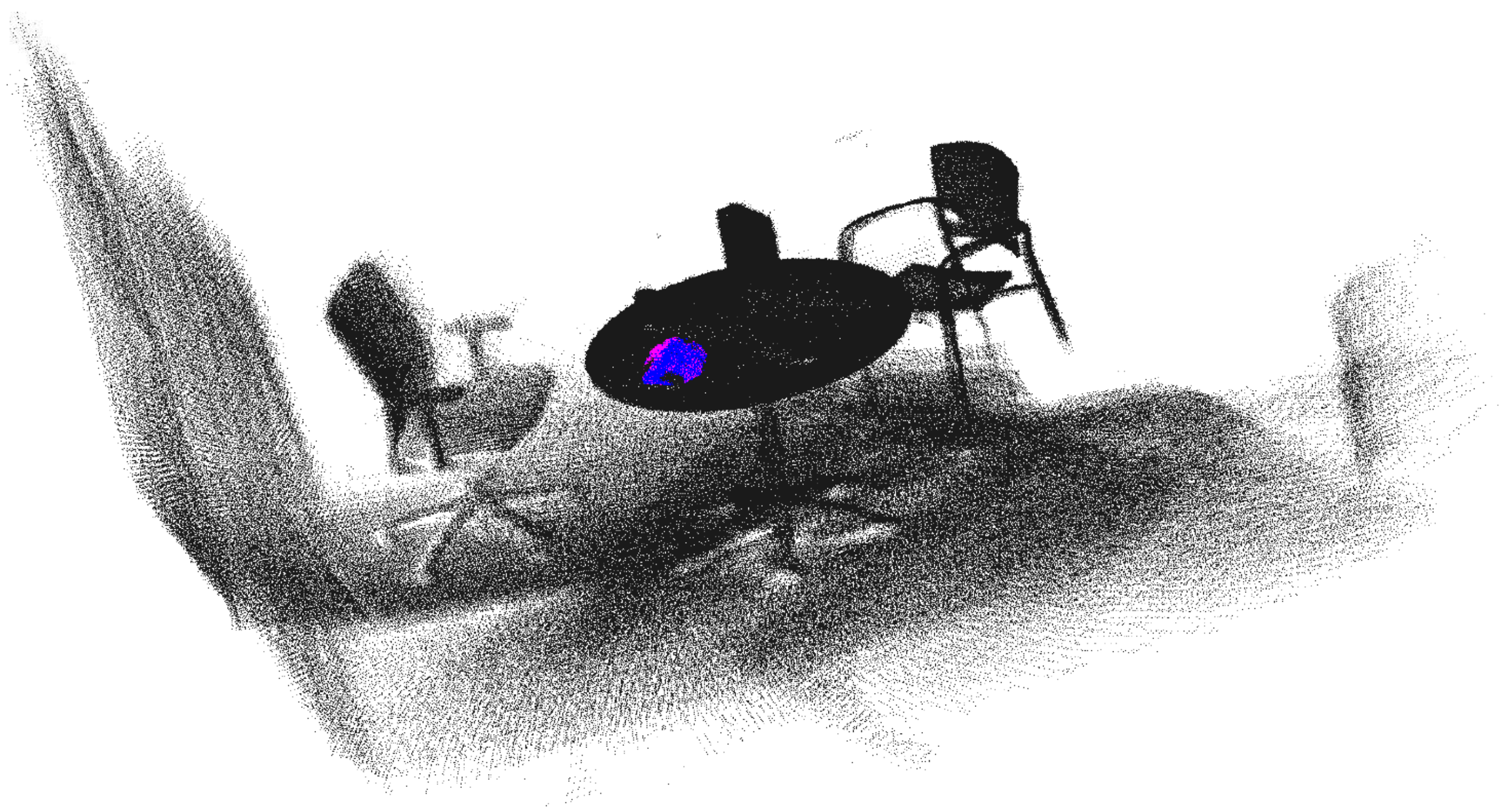}
\end{minipage}

\\

&&\footnotesize{$N=326$, 96.32\%}& & \,\footnotesize{$102.869^{\circ},1.769m,0.045s$}\, &\,\footnotesize{$\textbf{1.272}^{\circ},\textbf{0.017}m,12.510s$}\, &\,\footnotesize{$\textbf{1.272}^{\circ},\textbf{0.017}m,0.784s$}\,&\,\footnotesize{$\textbf{1.272}^{\circ},\textbf{0.017}m,\textbf{0.093}s$}\,

\\

\rotatebox{90}{\,\,\footnotesize{\textit{Scene-11, can}}\,}\,

& &

\begin{minipage}[t]{0.19\linewidth}
\centering
\includegraphics[width=1\linewidth]{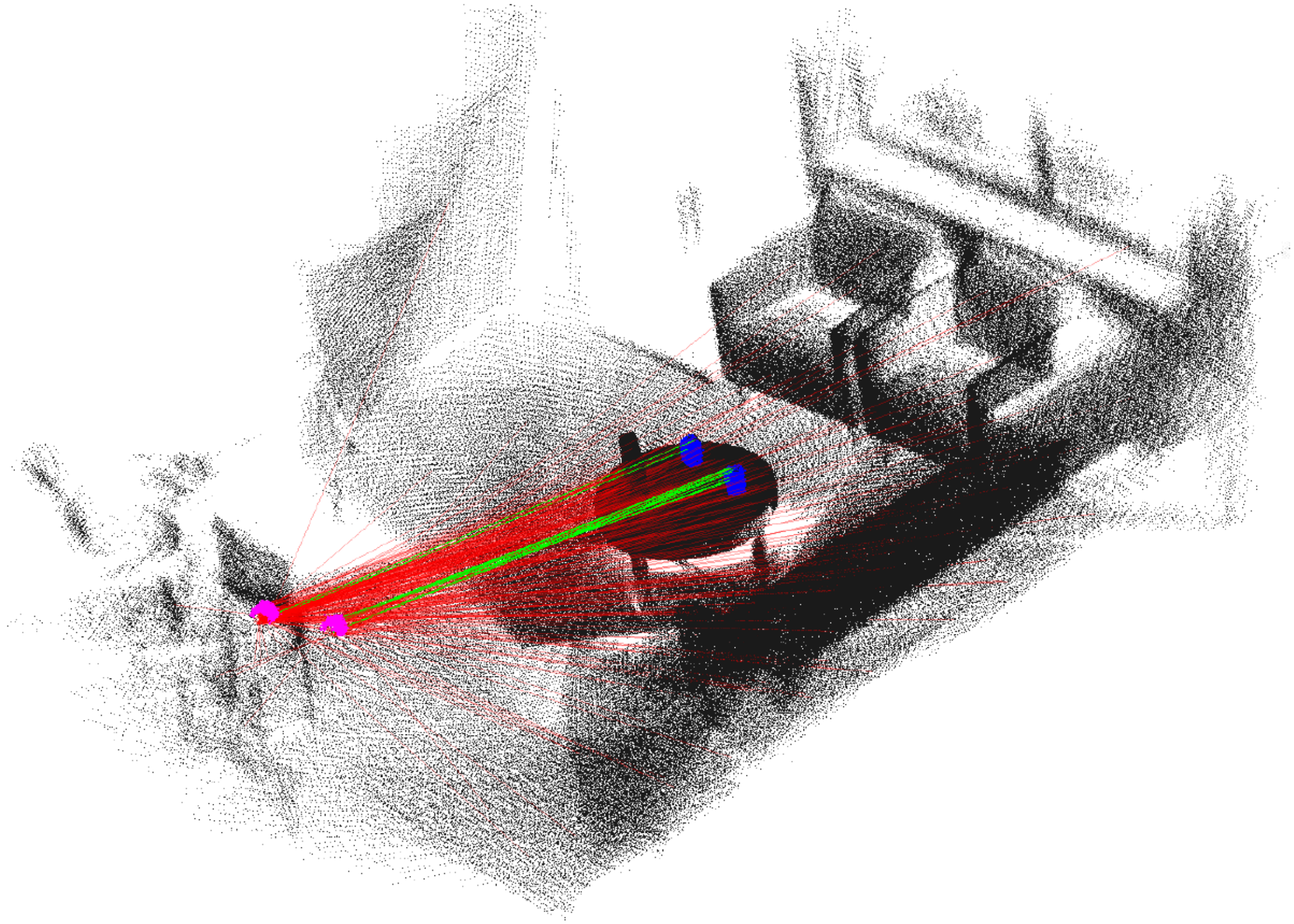}
\end{minipage}

& &

\begin{minipage}[t]{0.19\linewidth}
\centering
\includegraphics[width=1\linewidth]{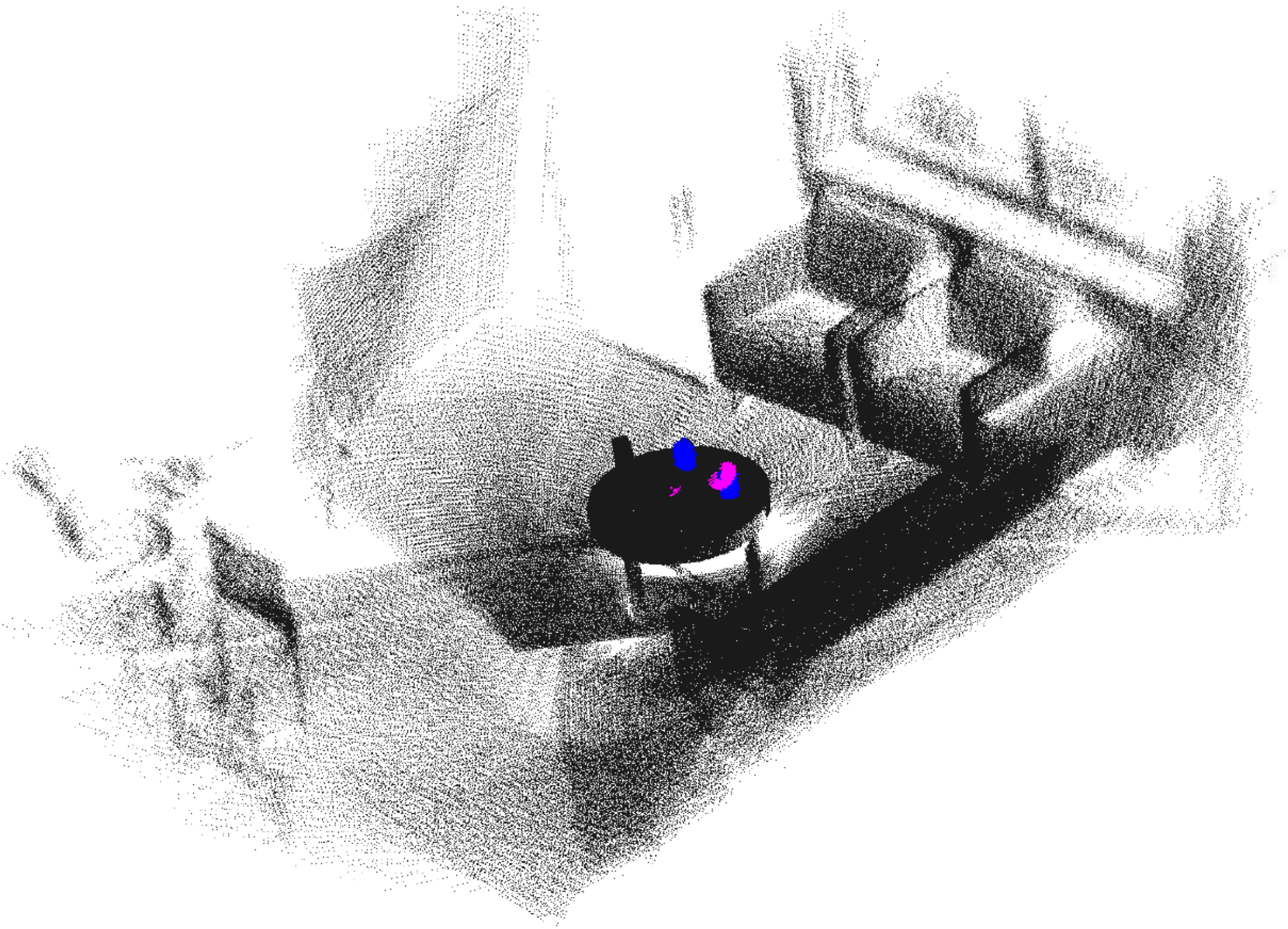}
\end{minipage}

&

\begin{minipage}[t]{0.19\linewidth}
\centering
\includegraphics[width=1\linewidth]{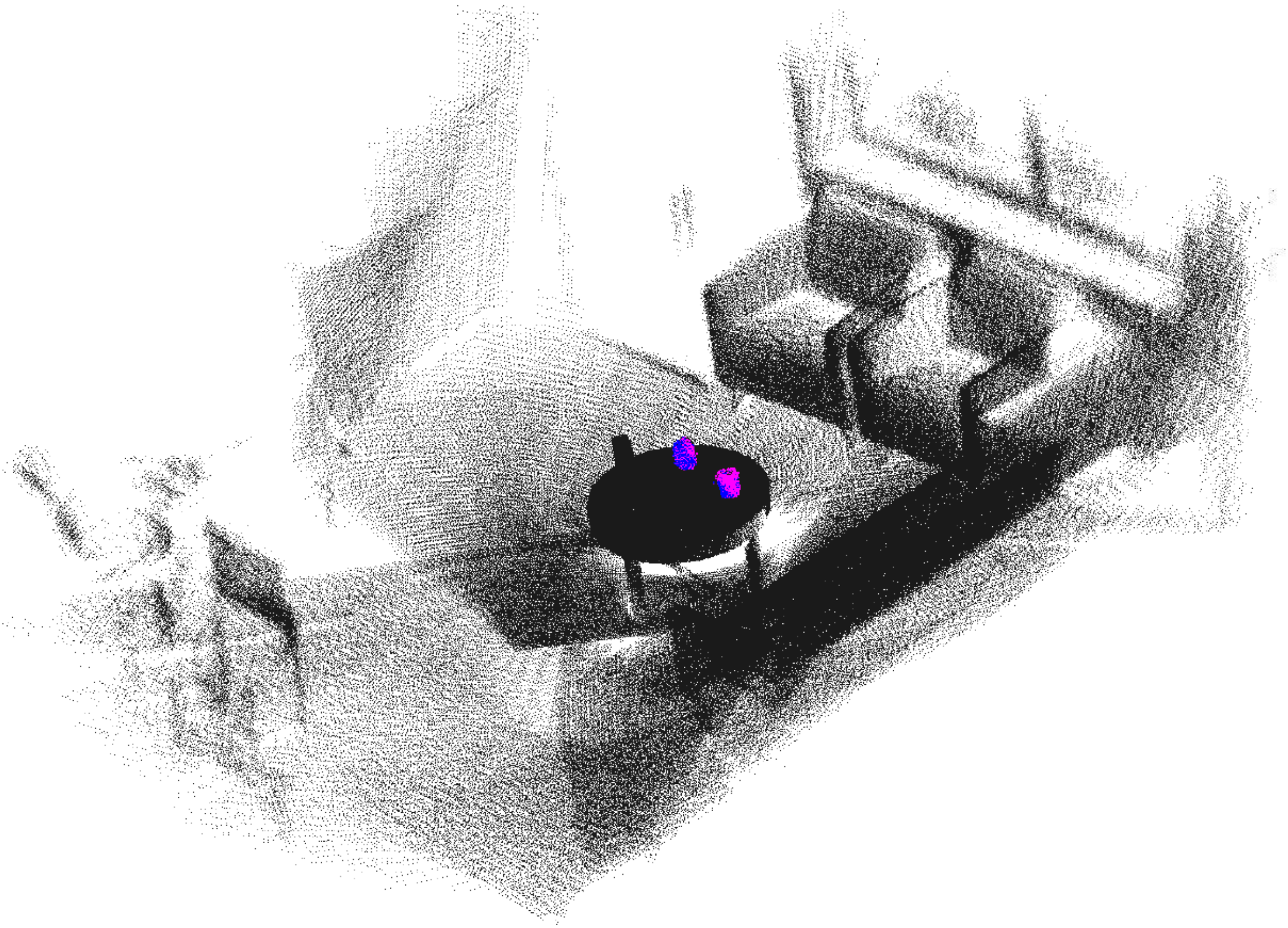}
\end{minipage}

&

\begin{minipage}[t]{0.19\linewidth}
\centering
\includegraphics[width=1\linewidth]{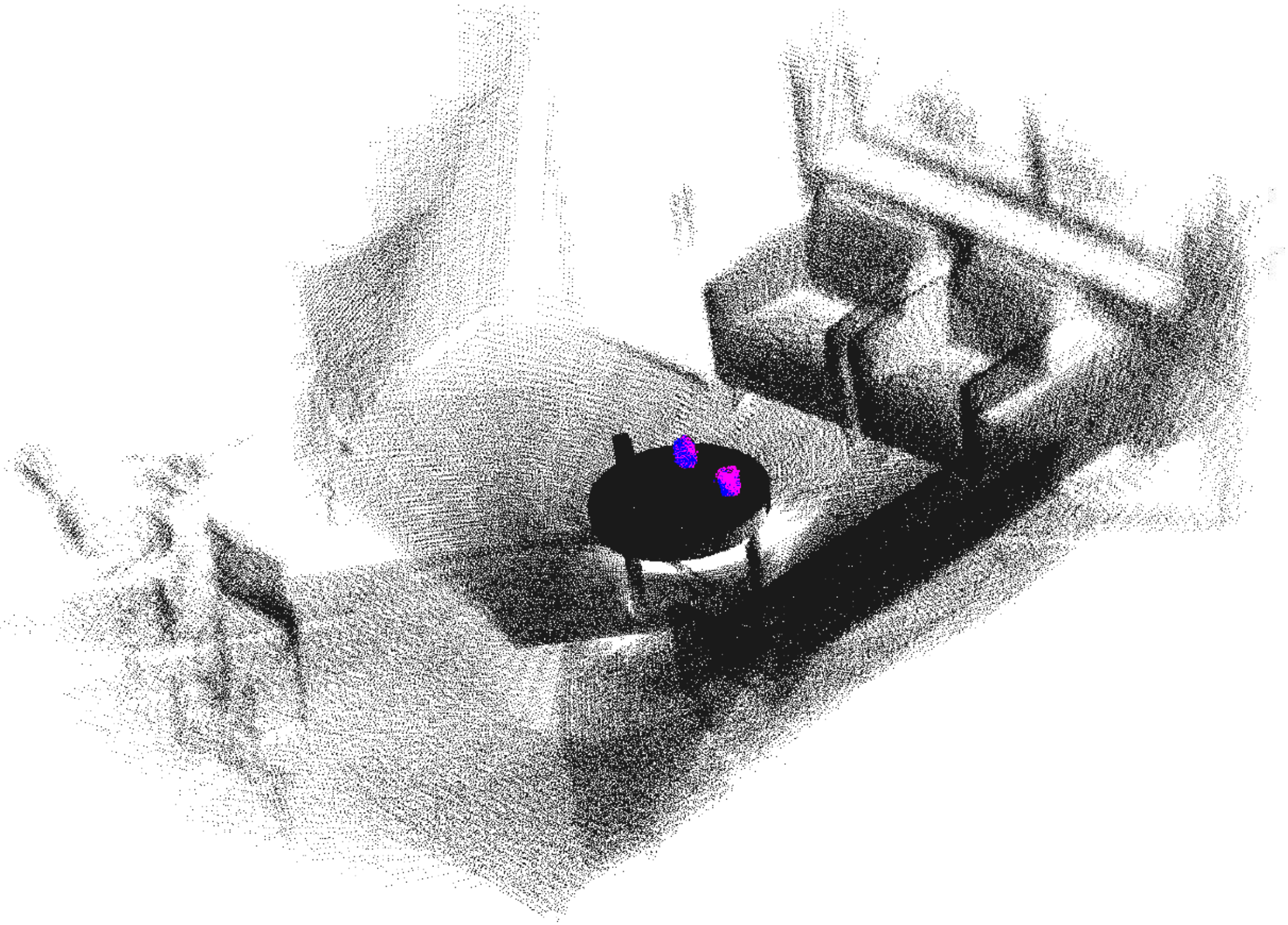}
\end{minipage}

&

\begin{minipage}[t]{0.19\linewidth}
\centering
\includegraphics[width=1\linewidth]{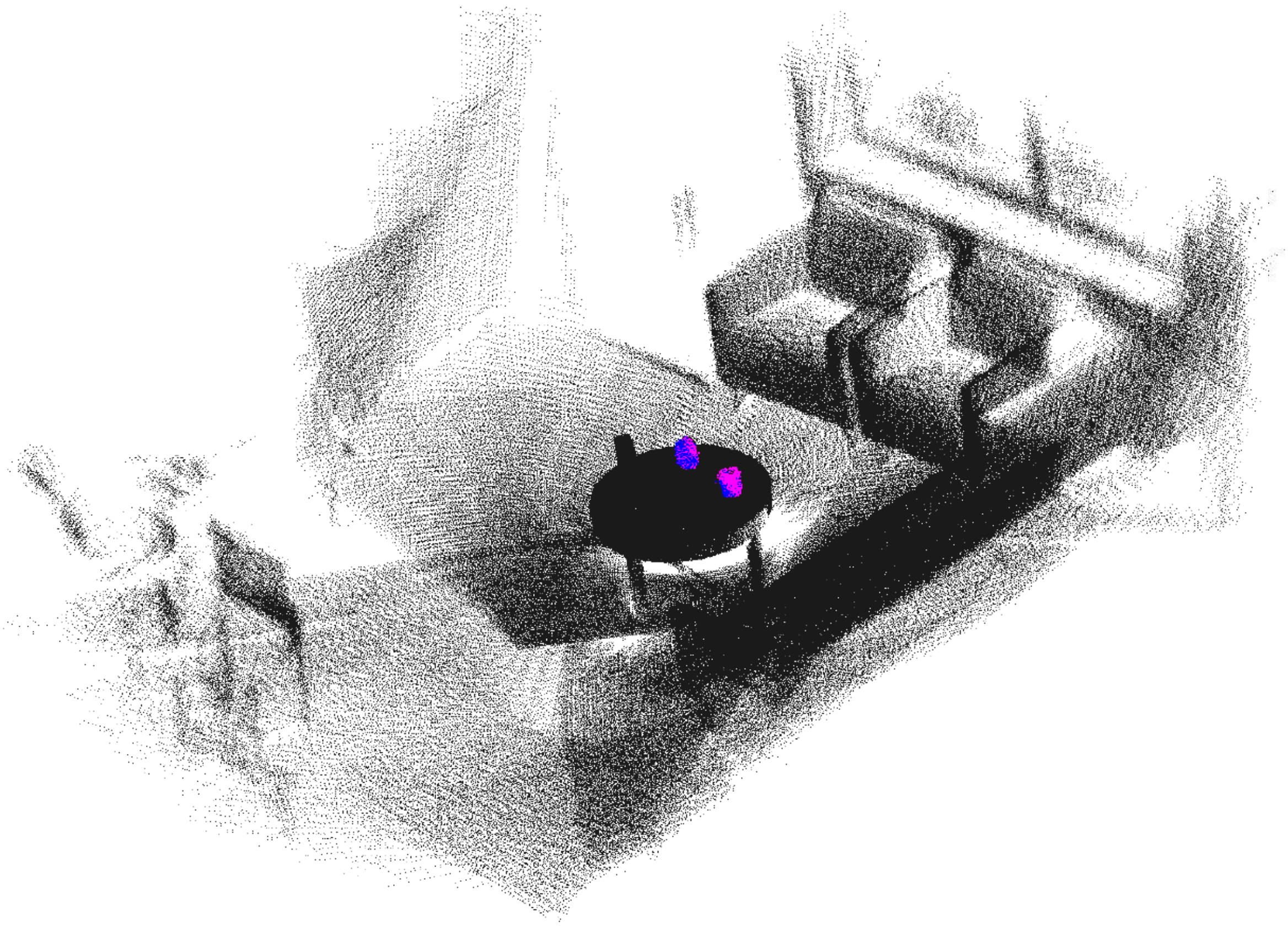}
\end{minipage}

\\

&&\footnotesize{$N=339$, 97.05\%}& & \,\footnotesize{$140.652^{\circ},2.390m,0.038s$}\, &\,\footnotesize{$121.956^{\circ},4.738m,18.808s$}\, &\,\footnotesize{$\textbf{0.582}^{\circ},\textbf{0.014}m,0.753s$}\,&\,\footnotesize{$\textbf{0.582}^{\circ},\textbf{0.014}m,\textbf{0.119}s$}\,

\\

\rotatebox{90}{\,\,\footnotesize{\textit{Scene-12, mug}}\,}\,

& &

\begin{minipage}[t]{0.19\linewidth}
\centering
\includegraphics[width=1\linewidth]{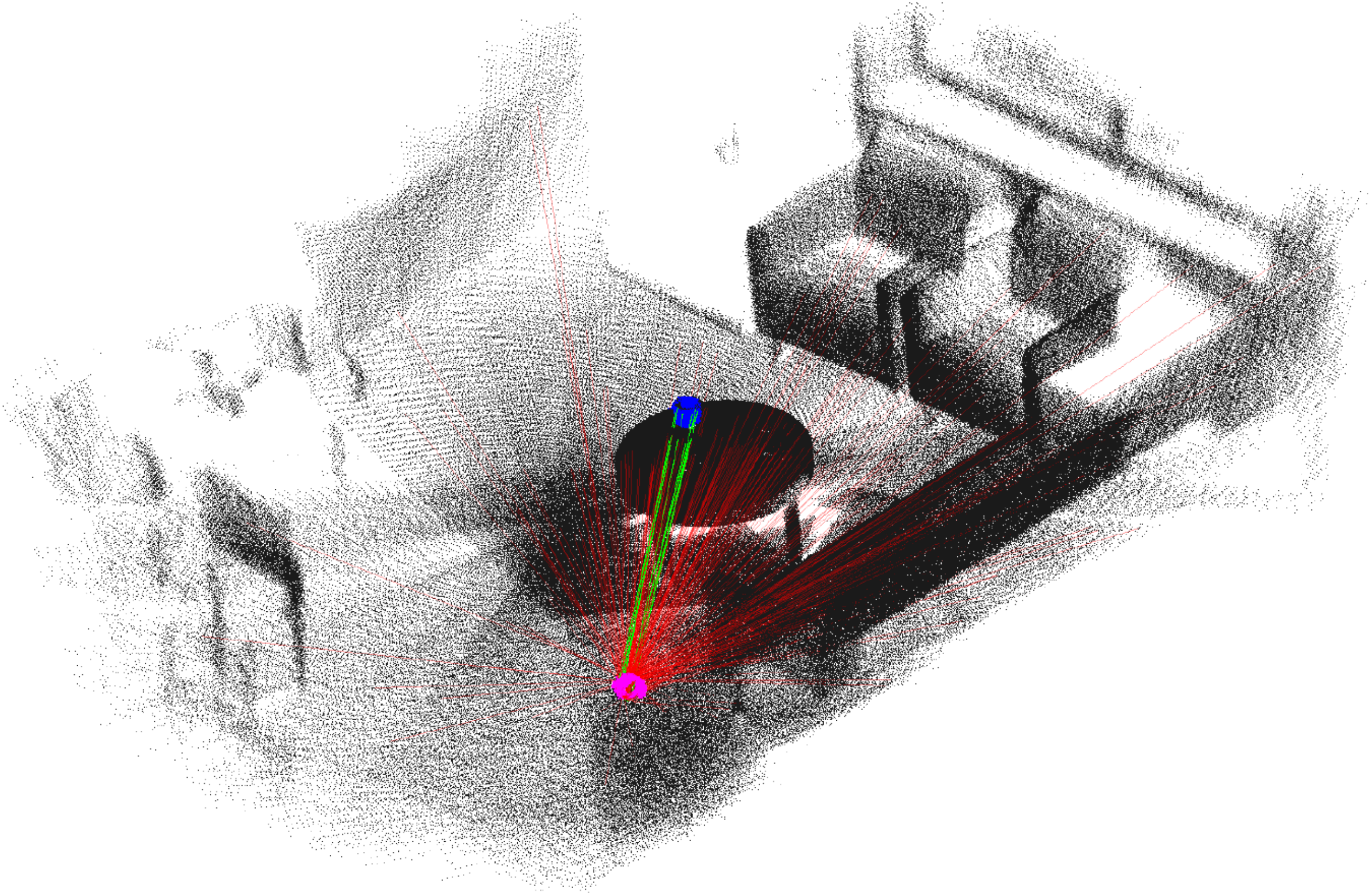}
\end{minipage}

& &

\begin{minipage}[t]{0.19\linewidth}
\centering
\includegraphics[width=1\linewidth]{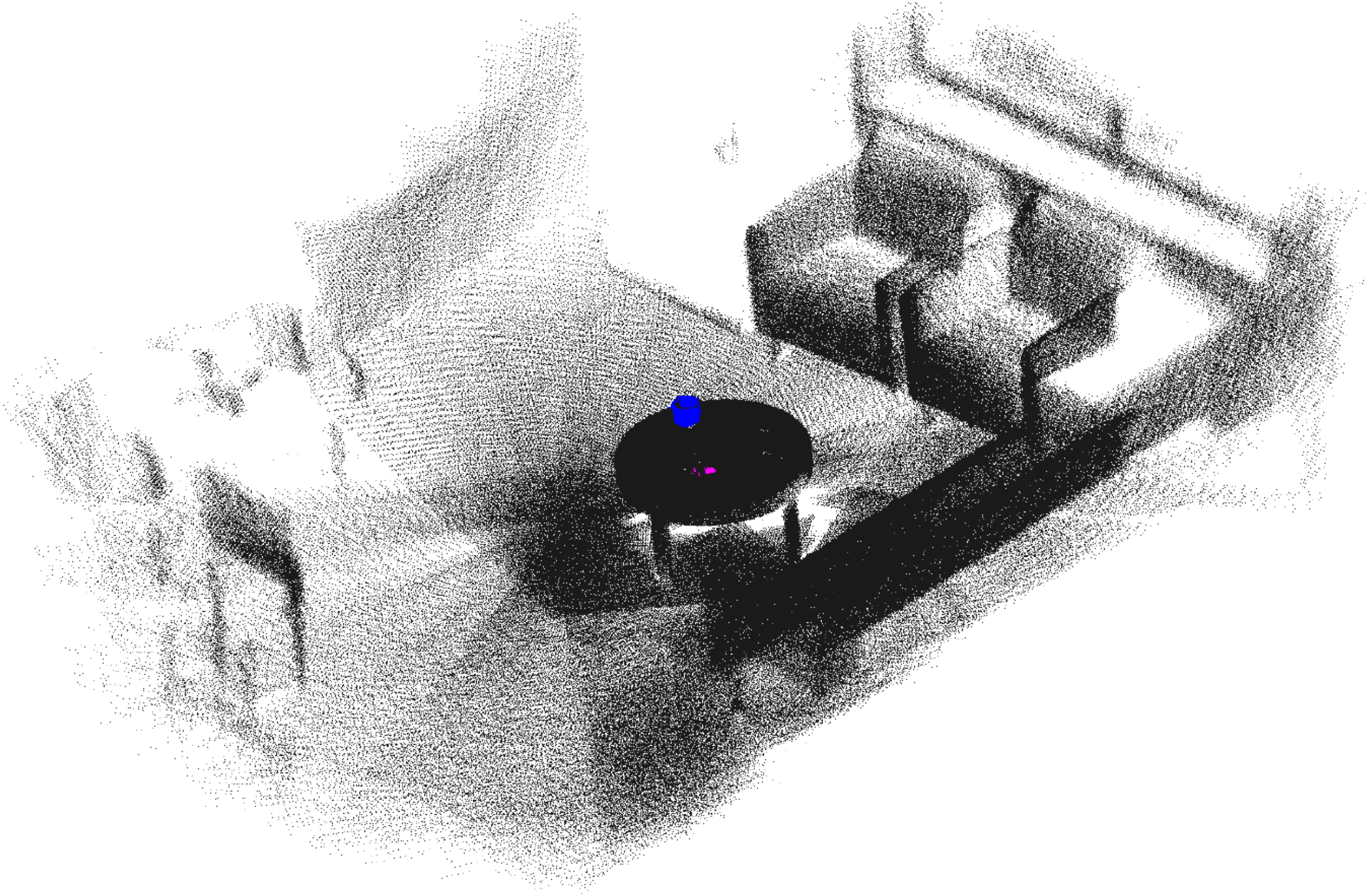}
\end{minipage}

&

\begin{minipage}[t]{0.19\linewidth}
\centering
\includegraphics[width=1\linewidth]{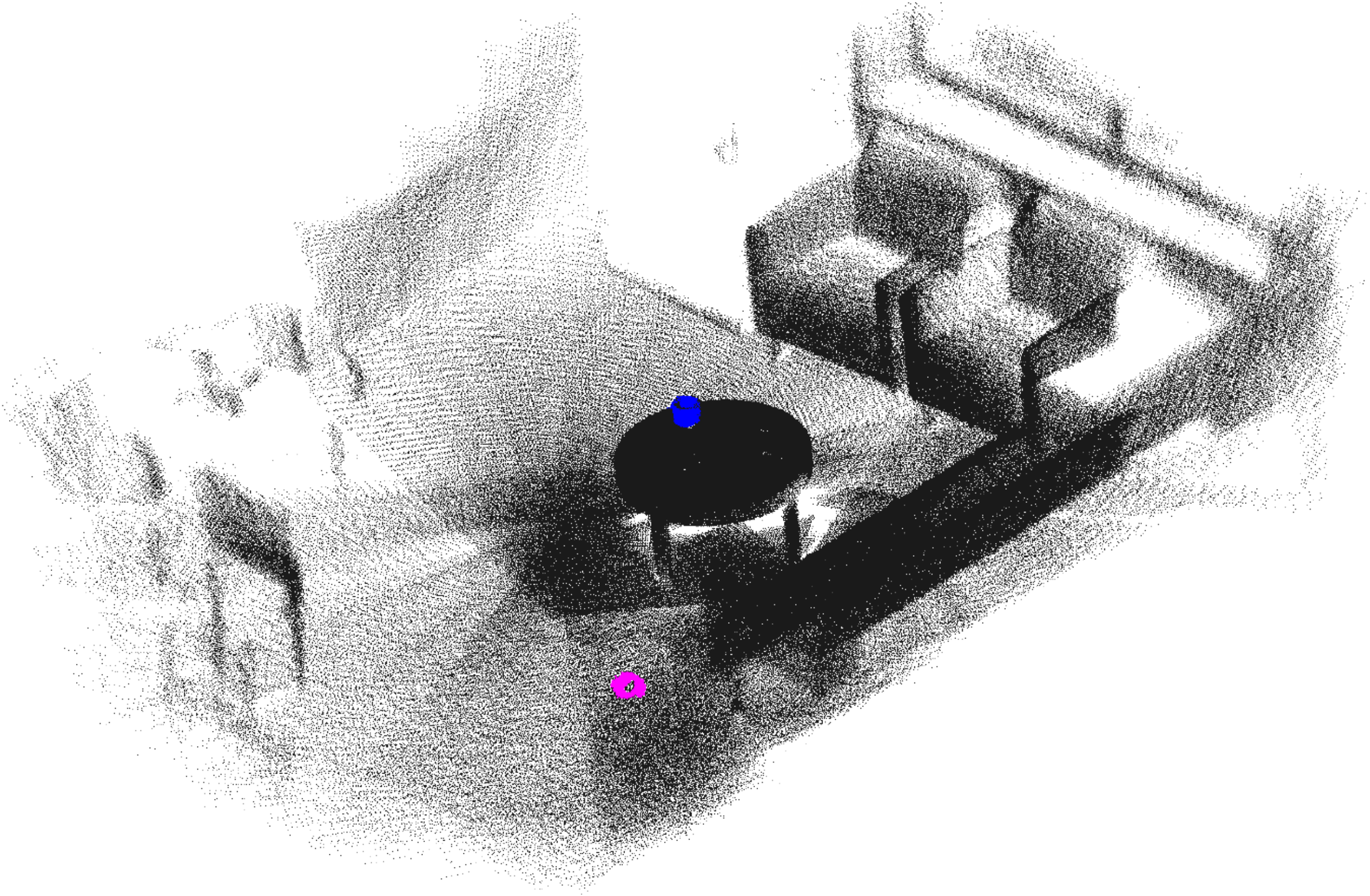}
\end{minipage}

&

\begin{minipage}[t]{0.19\linewidth}
\centering
\includegraphics[width=1\linewidth]{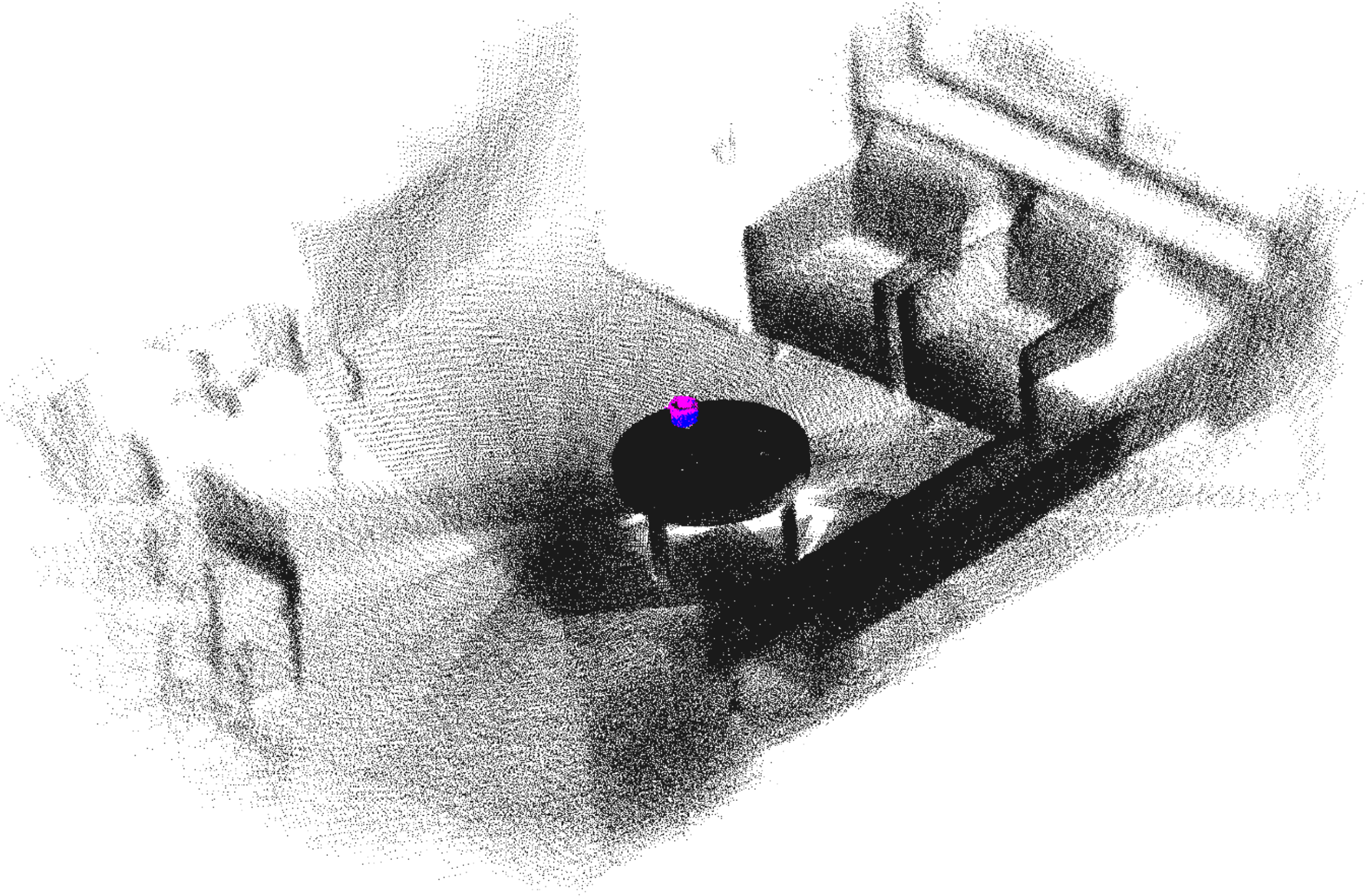}
\end{minipage}

&

\begin{minipage}[t]{0.19\linewidth}
\centering
\includegraphics[width=1\linewidth]{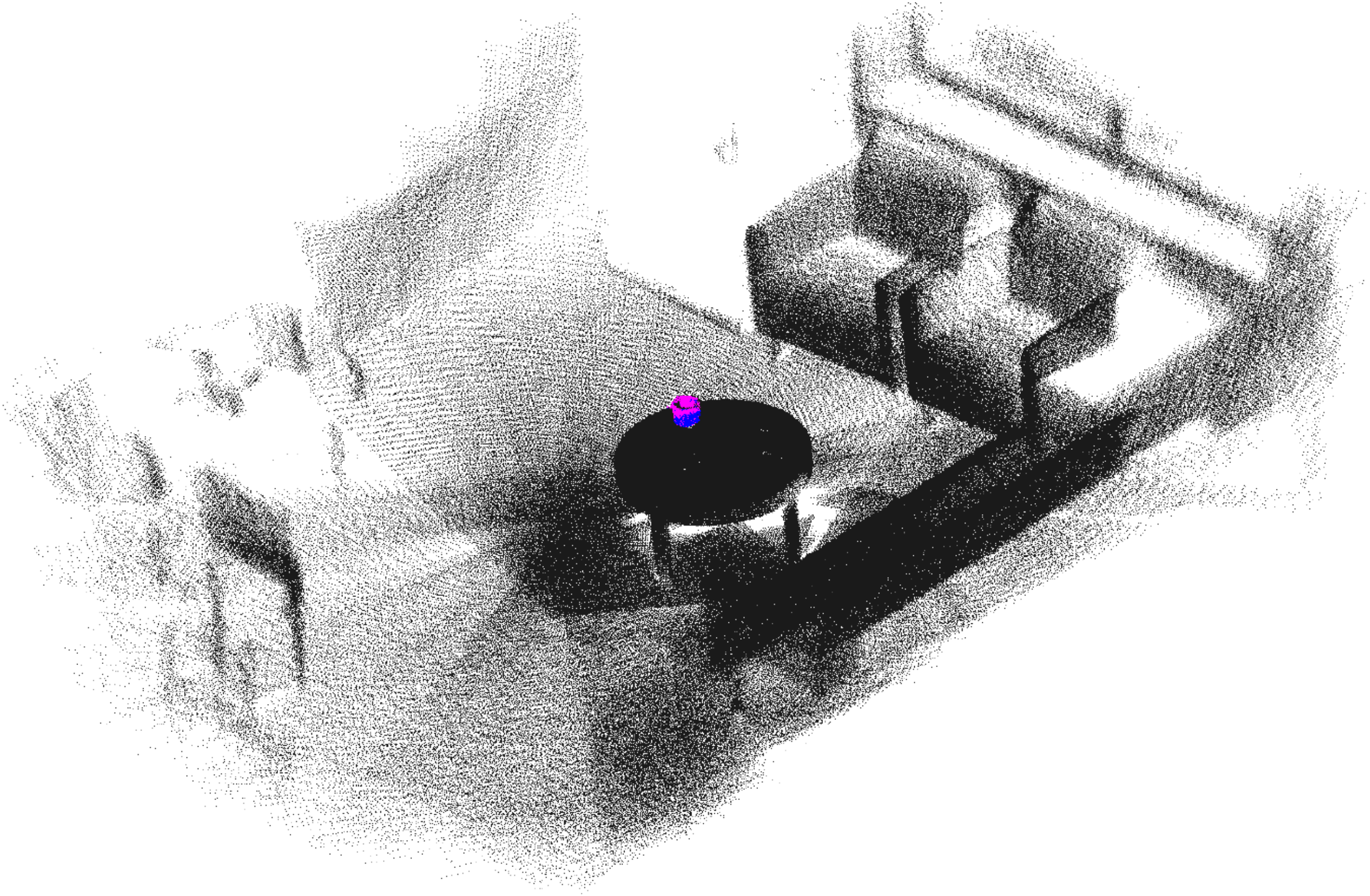}
\end{minipage}

\\

&&\footnotesize{$N=382$, 96.07\%}& & \,\footnotesize{$103.388^{\circ},1.950m,0.047s$}\, &\,\footnotesize{$\textbf{0.640}^{\circ},\textbf{0.006}m,17.814s$}\, &\,\footnotesize{$\textbf{0.640}^{\circ},\textbf{0.006}m,0.910s$}\,&\,\footnotesize{$\textbf{0.640}^{\circ},\textbf{0.006}m,\textbf{0.103}s$}\,

\\

\rotatebox{90}{\,\,\footnotesize{\textit{Scene-14, bowl}}\,}\,

& &

\begin{minipage}[t]{0.19\linewidth}
\centering
\includegraphics[width=1\linewidth]{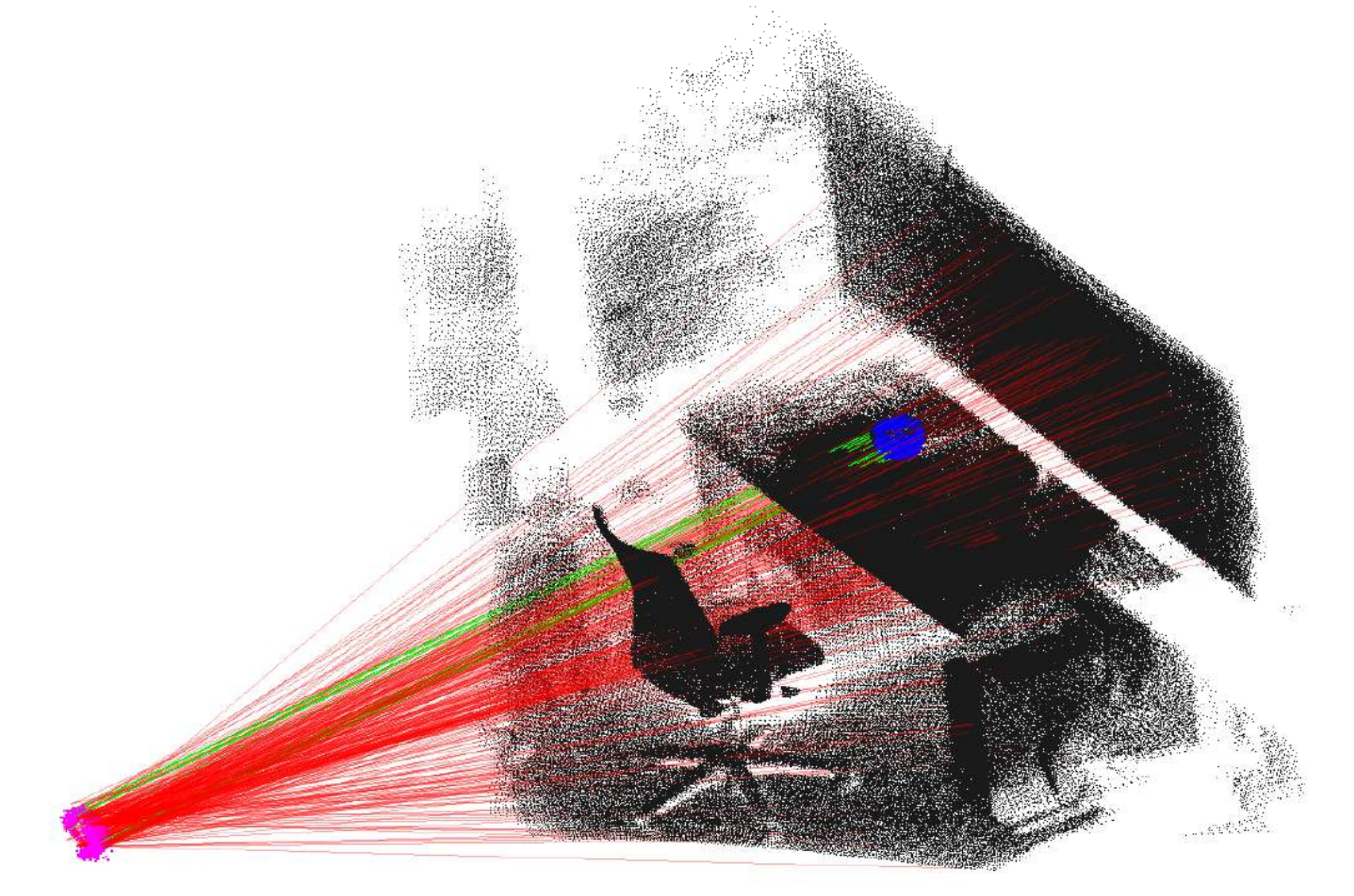}
\end{minipage}

& &

\begin{minipage}[t]{0.19\linewidth}
\centering
\includegraphics[width=1\linewidth]{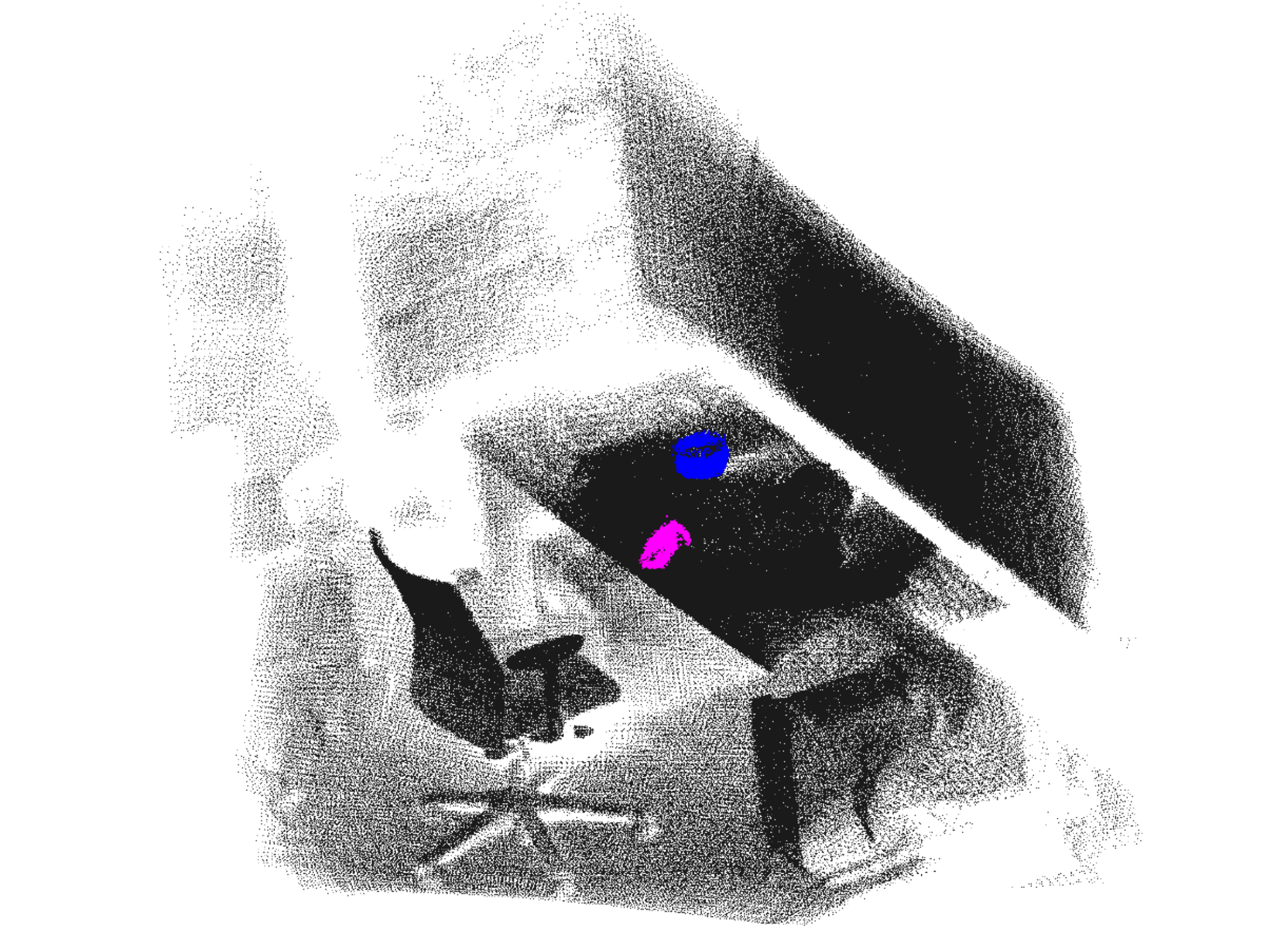}
\end{minipage}

&

\begin{minipage}[t]{0.19\linewidth}
\centering
\includegraphics[width=1\linewidth]{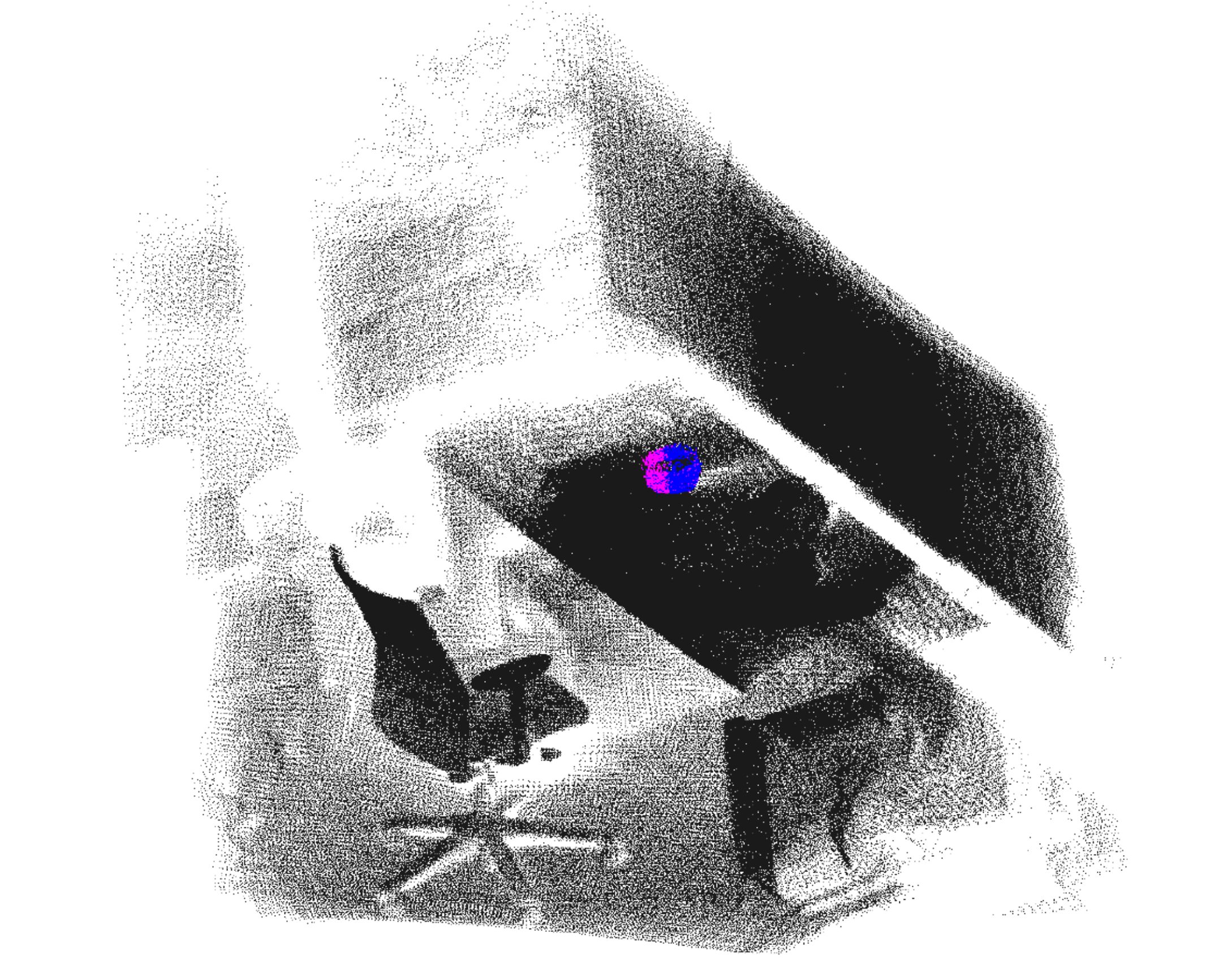}
\end{minipage}

&

\begin{minipage}[t]{0.19\linewidth}
\centering
\includegraphics[width=1\linewidth]{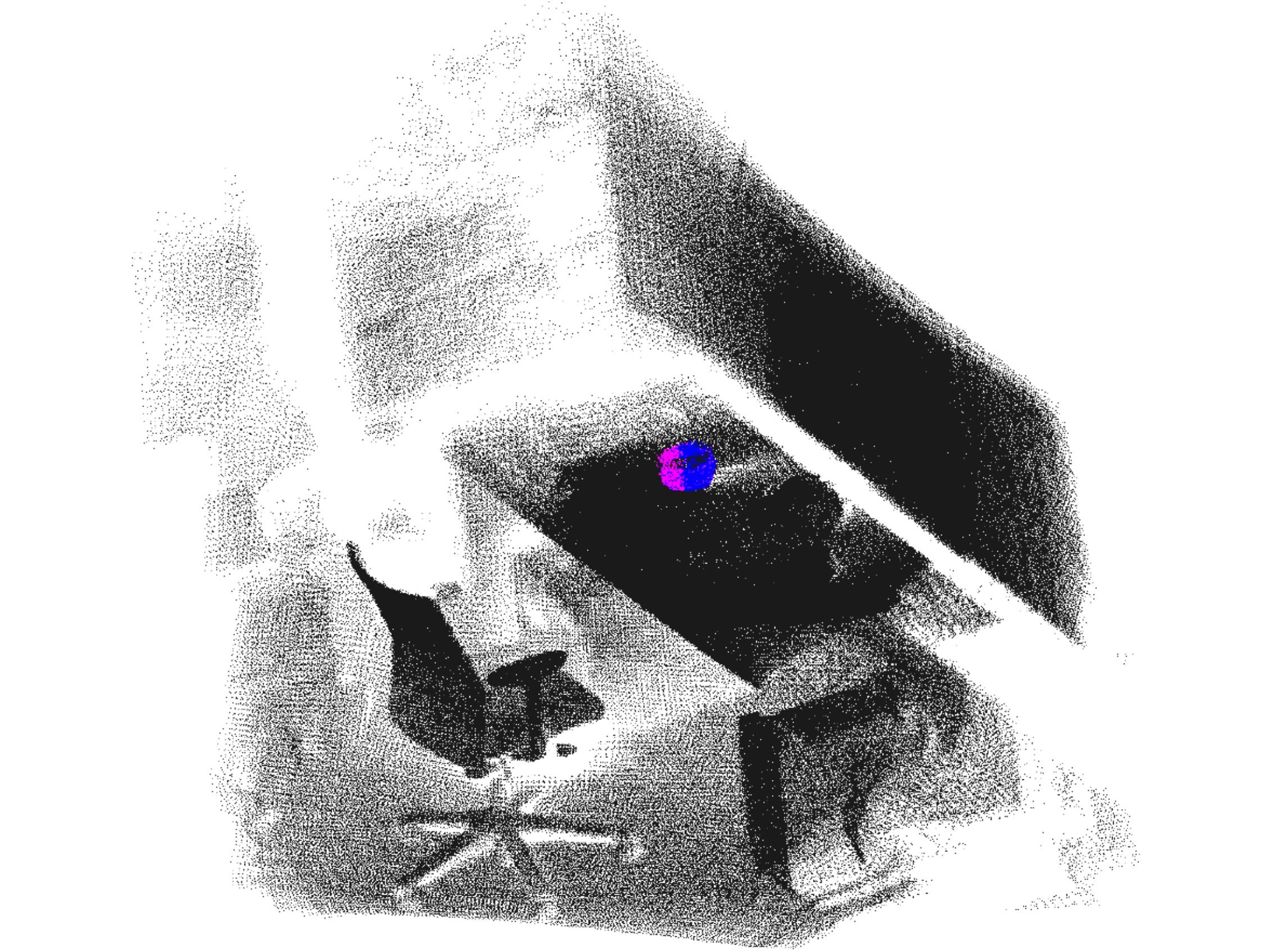}
\end{minipage}

&

\begin{minipage}[t]{0.19\linewidth}
\centering
\includegraphics[width=1\linewidth]{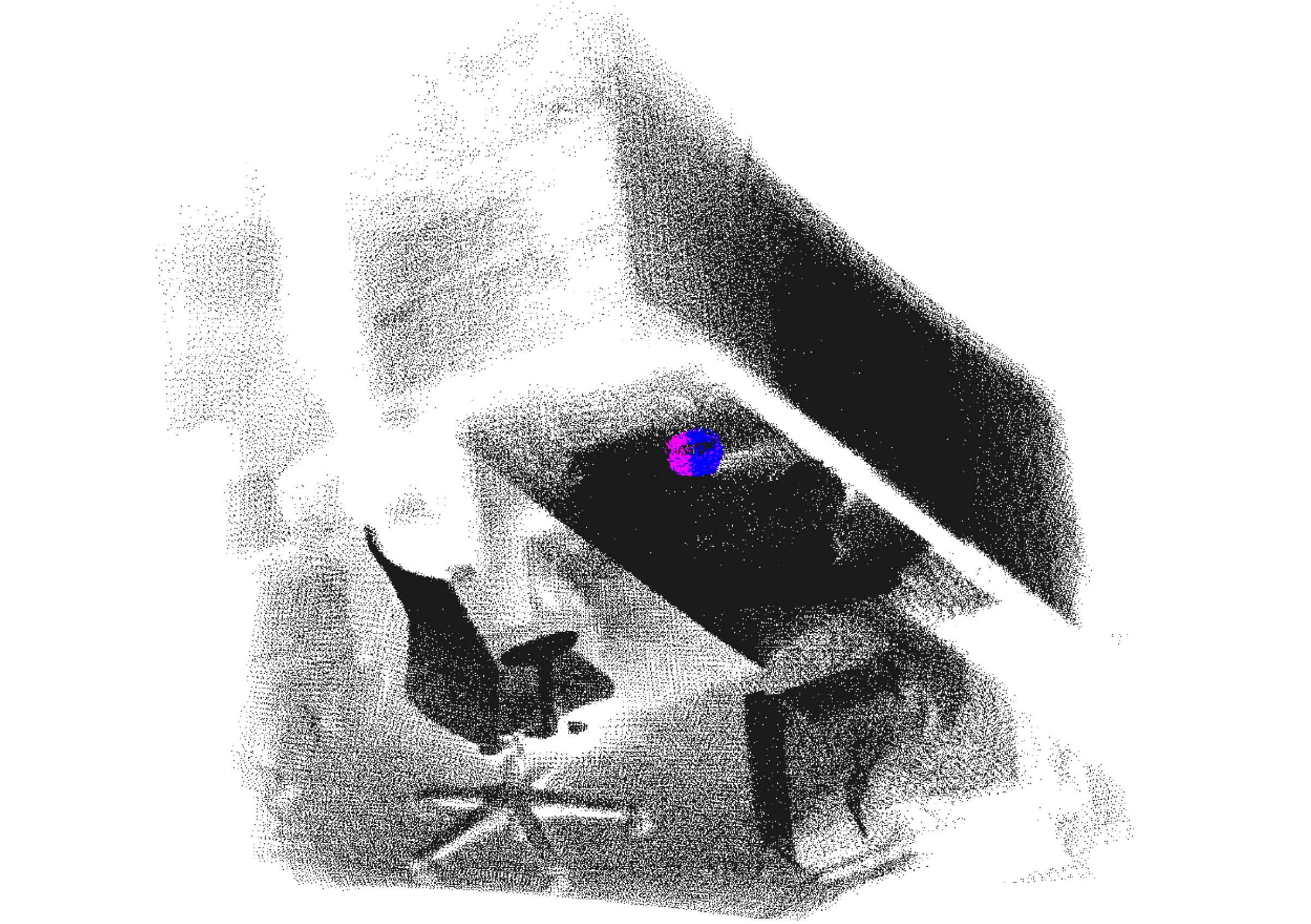}
\end{minipage}
\end{tabular}

\caption{Supplementary object localization and pose estimation results on the RGB-D scenes dataset~\cite{lai2011large}. The explicit organization of the results is similar to Fig.~\ref{obj-local1}. Best results are shown in \textbf{bold} font.}
\label{obj-local2}
\vspace{-3mm}
\end{figure*}

\subsection{Real Application: Object Localization}

We evaluate DANIEL in the real-world 3D object localization and pose estimation application. We adopt the large-scale RGB-D Scenes dataset~\cite{lai2011large}, where 3D object models (point clouds) in different shapes and different RGB-D scene scans are provided. For each pair of object and scene, we extract the point cloud of the object from the scene according to the ground-truth labels, and then change the pose of this object with a random transformation ($||\boldsymbol{t}||\leq 3$). Afterwards, we downsample the object and the scene and apply FPFH to build correspondences between the transformed object and the entire scene, which would easily trigger high outlier ratios. Then, we use one non-minimal solver: GNC-TLS, one RANSAC solver: FLO-RANSAC (with 10000 maximum iterations), one guaranteed outlier removal solver: GORE+RANSAC, and our solver DANIEL to estimate the relative pose (rigid transformation) between the object and scene, where noise is constantly set to $\sigma=0.001m$.

We conduct 10 tests over 10 scenes with 5 objects, including the \textit{coffee mug, cereal box, cap, soda can} and \textit{bowl}, where all these tests are with extreme outlier ratios (ranging from 95.16\% to 97.94\%). We report the qualitative and quantitative object localization results in Fig.~\ref{obj-local1} and~\ref{obj-local2}, where each figure shows the results of 5 tests in 5 scenes (with possibly different objects). We can see that in such a high-outlier environment, GNC-TLS fails to render reasonable results in all tests, while FLO-RANSAC, mostly running in tens of seconds, could only succeed in 40\% tests. GORE+RANSAC and our DANIEL are the two solvers that can successfully localize all the objects (to be specific, estimate the correct transformation and reproject the object back to the scene) in all the 10 tests. More importantly, DANIEL runs only in tens or hundreds of milliseconds and is significantly faster than GORE+RANSAC, showing the most outstanding performance.

\subsection{Scan Matching}

Scan matching, or called scene stitching, is another crucial application of point cloud registration, underlying the 3D reconstruction and SLAM technologies. We test DANIEL for scan matching based on the Microsoft 7-scenes dataset~\cite{shotton2013scene} that provides realistic RGB images with associated depth images. We deliberately select 10 pairs of scans with overlapping scenes from the \textit{red kitchen, office, chess, stairs} and \textit{heads}. Since FPFH may generate too many outliers on RGB-D data, 

\clearpage

\begin{figure*}[h]
\centering
\setlength\tabcolsep{0pt}
\addtolength{\tabcolsep}{-0.2pt}
\begin{tabular}{c|cc|c|c|c|c}
\quad &\,\footnotesize{Correspondences}\, &\,&  \footnotesize{GNC-TLS} & \footnotesize{FLO-RANSAC} & \footnotesize{GORE+RANSAC} & \footnotesize{DANIEL}
\\
\hline

 & \footnotesize{$N=1840$} && \footnotesize{\textcolor[rgb]{1,0,0}{Fail}$,\verb|\|,0.245s$} &\footnotesize{\textcolor[rgb]{0,0.8,0}{Succeed}$,0.308,8.960s$}&\footnotesize{\textcolor[rgb]{0,0.8,0}{Succeed}$,\textbf{0.283},11.727s$}&\footnotesize{\textcolor[rgb]{0,0.8,0}{Succeed}$,0.299,\textbf{0.182}s$}

\\

\rotatebox{90}{\,\,\footnotesize{\textit{red kitchen}}\,}\,
&
\,\,
\begin{minipage}[t]{0.1\linewidth}
\centering
\includegraphics[width=1\linewidth]{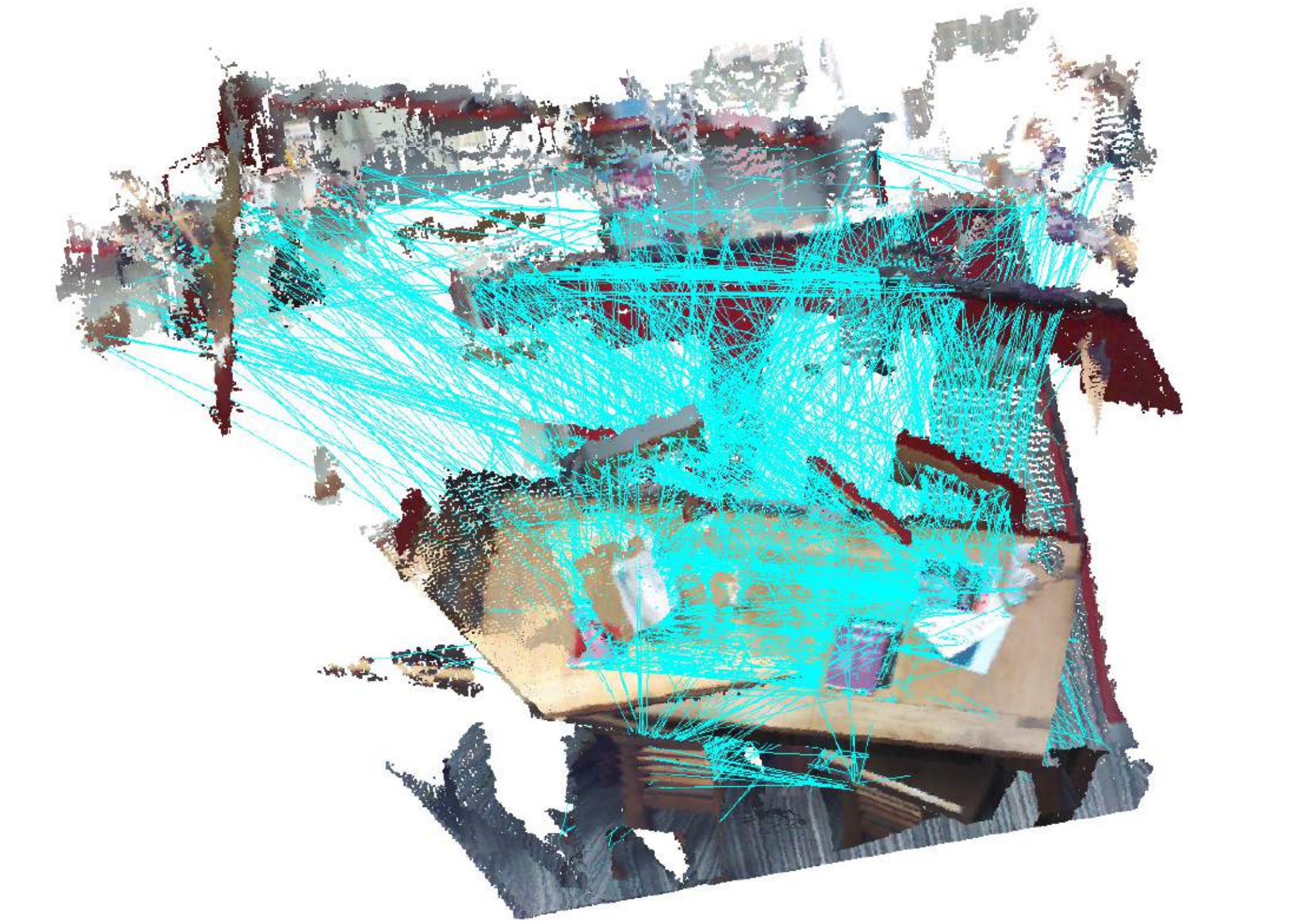}
\end{minipage}\,\,
& &
\,\,
\begin{minipage}[t]{0.19\linewidth}
\centering
\includegraphics[width=.48\linewidth]{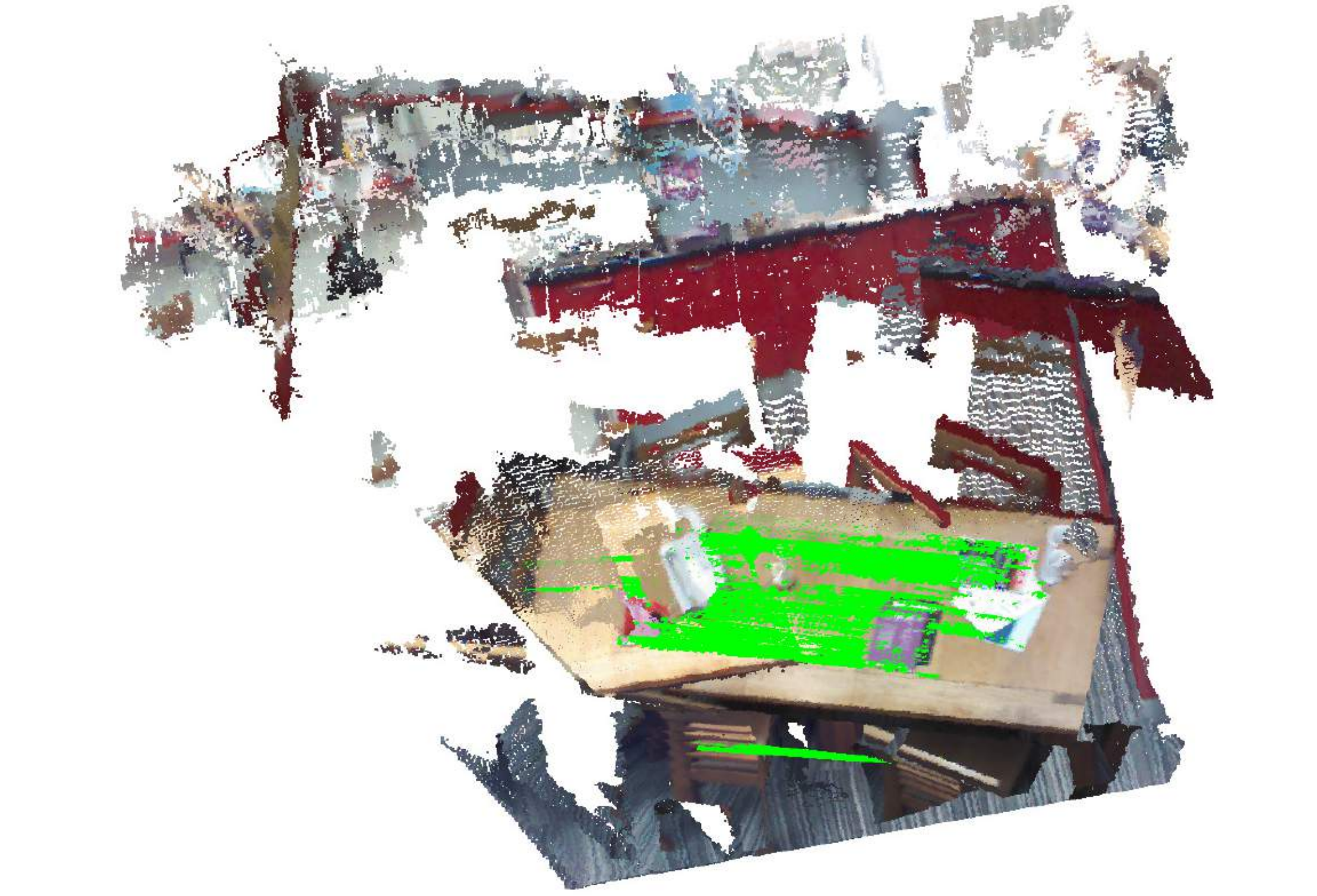}
\includegraphics[width=.48\linewidth]{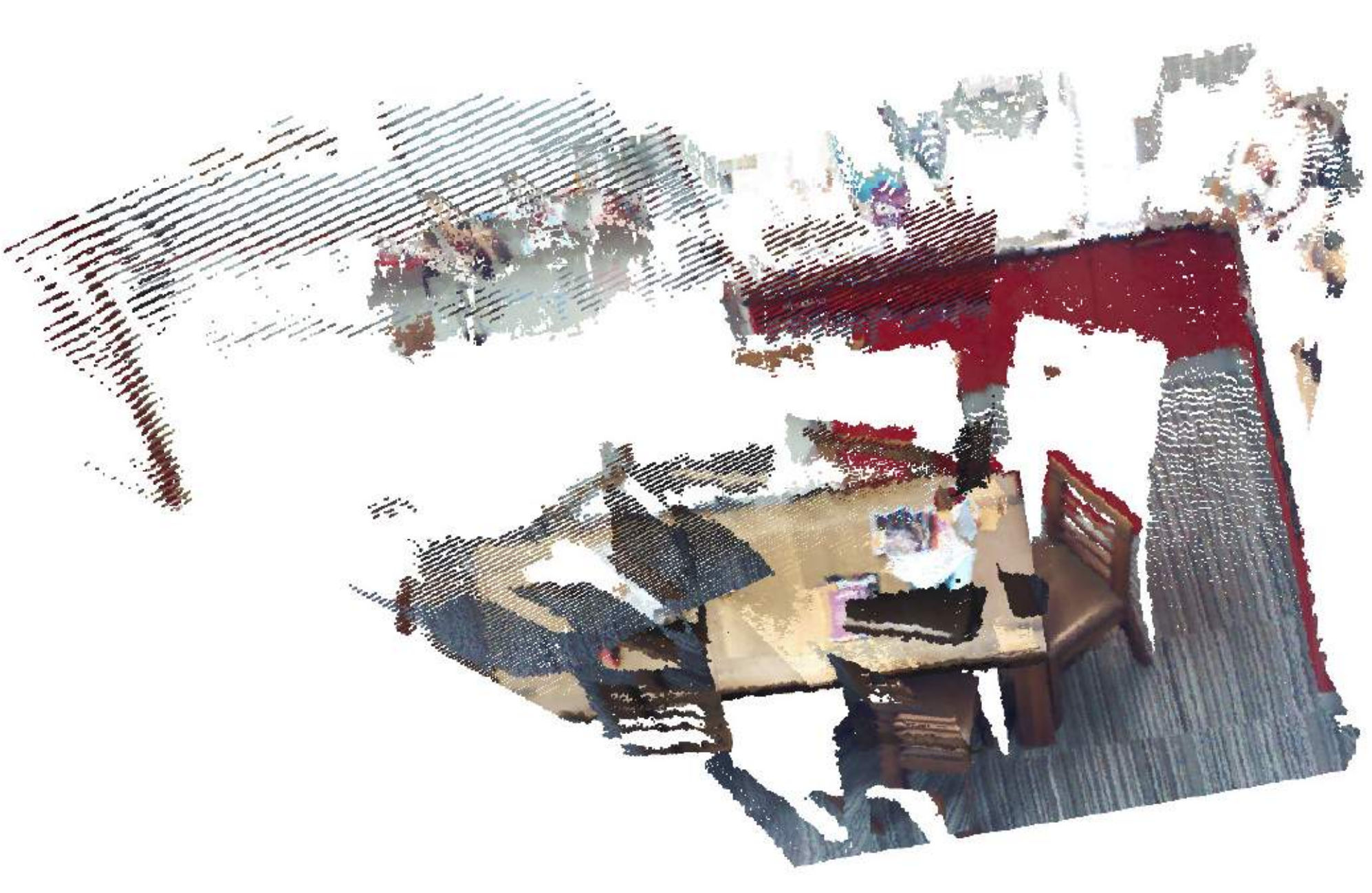}
\end{minipage}\,\,
&
\,\,
\begin{minipage}[t]{0.19\linewidth}
\centering
\includegraphics[width=.48\linewidth]{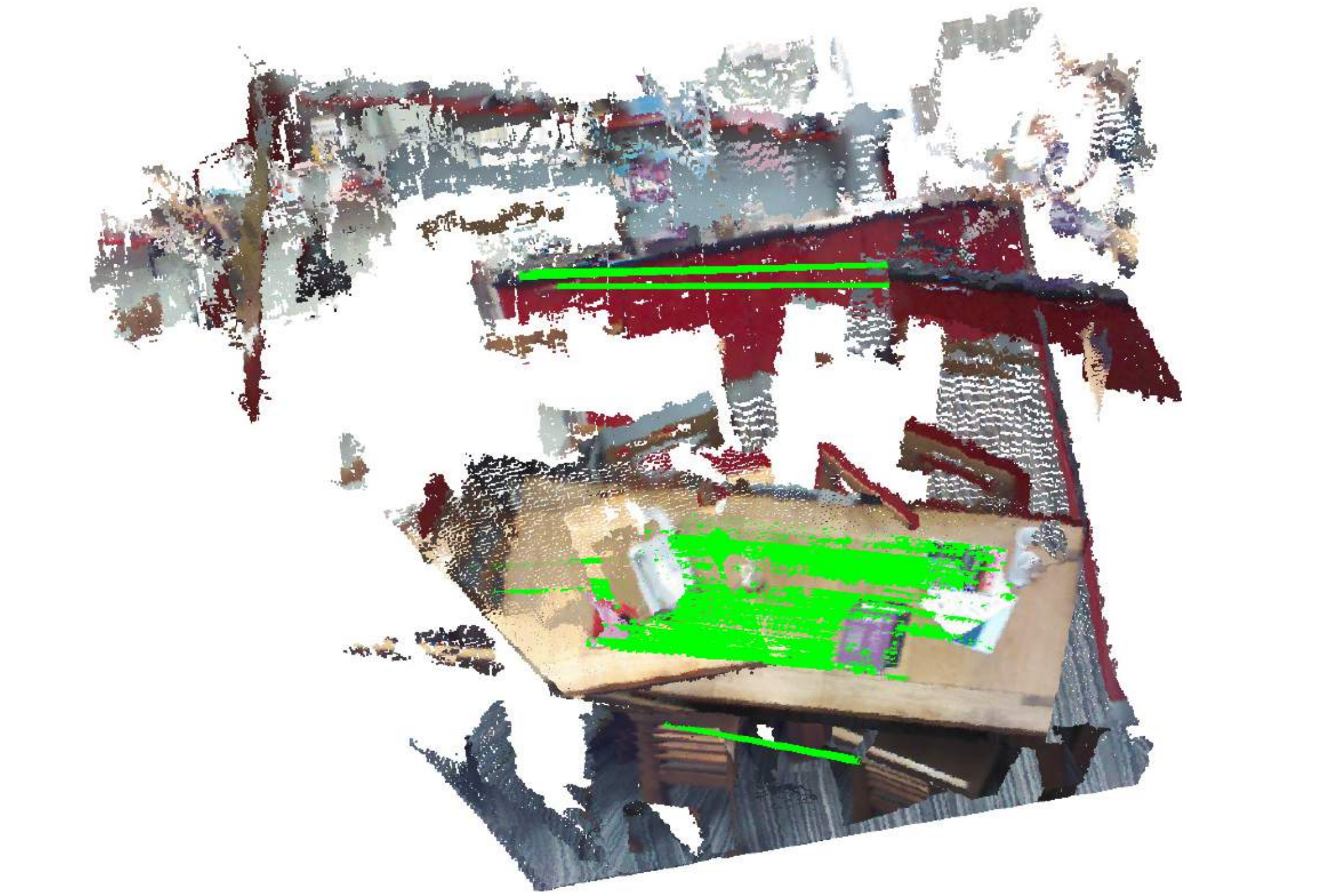}
\includegraphics[width=.48\linewidth]{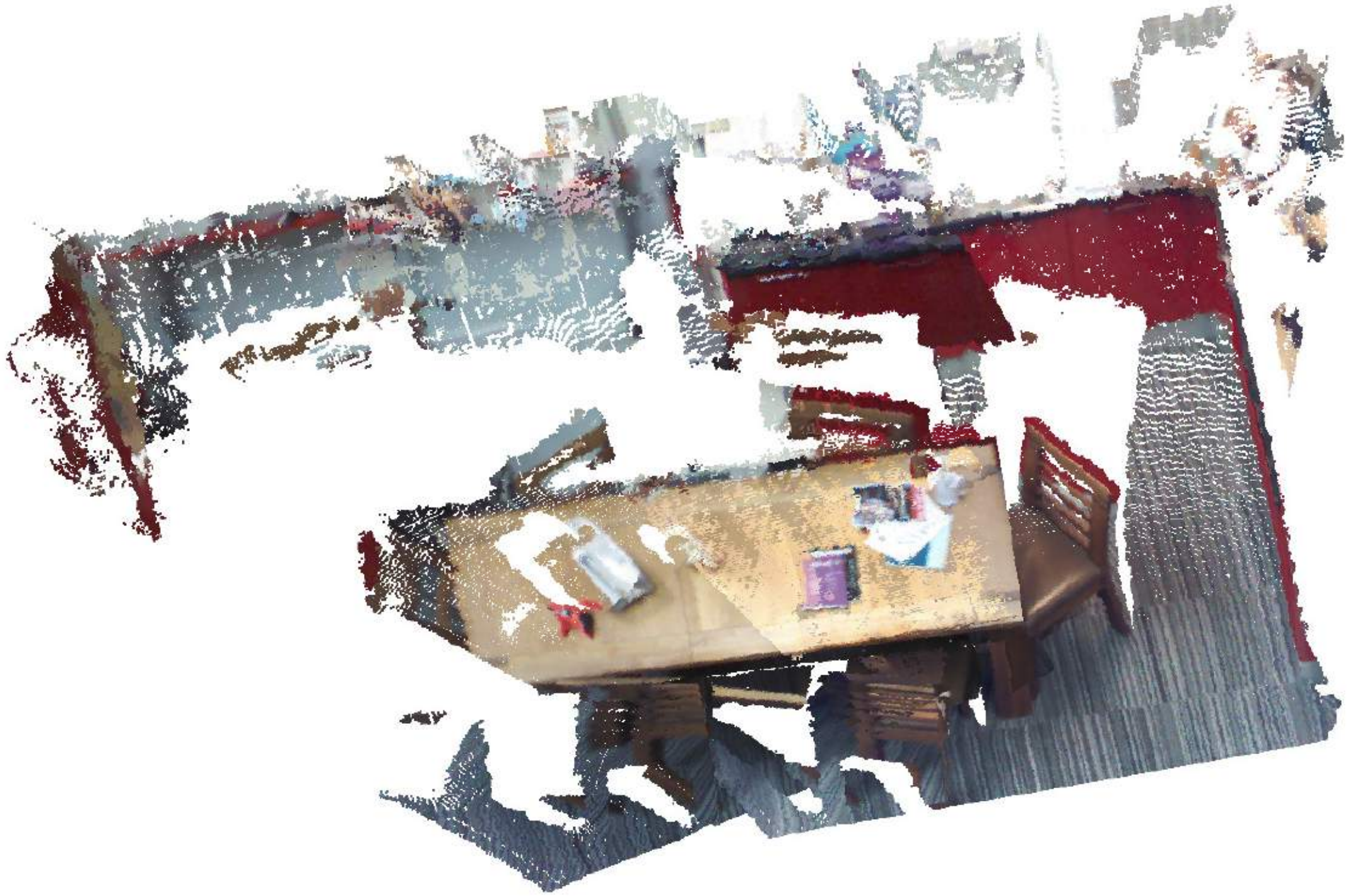}
\end{minipage}\,\,
&
\,\,
\begin{minipage}[t]{0.19\linewidth}
\centering
\includegraphics[width=.48\linewidth]{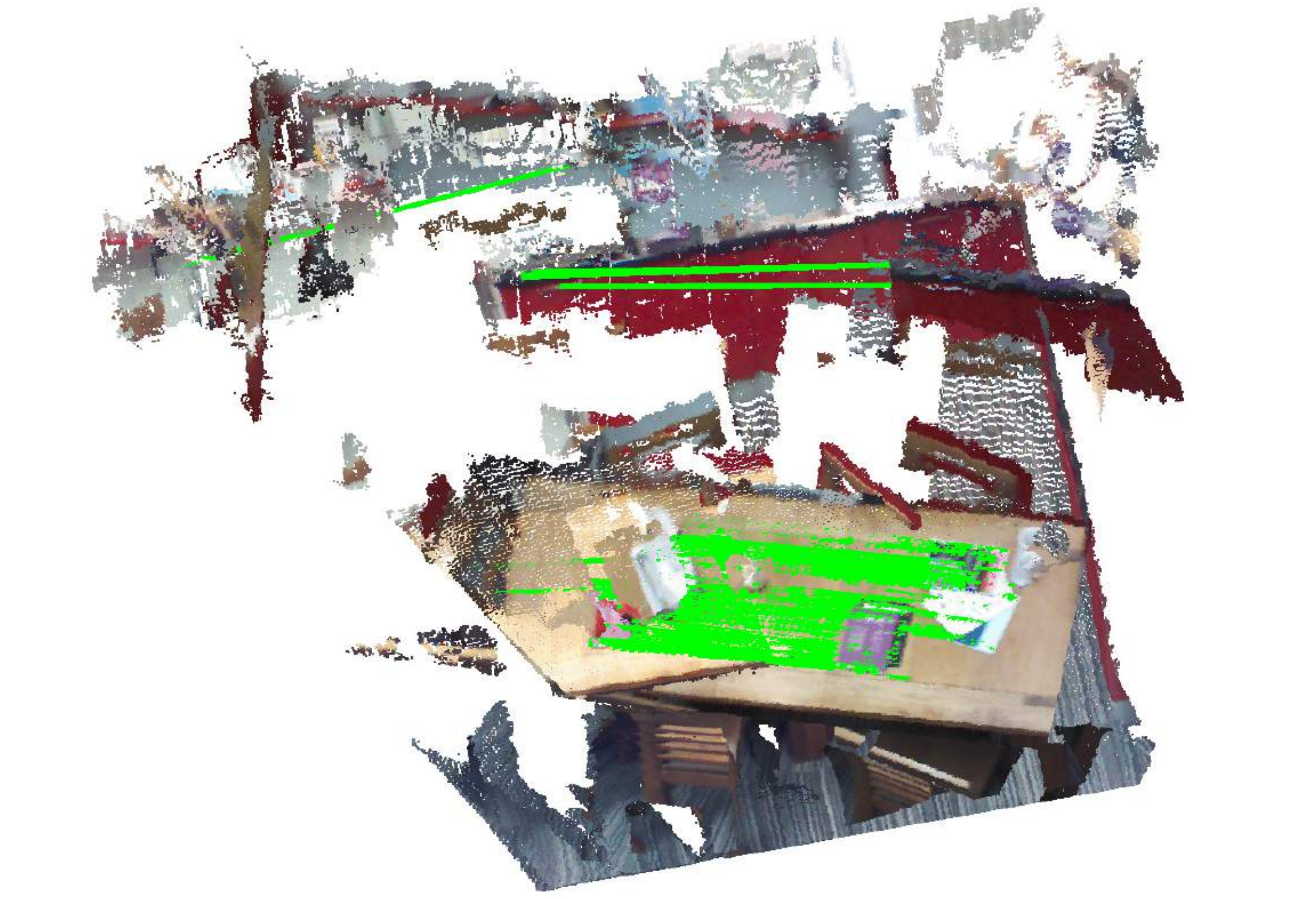}
\includegraphics[width=.48\linewidth]{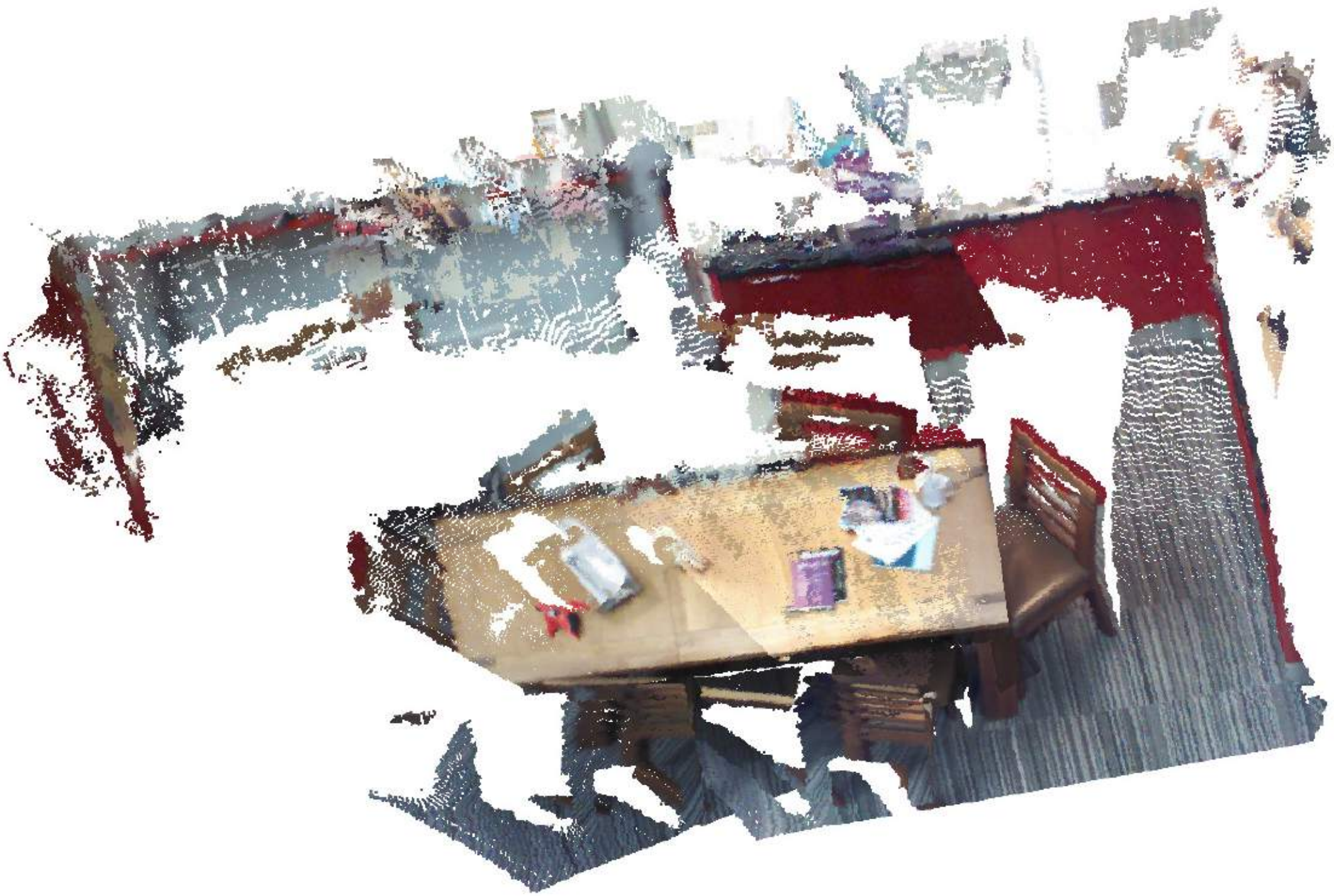}
\end{minipage}\,\,
&
\,\,
\begin{minipage}[t]{0.19\linewidth}
\centering
\includegraphics[width=.48\linewidth]{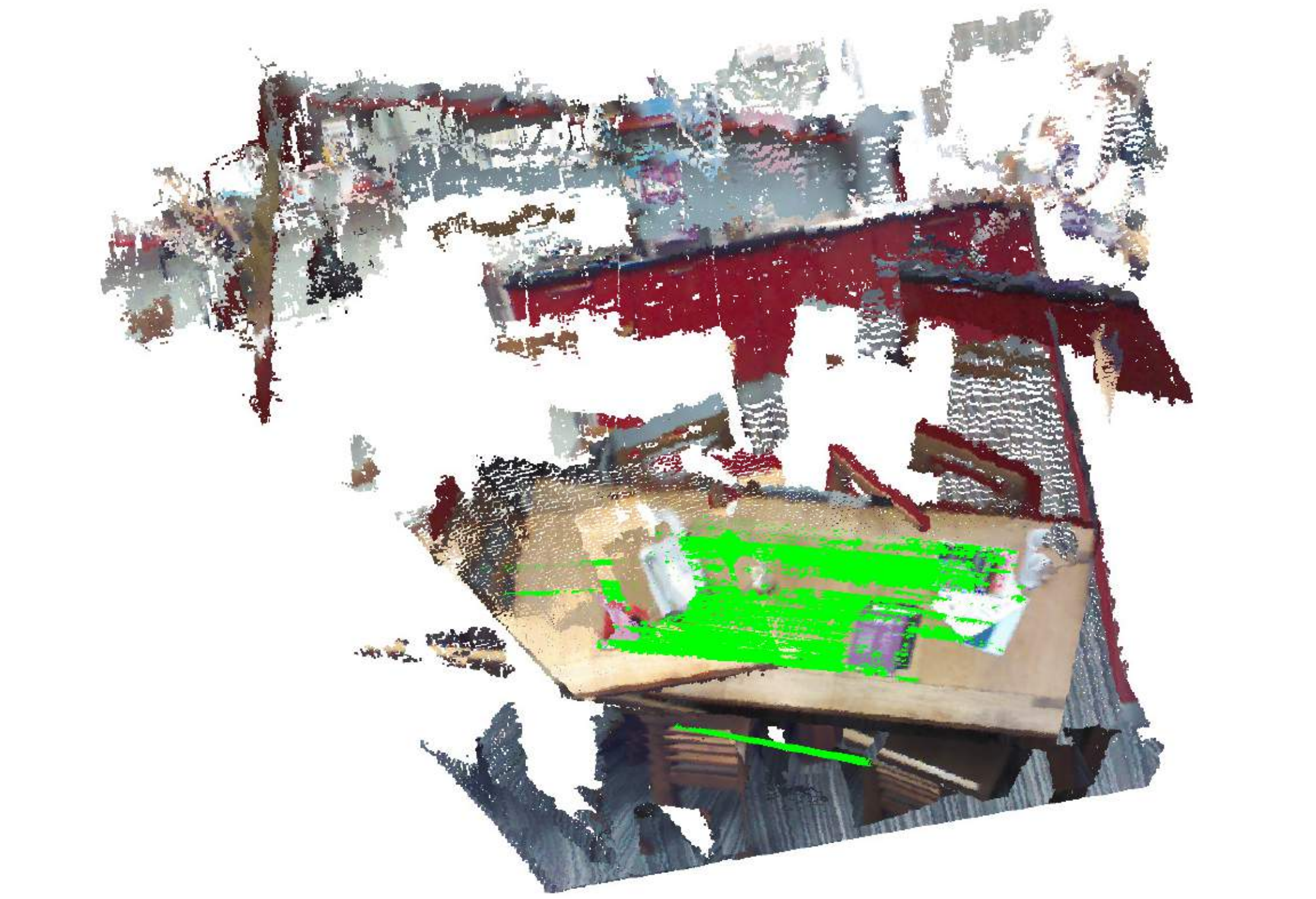}
\includegraphics[width=.48\linewidth]{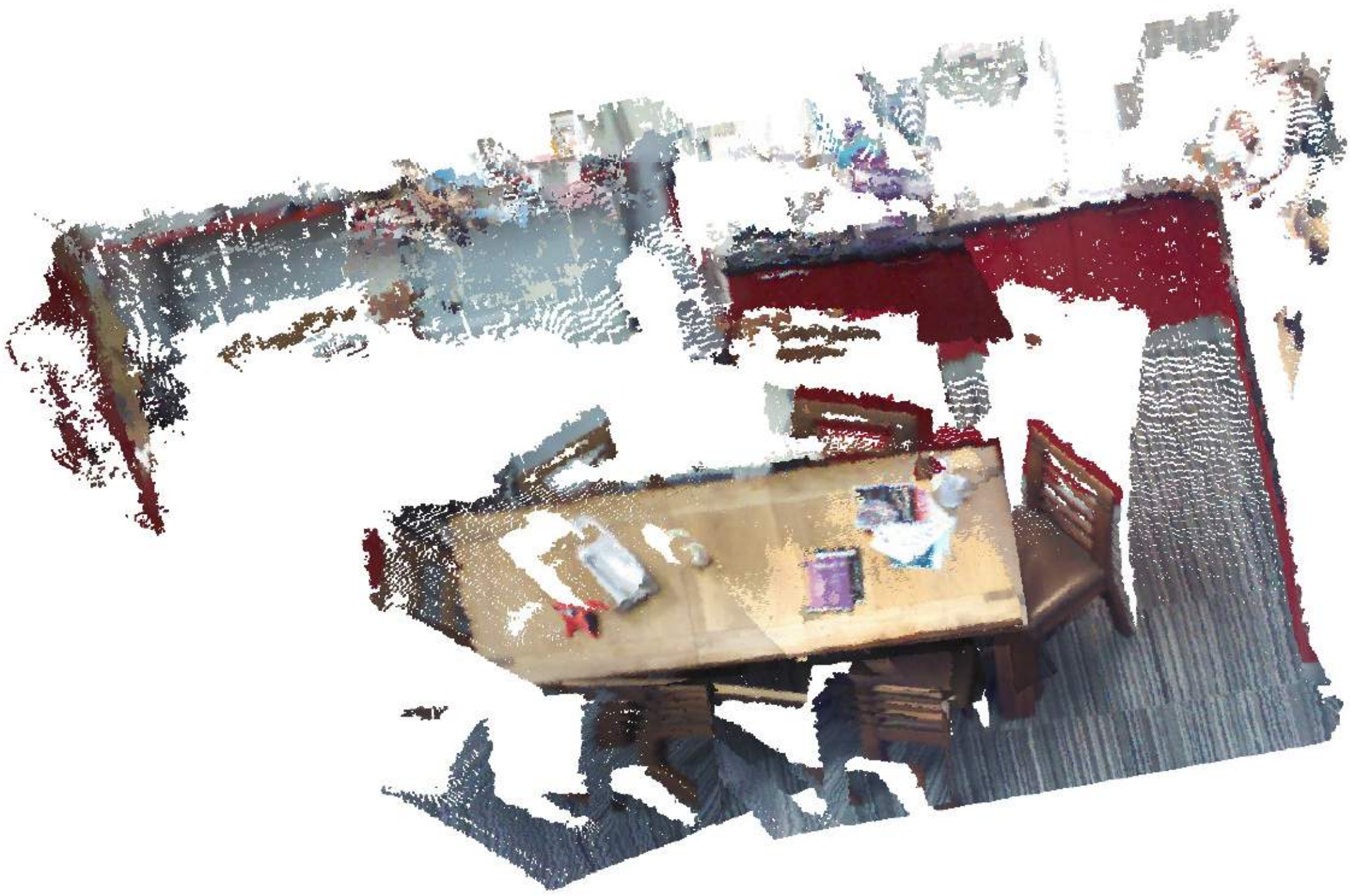}
\end{minipage}\,\,

\\

 & \footnotesize{$N=1273$} && \footnotesize{\textcolor[rgb]{1,0,0}{Fail}$,\verb|\|,0.118s$} &\footnotesize{\textcolor[rgb]{0,0.8,0}{Succeed}$,0.352,34.662s$}&\footnotesize{\textcolor[rgb]{0,0.8,0}{Succeed}$,0.343,8.657s$}&\footnotesize{\textcolor[rgb]{0,0.8,0}{Succeed}$,\textbf{0.335},\textbf{1.190}s$}

\\

\rotatebox{90}{\,\,\footnotesize{\textit{red kitchen}}\,}\,
&
\,\,
\begin{minipage}[t]{0.1\linewidth}
\centering
\includegraphics[width=1\linewidth]{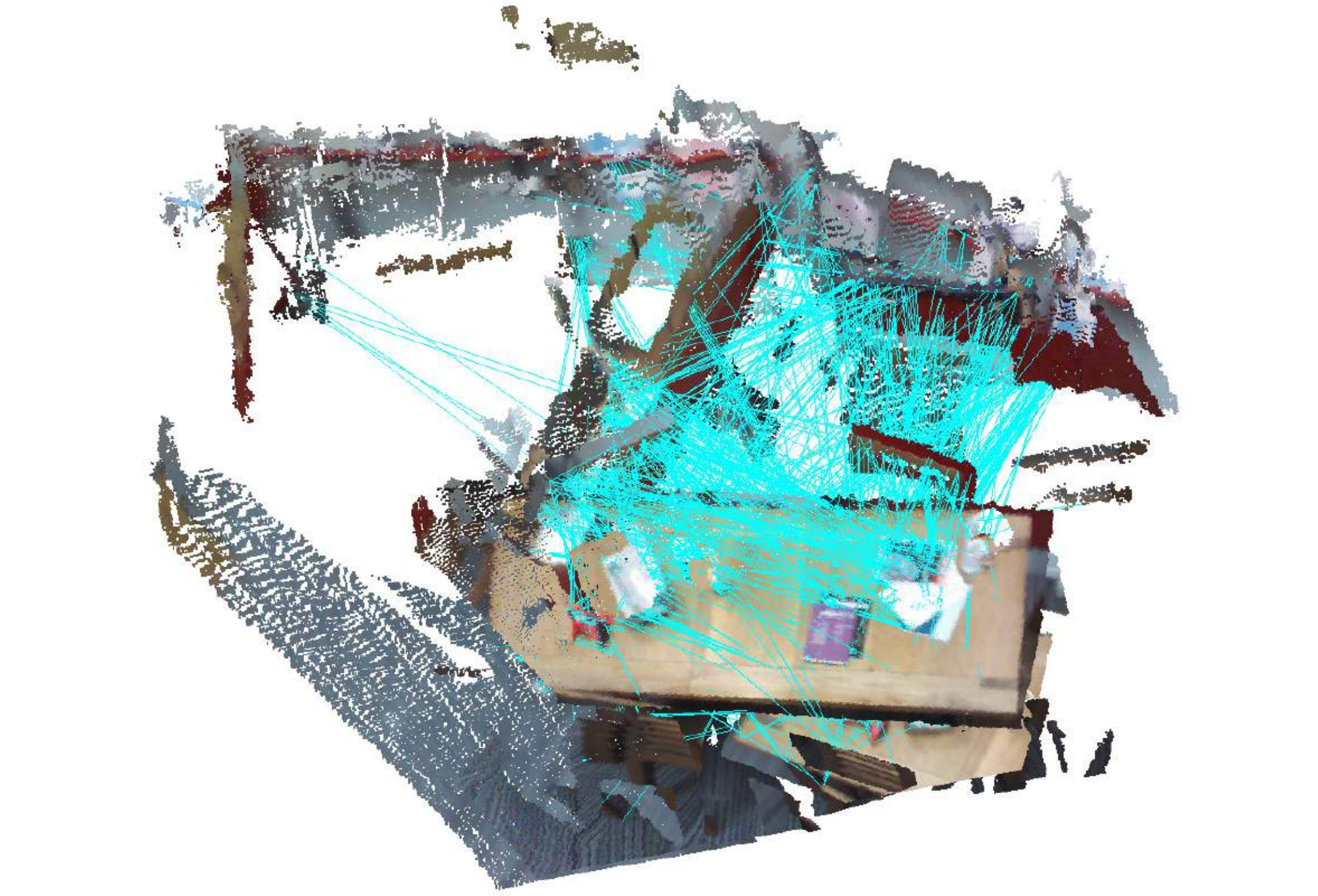}
\end{minipage}\,\,
& &
\,\,
\begin{minipage}[t]{0.19\linewidth}
\centering
\includegraphics[width=.48\linewidth]{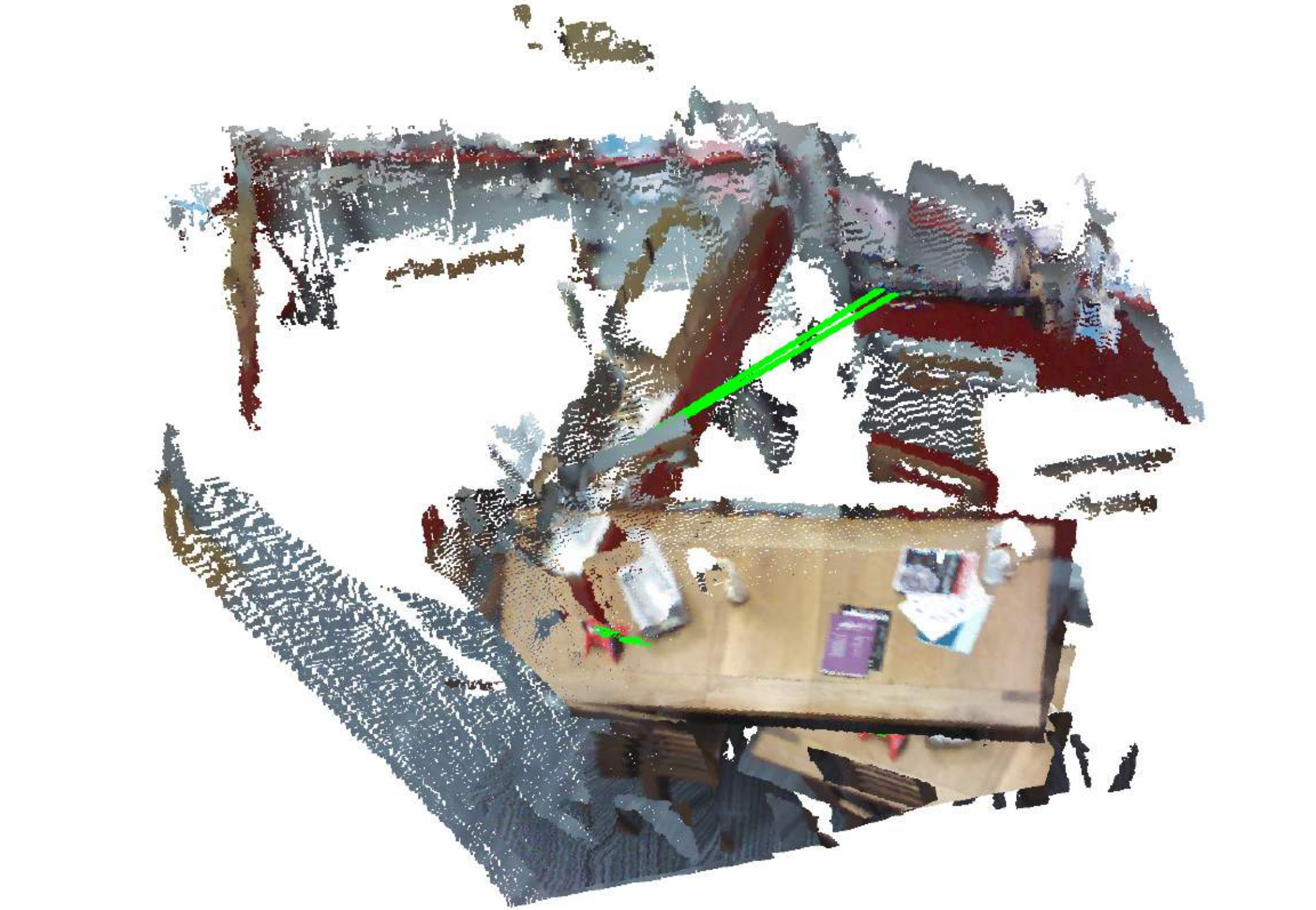}
\includegraphics[width=.48\linewidth]{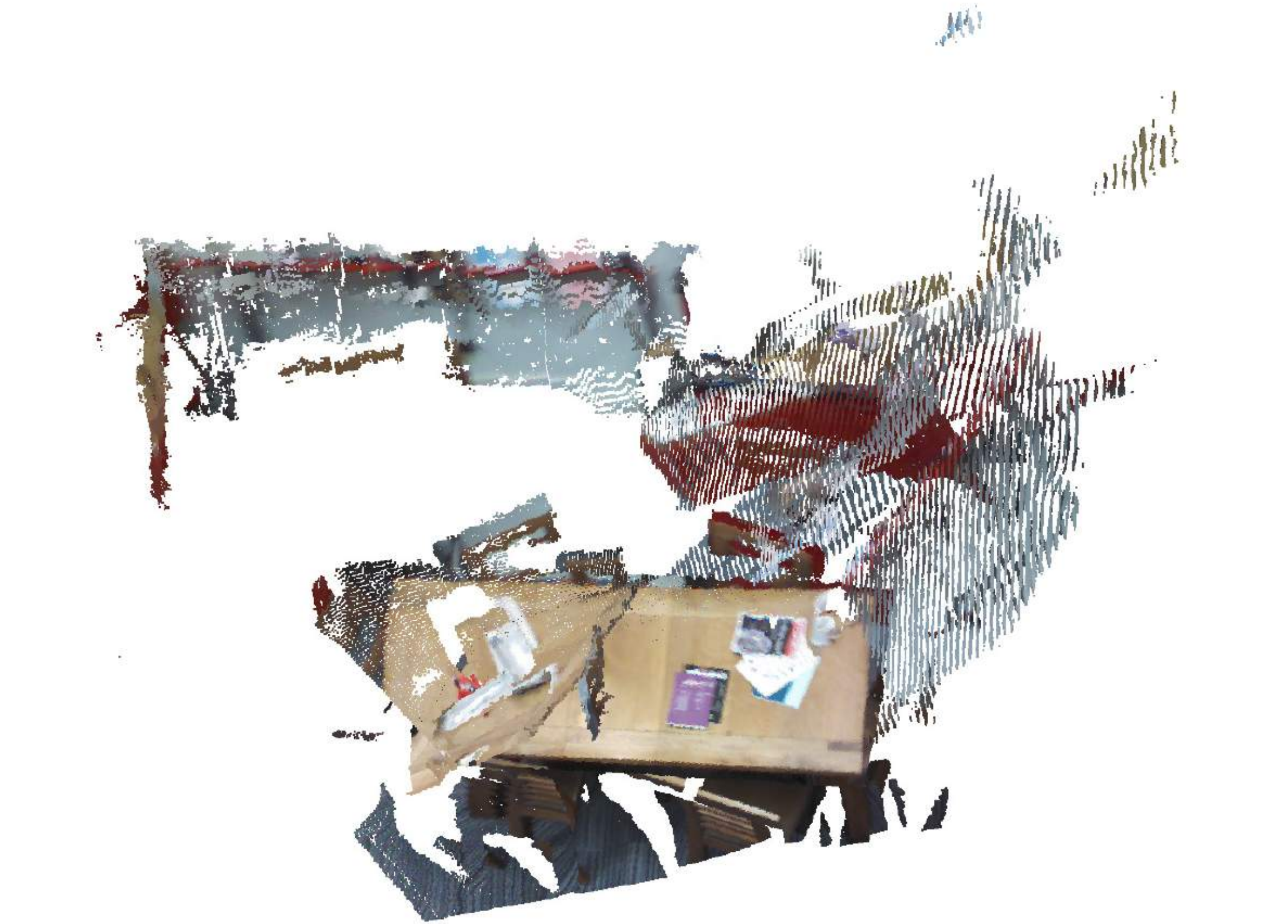}
\end{minipage}\,\,
&
\,\,
\begin{minipage}[t]{0.19\linewidth}
\centering
\includegraphics[width=.48\linewidth]{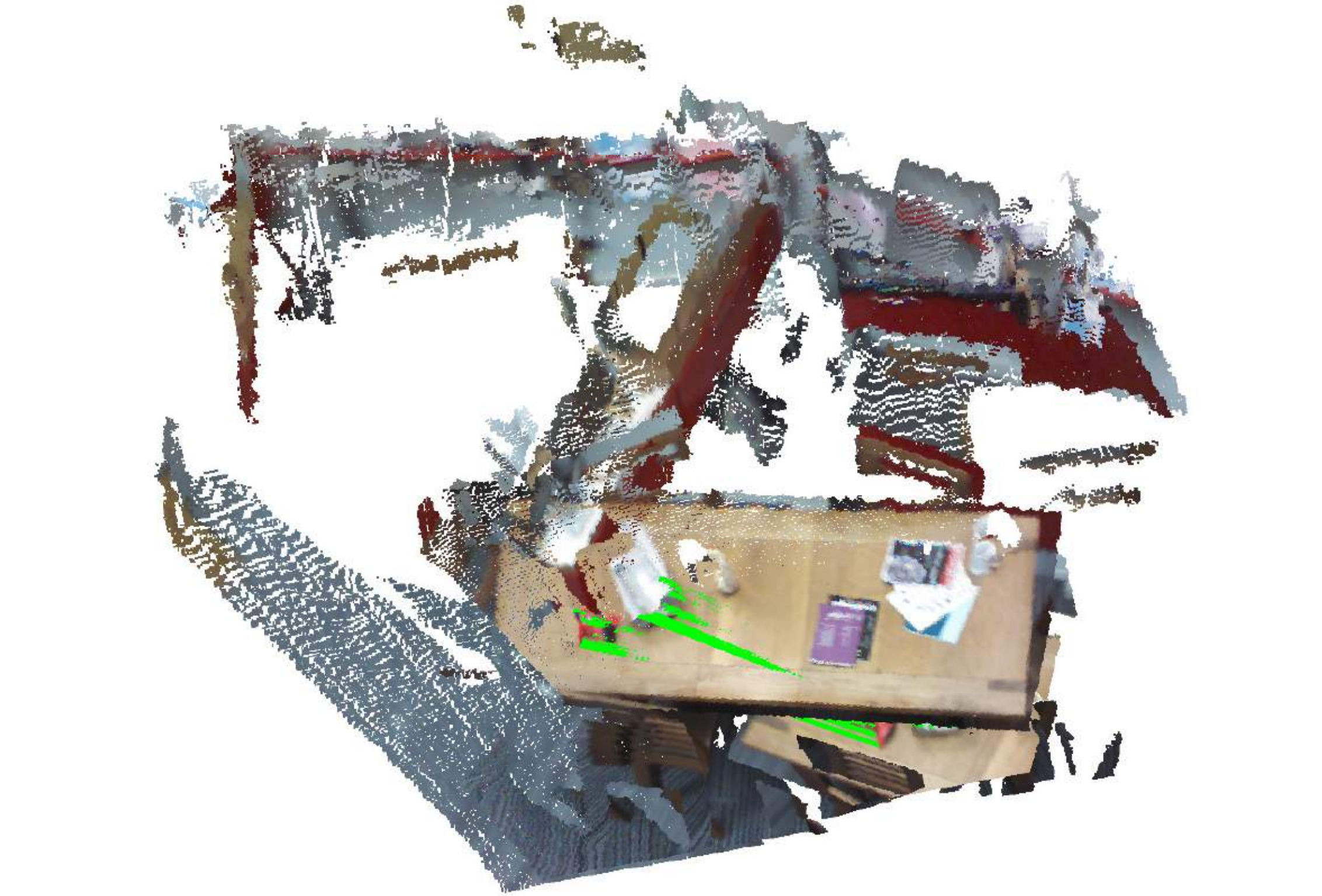}
\includegraphics[width=.48\linewidth]{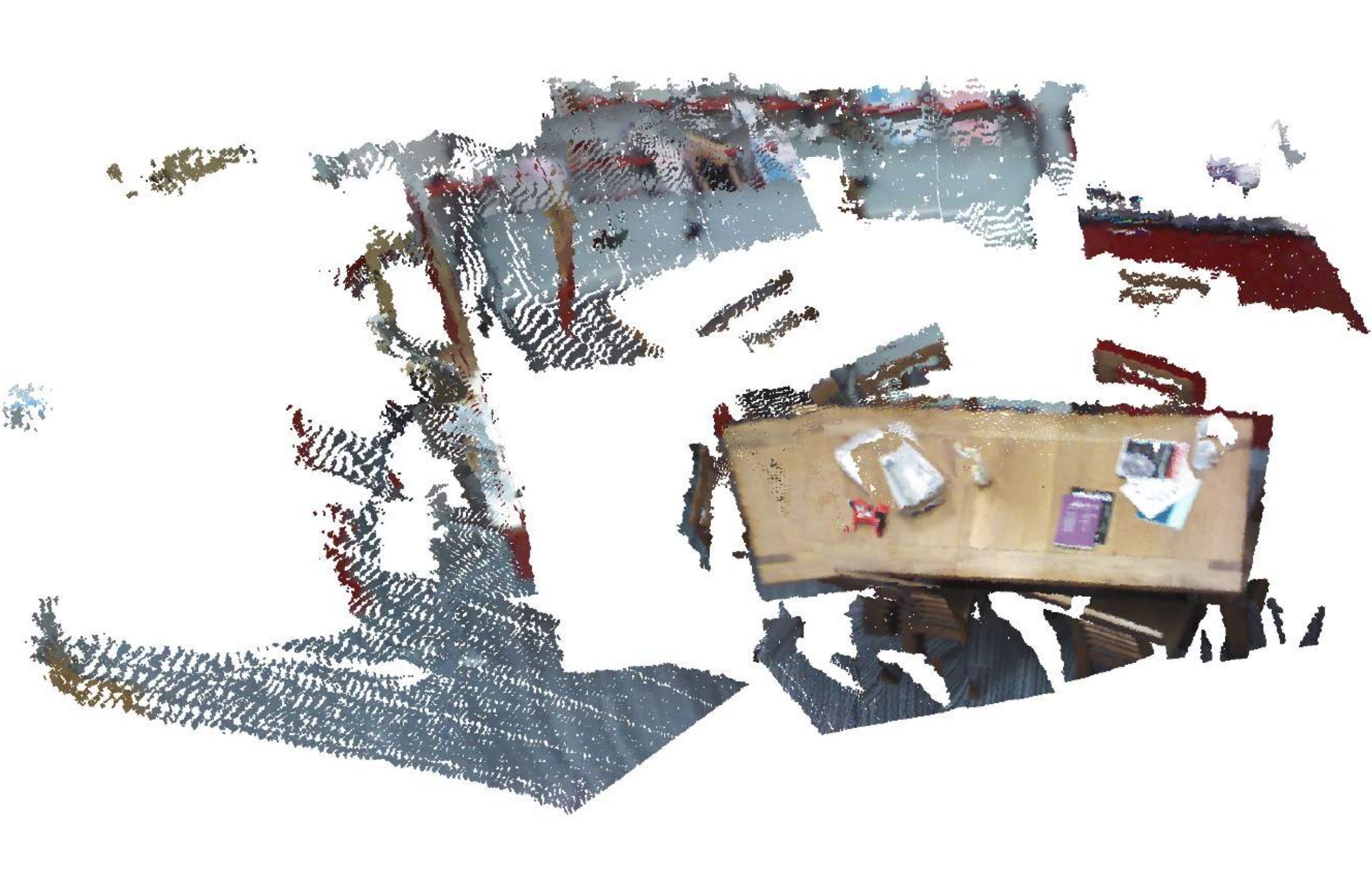}
\end{minipage}\,\,
&
\,\,
\begin{minipage}[t]{0.19\linewidth}
\centering
\includegraphics[width=.48\linewidth]{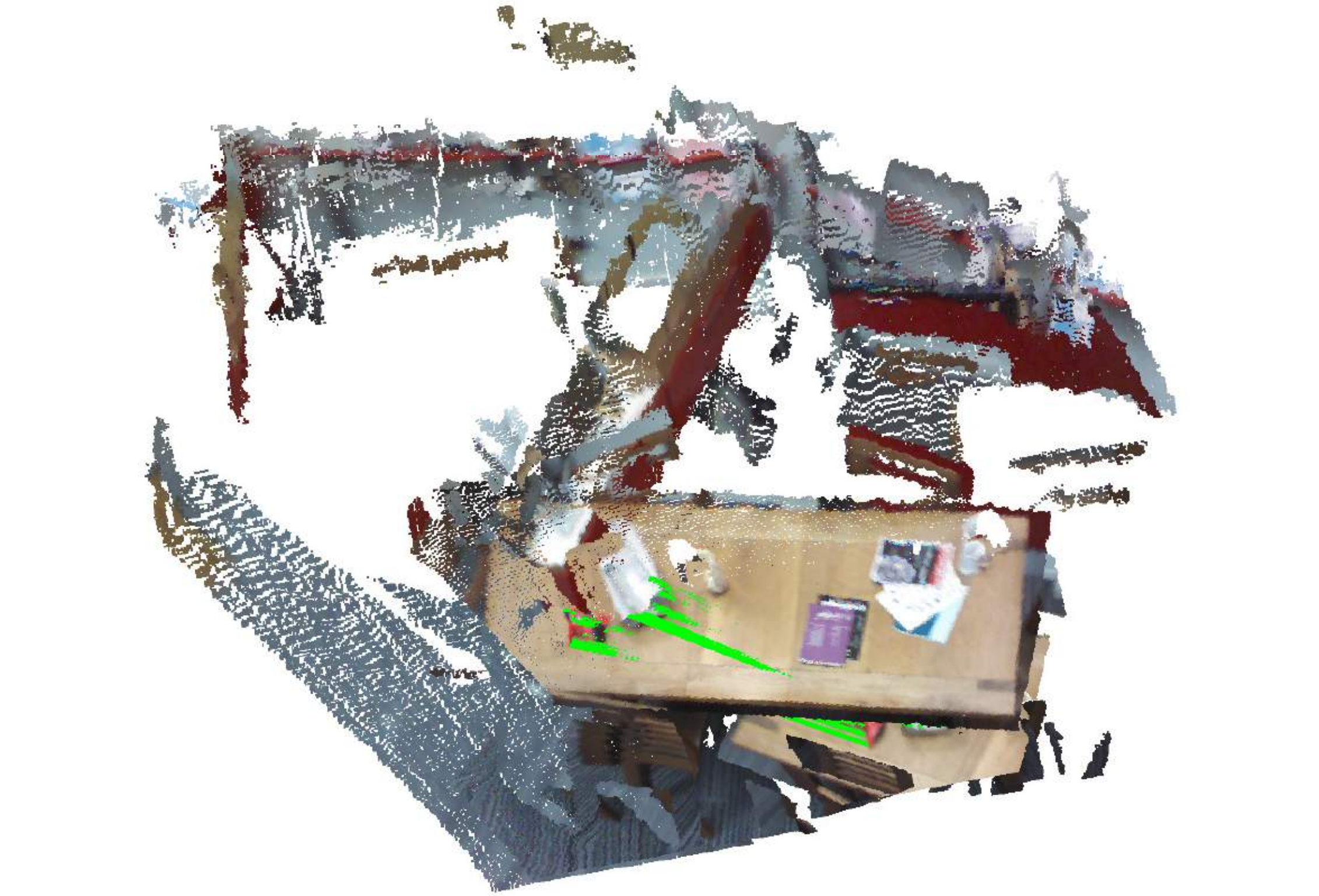}
\includegraphics[width=.48\linewidth]{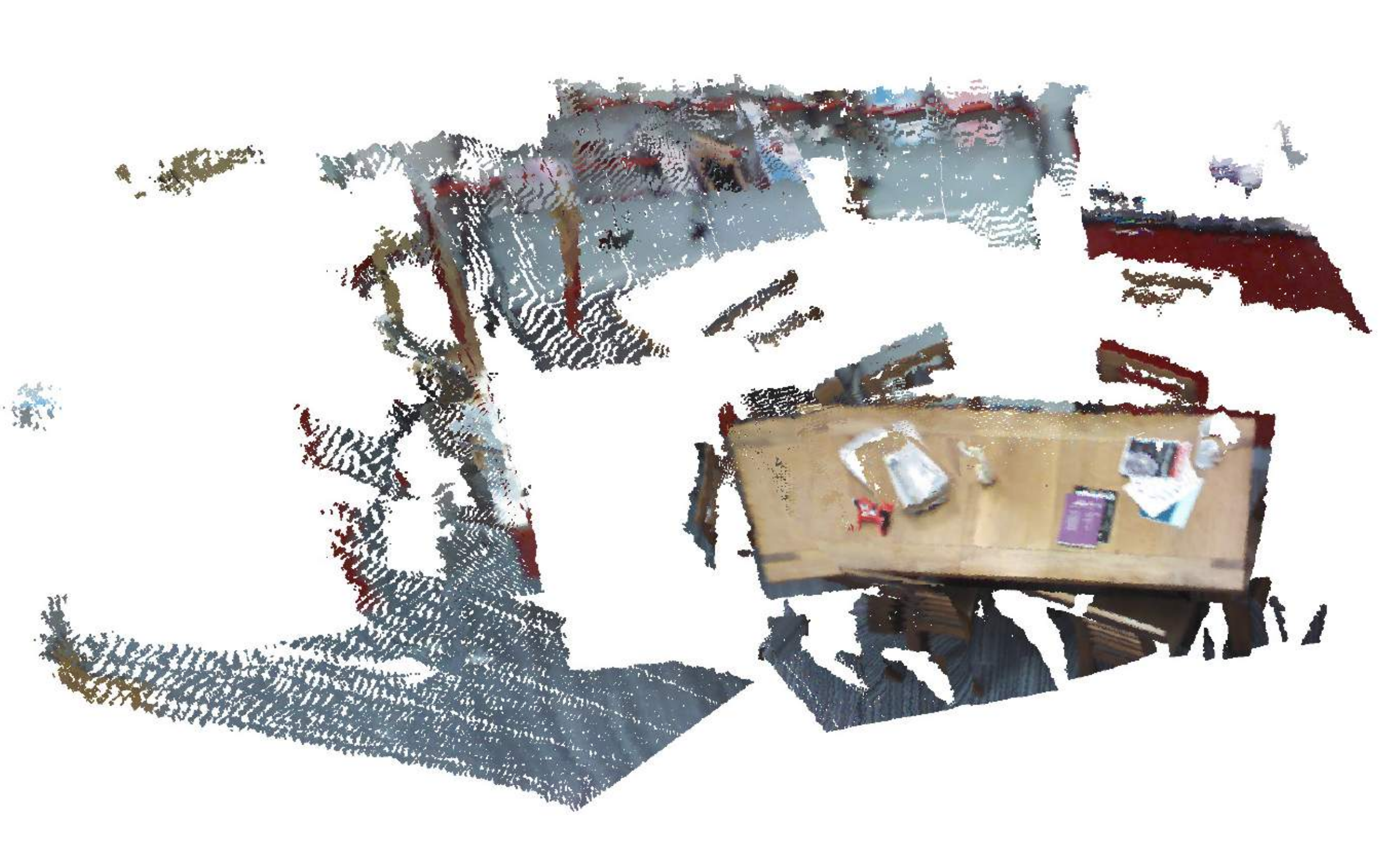}
\end{minipage}\,\,
&
\,\,
\begin{minipage}[t]{0.19\linewidth}
\centering
\includegraphics[width=.48\linewidth]{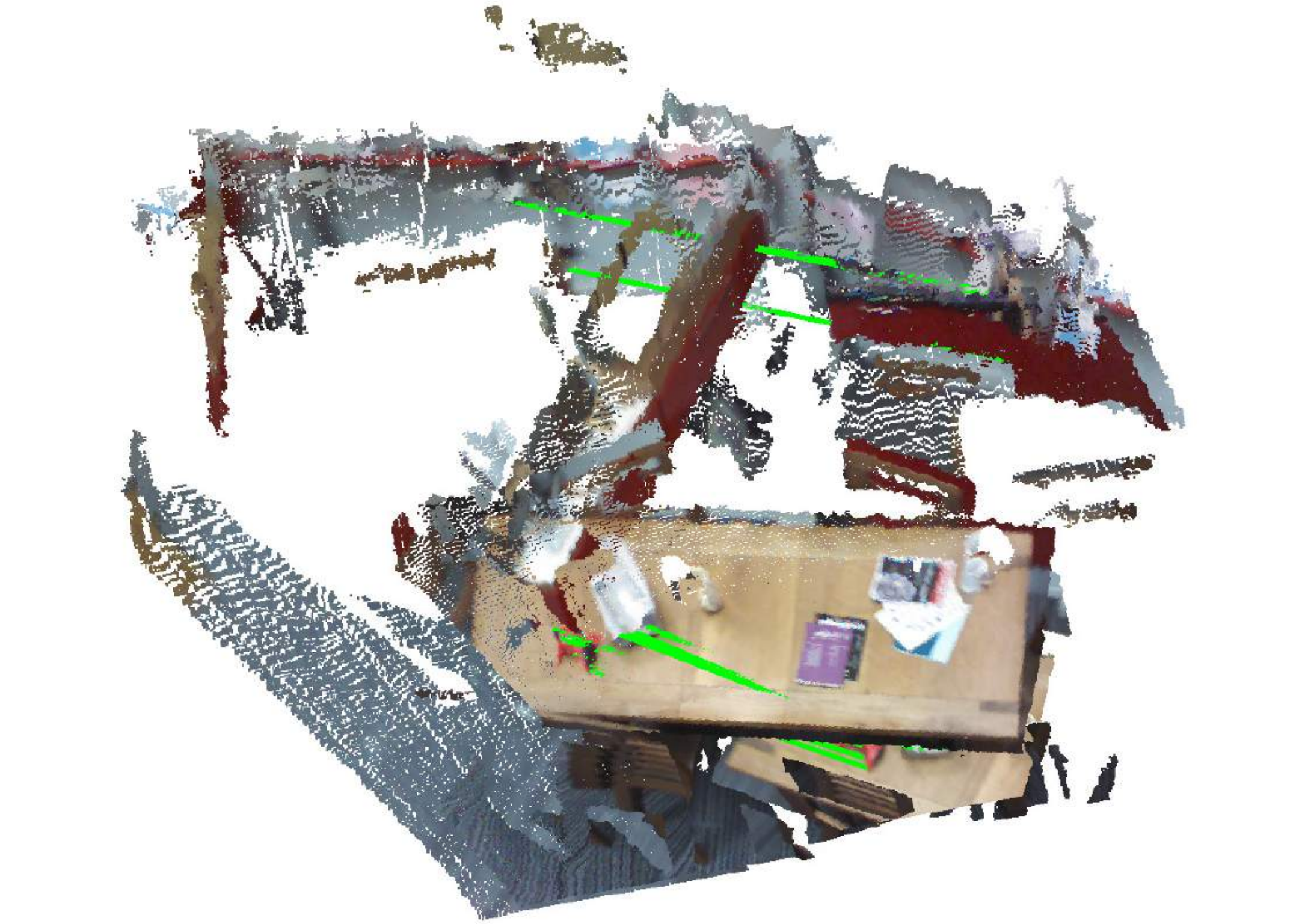}
\includegraphics[width=.48\linewidth]{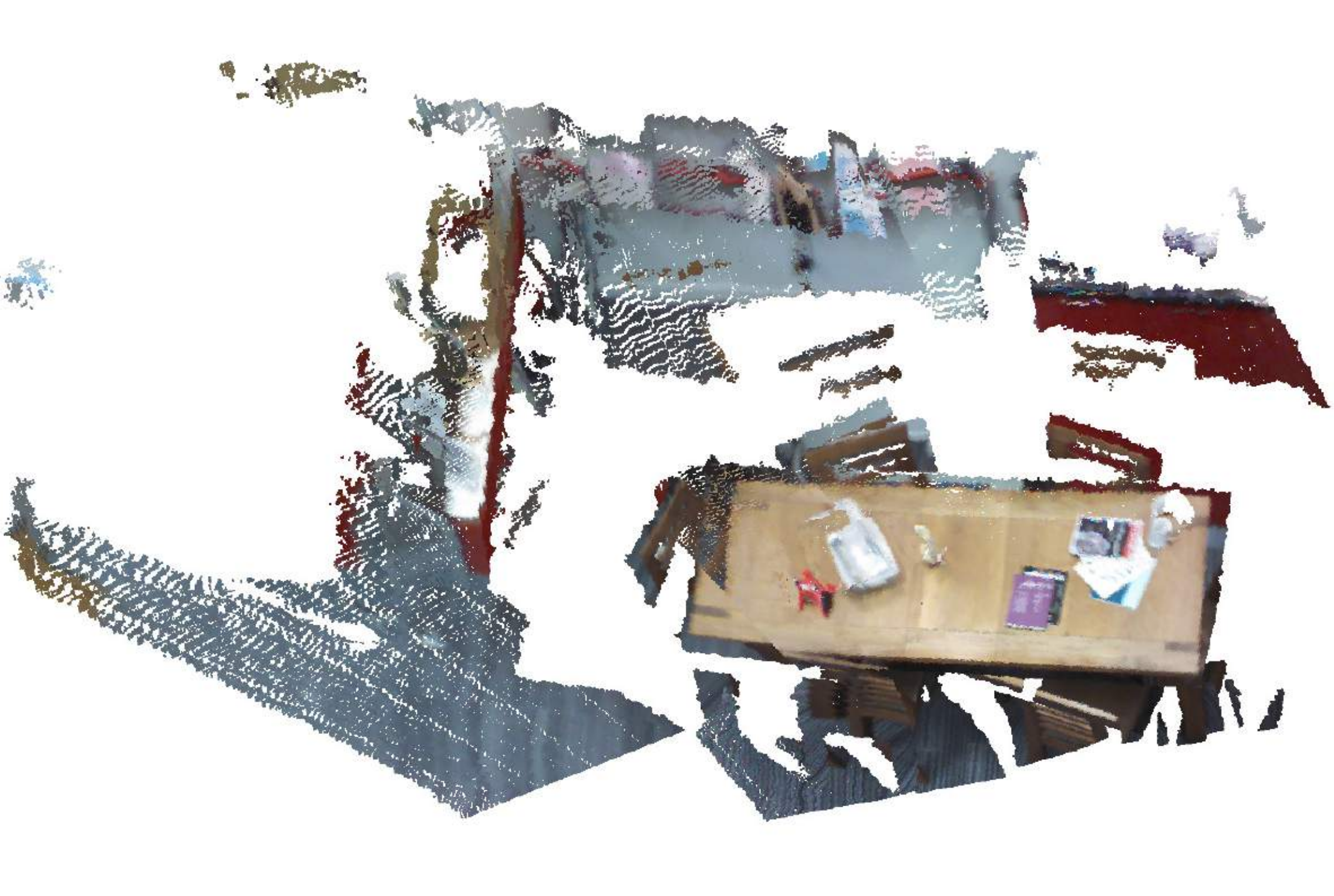}
\end{minipage}\,\,

\\

 & \footnotesize{$N=900$} && \footnotesize{\textcolor[rgb]{1,0,0}{Fail}$,\verb|\|,0.093s$} &\footnotesize{\textcolor[rgb]{0,0.8,0}{Succeed}$,0.438,12.606s$}&\footnotesize{\textcolor[rgb]{0,0.8,0}{Succeed}$,0.335,7.440s$}&\footnotesize{\textcolor[rgb]{0,0.8,0}{Succeed}$,\textbf{0.314},\textbf{0.177}s$}

\\
\rotatebox{90}{\,\,\footnotesize{\textit{red kitchen}}\,}\,
&
\,\,
\begin{minipage}[t]{0.1\linewidth}
\centering
\includegraphics[width=1\linewidth]{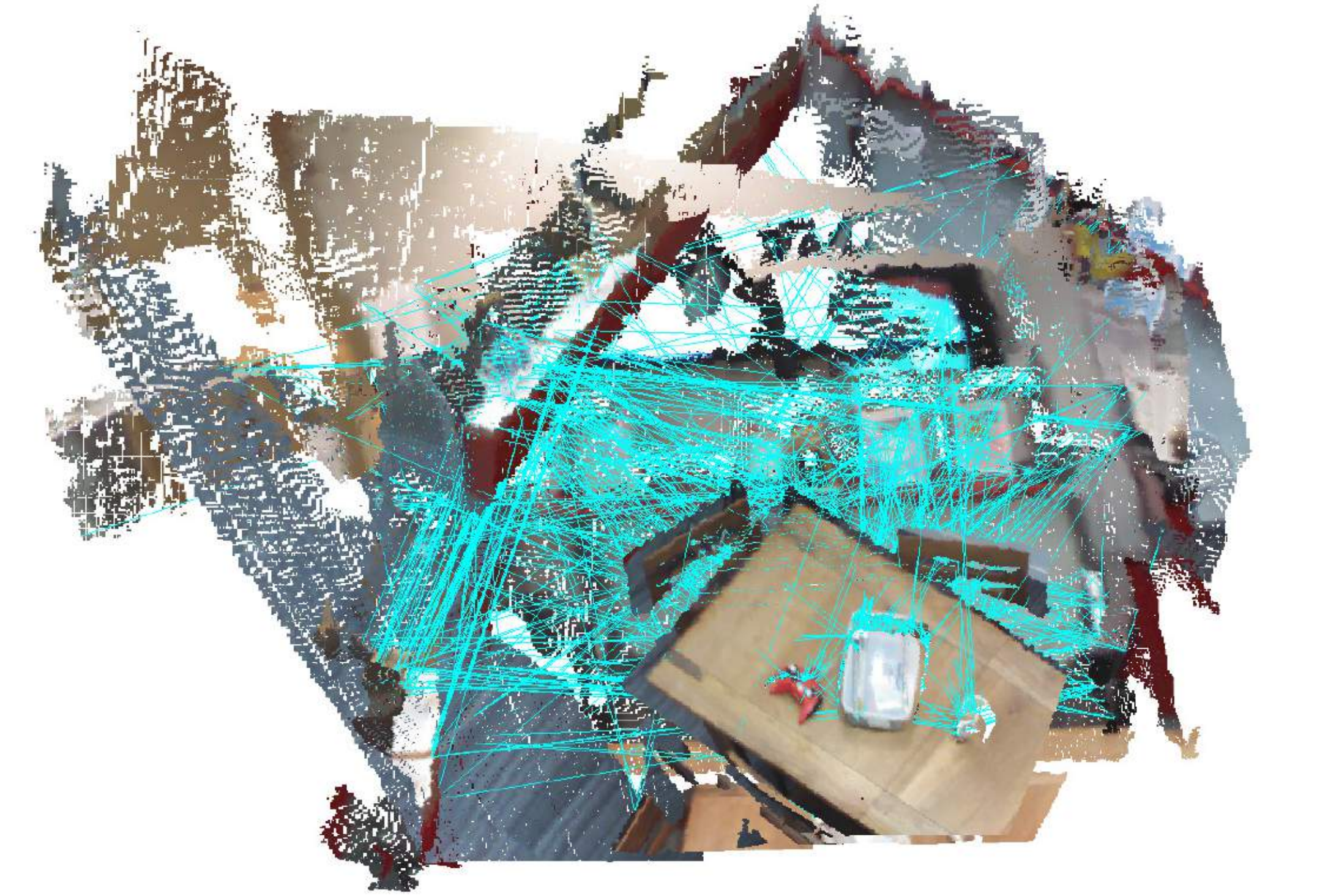}
\end{minipage}\,\,
& &
\,\,
\begin{minipage}[t]{0.19\linewidth}
\centering
\includegraphics[width=.48\linewidth]{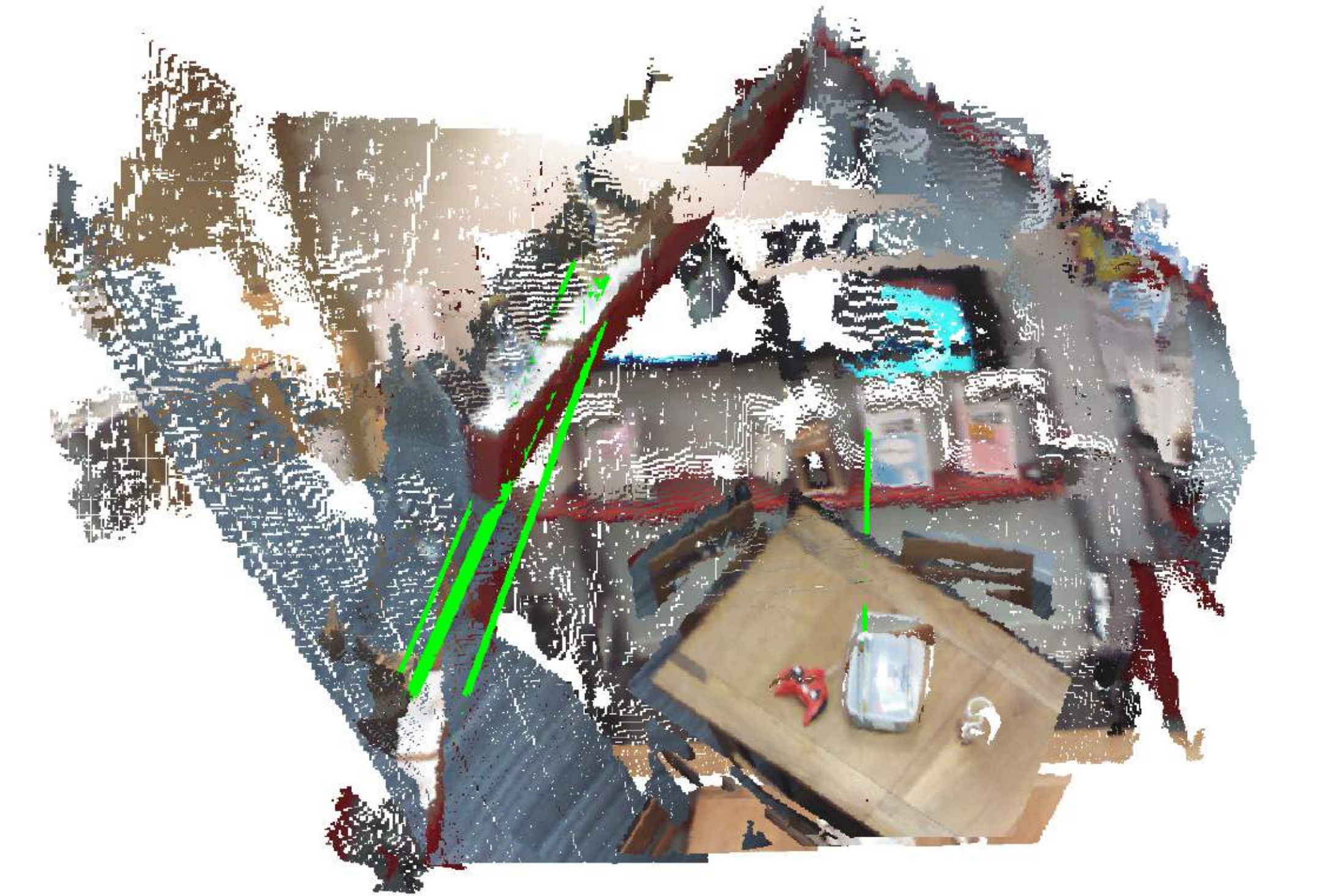}
\includegraphics[width=.48\linewidth]{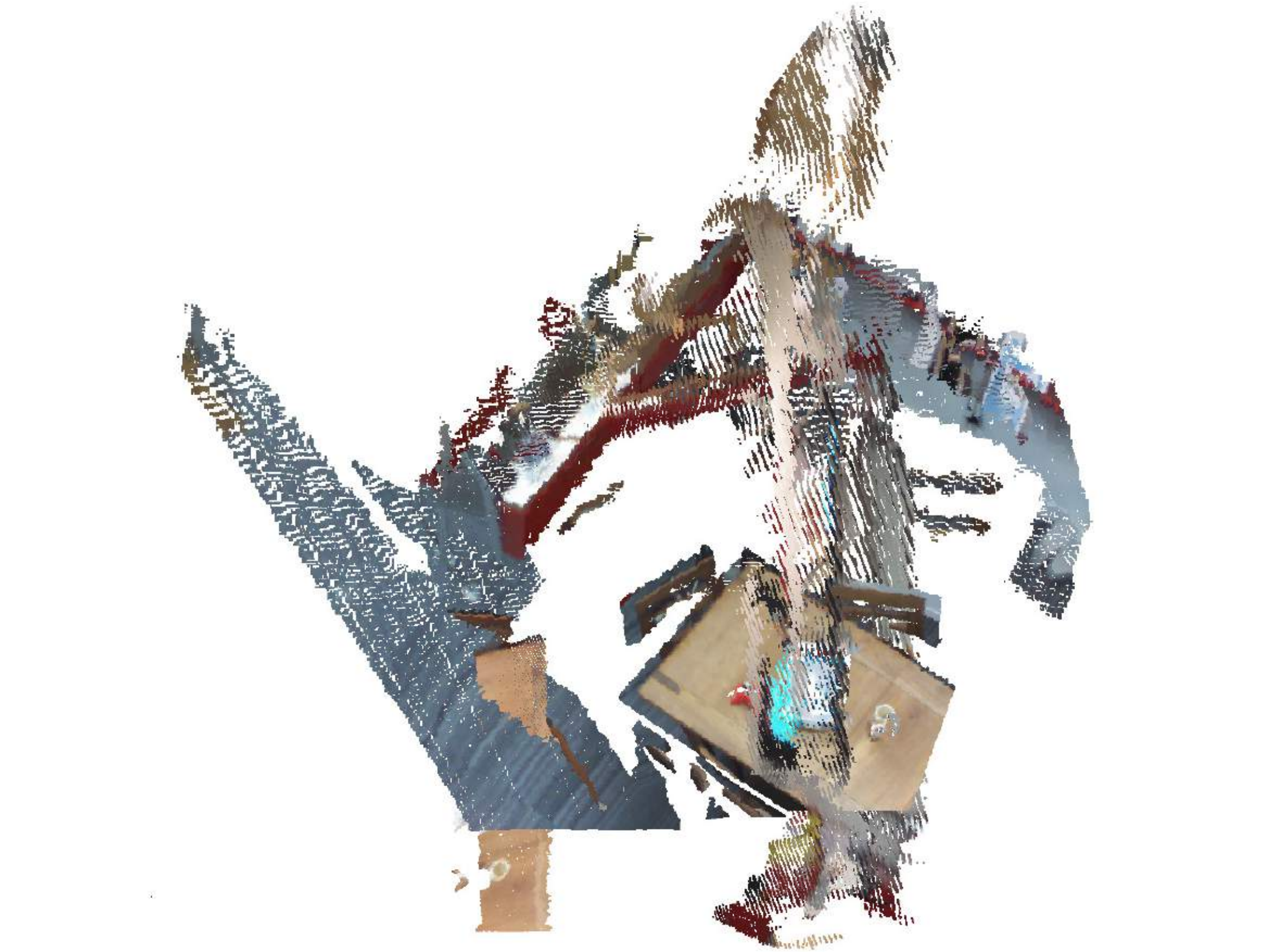}
\end{minipage}\,\,
&
\,\,
\begin{minipage}[t]{0.19\linewidth}
\centering
\includegraphics[width=.48\linewidth]{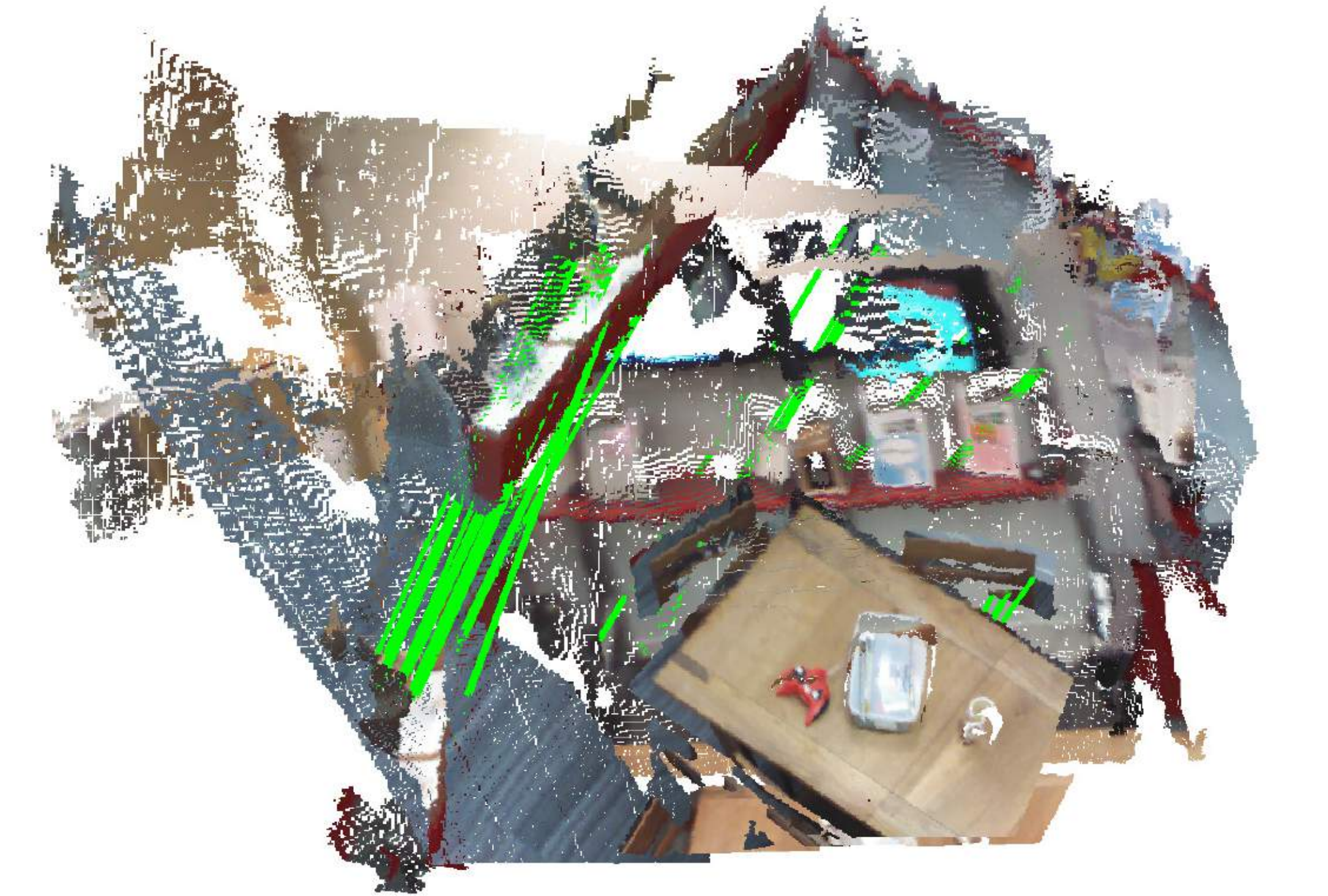}
\includegraphics[width=.48\linewidth]{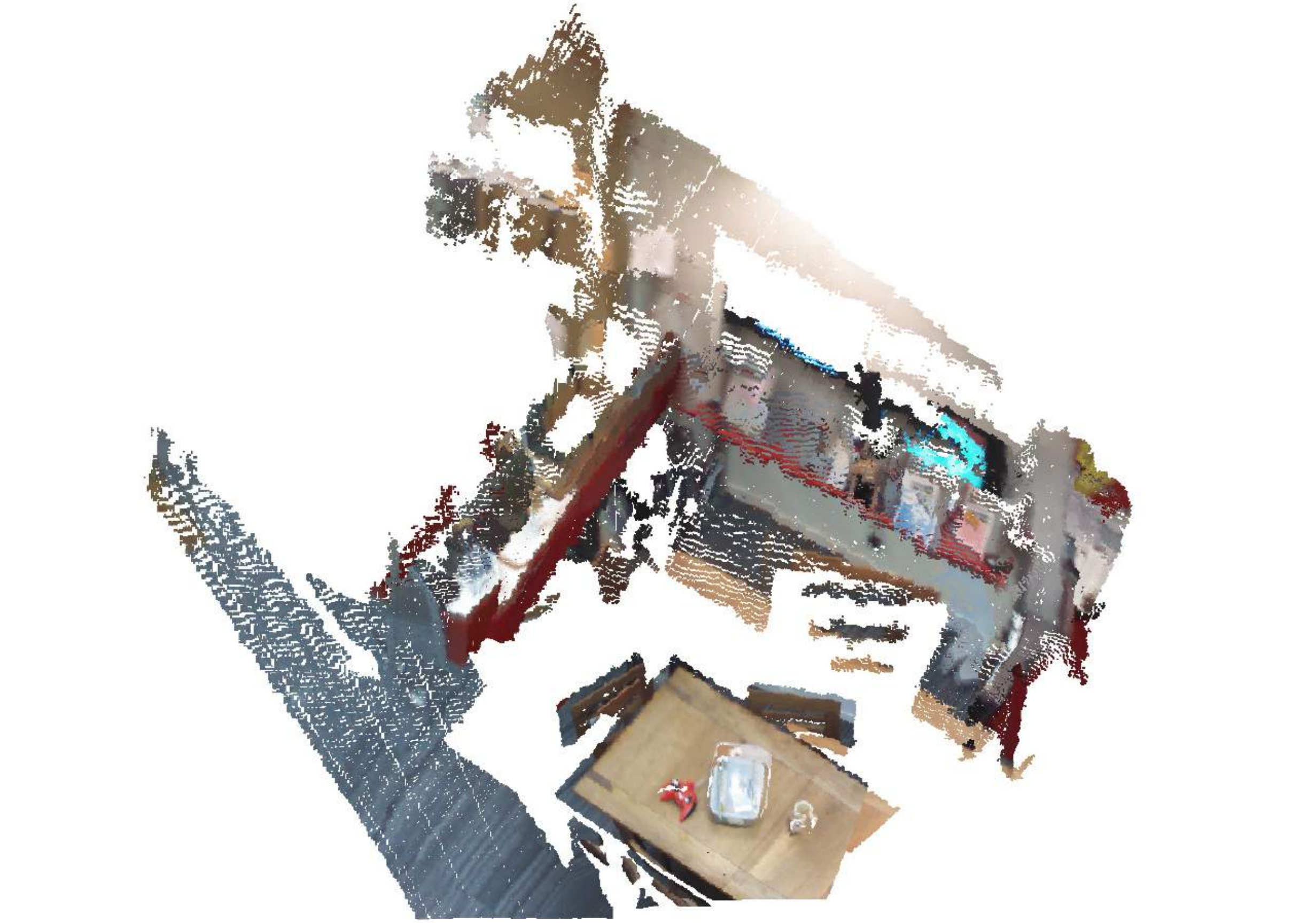}
\end{minipage}\,\,
&
\,\,
\begin{minipage}[t]{0.19\linewidth}
\centering
\includegraphics[width=.48\linewidth]{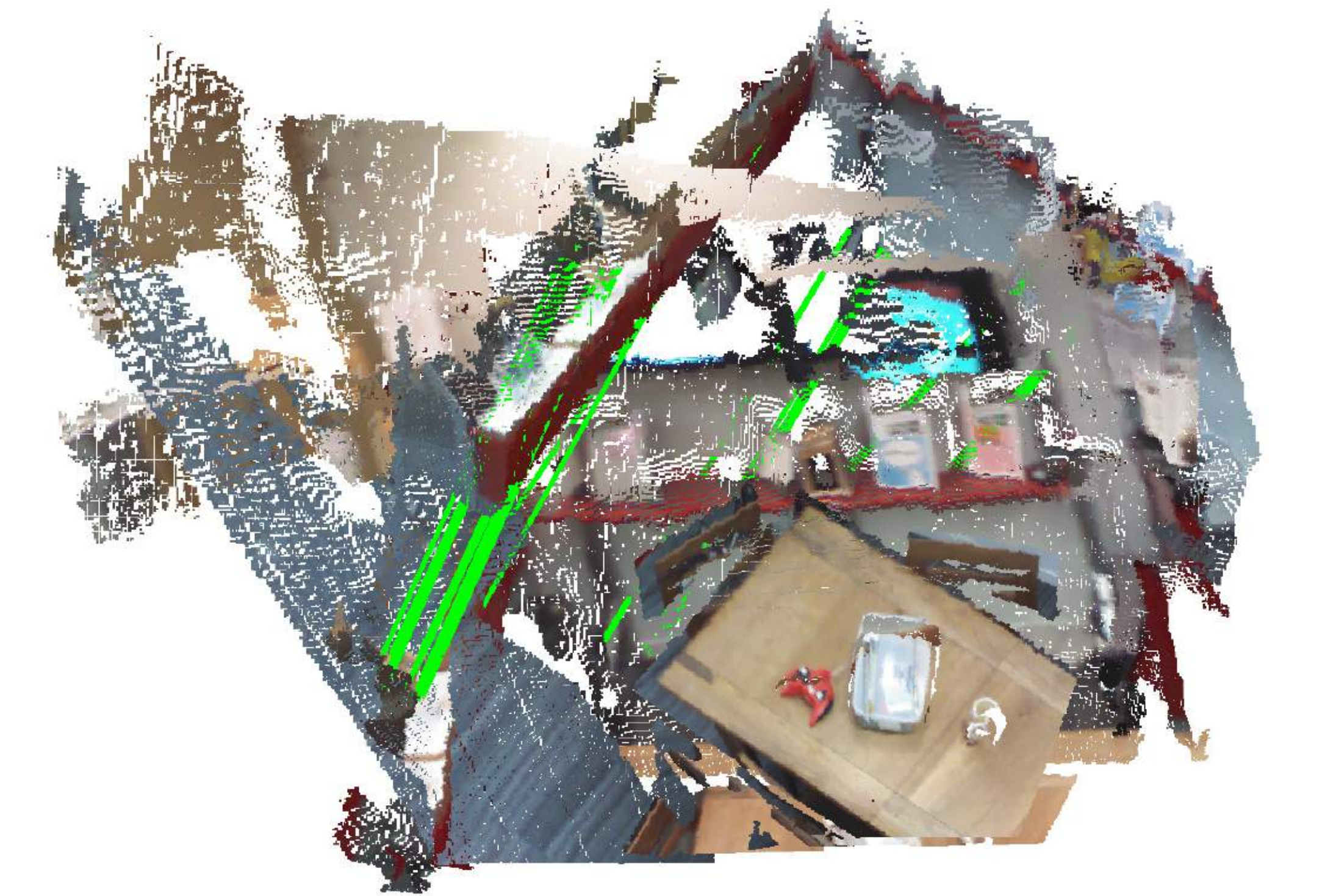}
\includegraphics[width=.48\linewidth]{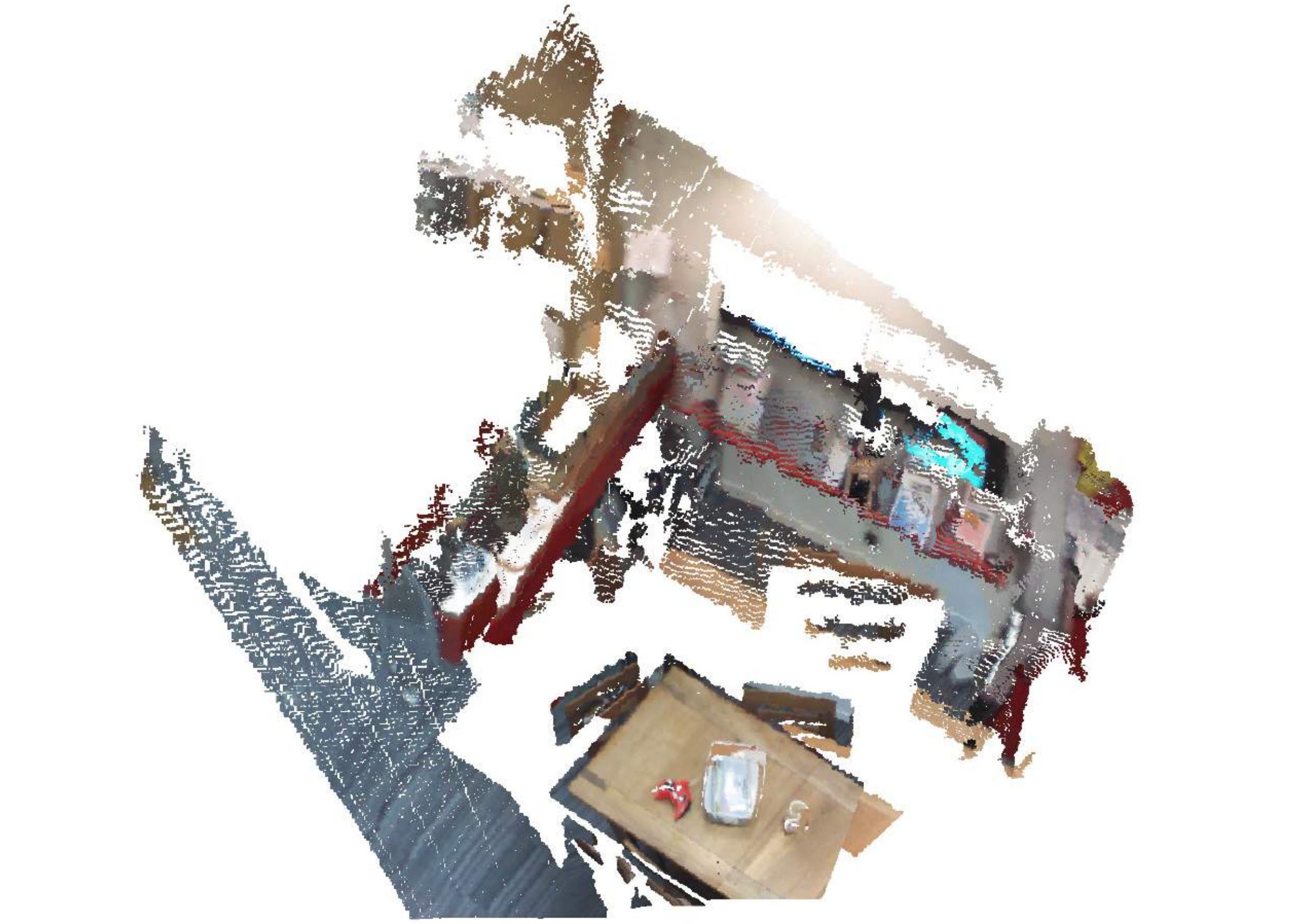}
\end{minipage}\,\,
&
\,\,
\begin{minipage}[t]{0.19\linewidth}
\centering
\includegraphics[width=.48\linewidth]{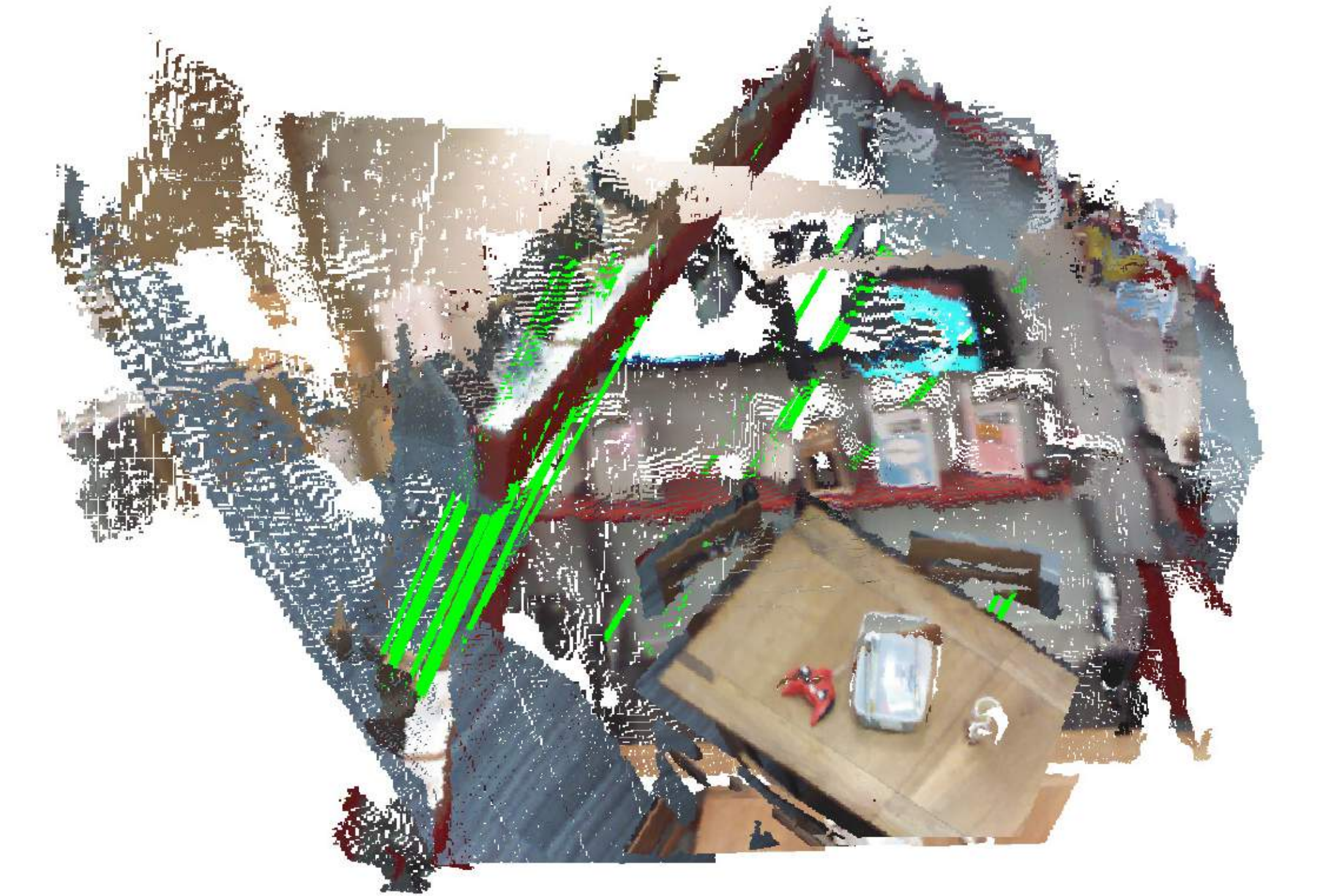}
\includegraphics[width=.48\linewidth]{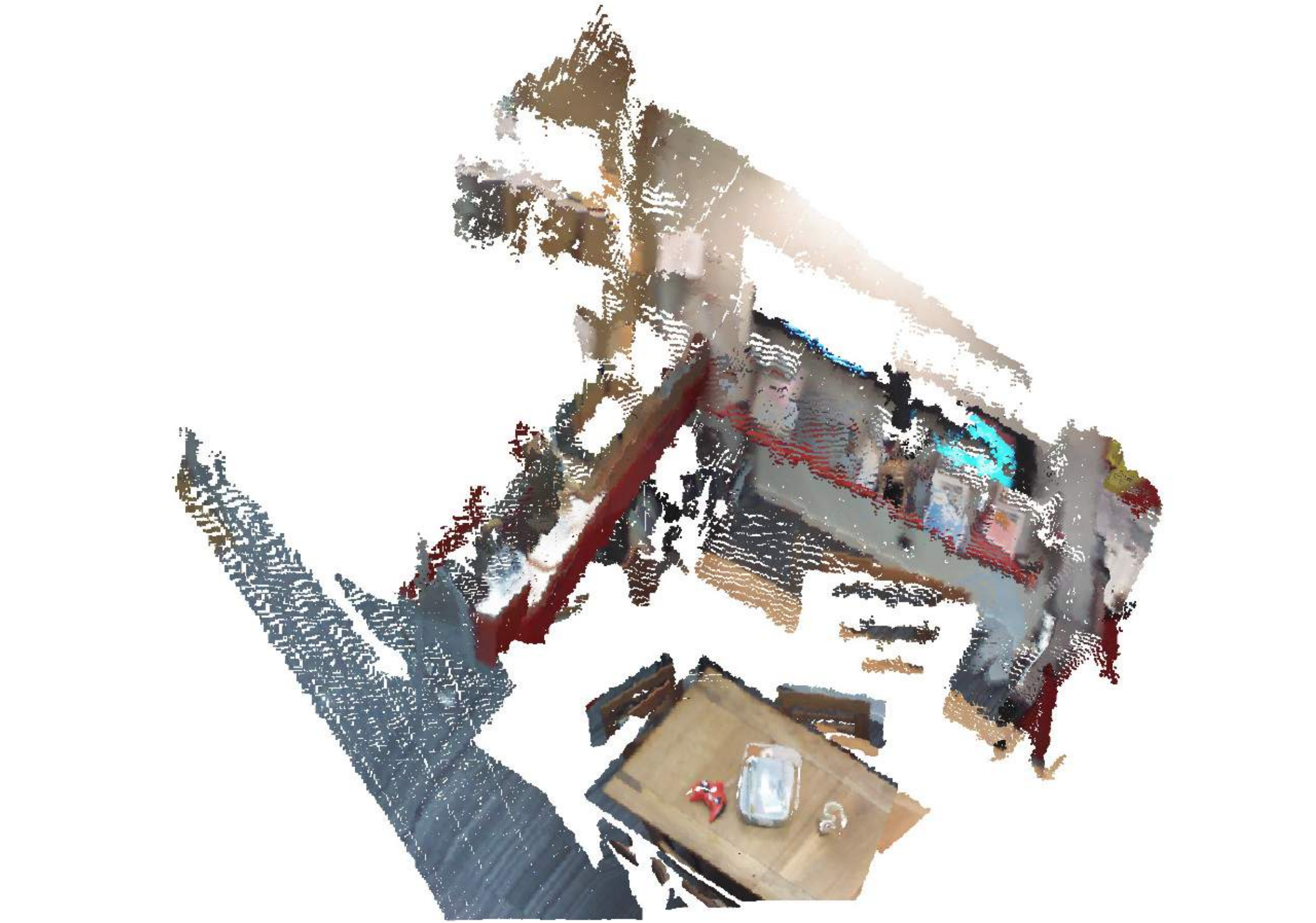}
\end{minipage}\,\,

\\

 & \footnotesize{$N=1984$} && \footnotesize{\textcolor[rgb]{0,0.8,0}{Succeed}$,0.314,0.291s$} &\footnotesize{\textcolor[rgb]{0,0.8,0}{Succeed}$,0.516,8.044s$}&\footnotesize{\textcolor[rgb]{0,0.8,0}{Succeed}$,0.323,74.158s$}&\footnotesize{\textcolor[rgb]{0,0.8,0}{Succeed}$,\textbf{0.301},\textbf{0.272}s$}

\\
\rotatebox{90}{\,\,\footnotesize{\textit{red kitchen}}\,}\,
&
\,\,
\begin{minipage}[t]{0.1\linewidth}
\centering
\includegraphics[width=1\linewidth]{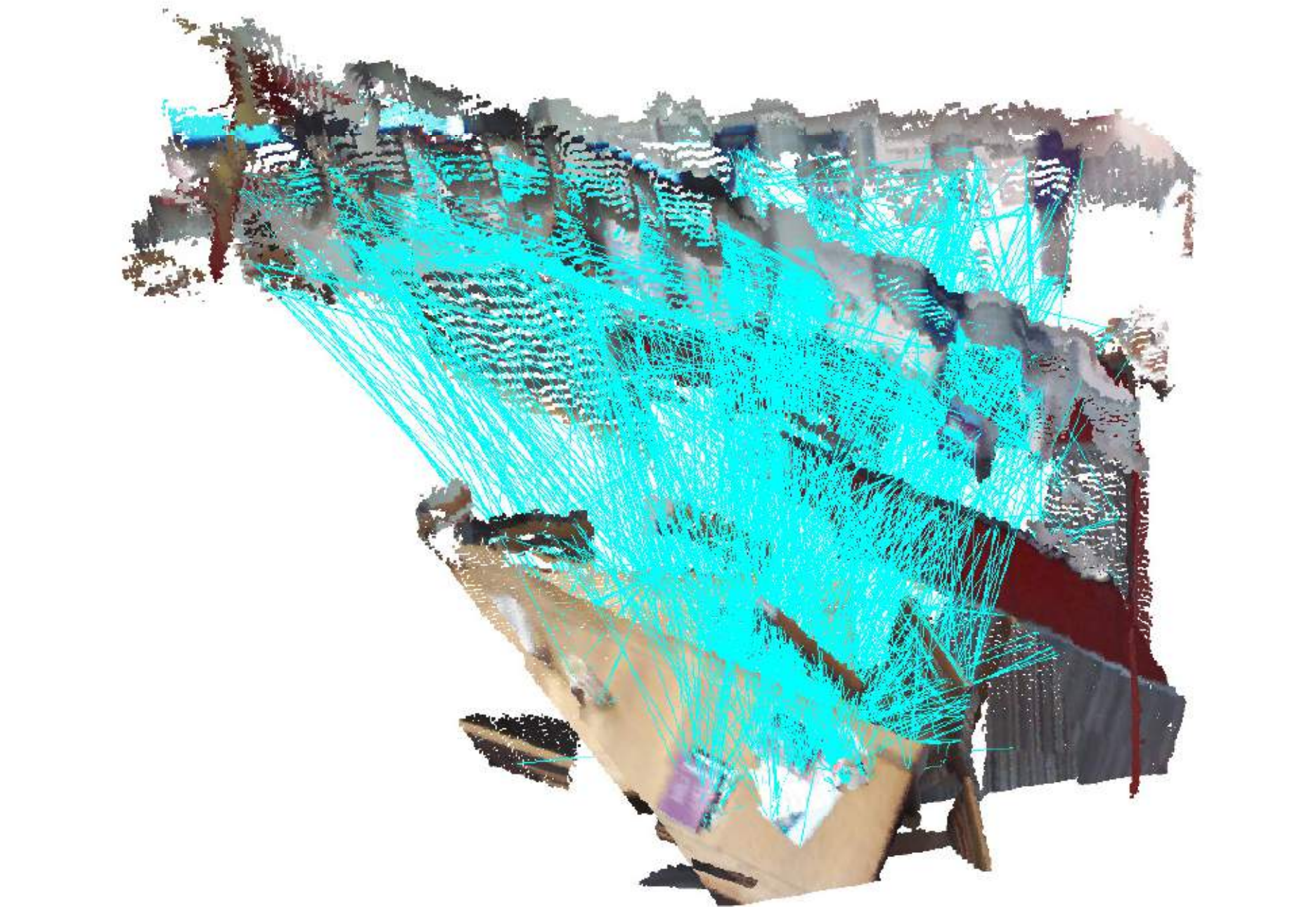}
\end{minipage}\,\,
& &
\,\,
\begin{minipage}[t]{0.19\linewidth}
\centering
\includegraphics[width=.48\linewidth]{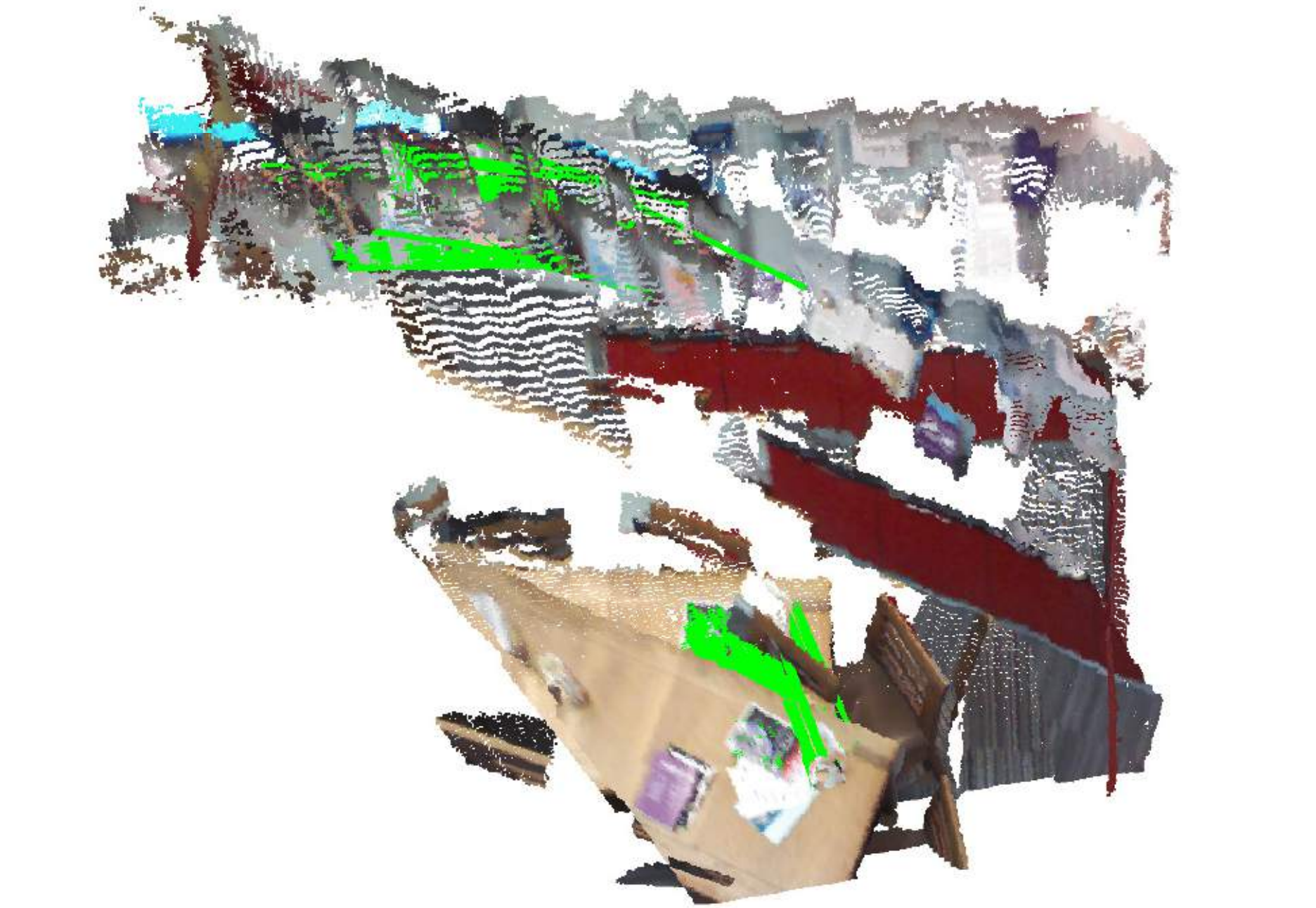}
\includegraphics[width=.48\linewidth]{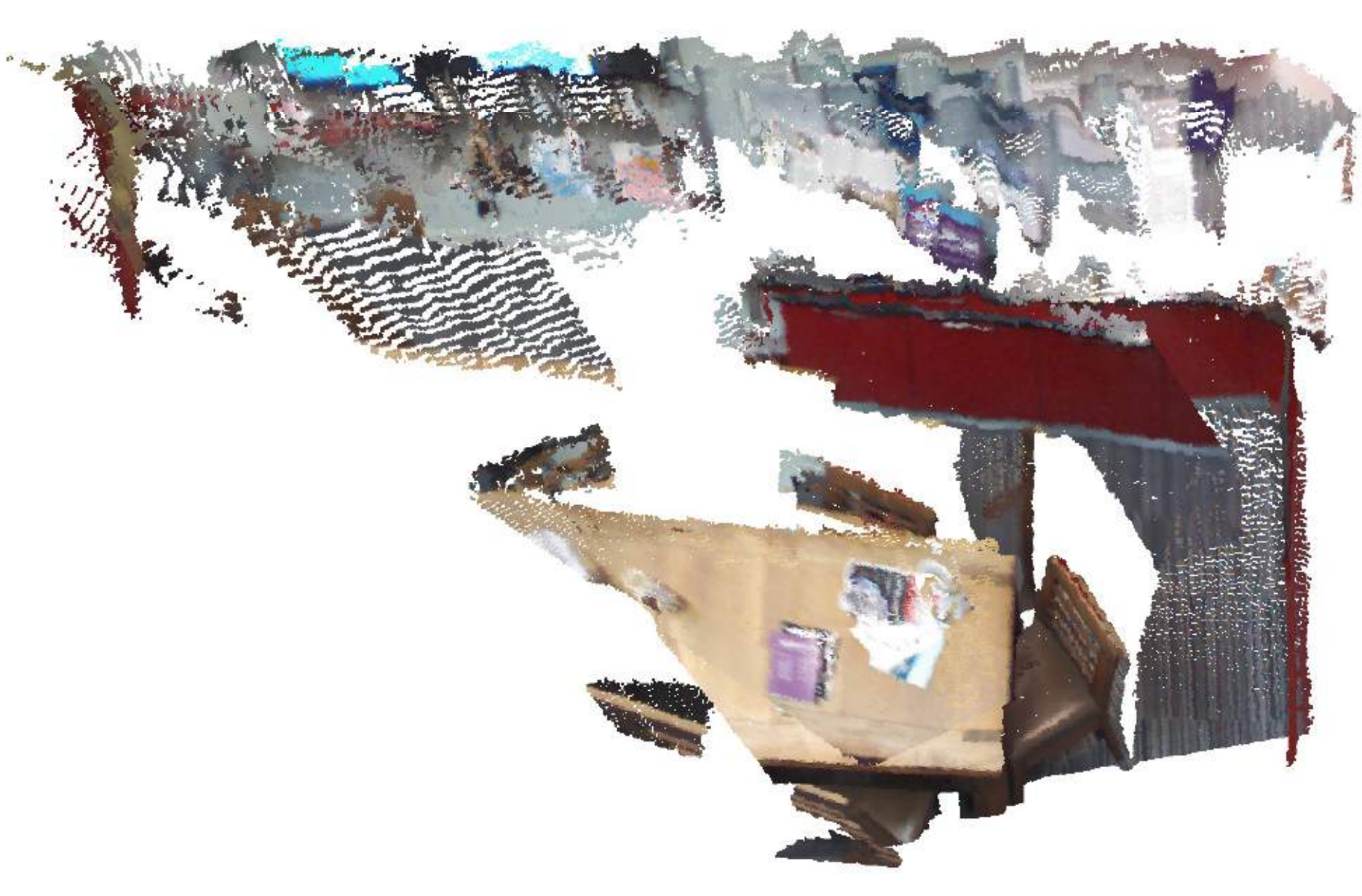}
\end{minipage}\,\,
&
\,\,
\begin{minipage}[t]{0.19\linewidth}
\centering
\includegraphics[width=.48\linewidth]{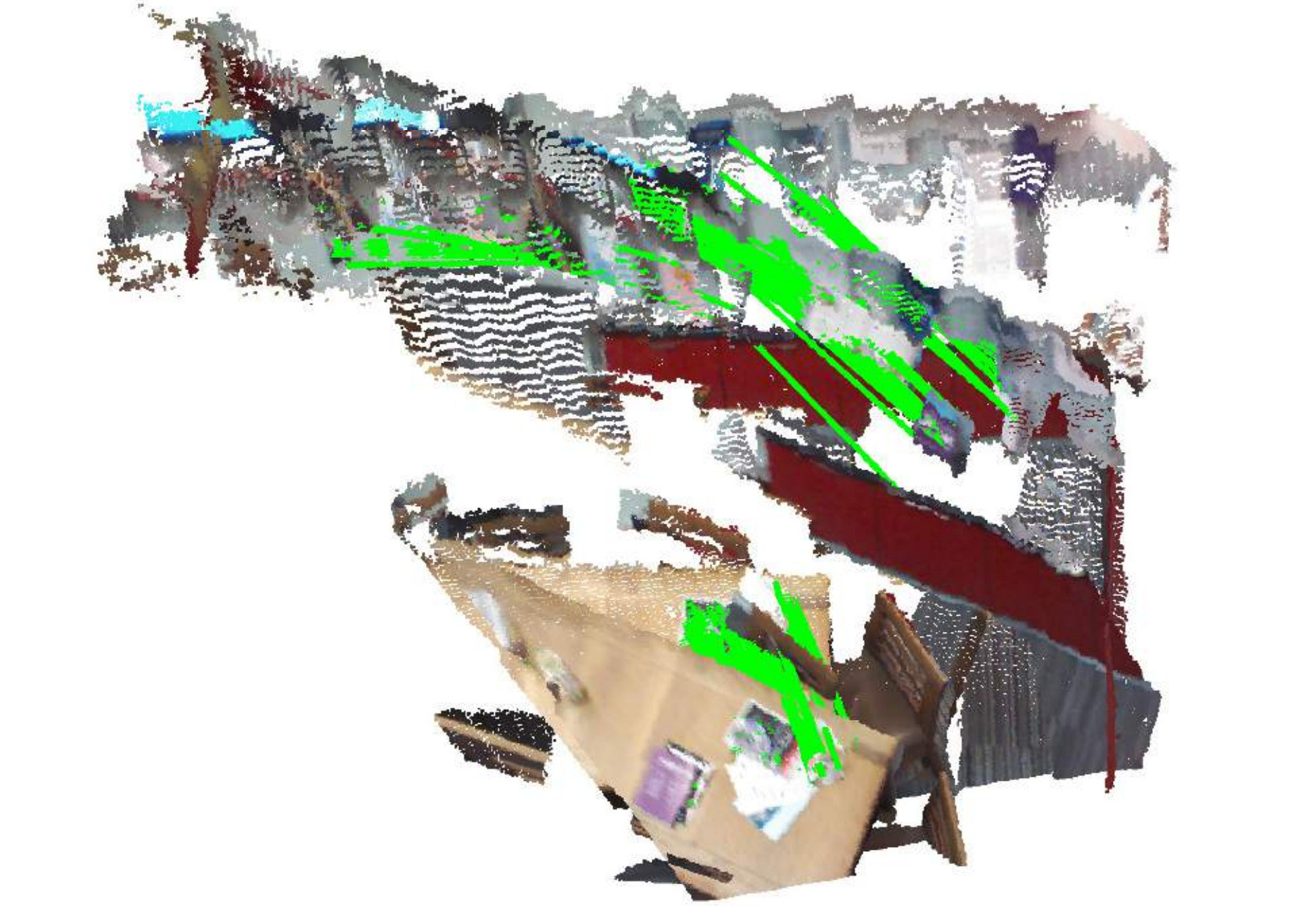}
\includegraphics[width=.48\linewidth]{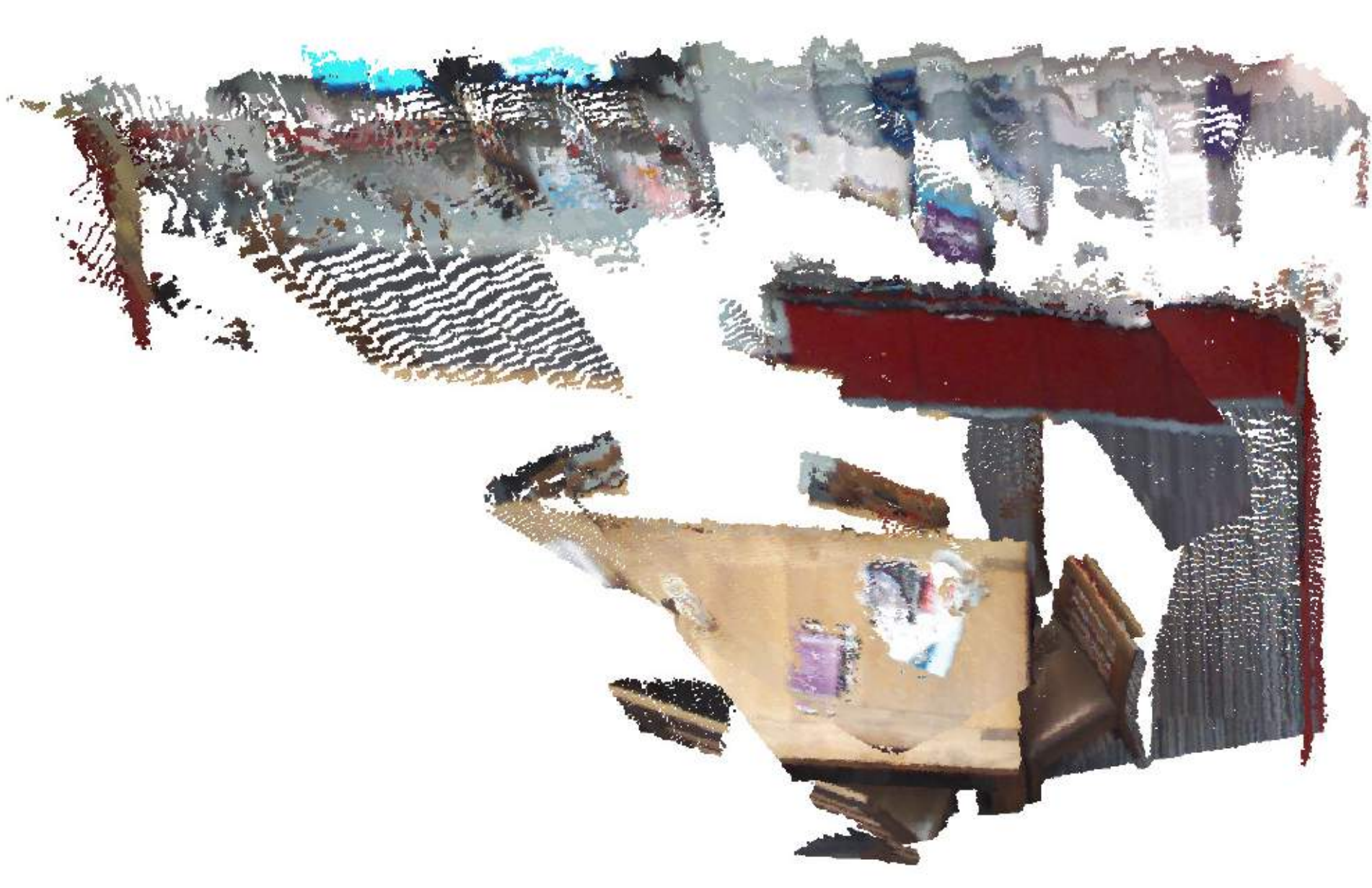}
\end{minipage}\,\,
&
\,\,
\begin{minipage}[t]{0.19\linewidth}
\centering
\includegraphics[width=.48\linewidth]{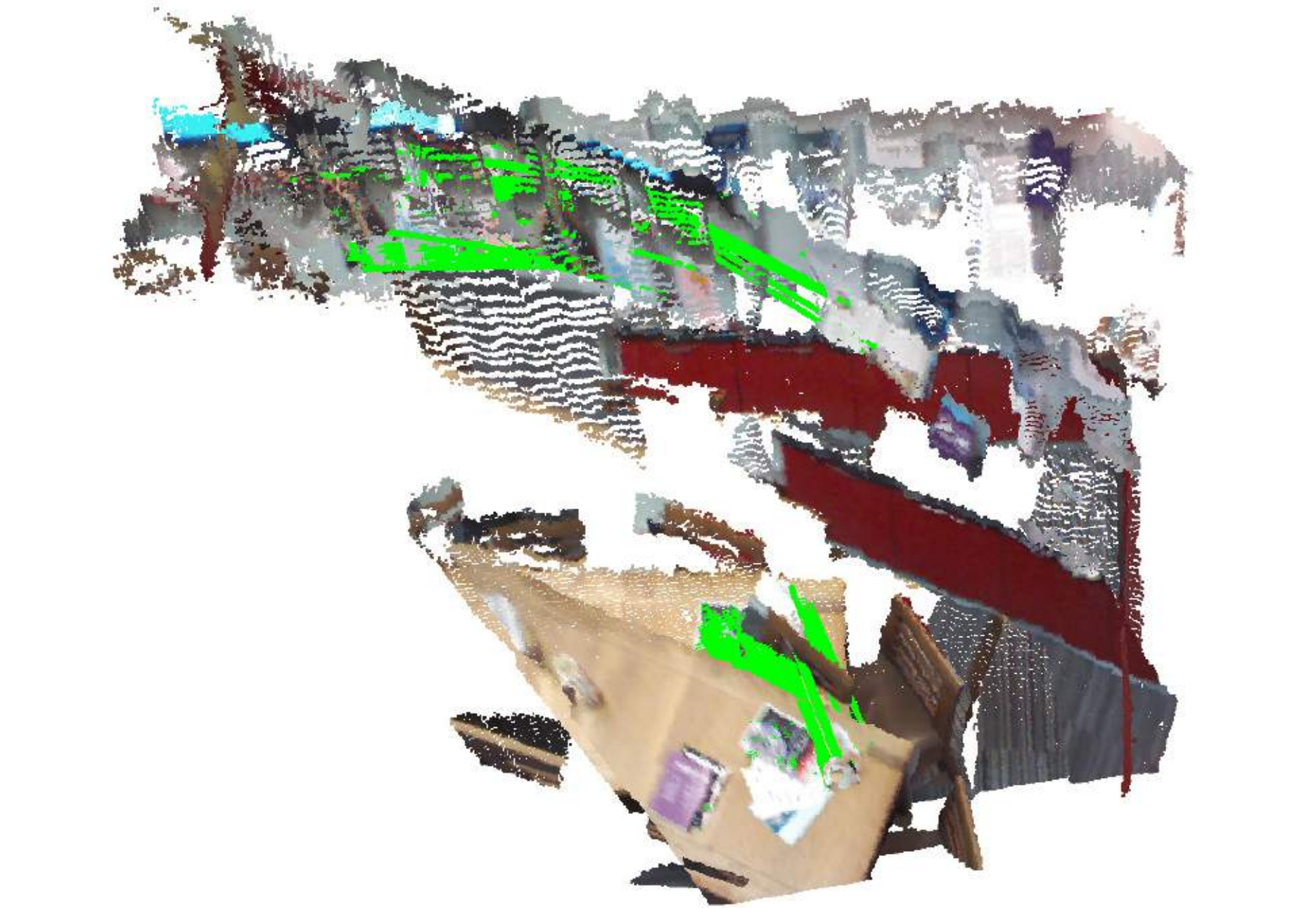}
\includegraphics[width=.48\linewidth]{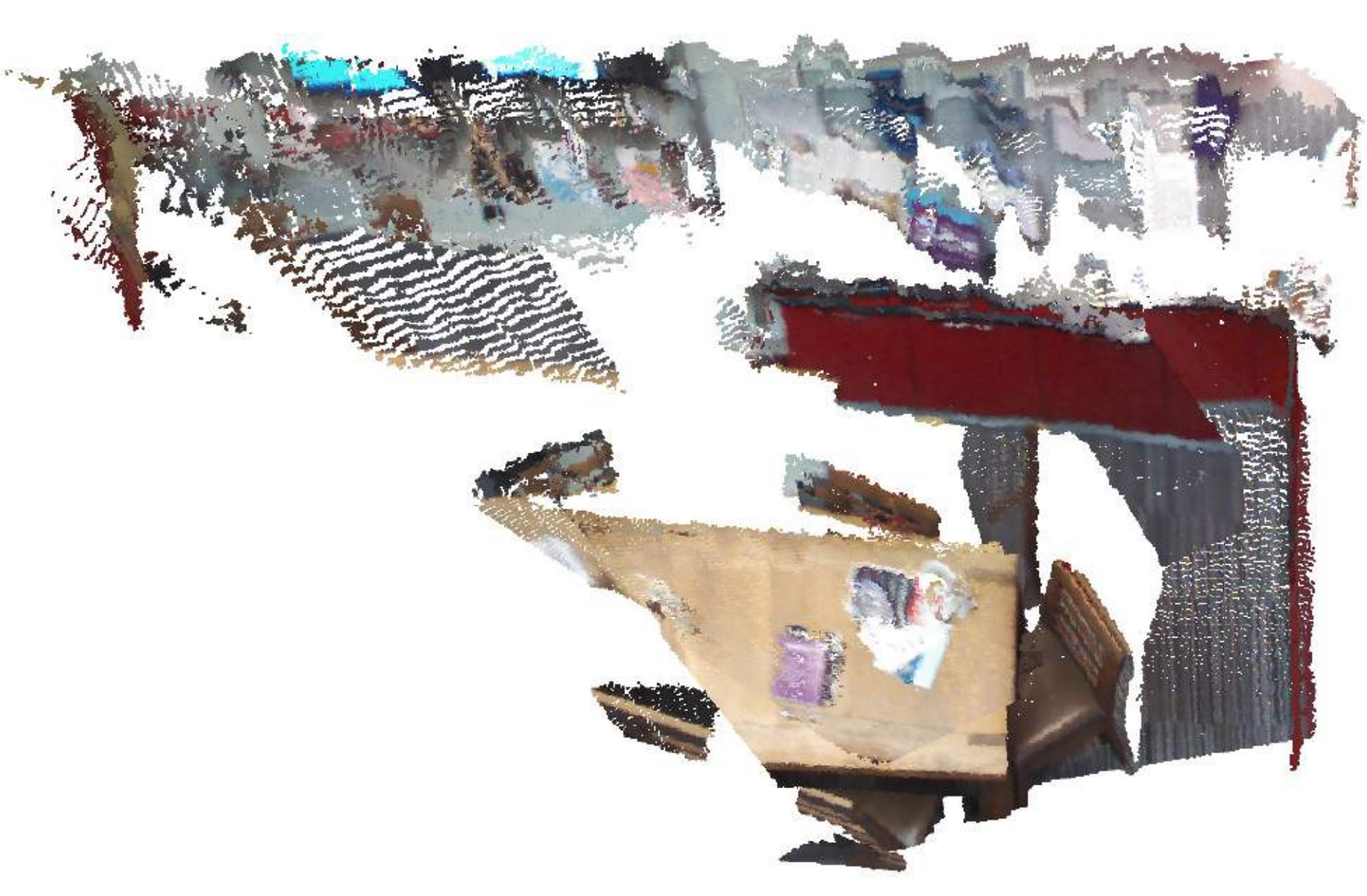}
\end{minipage}\,\,
&
\,\,
\begin{minipage}[t]{0.19\linewidth}
\centering
\includegraphics[width=.48\linewidth]{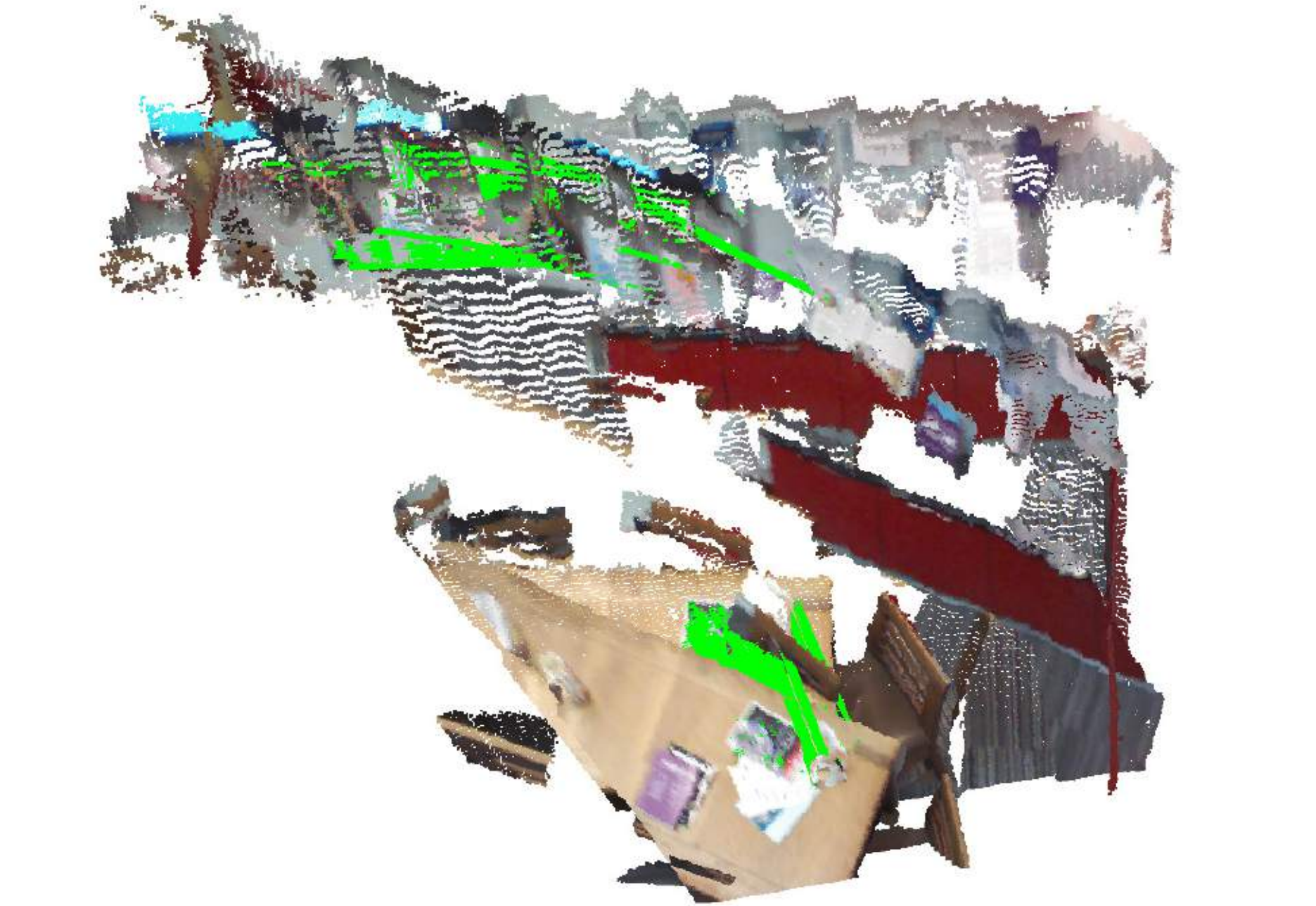}
\includegraphics[width=.48\linewidth]{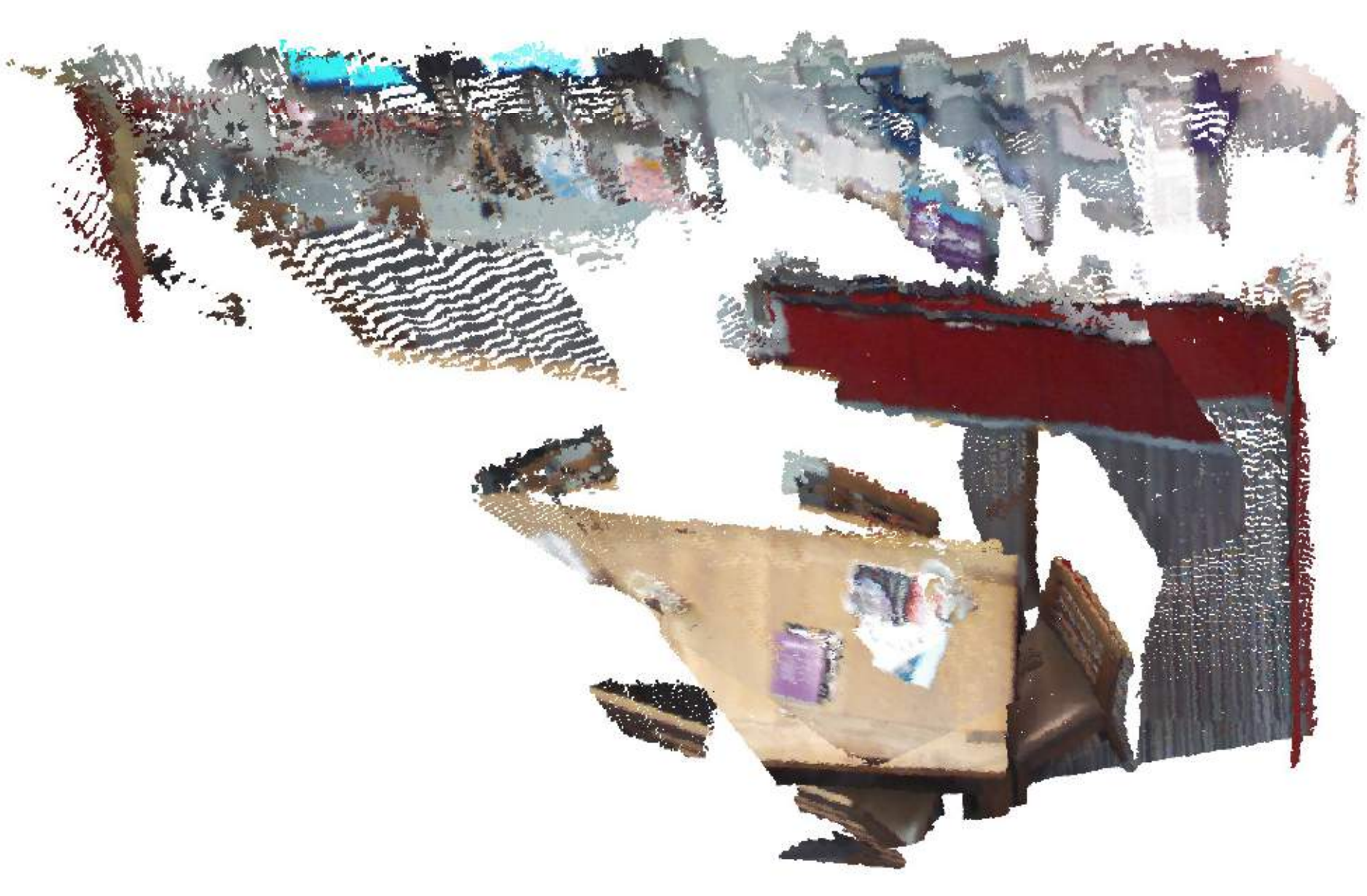}
\end{minipage}\,\,
\\

 & \footnotesize{$N=1823$} && \footnotesize{\textcolor[rgb]{1,0,0}{Fail}$,\verb|\|,0.199s$} &\footnotesize{\textcolor[rgb]{1,0,0}{Fail}$,\verb|\|,49.098s$}&\footnotesize{\textcolor[rgb]{1,0,0}{Fail}$,\verb|\|,63.692s$}&\footnotesize{\textcolor[rgb]{0,0.8,0}{Succeed}$,\textbf{0.278},\textbf{0.781}s$}

\\
\rotatebox{90}{\,\,\footnotesize{\textit{red kitchen}}\,}\,
&
\,\,
\begin{minipage}[t]{0.1\linewidth}
\centering
\includegraphics[width=1\linewidth]{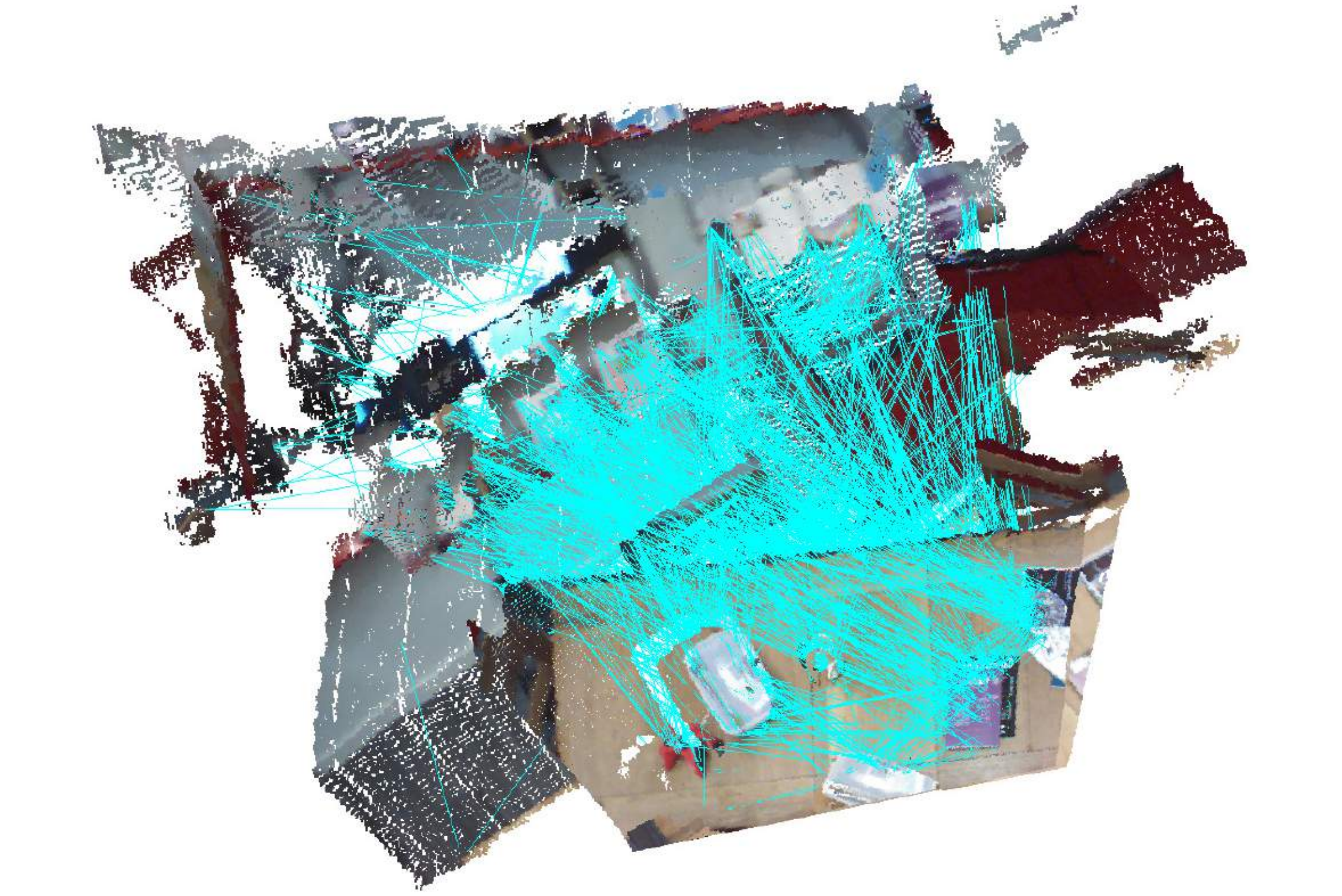}
\end{minipage}\,\,
& &
\,\,
\begin{minipage}[t]{0.19\linewidth}
\centering
\includegraphics[width=.48\linewidth]{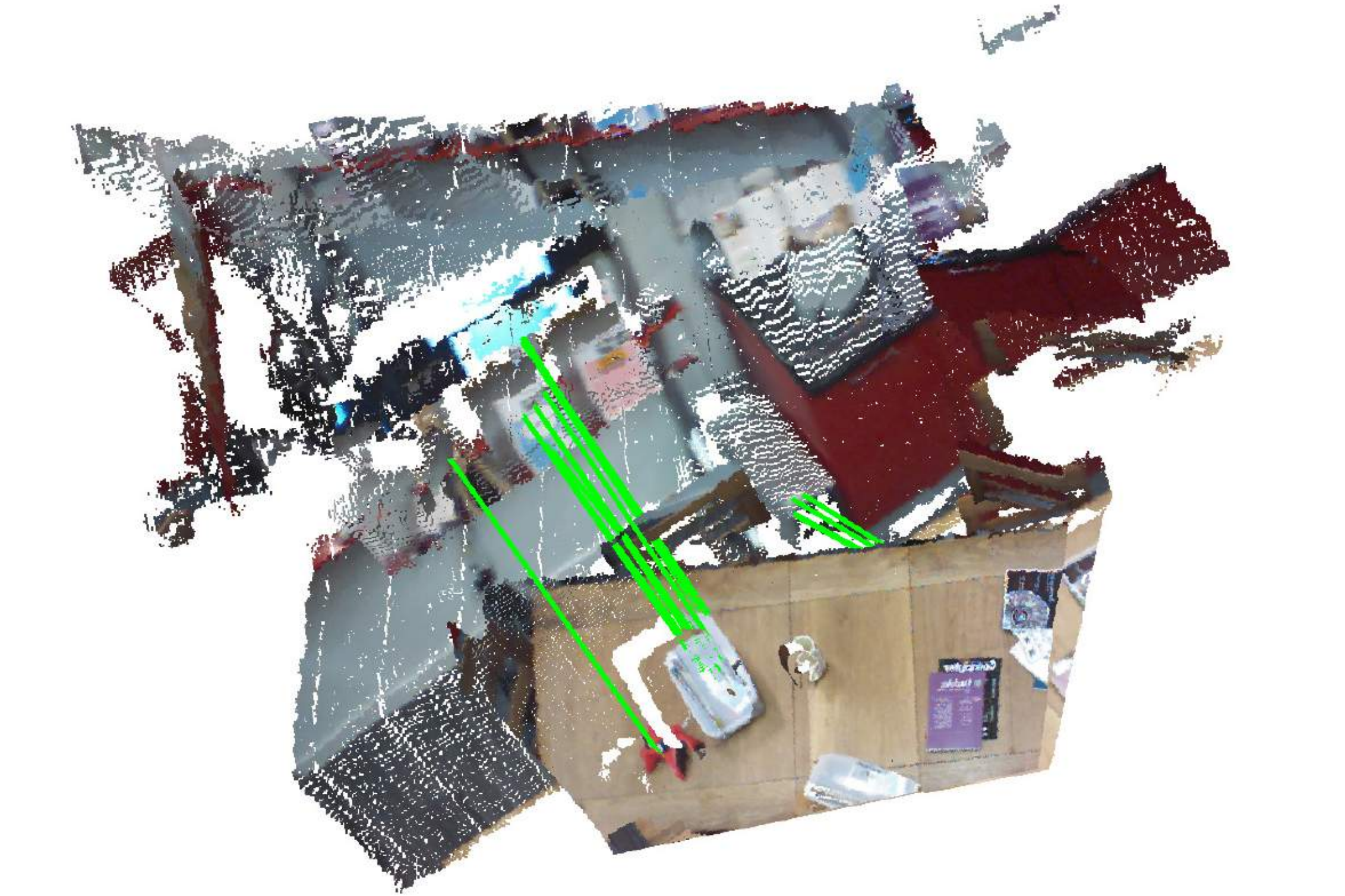}
\includegraphics[width=.48\linewidth]{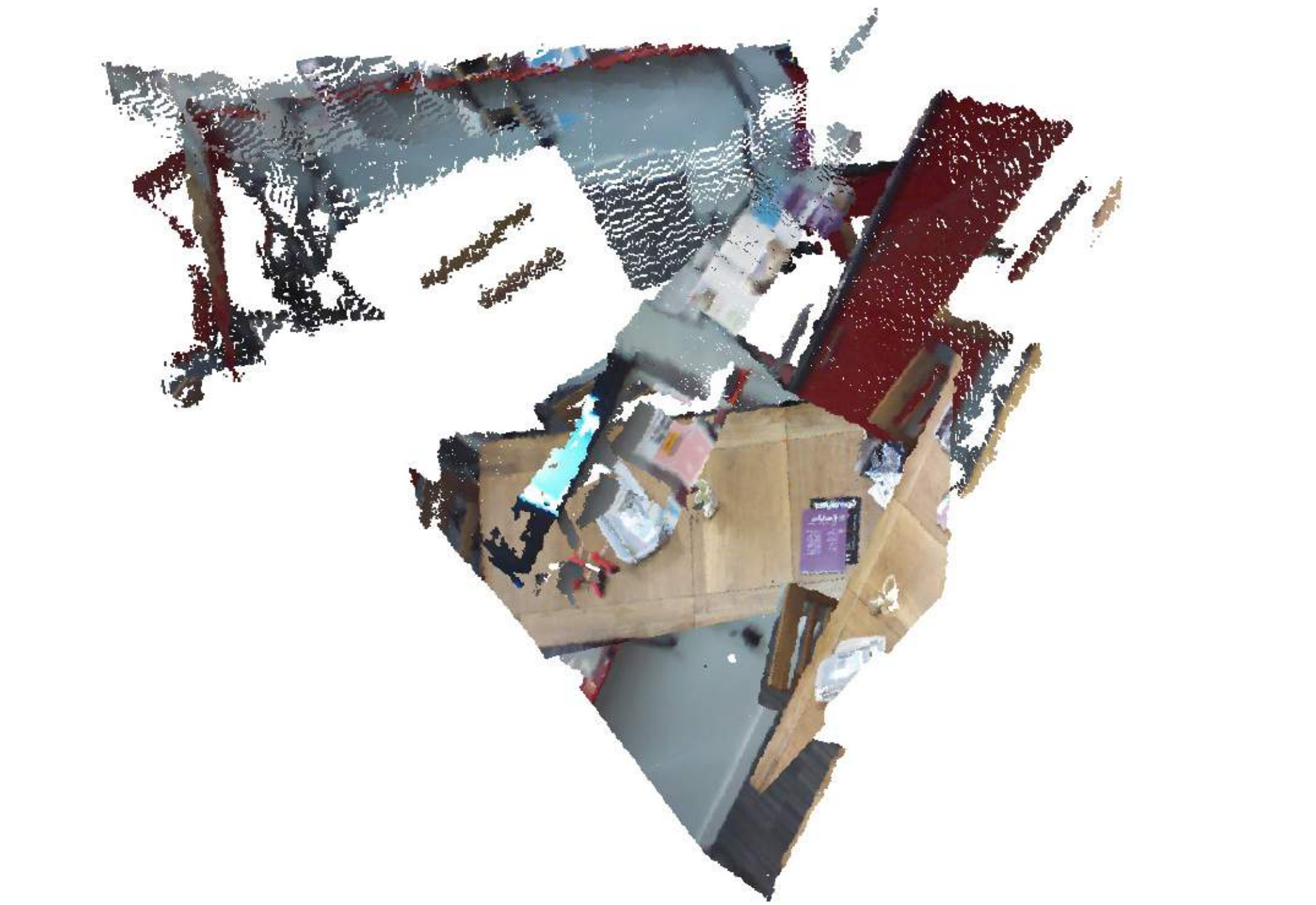}
\end{minipage}\,\,
&
\,\,
\begin{minipage}[t]{0.19\linewidth}
\centering
\includegraphics[width=.48\linewidth]{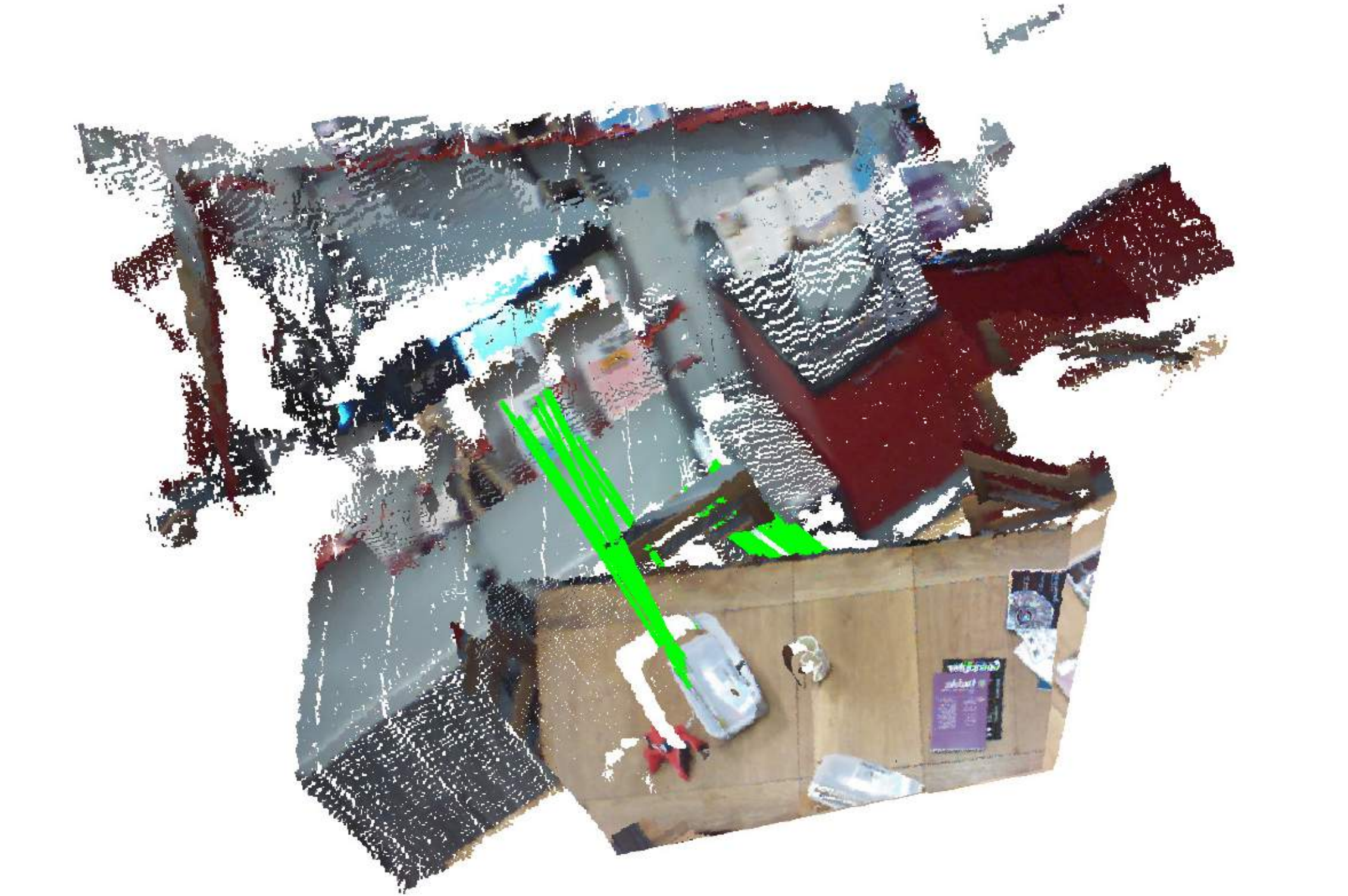}
\includegraphics[width=.48\linewidth]{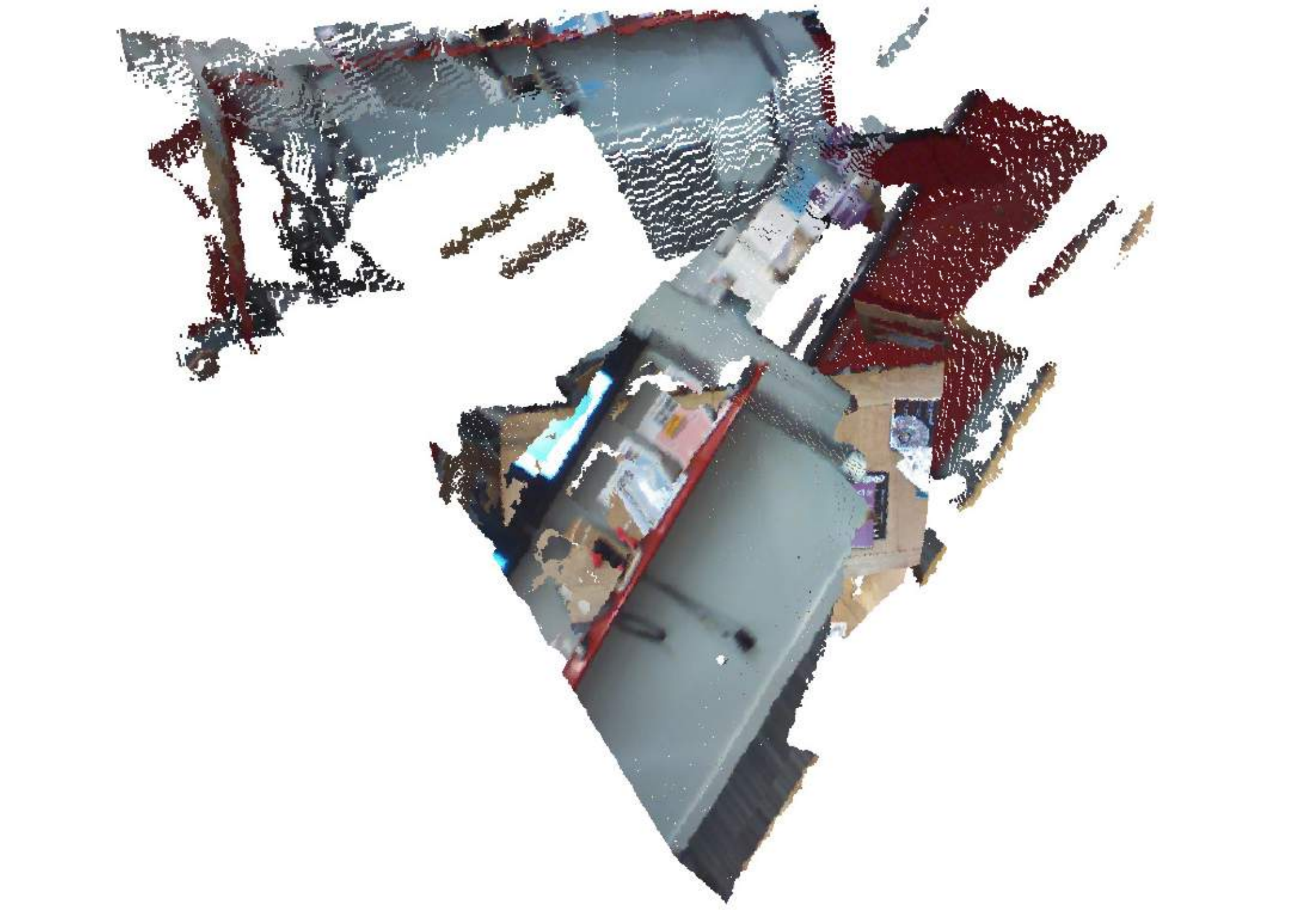}
\end{minipage}\,\,
&
\,\,
\begin{minipage}[t]{0.19\linewidth}
\centering
\includegraphics[width=.48\linewidth]{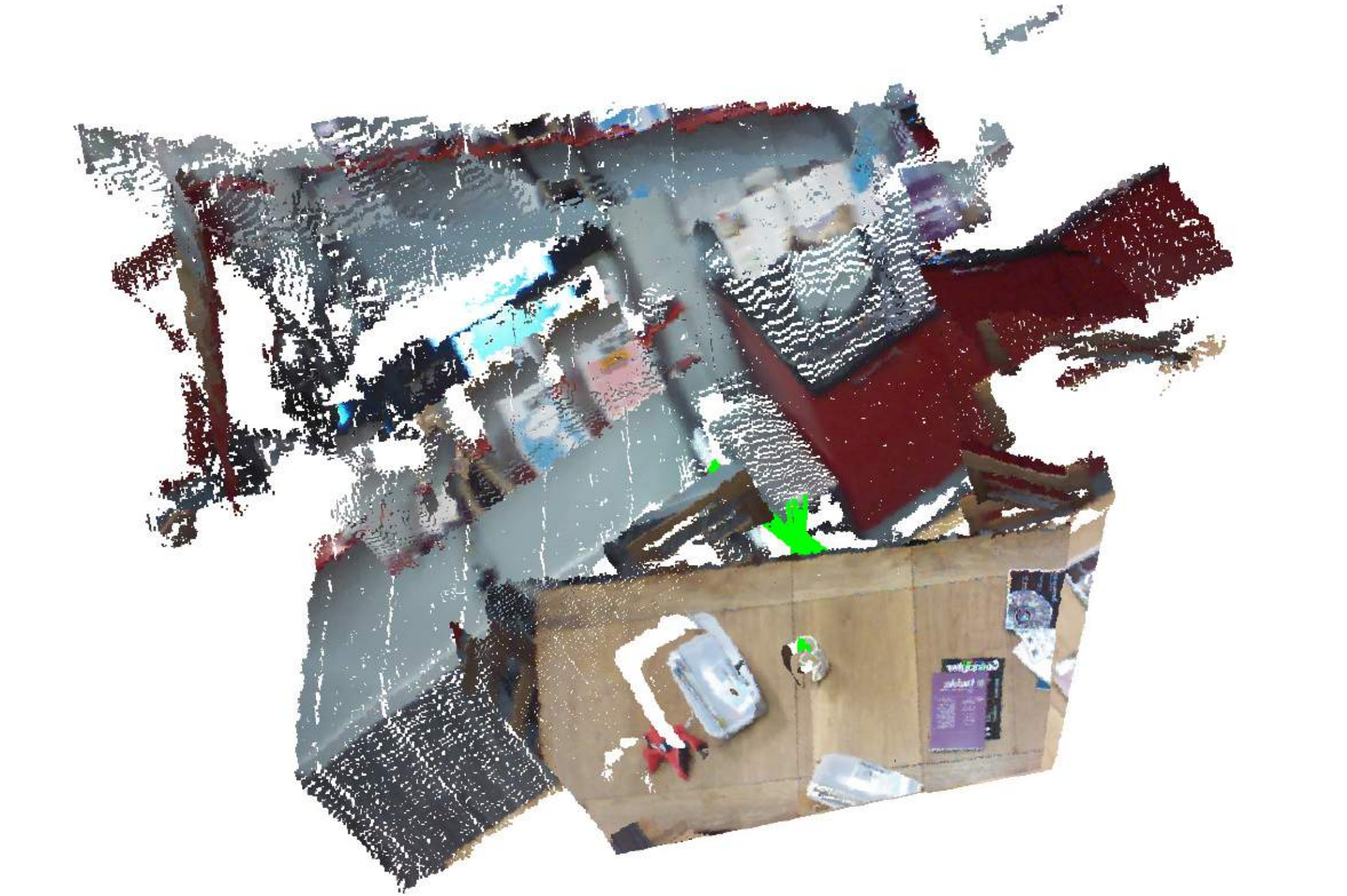}
\includegraphics[width=.48\linewidth]{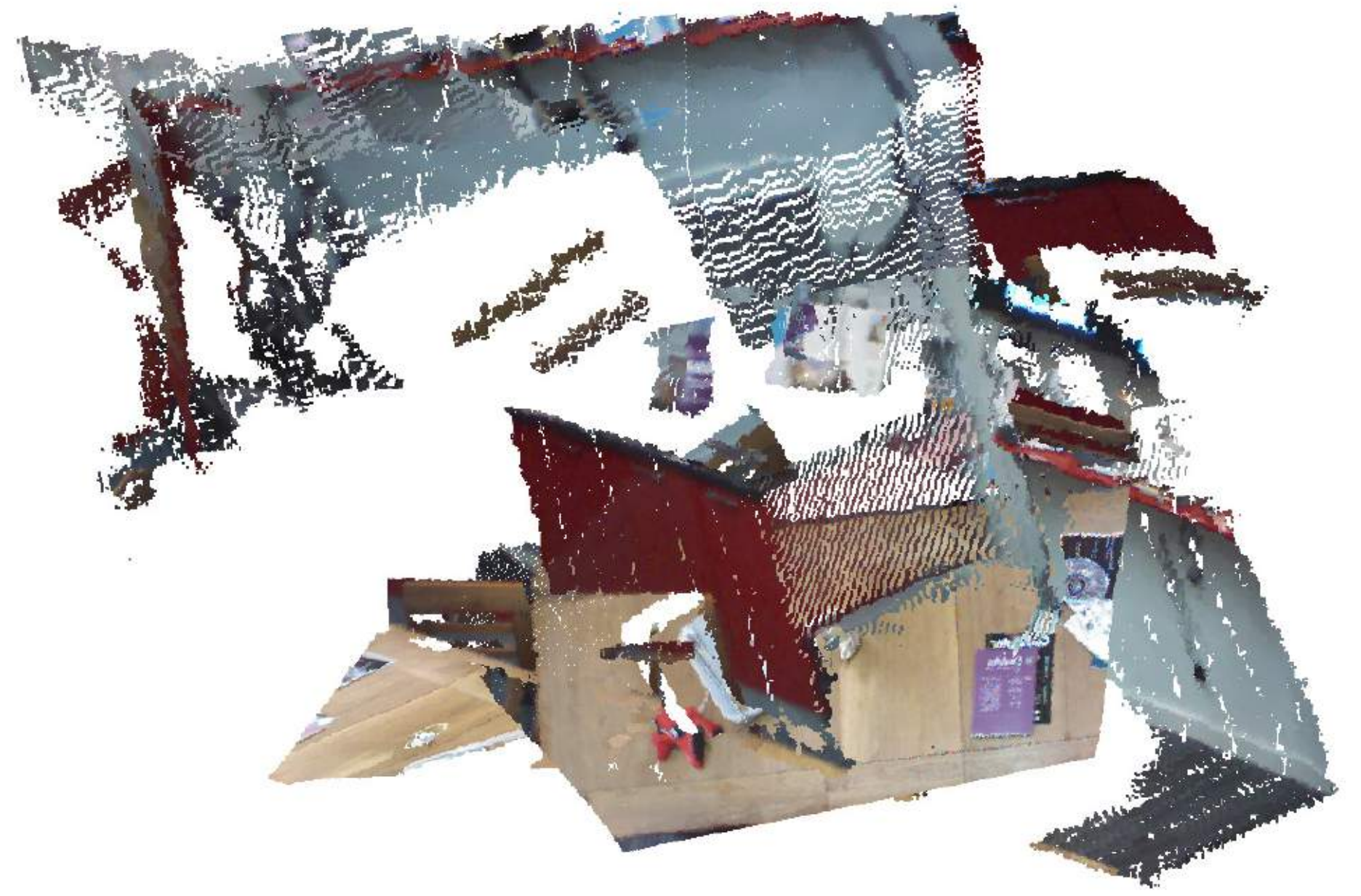}
\end{minipage}\,\,
&
\,\,
\begin{minipage}[t]{0.19\linewidth}
\centering
\includegraphics[width=.48\linewidth]{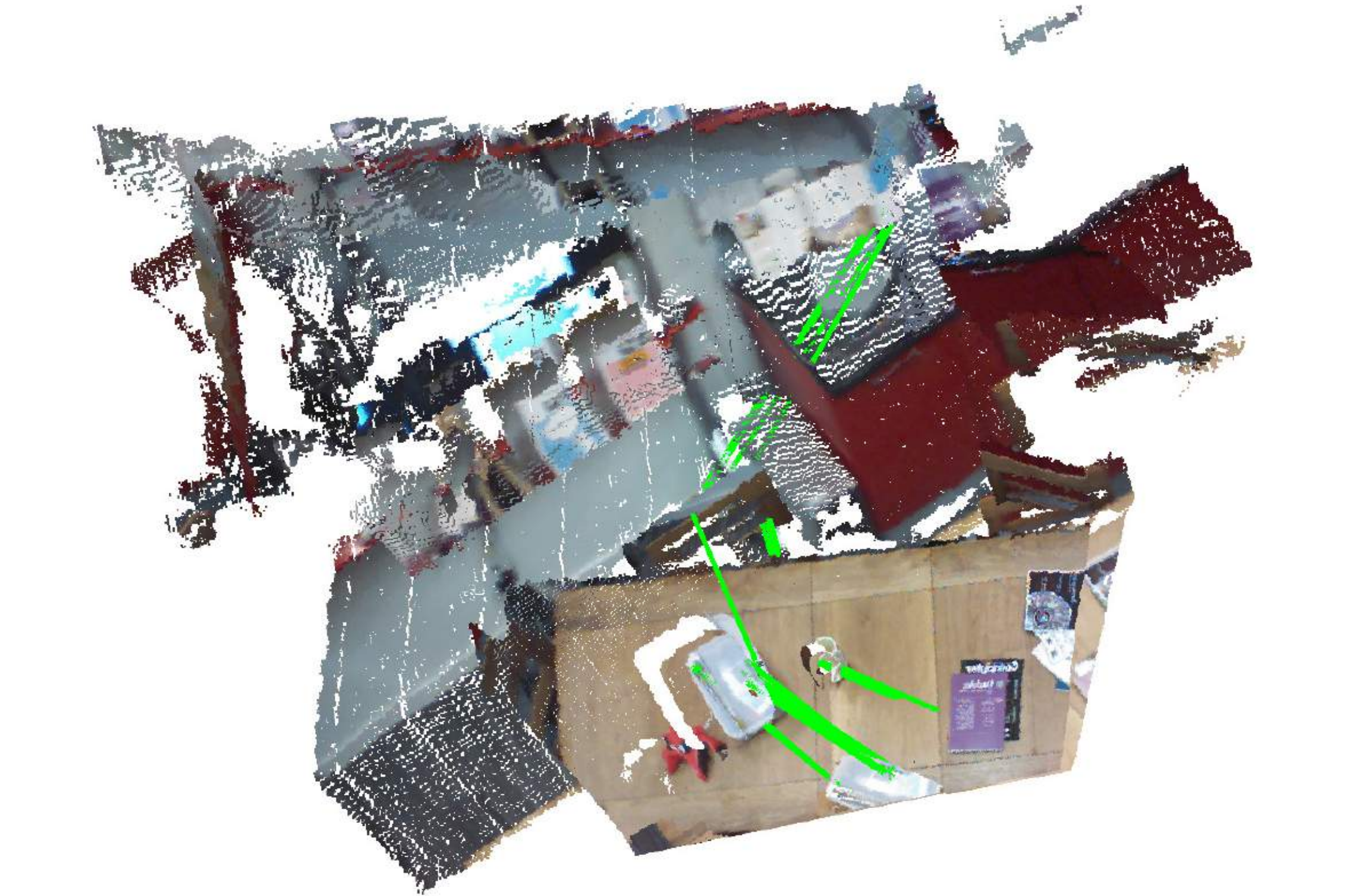}
\includegraphics[width=.48\linewidth]{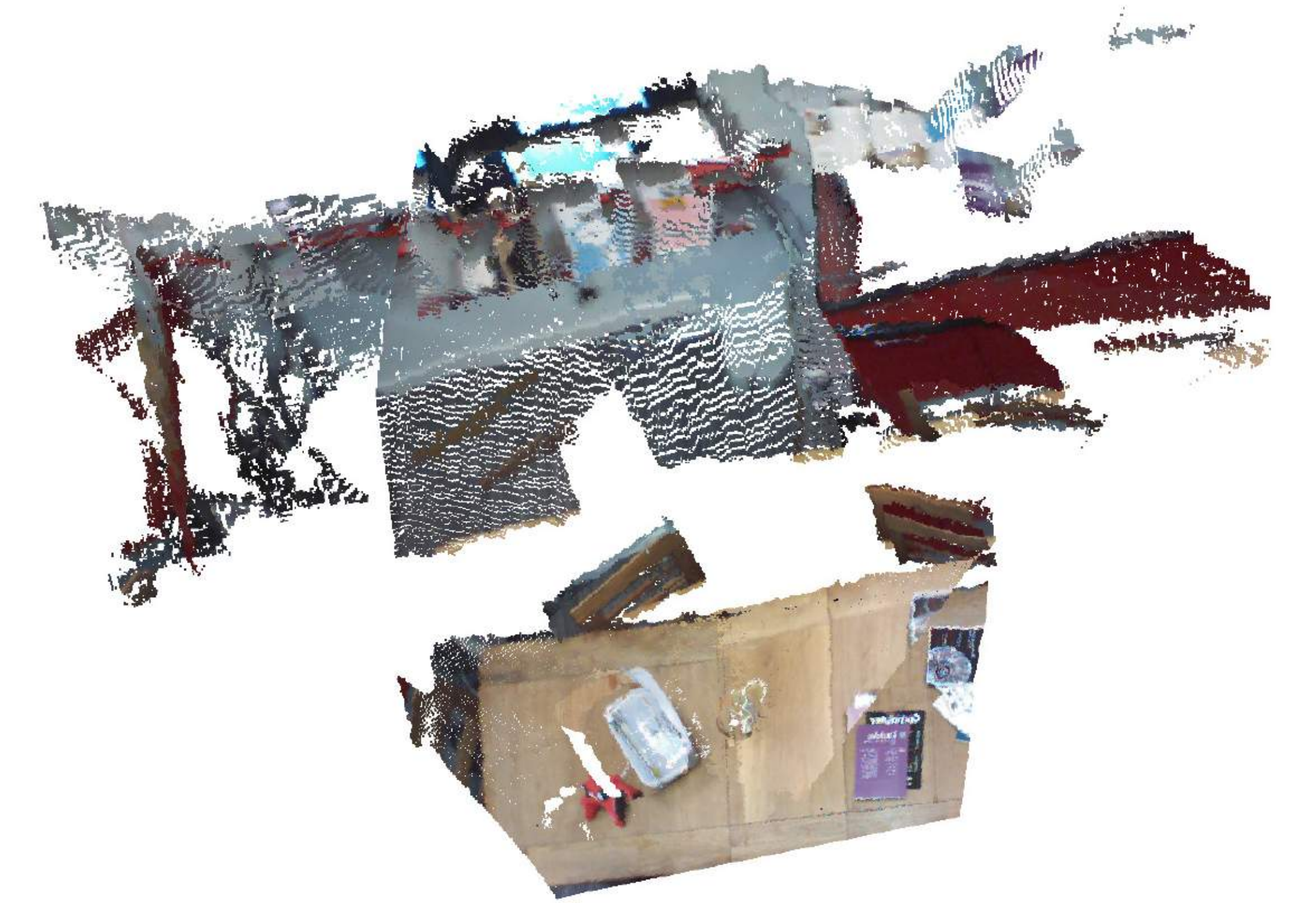}
\end{minipage}\,\,
\\

 & \footnotesize{$N=2002$} && \footnotesize{\textcolor[rgb]{0,0.8,0}{Succeed}$,0.288,0.293s$} &\footnotesize{\textcolor[rgb]{0,0.8,0}{Succeed}$,0.502,19.645s$}&\footnotesize{\textcolor[rgb]{0,0.8,0}{Succeed}$,0.292,70.343s$}&\footnotesize{\textcolor[rgb]{0,0.8,0}{Succeed}$,\textbf{0.287},\textbf{0.278}s$}

\\
\rotatebox{90}{\,\,\footnotesize{\textit{chess}}\,}\,
&
\,\,
\begin{minipage}[t]{0.1\linewidth}
\centering
\includegraphics[width=1\linewidth]{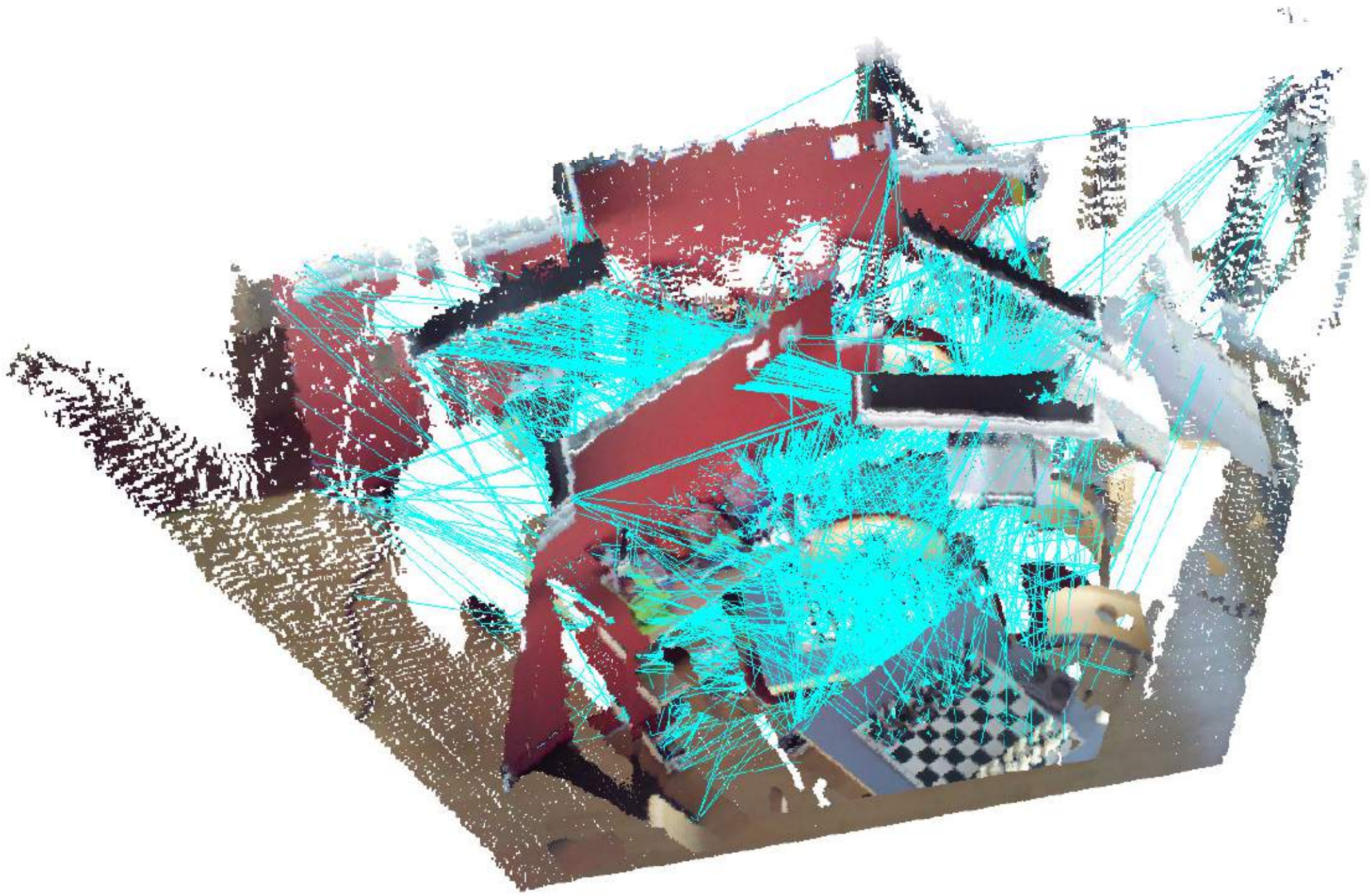}
\end{minipage}\,\,
& &
\,\,
\begin{minipage}[t]{0.19\linewidth}
\centering
\includegraphics[width=.48\linewidth]{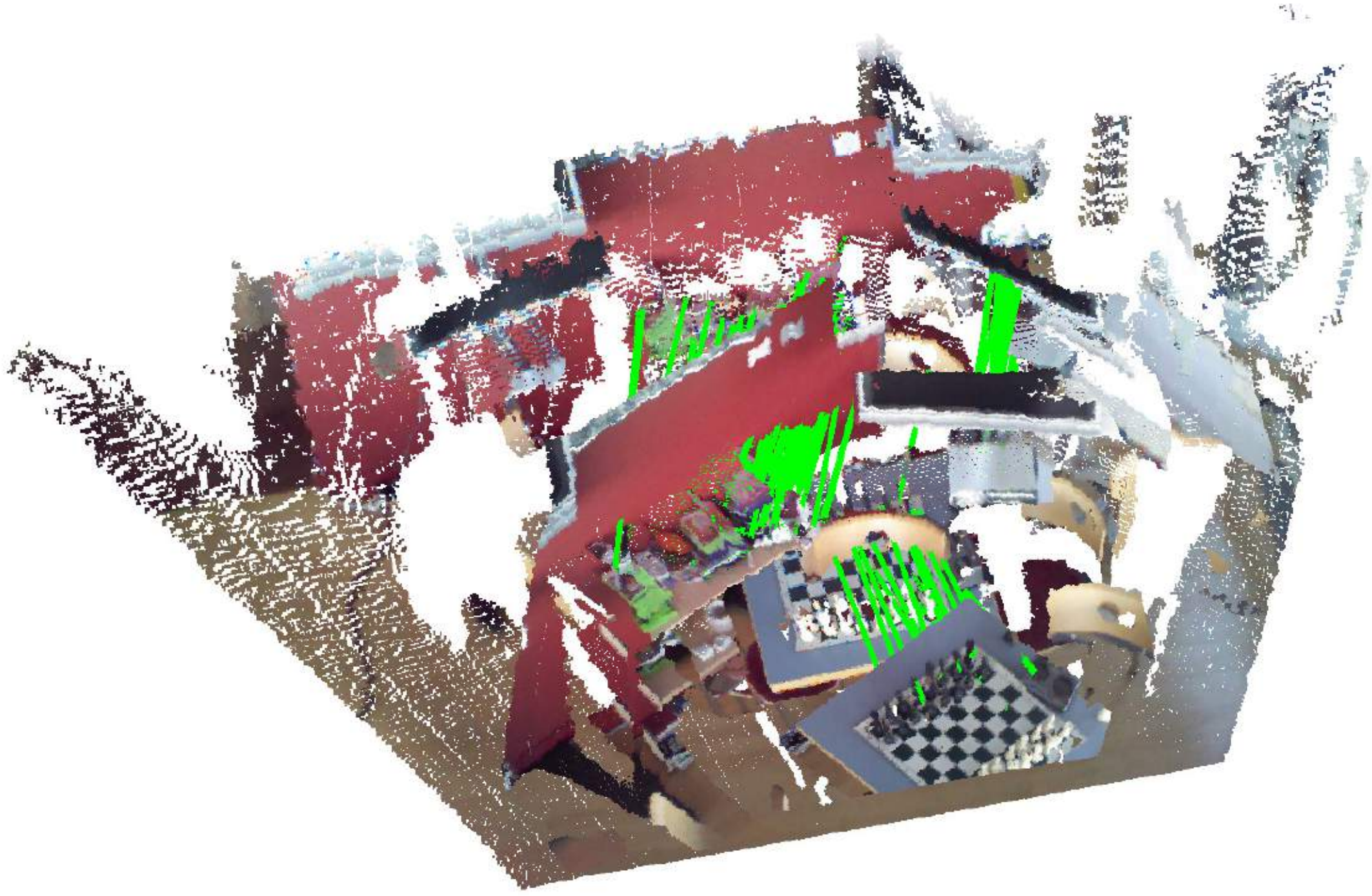}
\includegraphics[width=.48\linewidth]{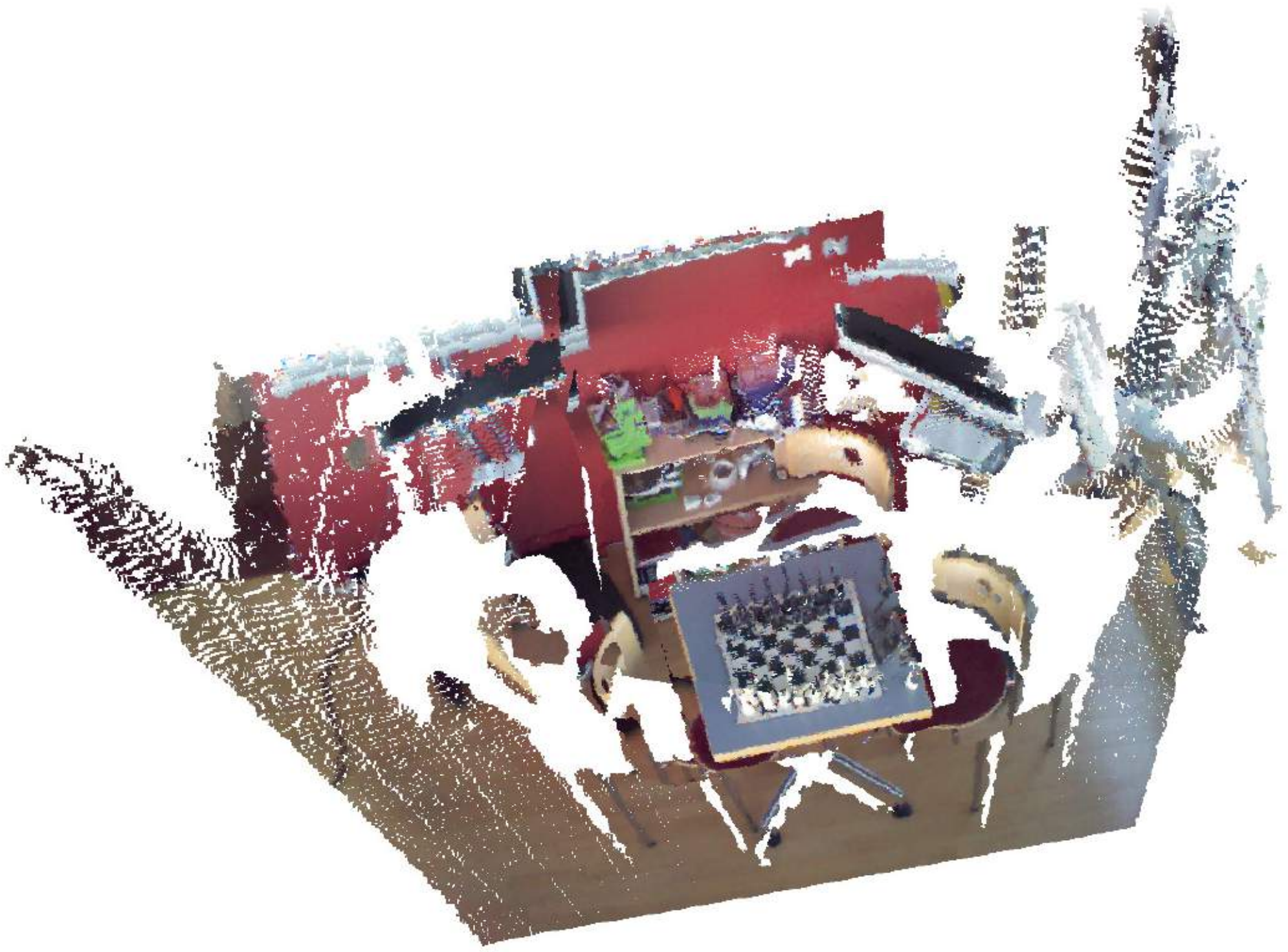}
\end{minipage}\,\,
&
\,\,
\begin{minipage}[t]{0.19\linewidth}
\centering
\includegraphics[width=.48\linewidth]{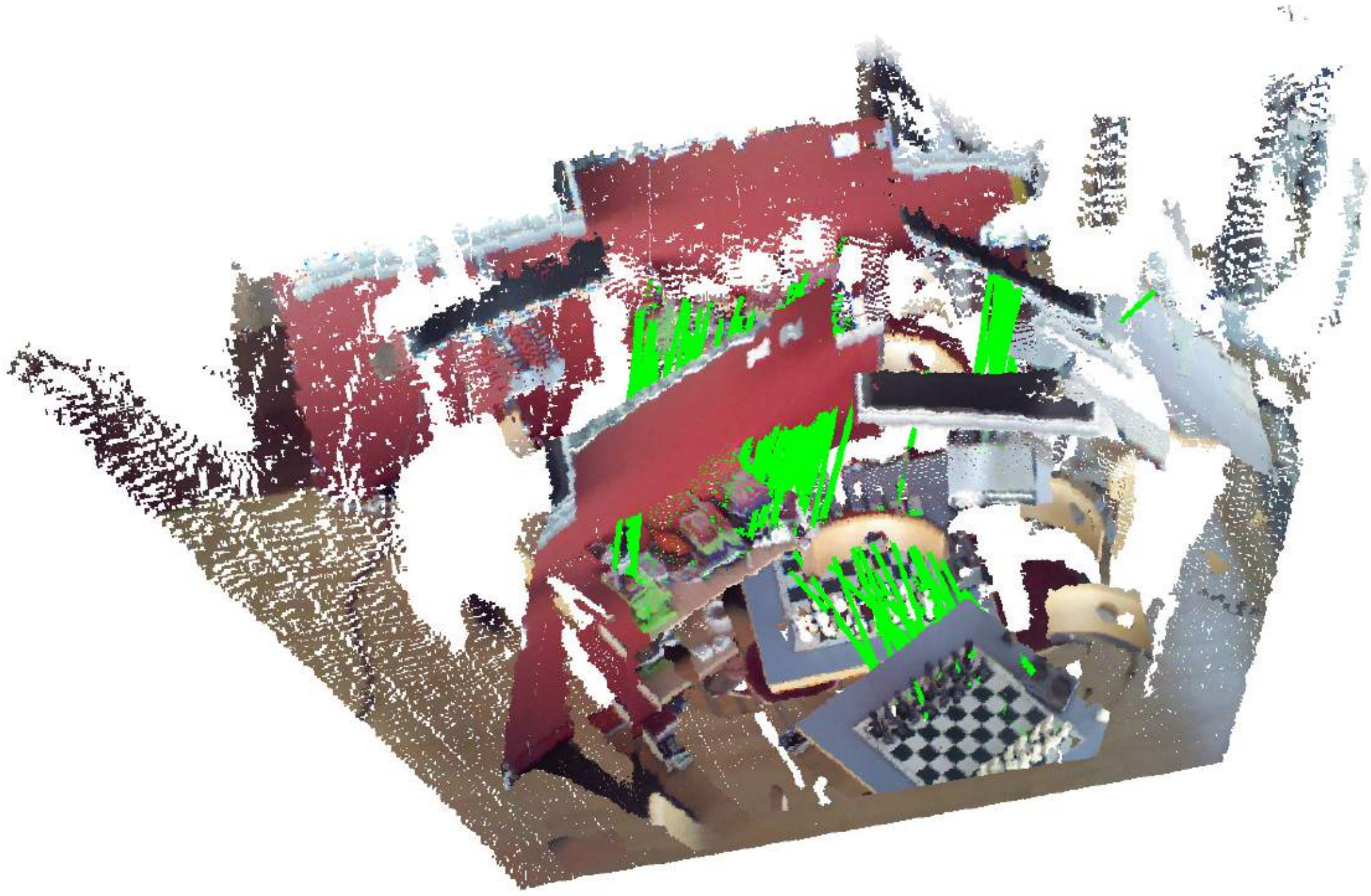}
\includegraphics[width=.48\linewidth]{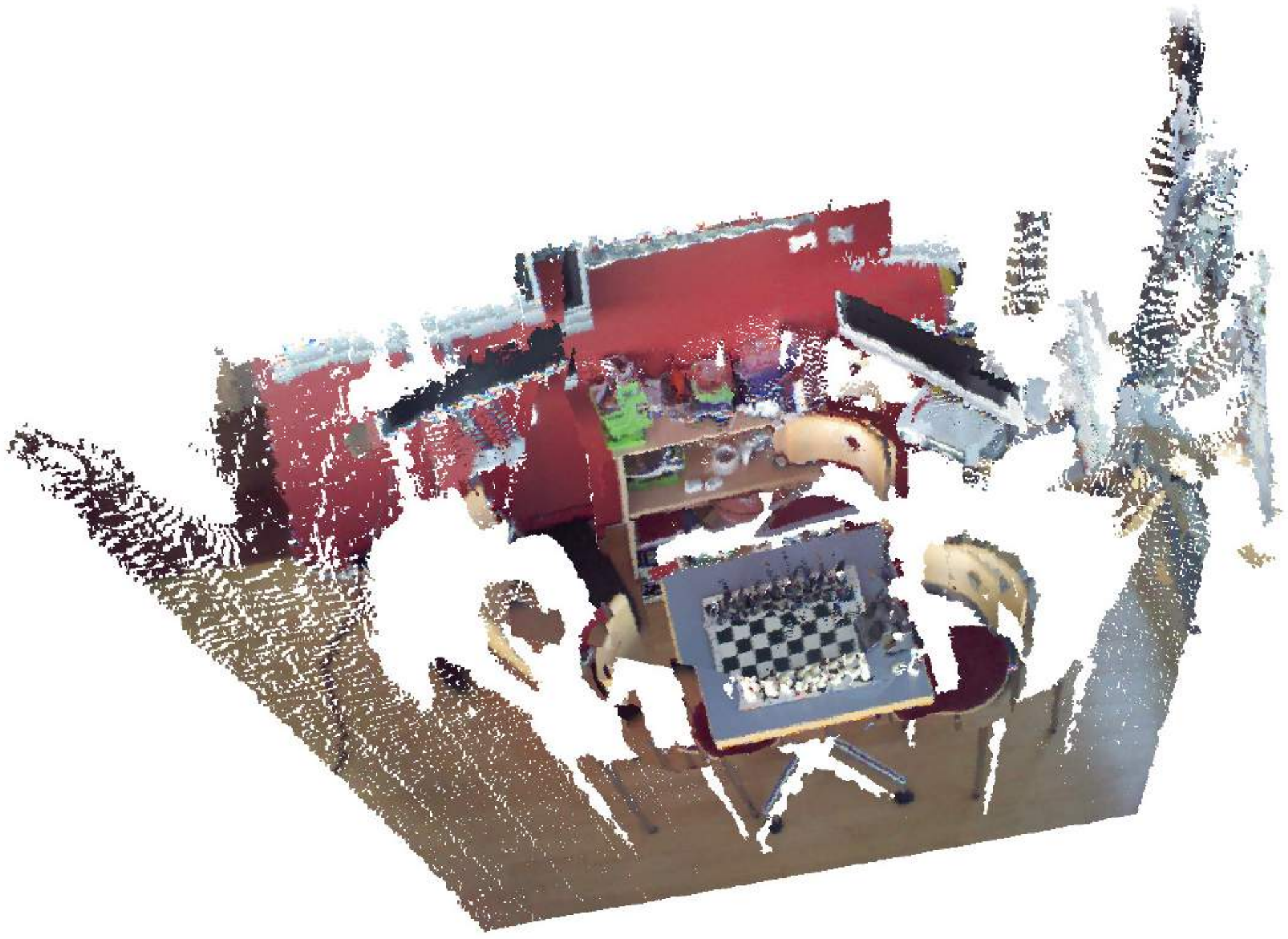}
\end{minipage}\,\,
&
\,\,
\begin{minipage}[t]{0.19\linewidth}
\centering
\includegraphics[width=.48\linewidth]{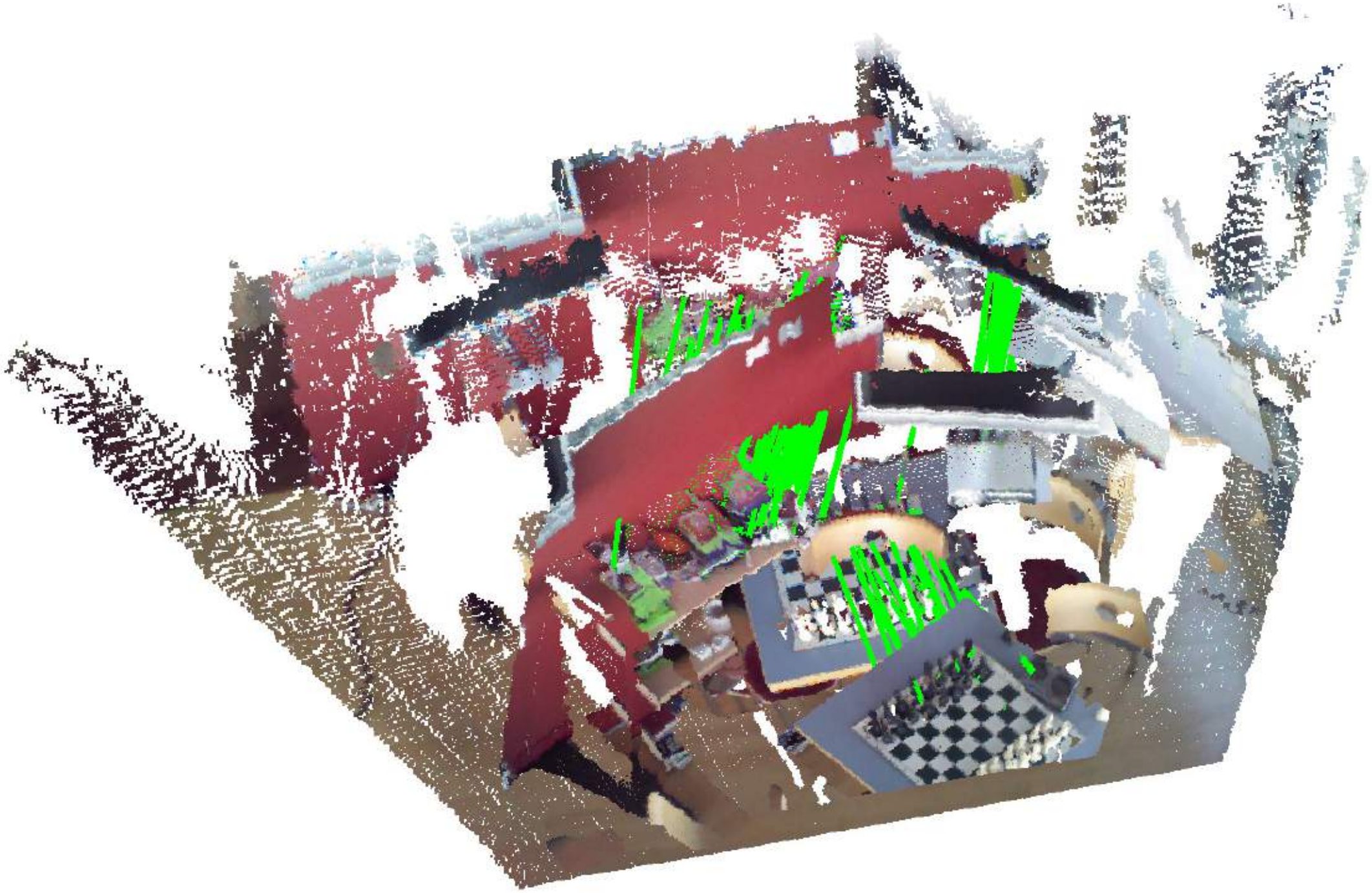}
\includegraphics[width=.48\linewidth]{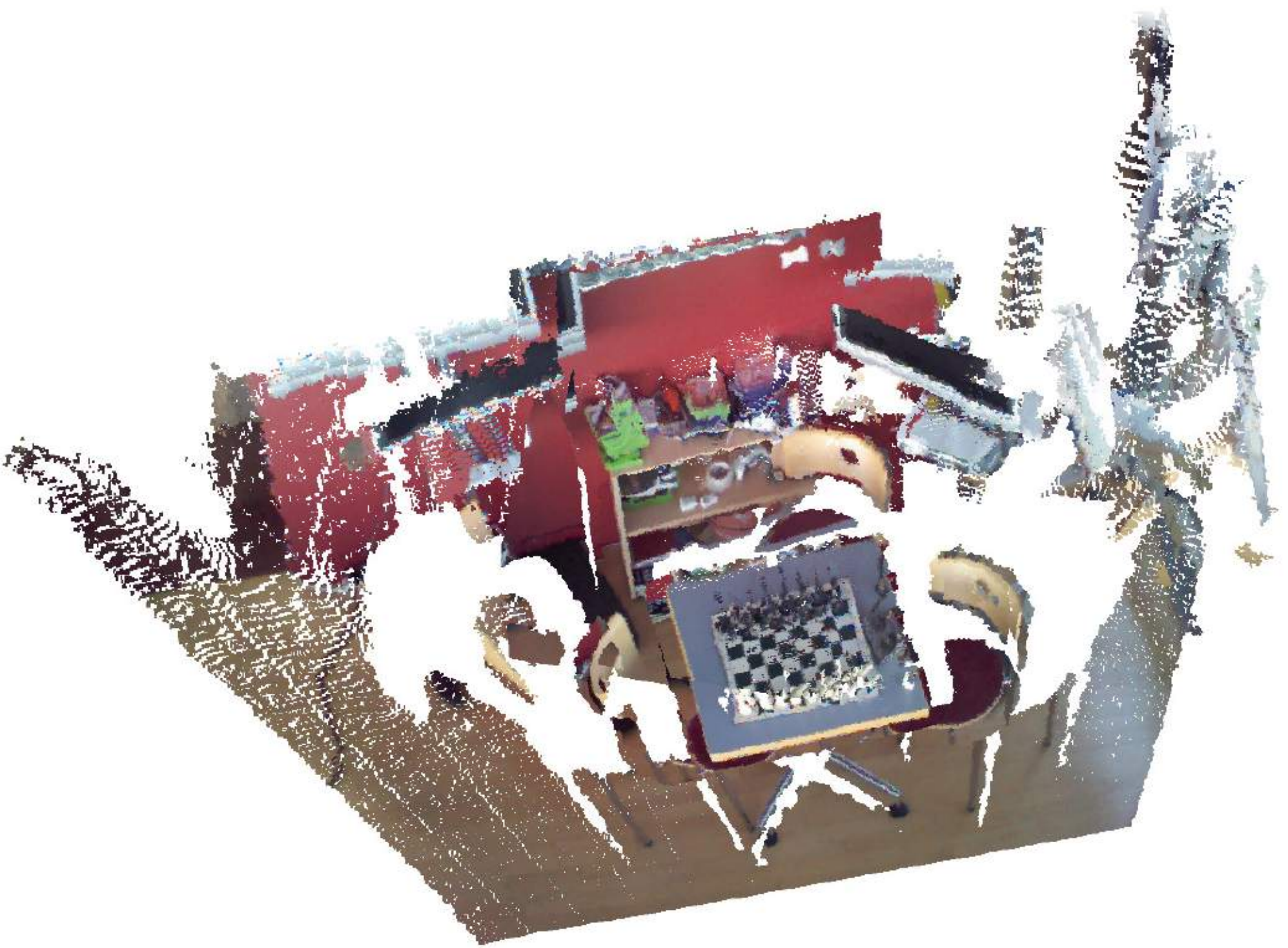}
\end{minipage}\,\,
&
\,\,
\begin{minipage}[t]{0.19\linewidth}
\centering
\includegraphics[width=.48\linewidth]{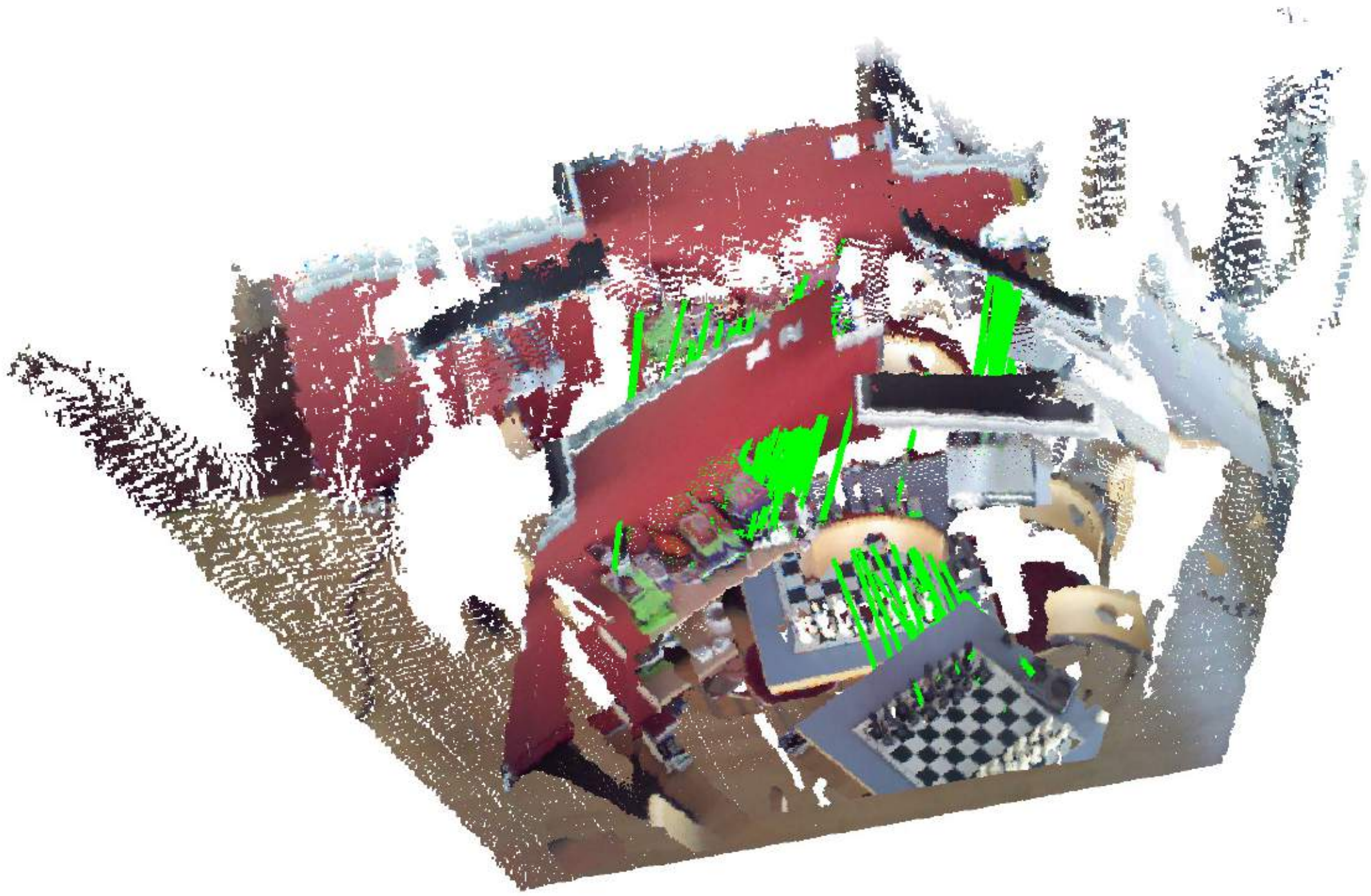}
\includegraphics[width=.48\linewidth]{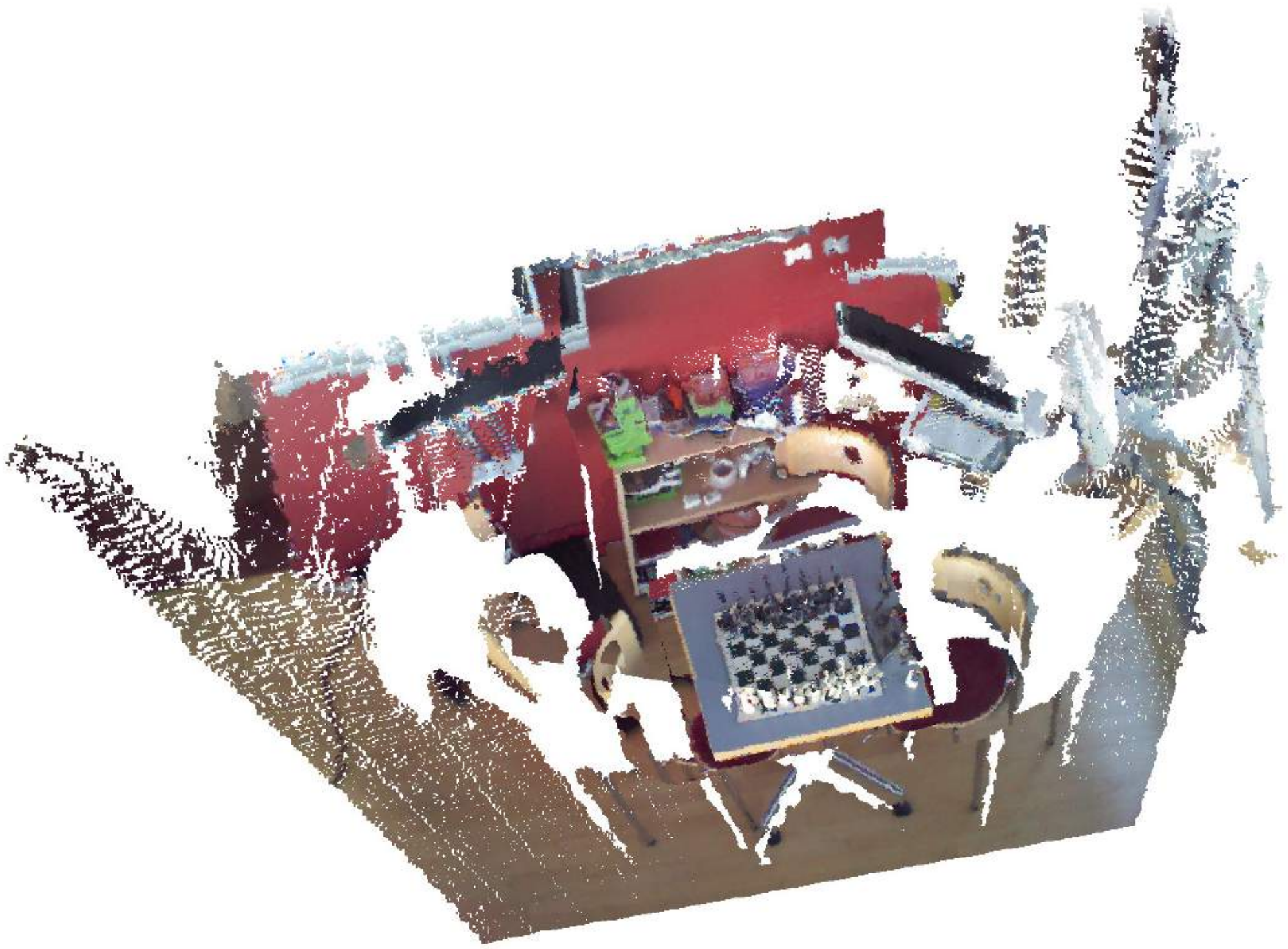}
\end{minipage}\,\,
\\

 & \footnotesize{$N=1822$} && \footnotesize{\textcolor[rgb]{1,0,0}{Fail}$,\verb|\|,0.262s$} &\footnotesize{\textcolor[rgb]{1,0,0}{Fail}$,\verb|\|,48.811$}&\footnotesize{\textcolor[rgb]{0,0.8,0}{Succeed}$,0.326,12.171s$}&\footnotesize{\textcolor[rgb]{0,0.8,0}{Succeed}$,\textbf{0.324},\textbf{0.784}s$}

\\
\rotatebox{90}{\,\,\footnotesize{\textit{office}}\,}\,
&
\,\,
\begin{minipage}[t]{0.1\linewidth}
\centering
\includegraphics[width=1\linewidth]{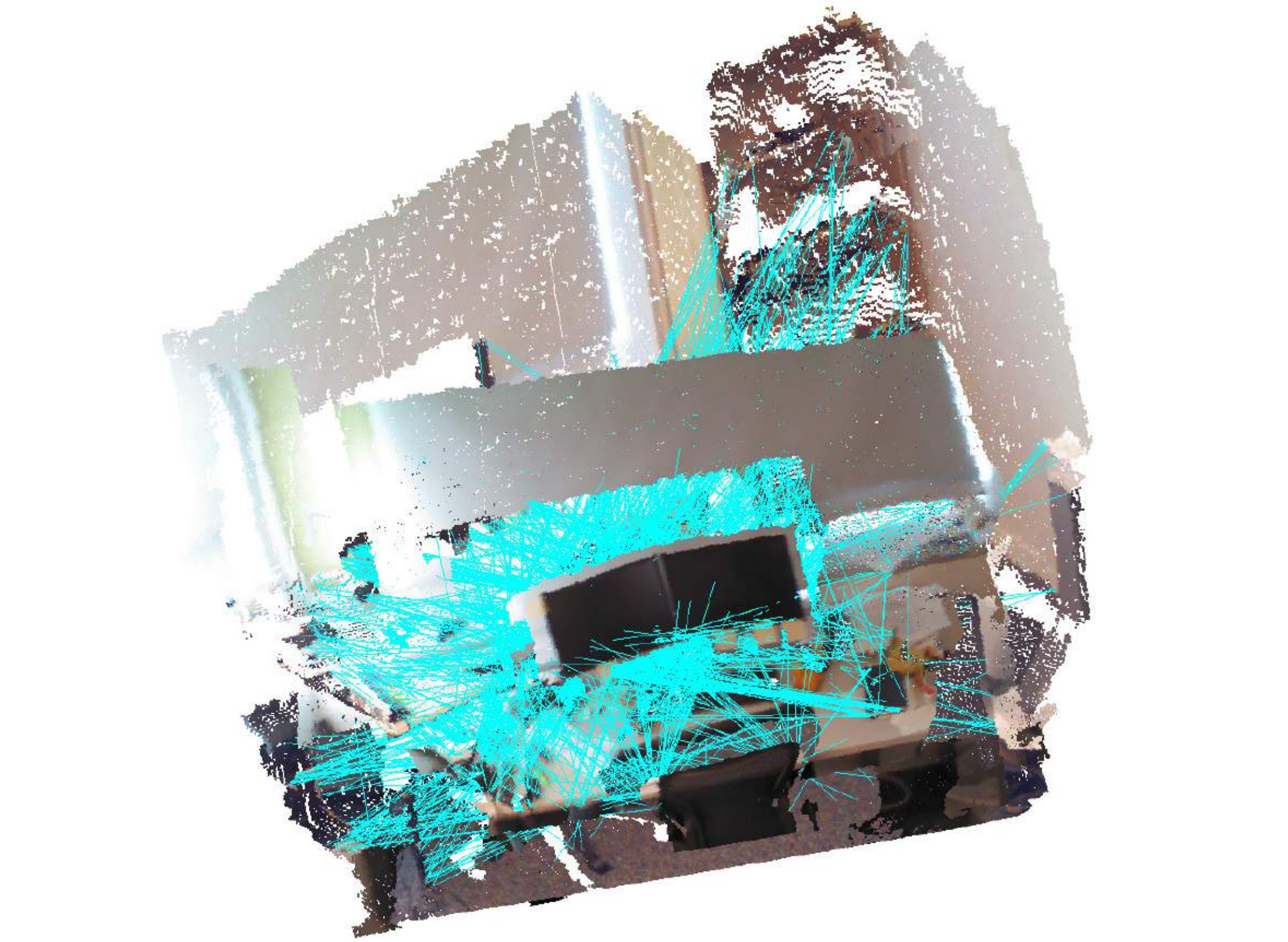}
\end{minipage}\,\,
& &
\,\,
\begin{minipage}[t]{0.19\linewidth}
\centering
\includegraphics[width=.48\linewidth]{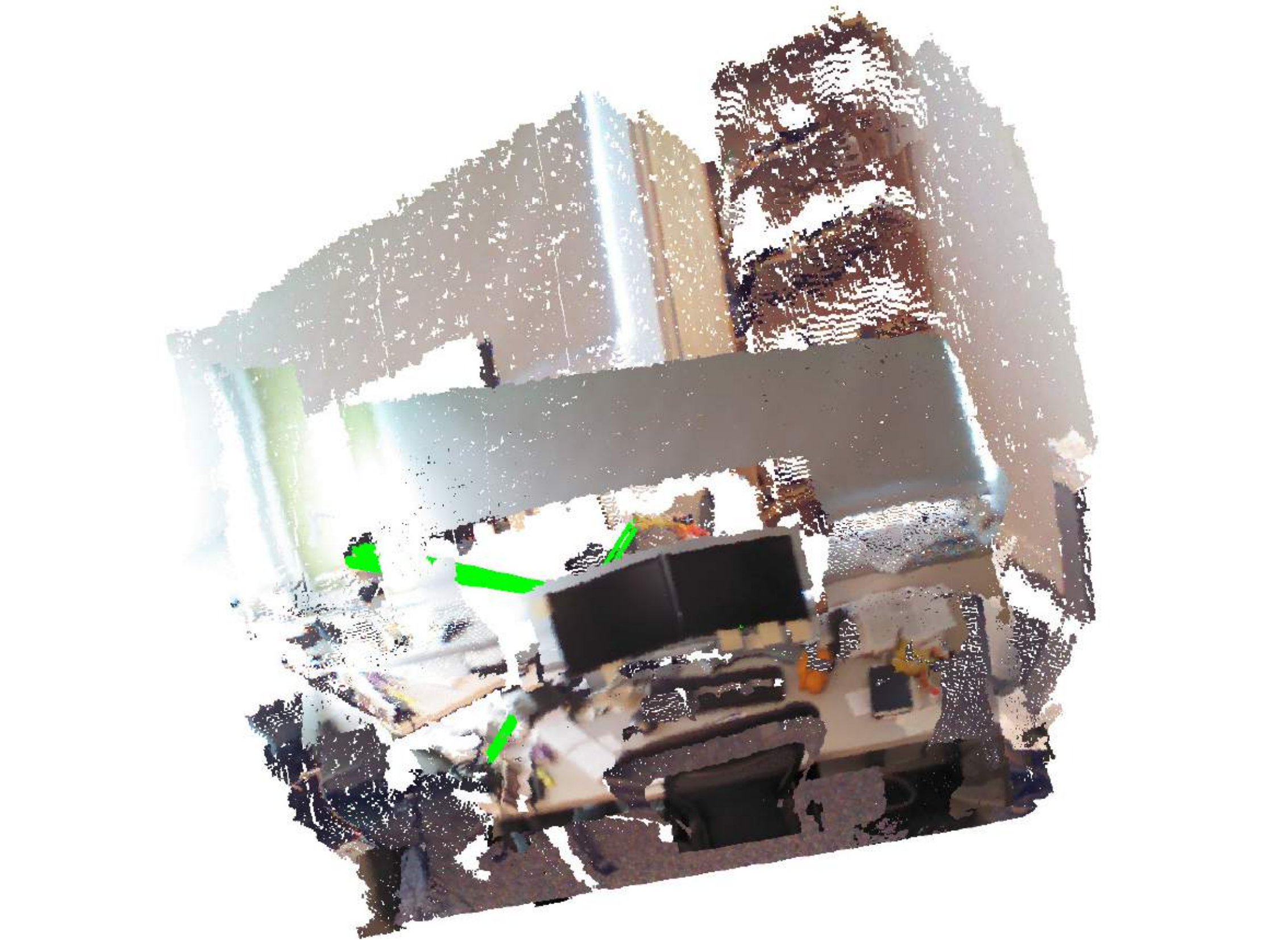}
\includegraphics[width=.48\linewidth]{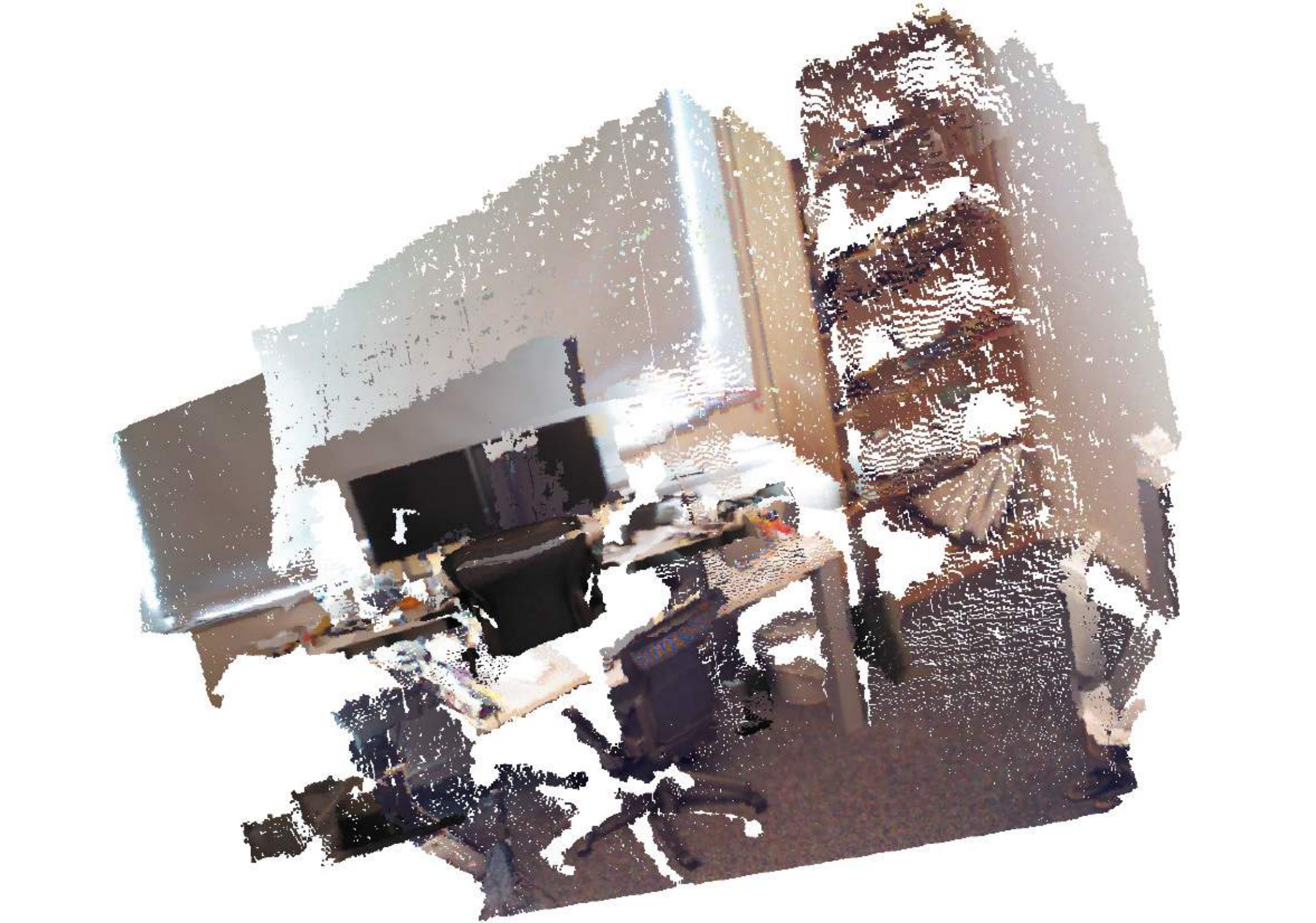}
\end{minipage}\,\,
&
\,\,
\begin{minipage}[t]{0.19\linewidth}
\centering
\includegraphics[width=.48\linewidth]{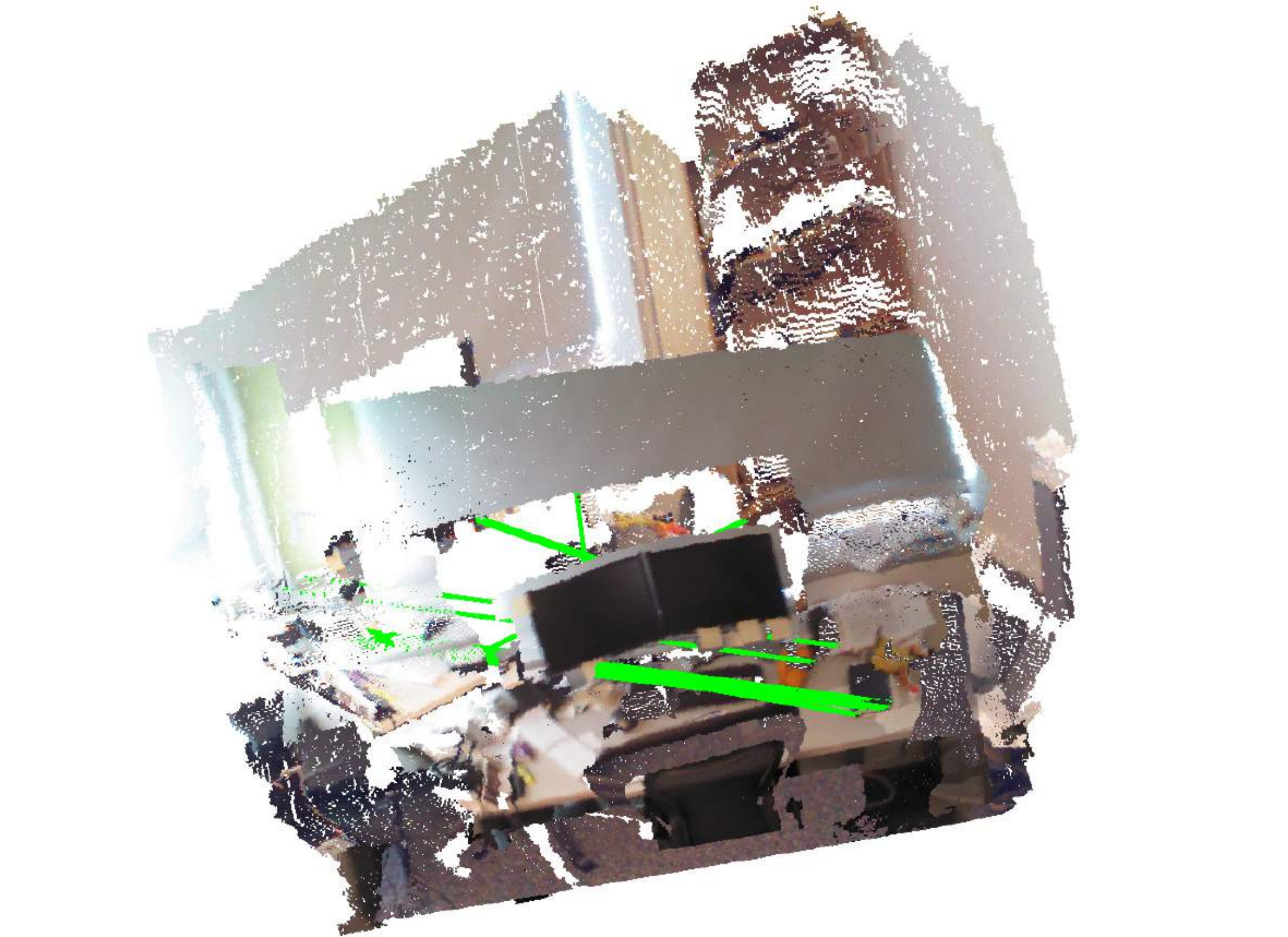}
\includegraphics[width=.48\linewidth]{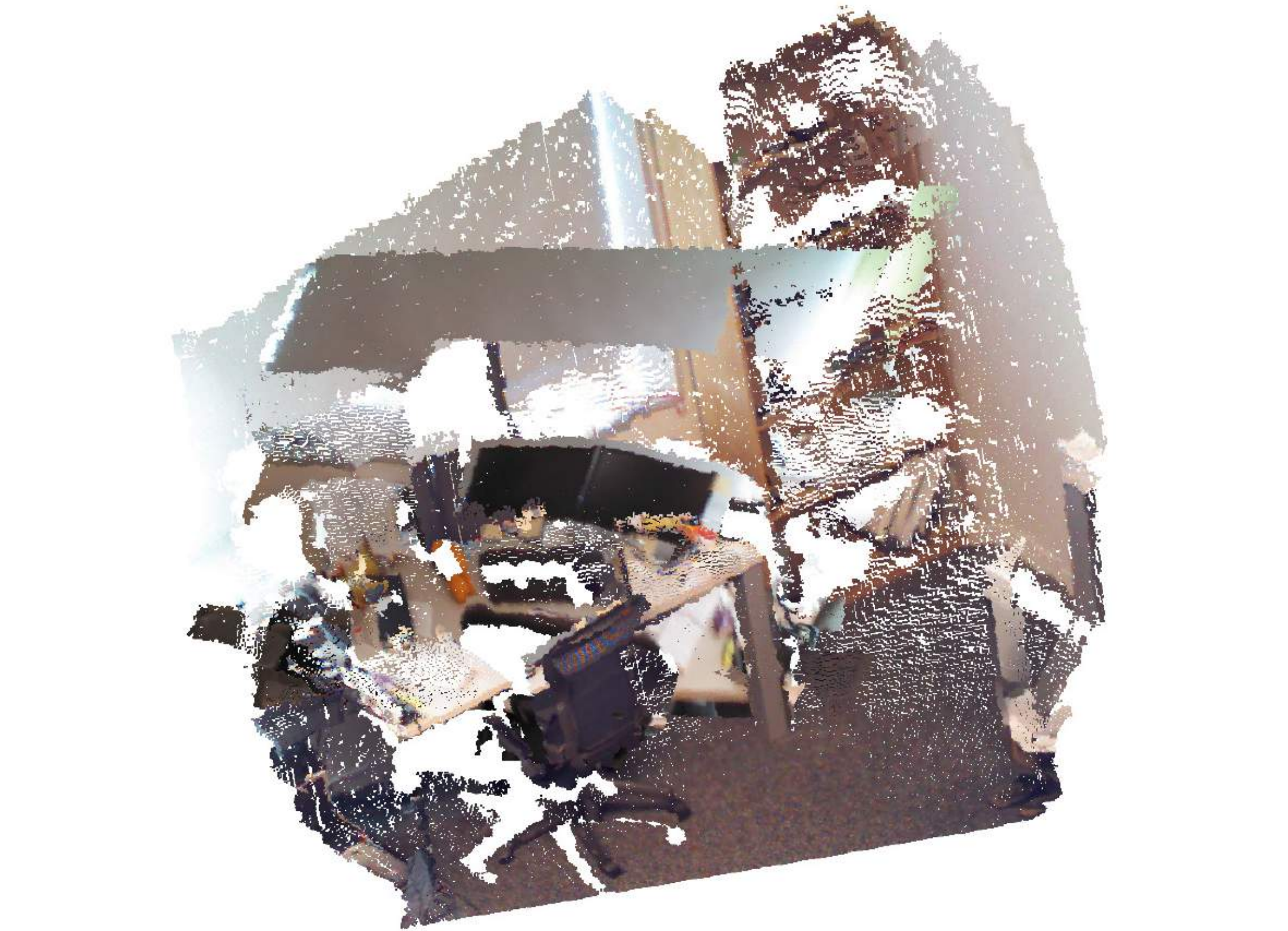}
\end{minipage}\,\,
&
\,\,
\begin{minipage}[t]{0.19\linewidth}
\centering
\includegraphics[width=.48\linewidth]{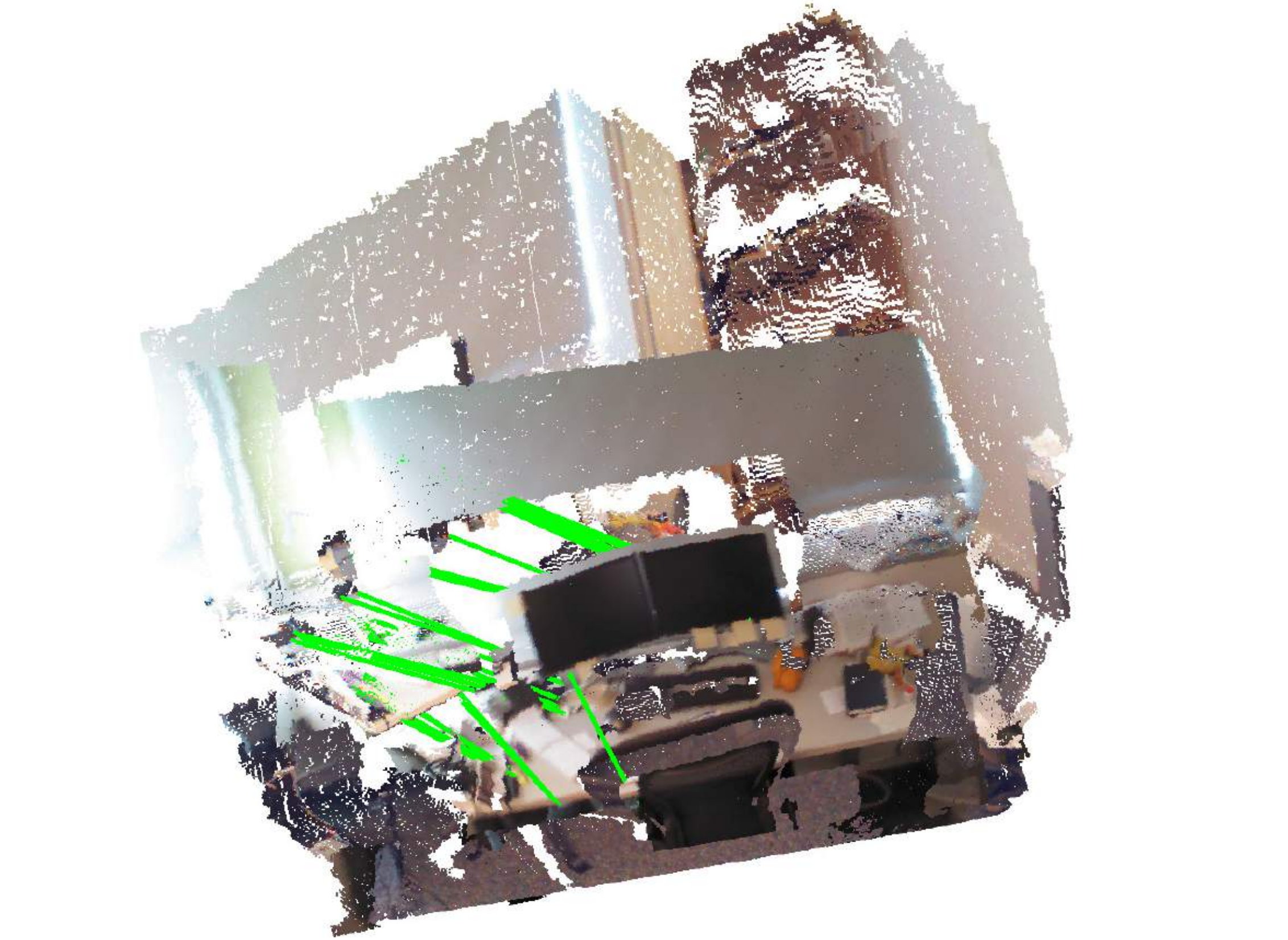}
\includegraphics[width=.48\linewidth]{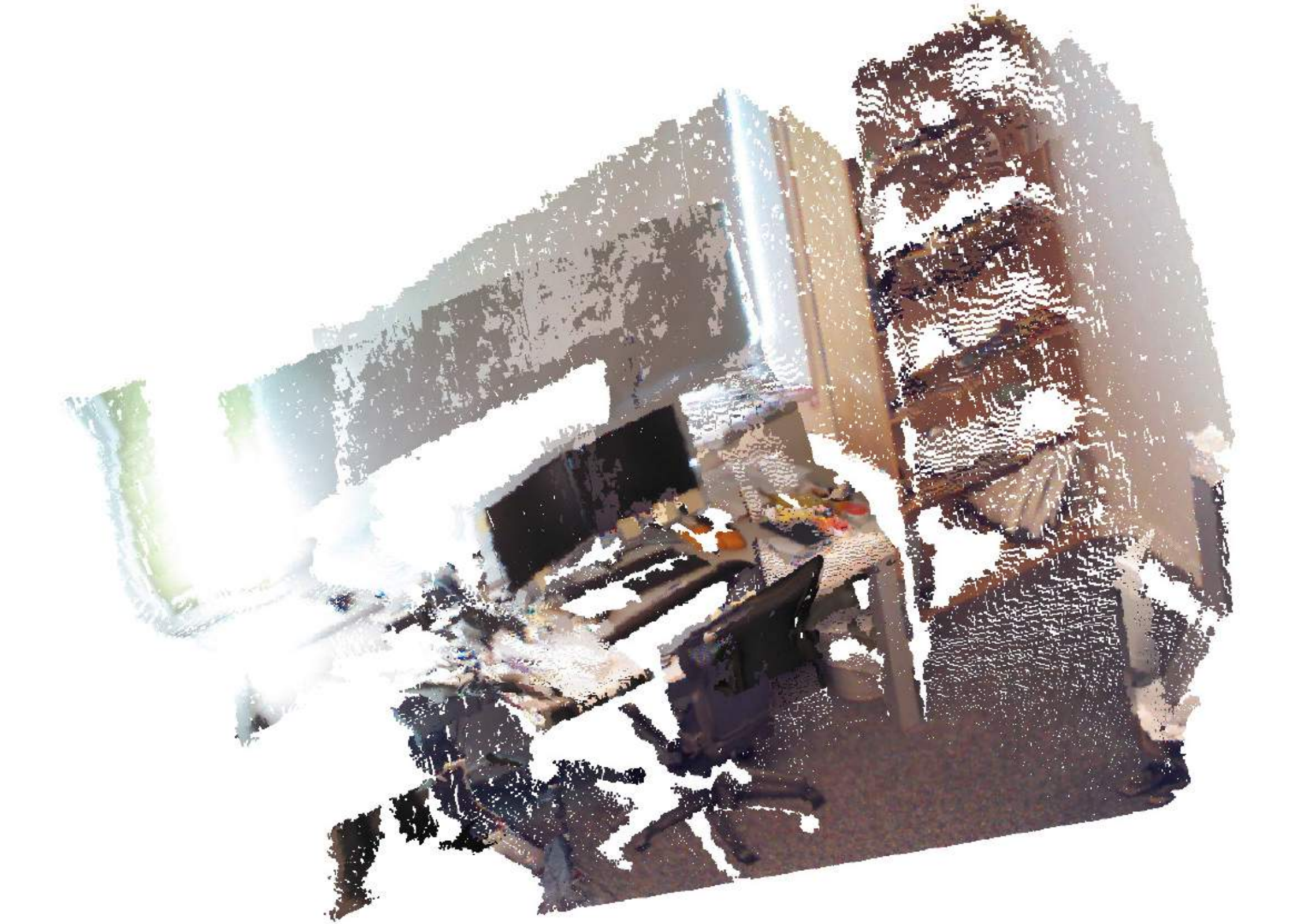}
\end{minipage}\,\,
&
\,\,
\begin{minipage}[t]{0.19\linewidth}
\centering
\includegraphics[width=.48\linewidth]{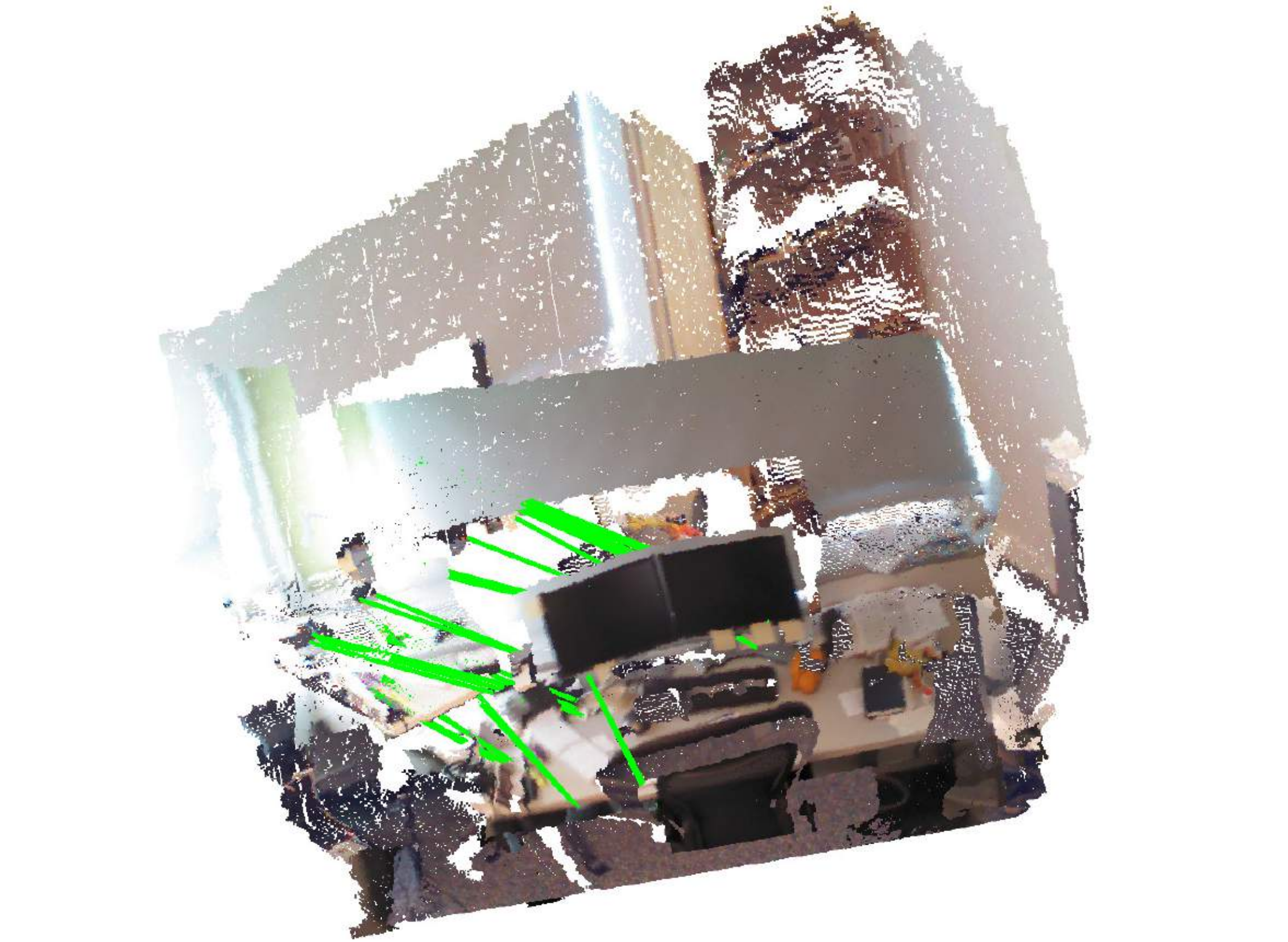}
\includegraphics[width=.48\linewidth]{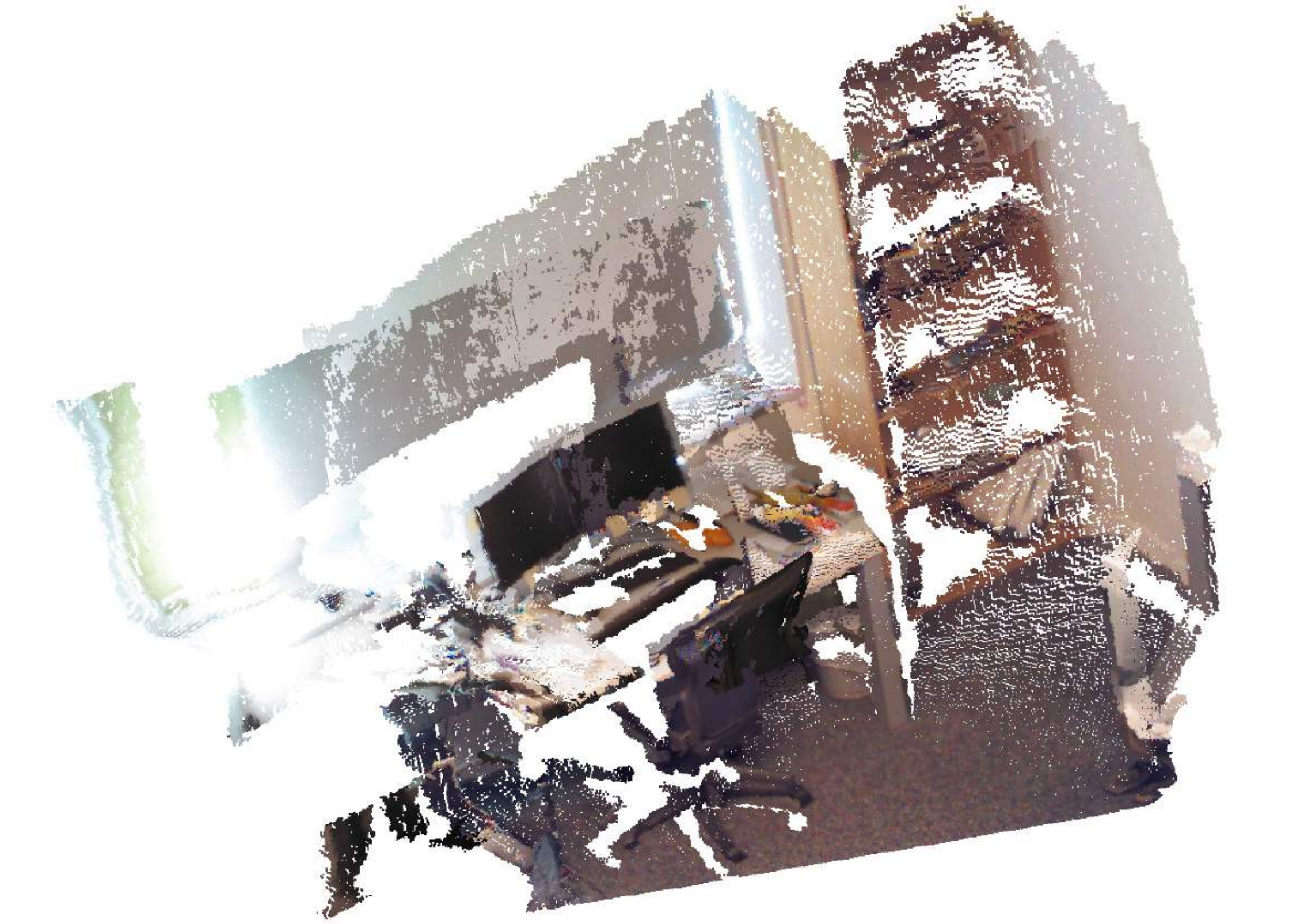}
\end{minipage}\,\,
\\

 & \footnotesize{$N=985$} && \footnotesize{\textcolor[rgb]{1,0,0}{Fail}$,\verb|\|,0.112s$} &\footnotesize{\textcolor[rgb]{0,0.8,0}{Succeed}$,0.373,26.628$}&\footnotesize{\textcolor[rgb]{0,0.8,0}{Succeed}$,0.360,9.035s$}&\footnotesize{\textcolor[rgb]{0,0.8,0}{Succeed}$,\textbf{0.334},\textbf{0.249}s$}

\\
\rotatebox{90}{\,\,\footnotesize{\textit{office}}\,}\,
&
\,\,
\begin{minipage}[t]{0.1\linewidth}
\centering
\includegraphics[width=1\linewidth]{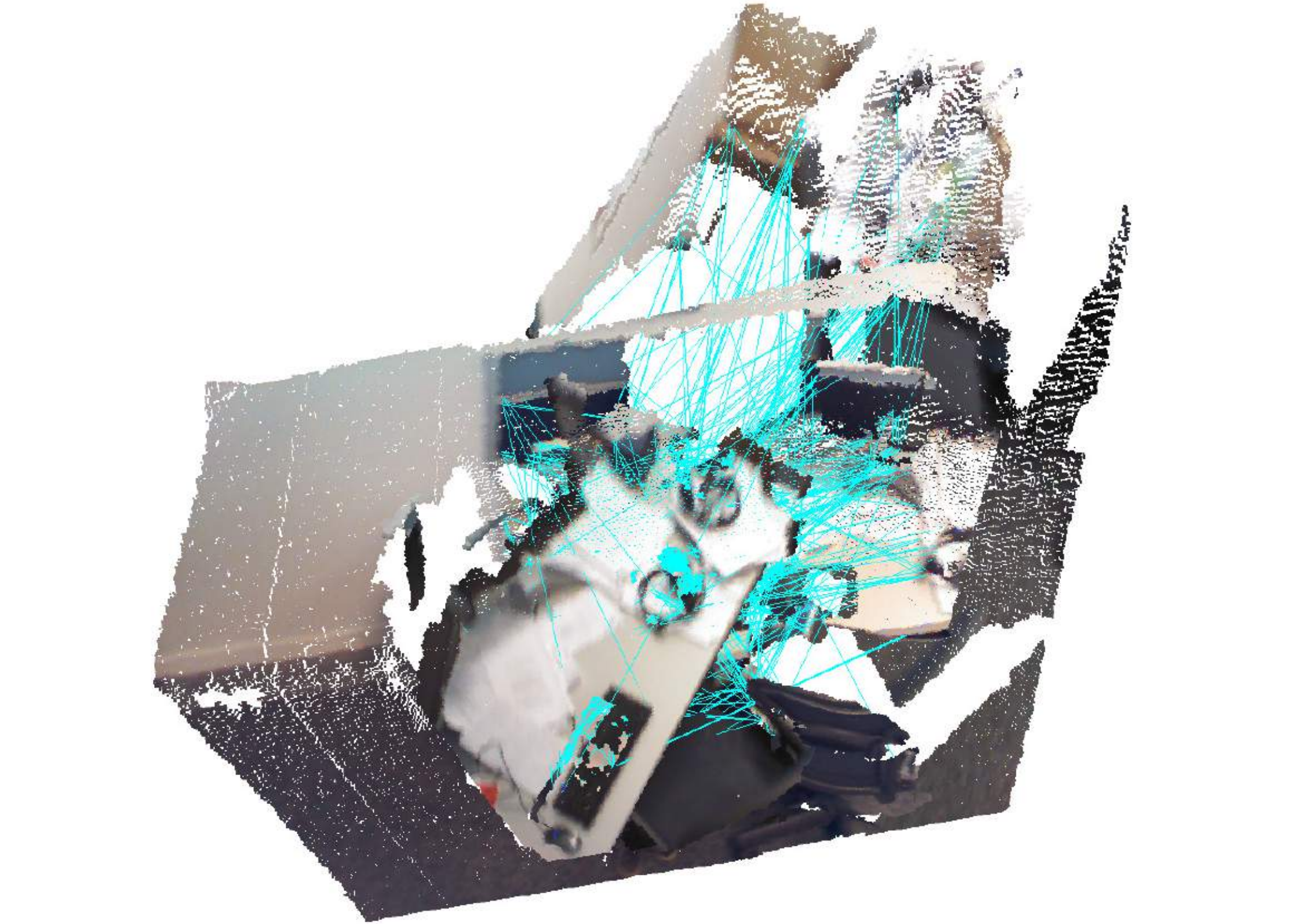}
\end{minipage}\,\,
& &
\,\,
\begin{minipage}[t]{0.19\linewidth}
\centering
\includegraphics[width=.48\linewidth]{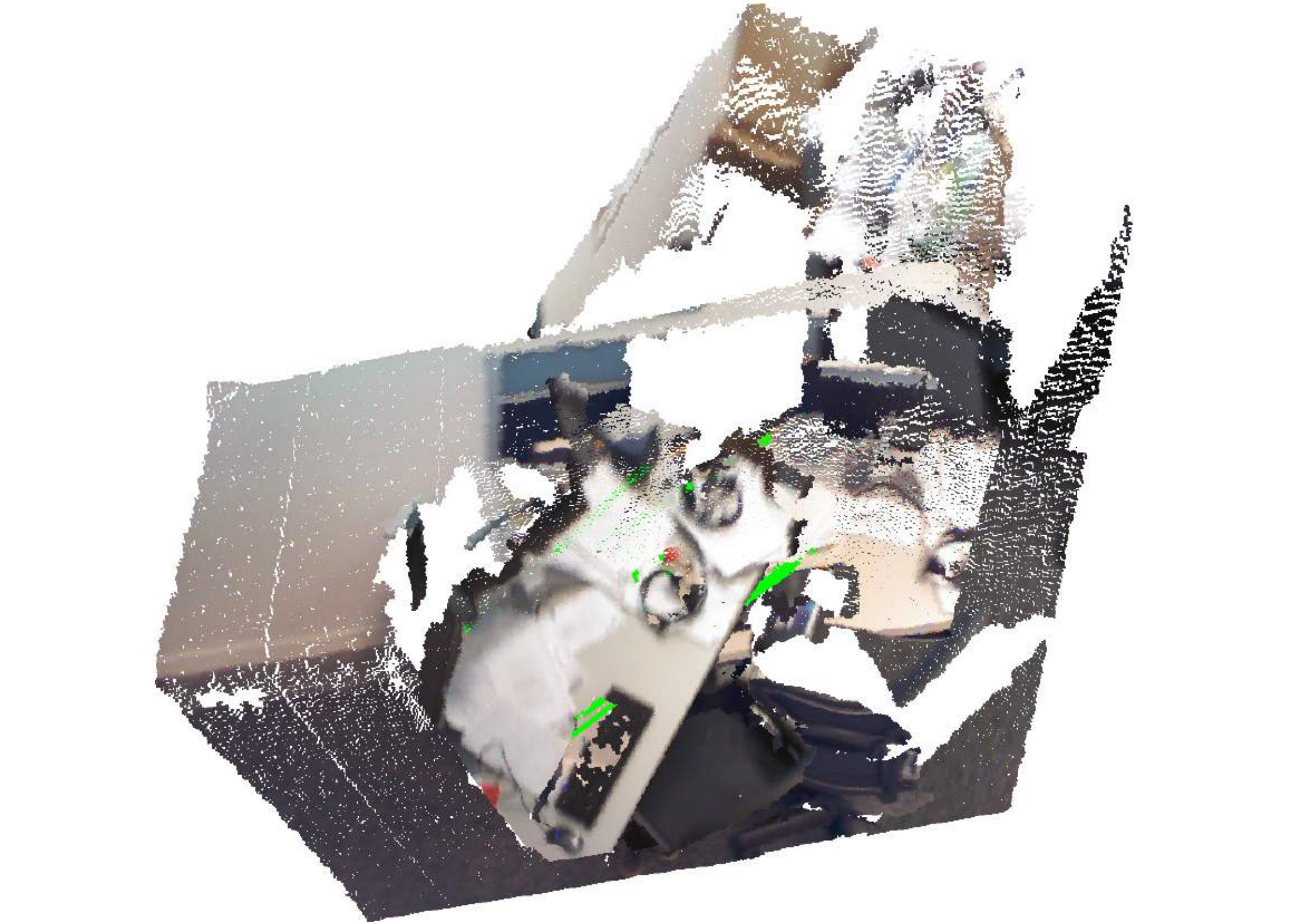}
\includegraphics[width=.48\linewidth]{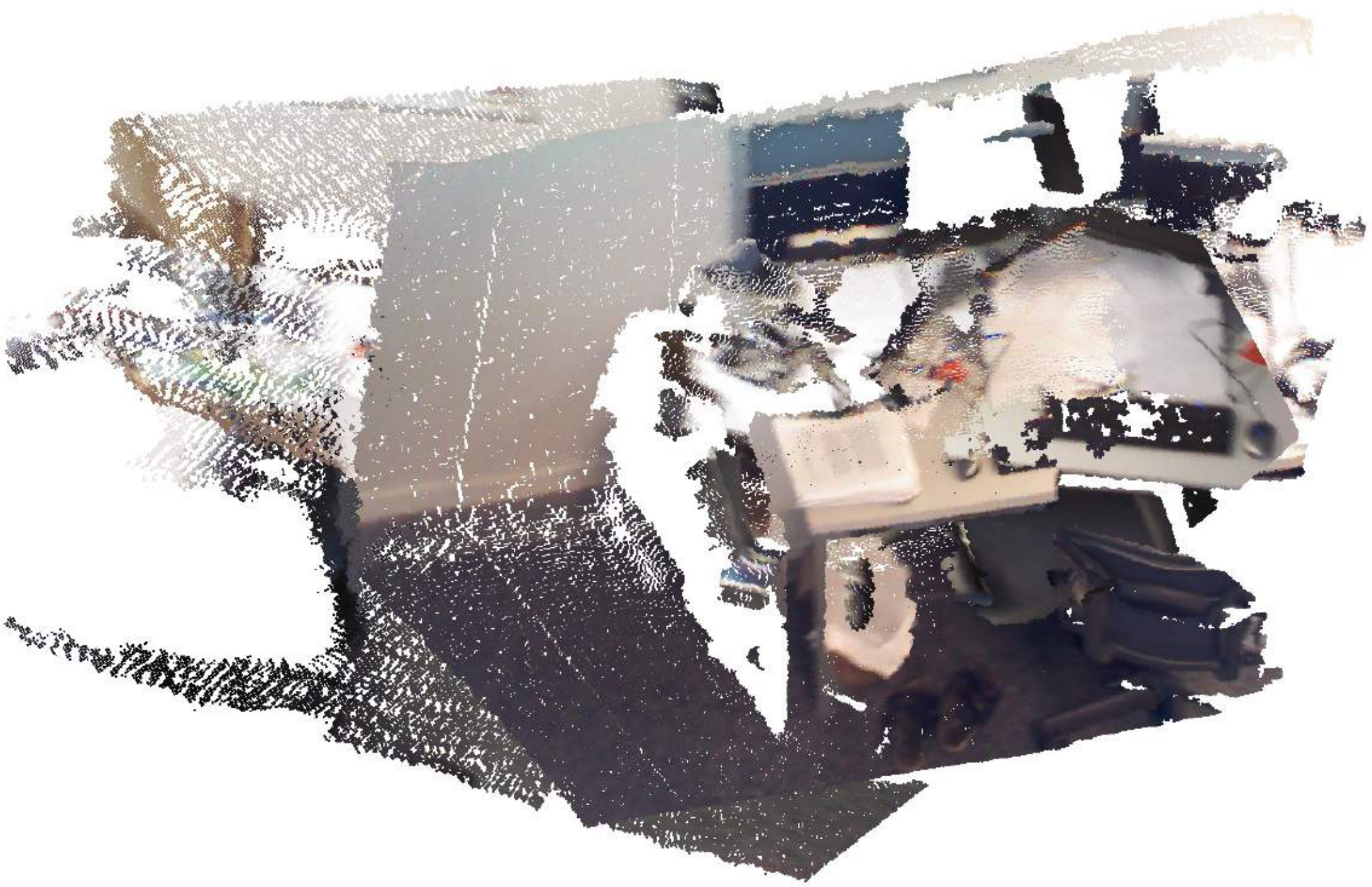}
\end{minipage}\,\,
&
\,\,
\begin{minipage}[t]{0.19\linewidth}
\centering
\includegraphics[width=.48\linewidth]{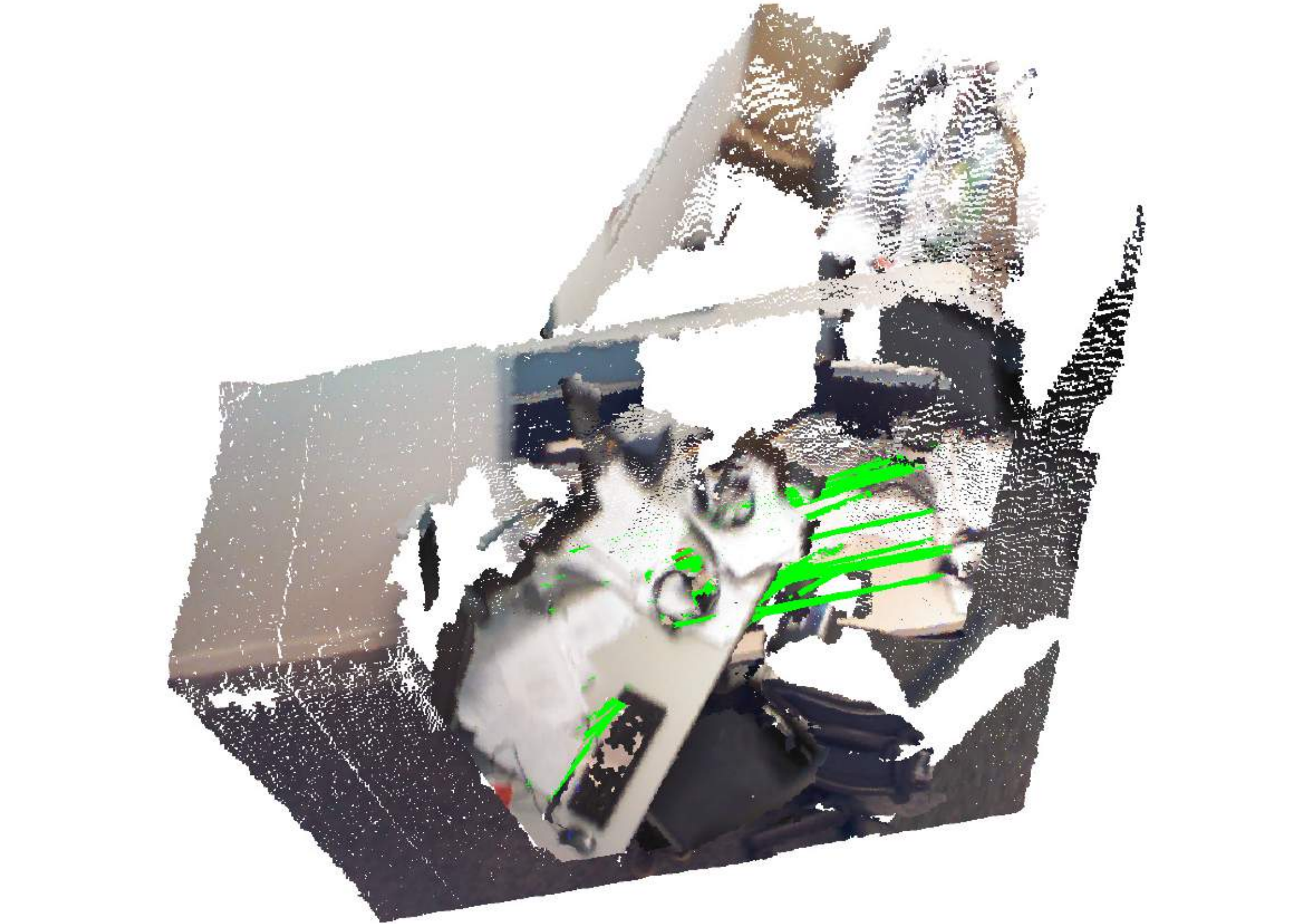}
\includegraphics[width=.48\linewidth]{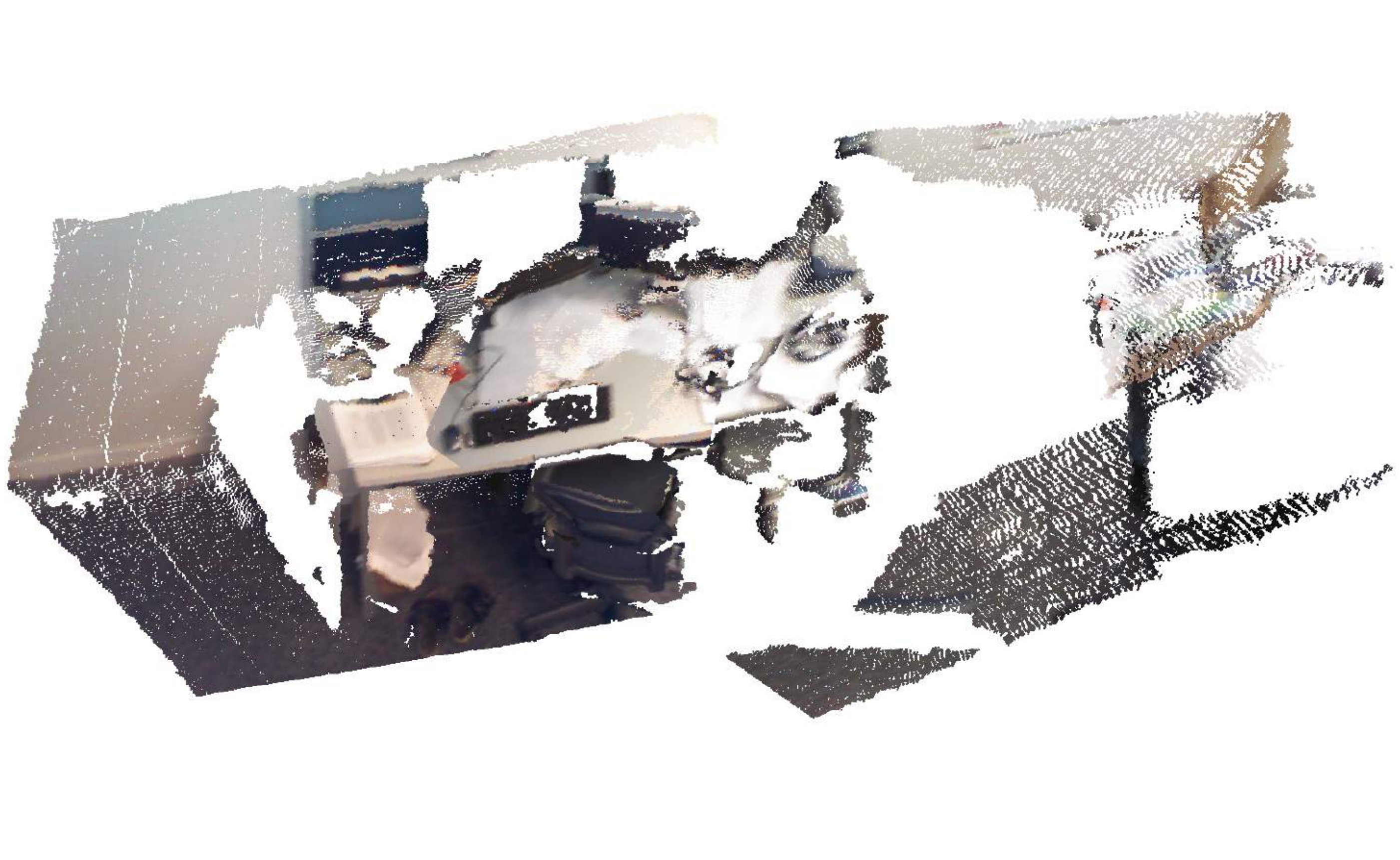}
\end{minipage}\,\,
&
\,\,
\begin{minipage}[t]{0.19\linewidth}
\centering
\includegraphics[width=.48\linewidth]{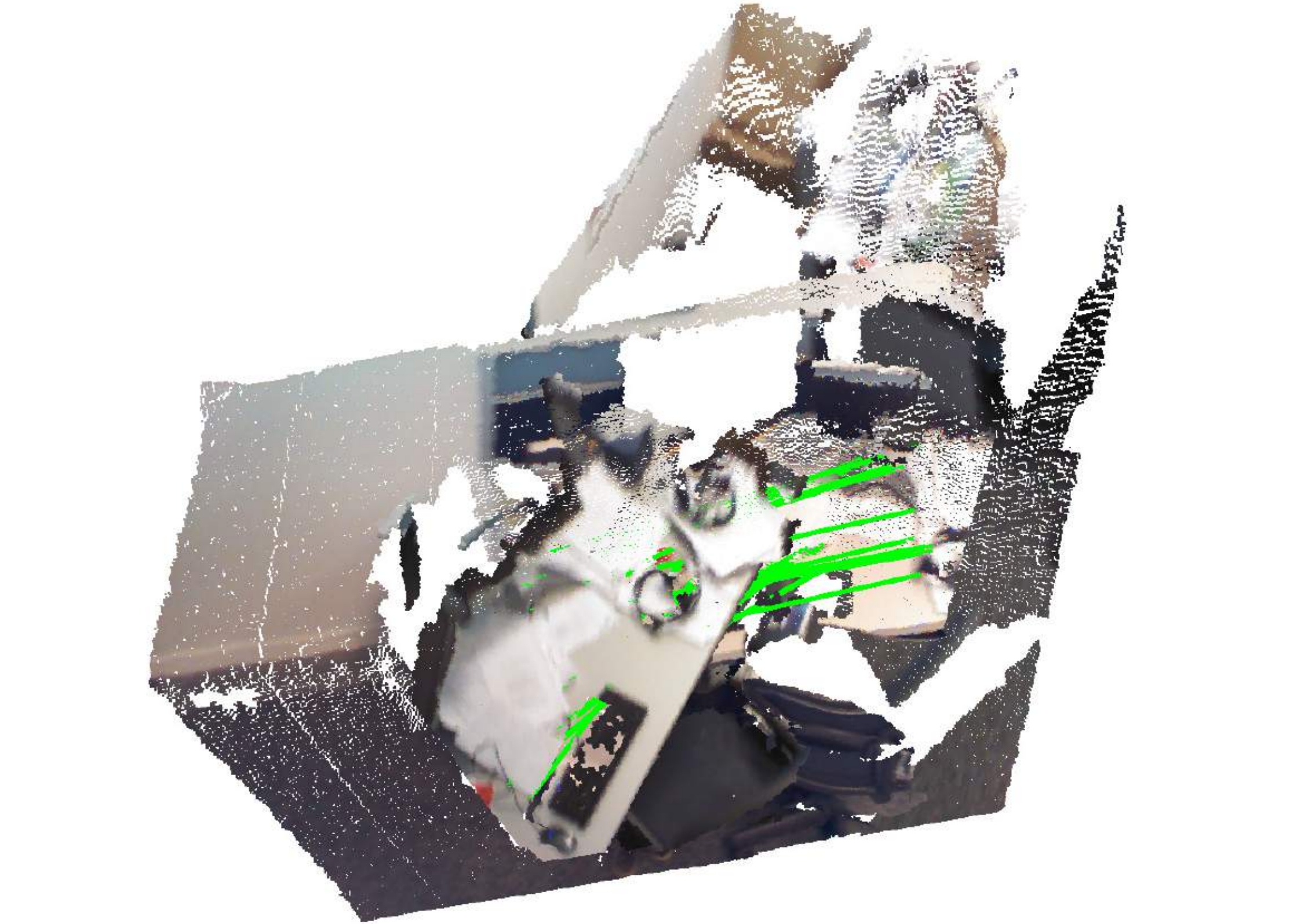}
\includegraphics[width=.48\linewidth]{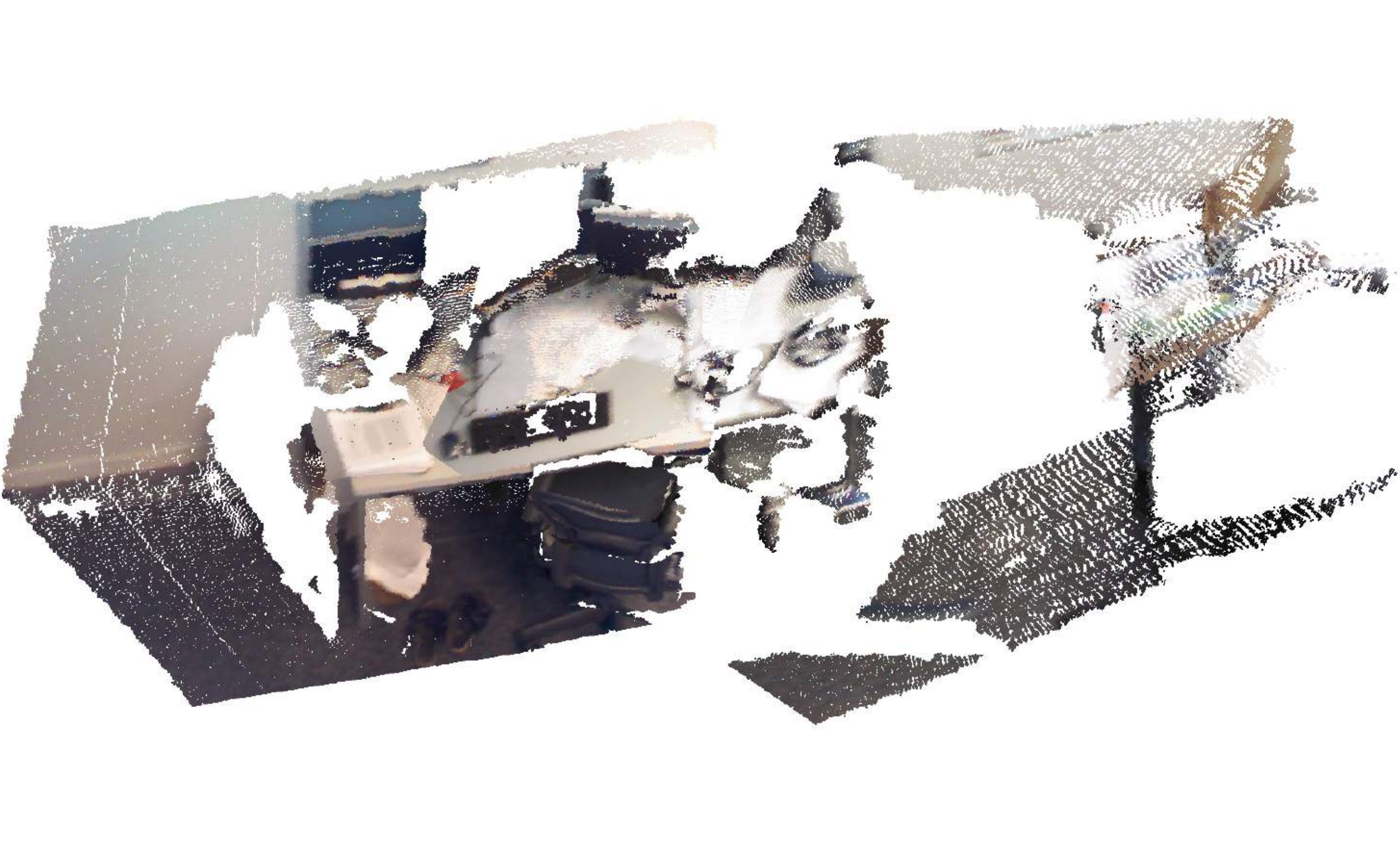}
\end{minipage}\,\,
&
\,\,
\begin{minipage}[t]{0.19\linewidth}
\centering
\includegraphics[width=.48\linewidth]{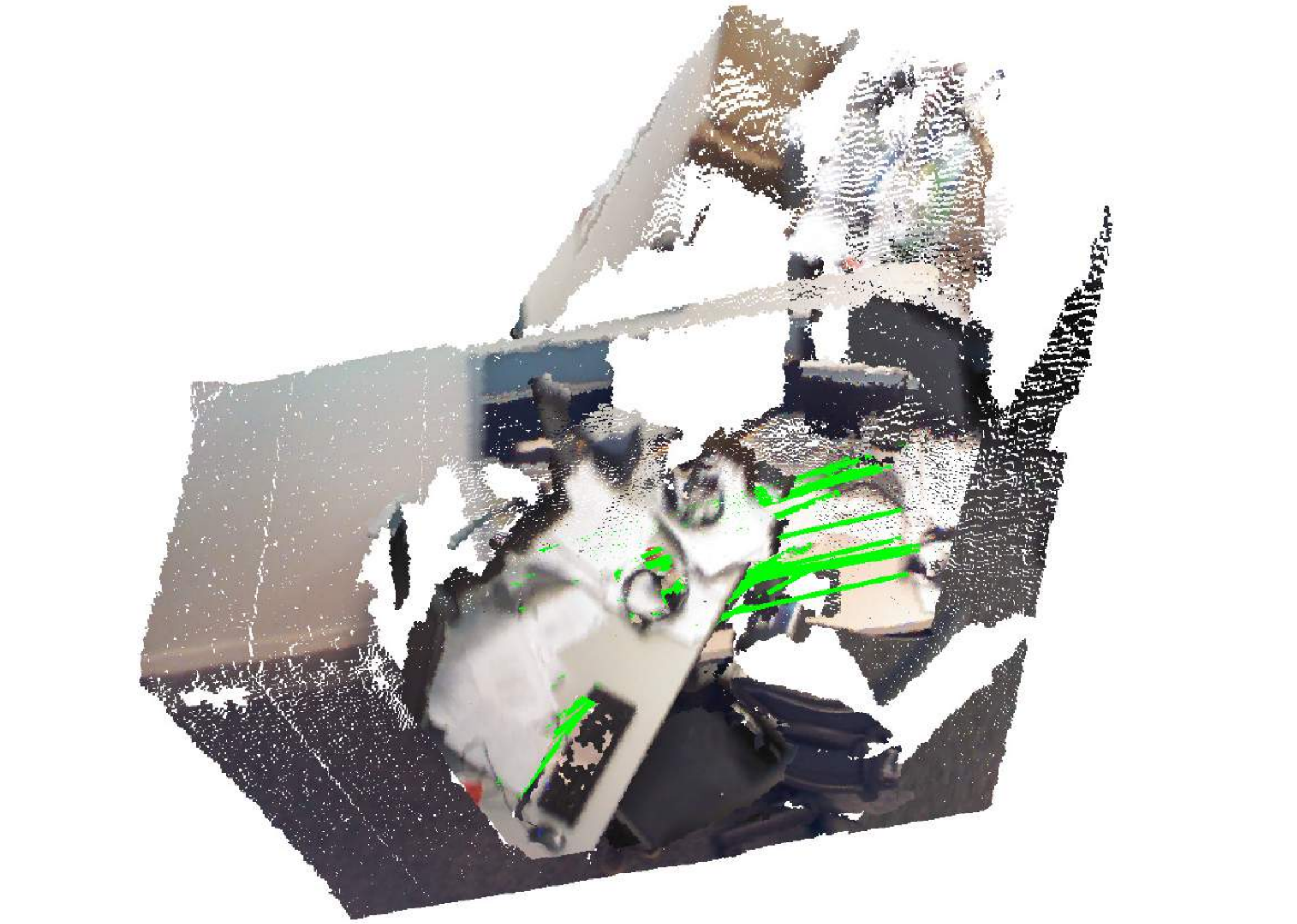}
\includegraphics[width=.48\linewidth]{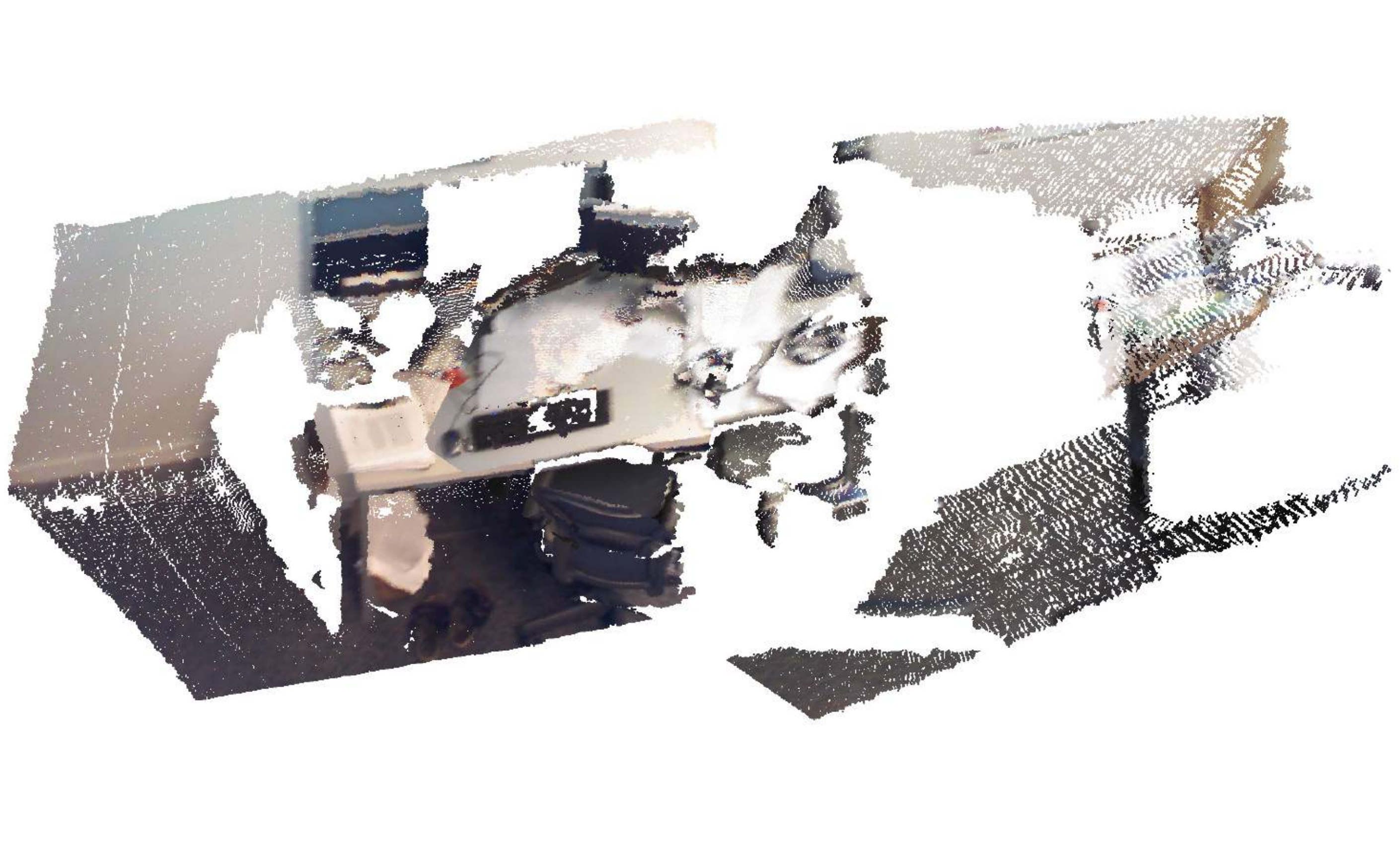}
\end{minipage}\,\,
\\

 & \footnotesize{$N=244$} && \footnotesize{\textcolor[rgb]{1,0,0}{Fail}$,\verb|\|,0.036s$} &\footnotesize{\textcolor[rgb]{0,0.8,0}{Succeed}$,\textbf{0.430},0.706$}&\footnotesize{\textcolor[rgb]{0,0.8,0}{Succeed}$,0.330,0.418s$}&\footnotesize{\textcolor[rgb]{0,0.8,0}{Succeed}$,\textbf{0.430},\textbf{0.039}s$}
\\
\rotatebox{90}{\,\,\footnotesize{\textit{stair}}\,}\,
&
\,\,
\begin{minipage}[t]{0.1\linewidth}
\centering
\includegraphics[width=1\linewidth]{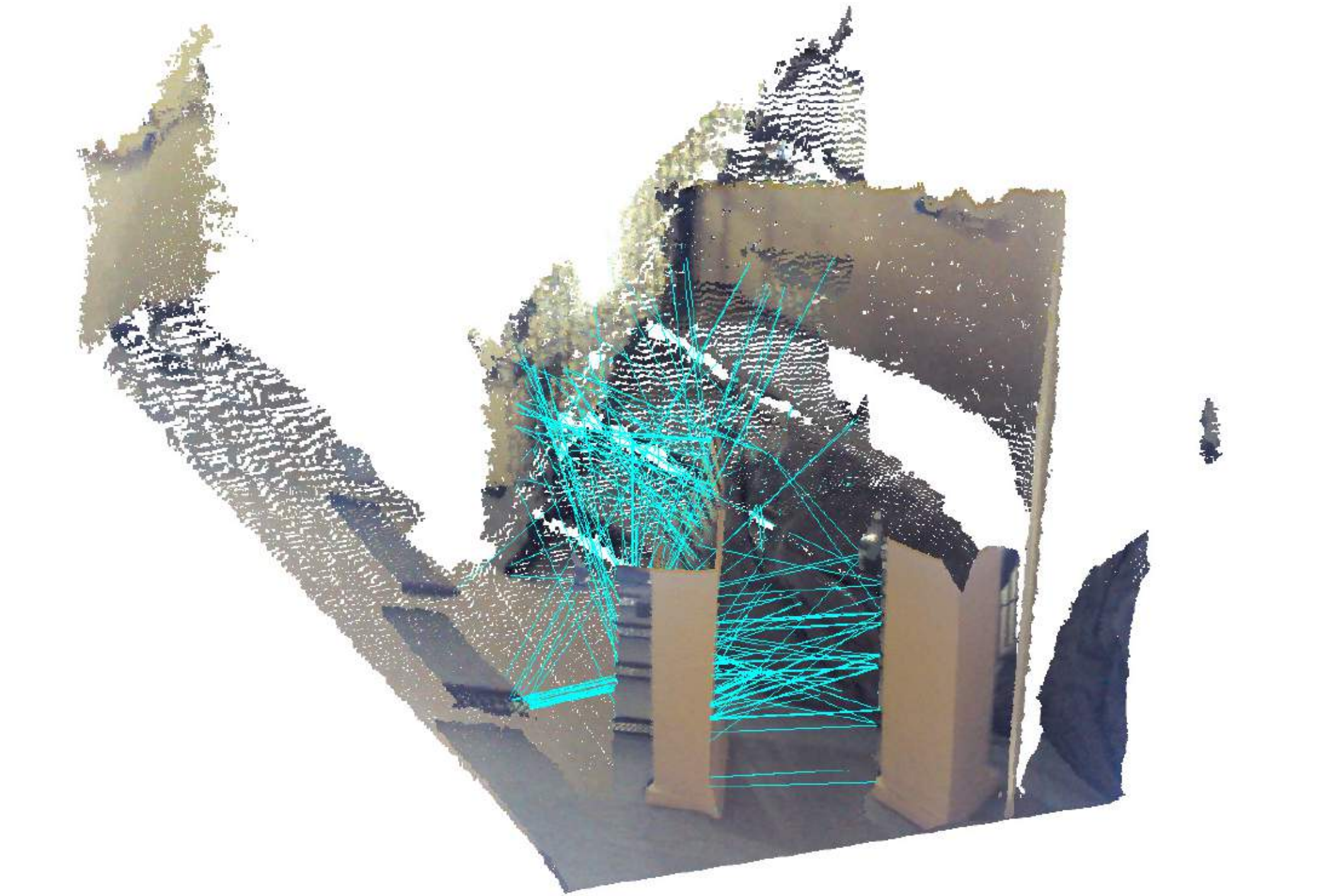}
\end{minipage}\,\,
& &
\,\,
\begin{minipage}[t]{0.19\linewidth}
\centering
\includegraphics[width=.48\linewidth]{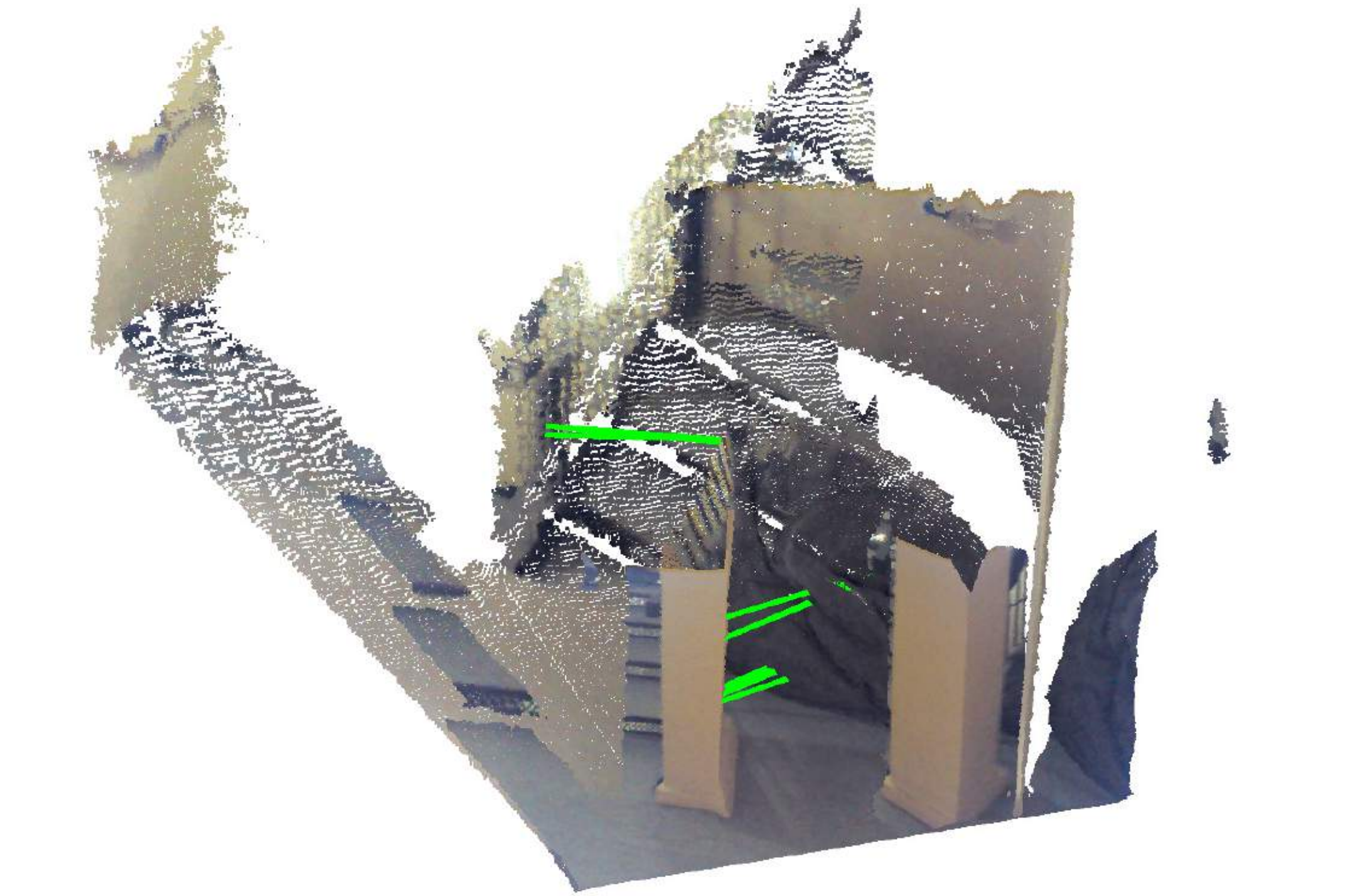}
\includegraphics[width=.48\linewidth]{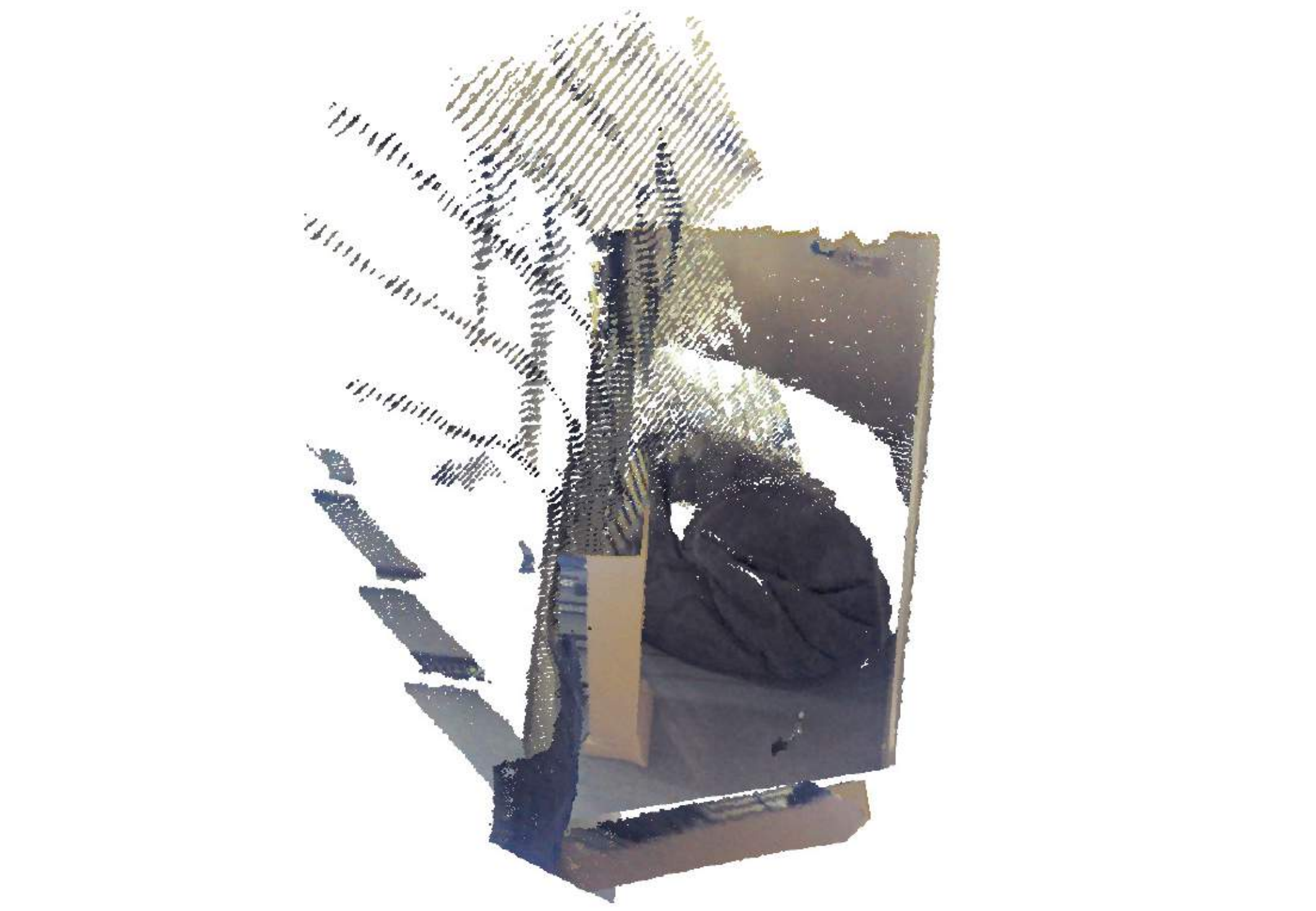}
\end{minipage}\,\,
&
\,\,
\begin{minipage}[t]{0.19\linewidth}
\centering
\includegraphics[width=.48\linewidth]{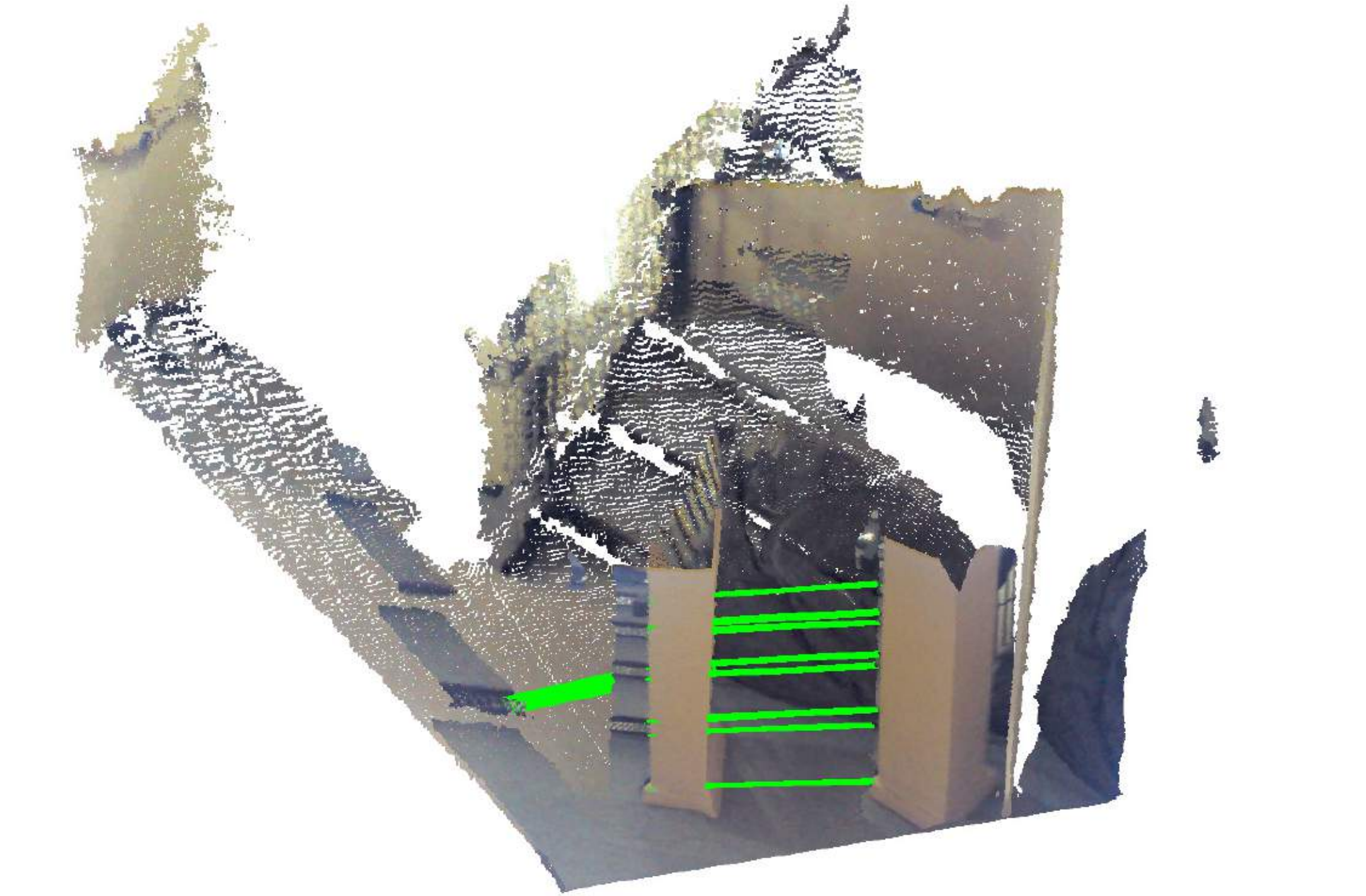}
\includegraphics[width=.48\linewidth]{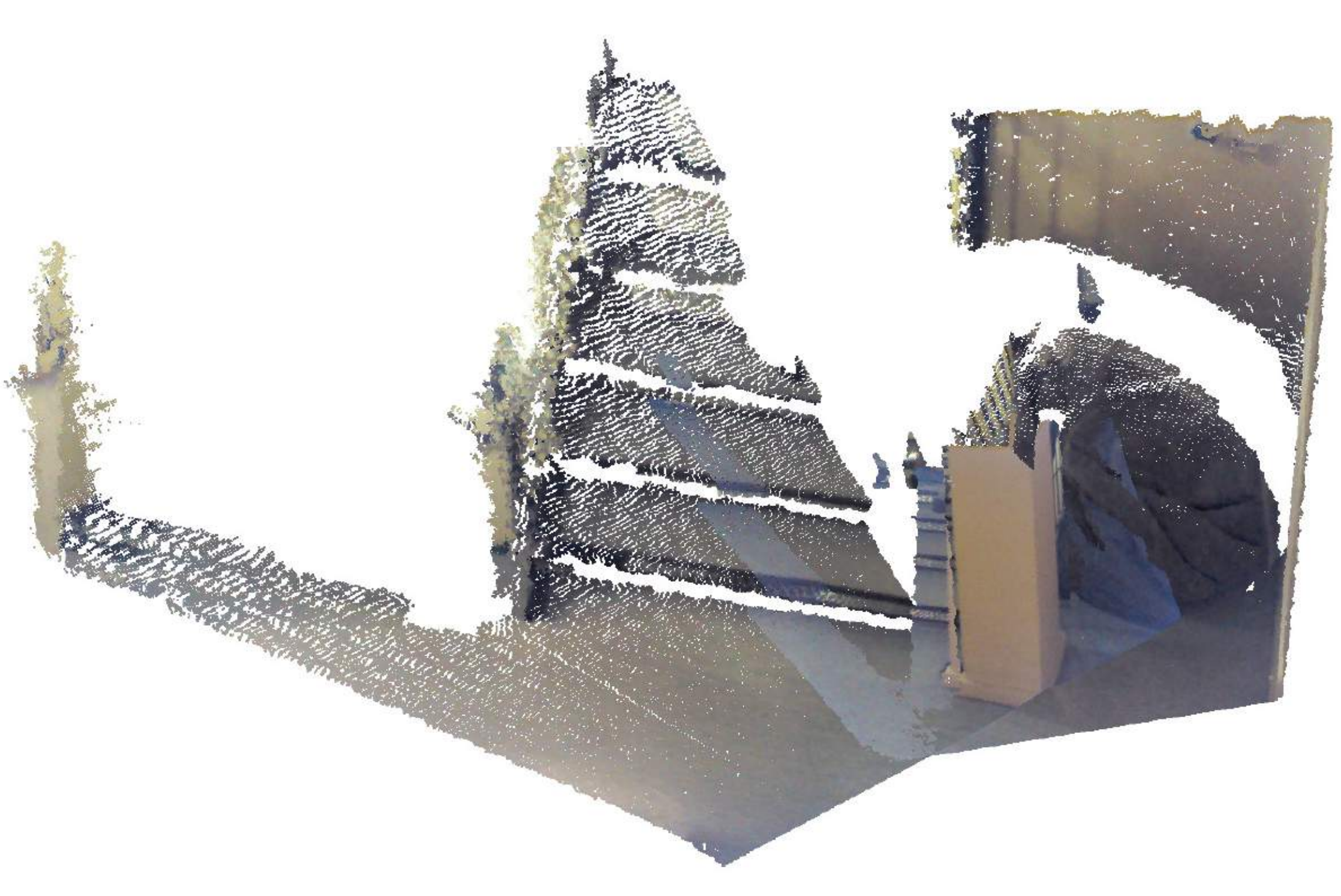}
\end{minipage}\,\,
&
\,\,
\begin{minipage}[t]{0.19\linewidth}
\centering
\includegraphics[width=.48\linewidth]{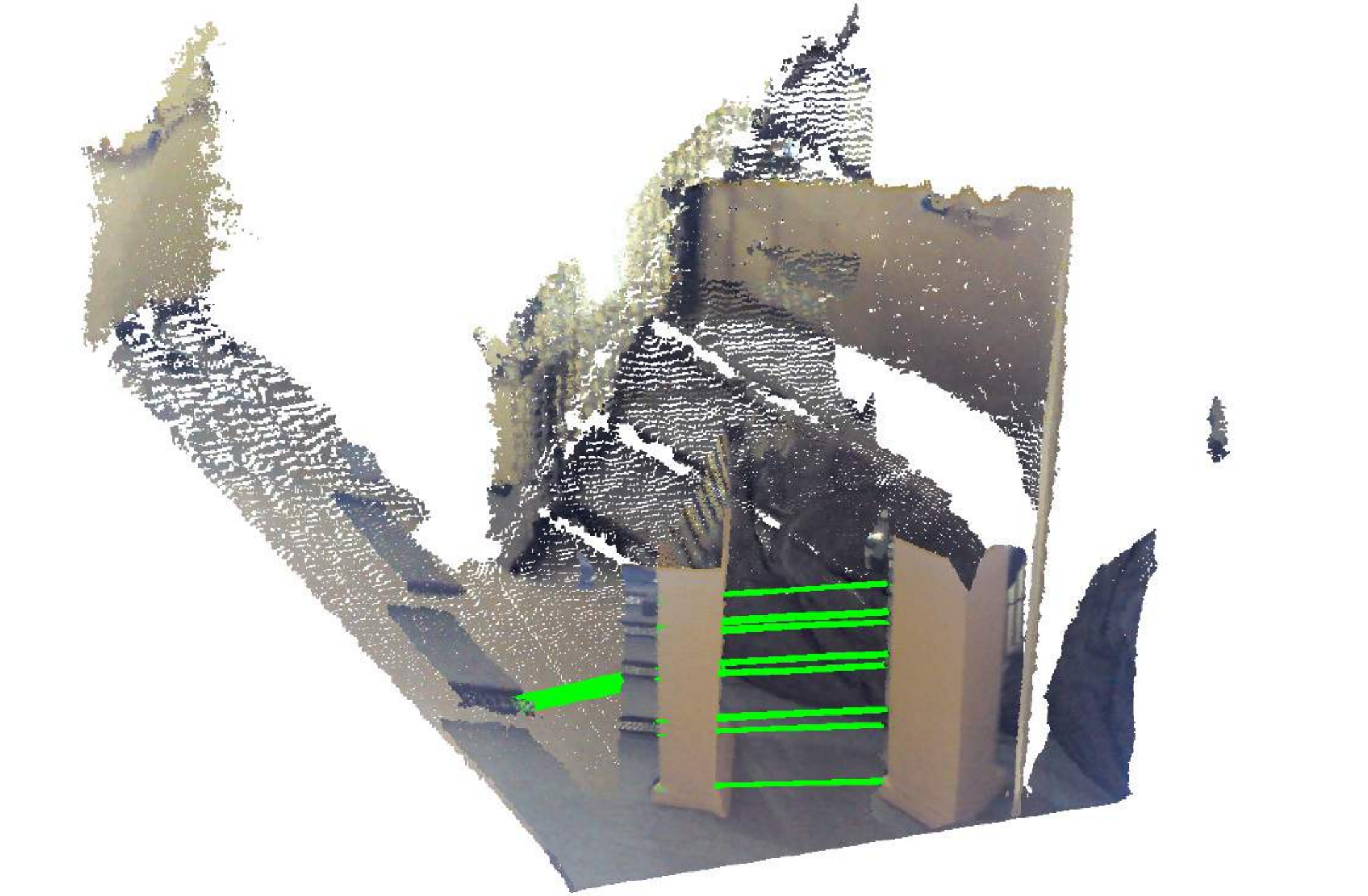}
\includegraphics[width=.48\linewidth]{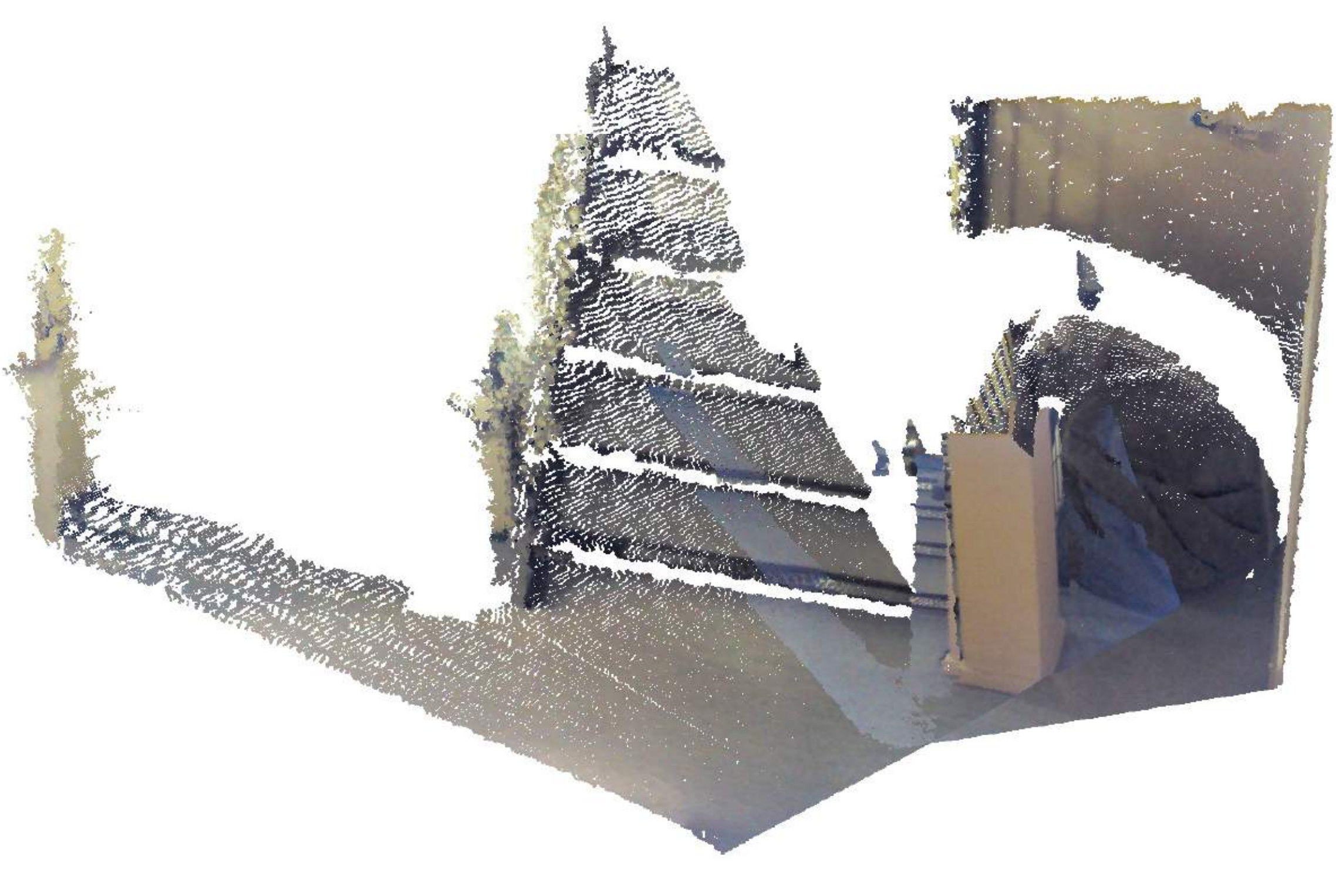}
\end{minipage}\,\,
&
\,\,
\begin{minipage}[t]{0.19\linewidth}
\centering
\includegraphics[width=.48\linewidth]{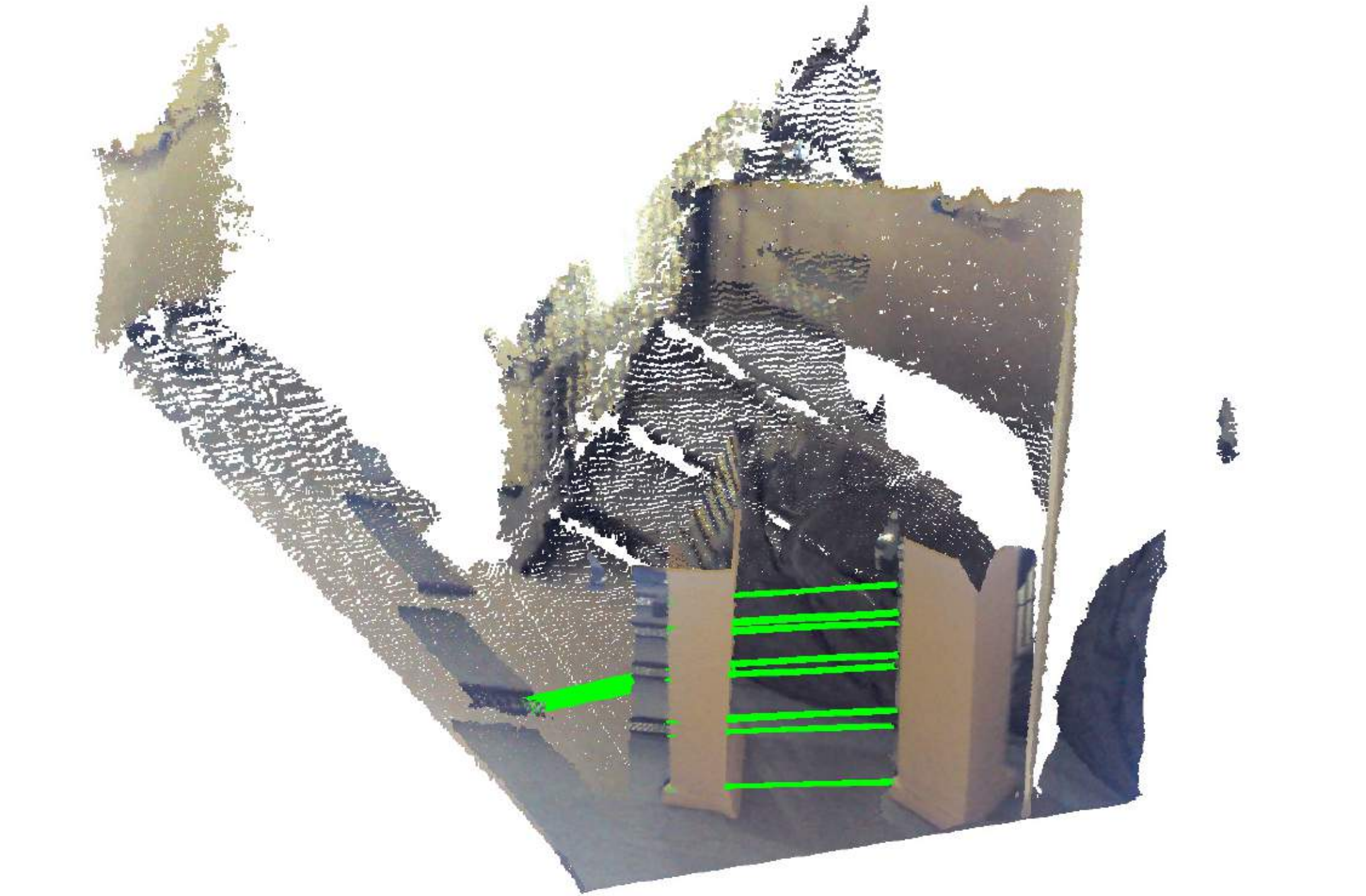}
\includegraphics[width=.48\linewidth]{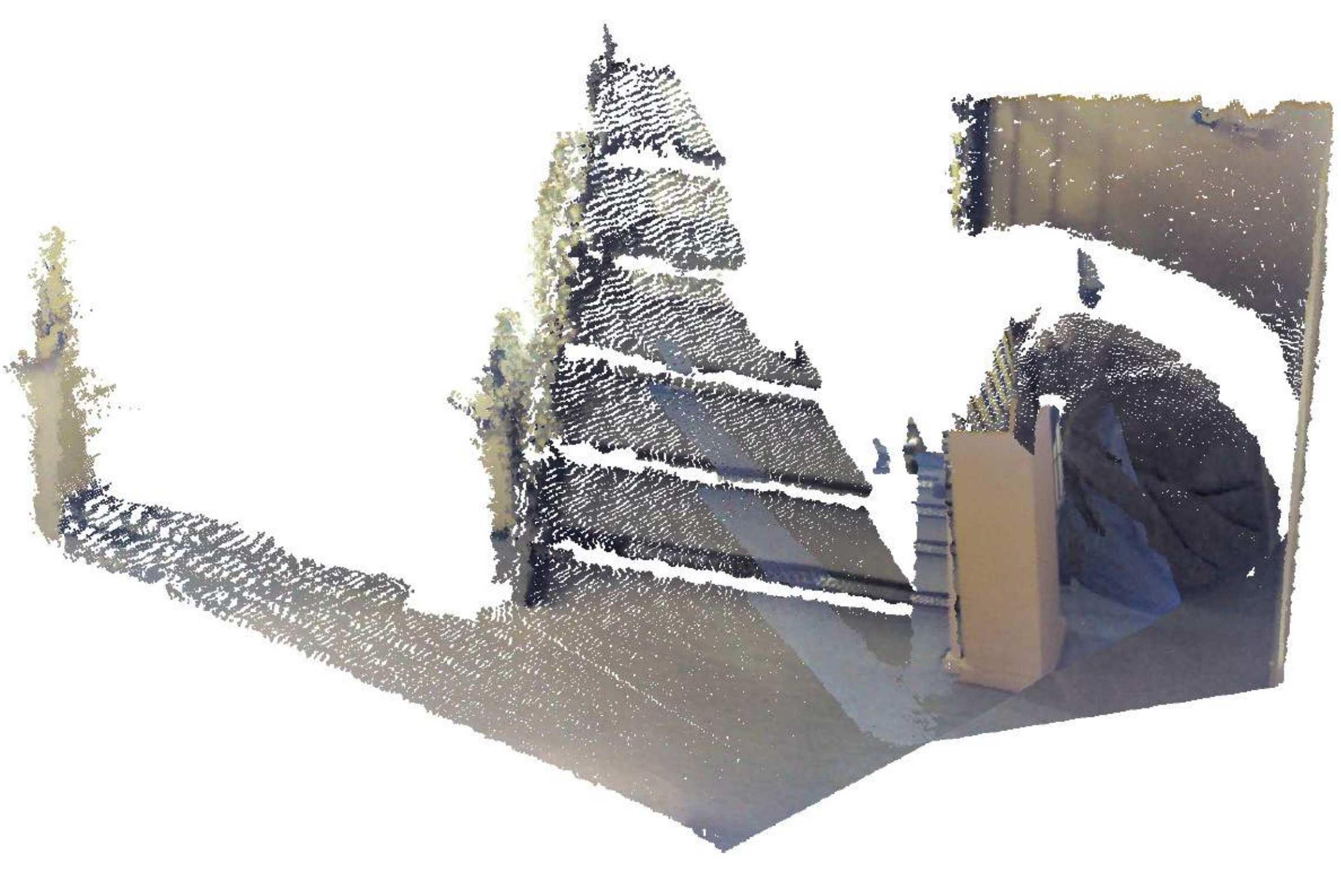}
\end{minipage}\,\,
\\

 & \footnotesize{$N=2001$} && \footnotesize{\textcolor[rgb]{0,0.8,0}{Succeed}$,\textbf{0.224},0.294s$} &\footnotesize{\textcolor[rgb]{0,0.8,0}{Succeed}$,0.262,5.414$}&\footnotesize{\textcolor[rgb]{0,0.8,0}{Succeed}$,0.225,49.971s$}&\footnotesize{\textcolor[rgb]{0,0.8,0}{Succeed}$,{0.227},\textbf{0.269}s$}

\\
\rotatebox{90}{\,\,\footnotesize{\textit{heads}}\,}\,
&
\,\,
\begin{minipage}[t]{0.1\linewidth}
\centering
\includegraphics[width=1\linewidth]{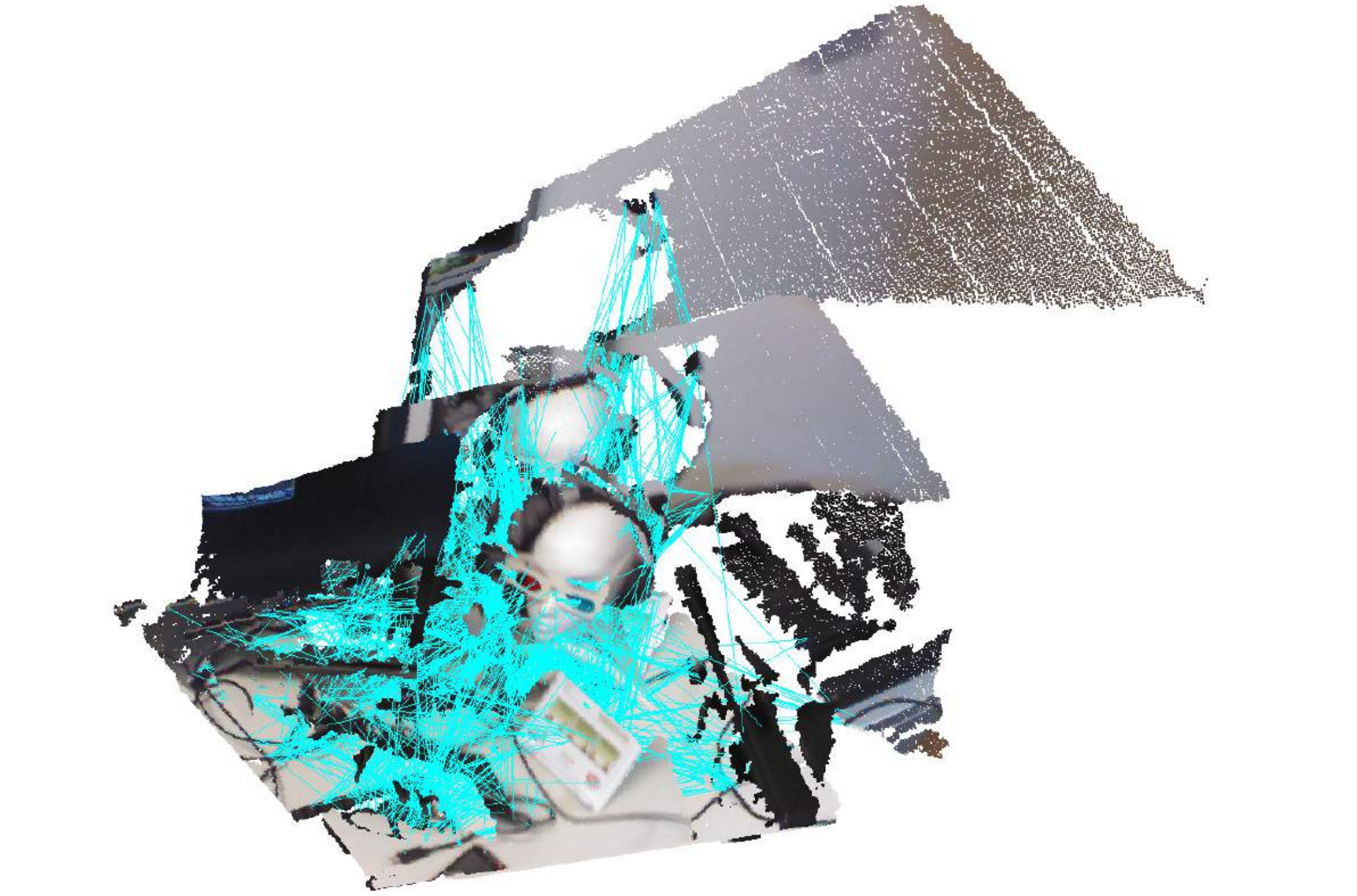}
\end{minipage}\,\,
& &
\,\,
\begin{minipage}[t]{0.19\linewidth}
\centering
\includegraphics[width=.48\linewidth]{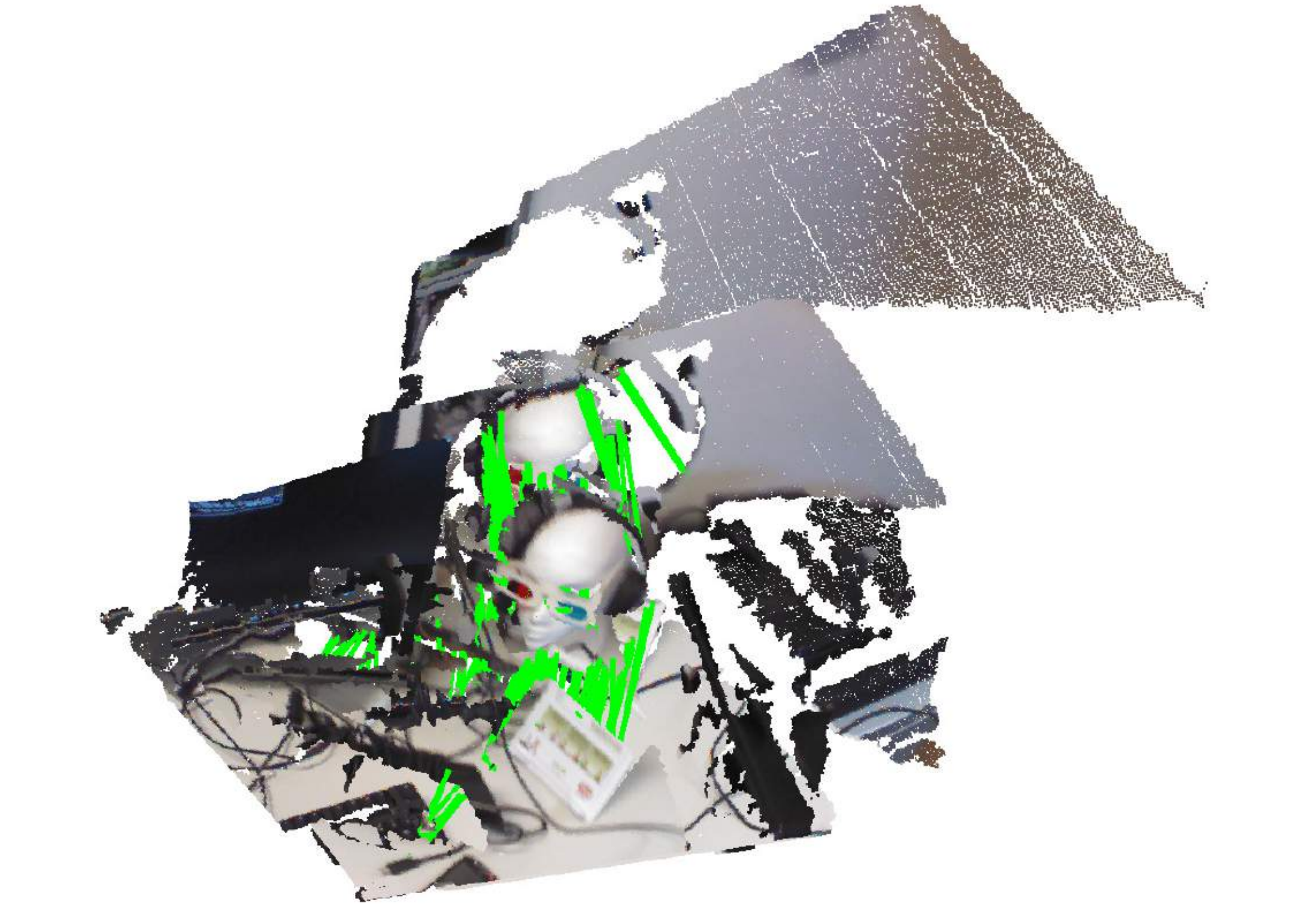}
\includegraphics[width=.48\linewidth]{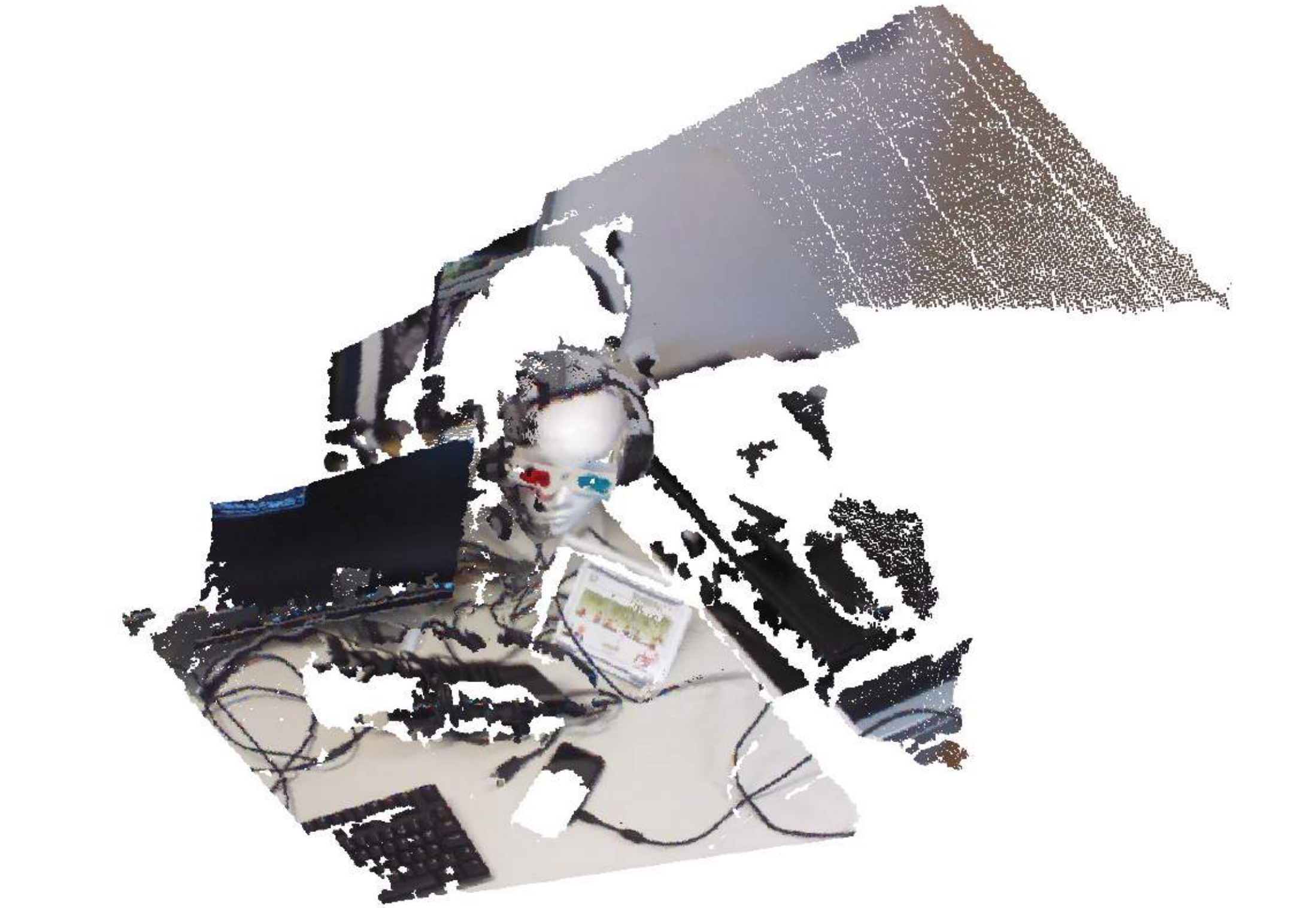}
\end{minipage}\,\,
&
\,\,
\begin{minipage}[t]{0.19\linewidth}
\centering
\includegraphics[width=.48\linewidth]{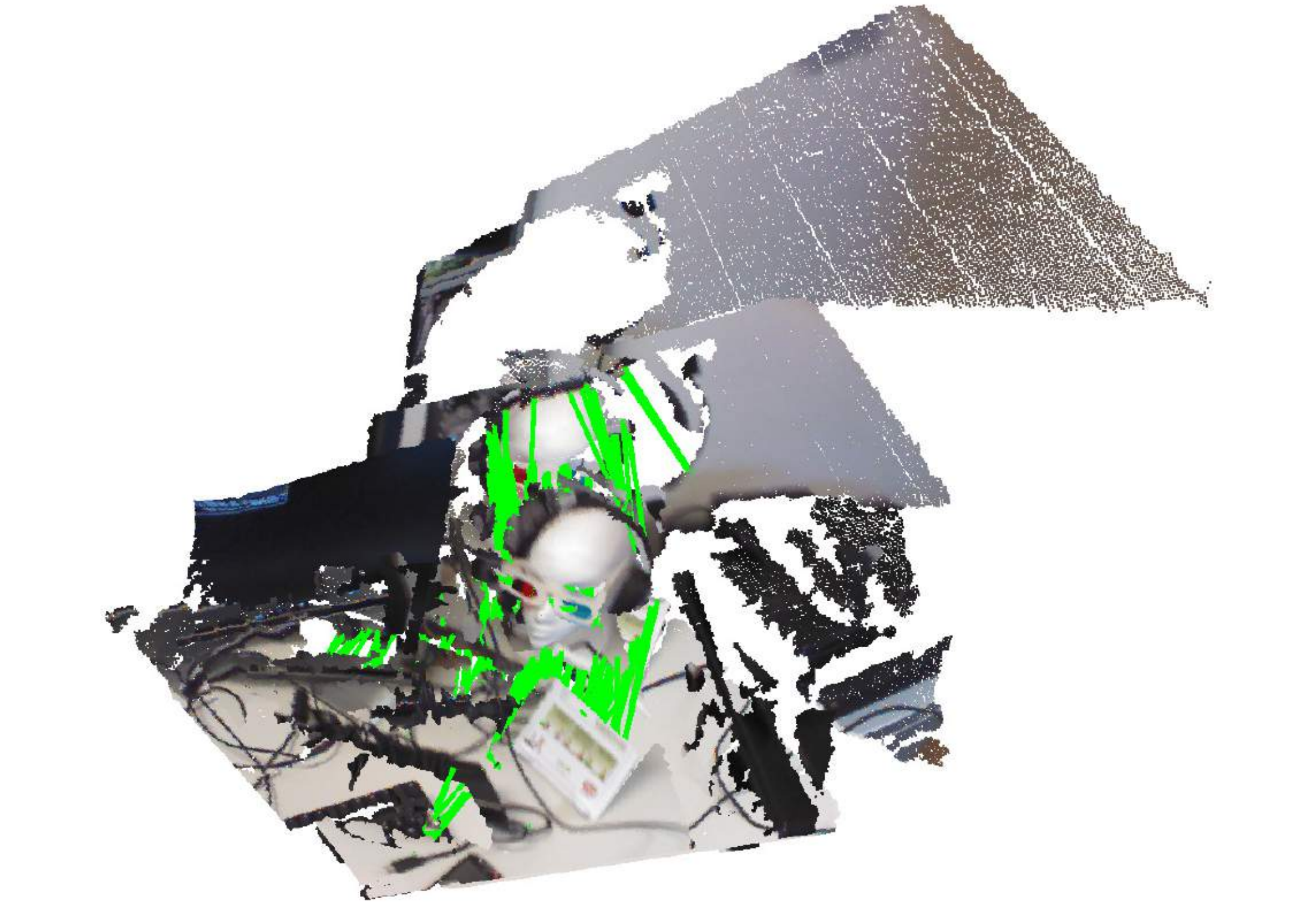}
\includegraphics[width=.48\linewidth]{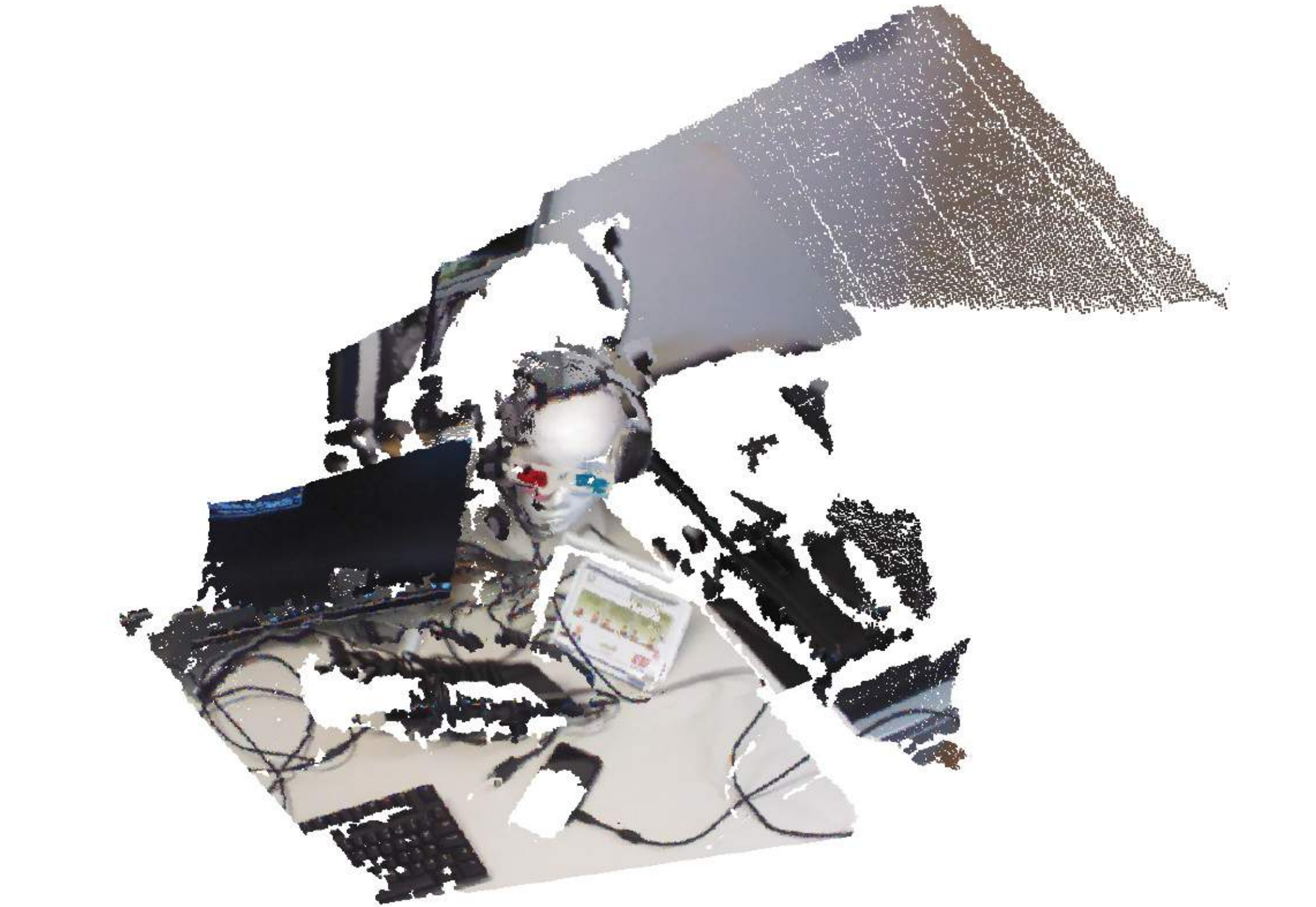}
\end{minipage}\,\,
&
\,\,
\begin{minipage}[t]{0.19\linewidth}
\centering
\includegraphics[width=.48\linewidth]{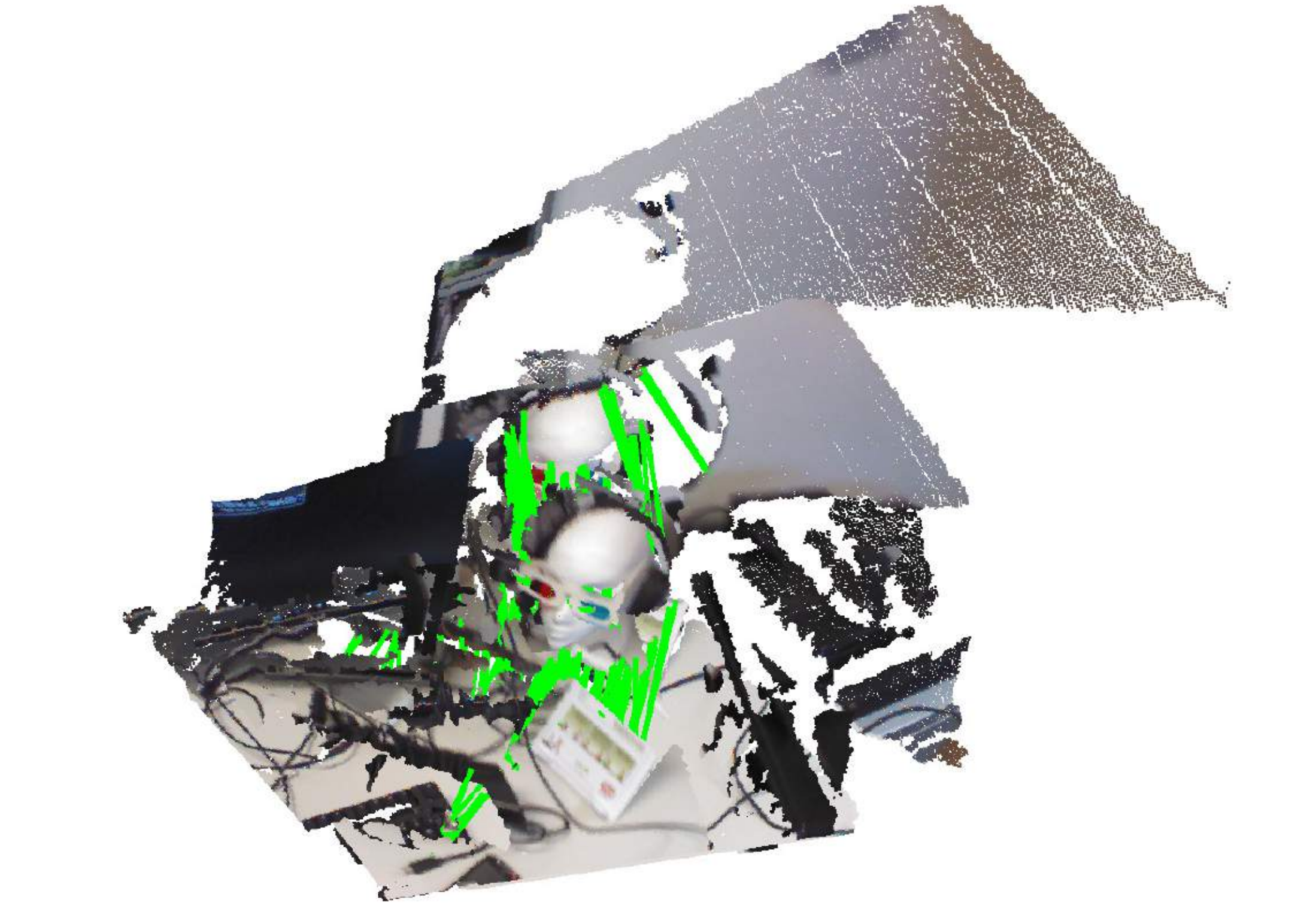}
\includegraphics[width=.48\linewidth]{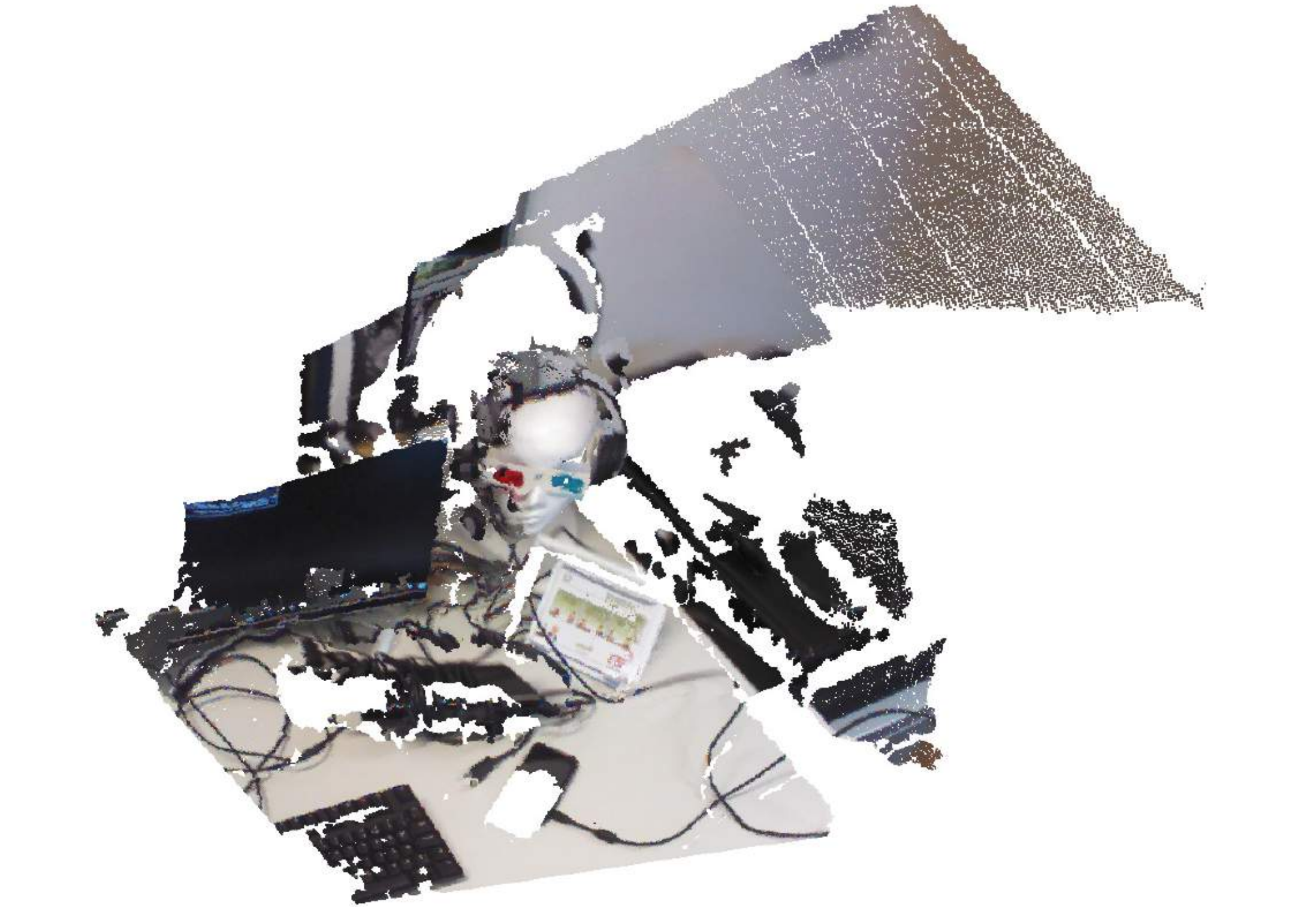}
\end{minipage}\,\,
&
\,\,
\begin{minipage}[t]{0.19\linewidth}
\centering
\includegraphics[width=.48\linewidth]{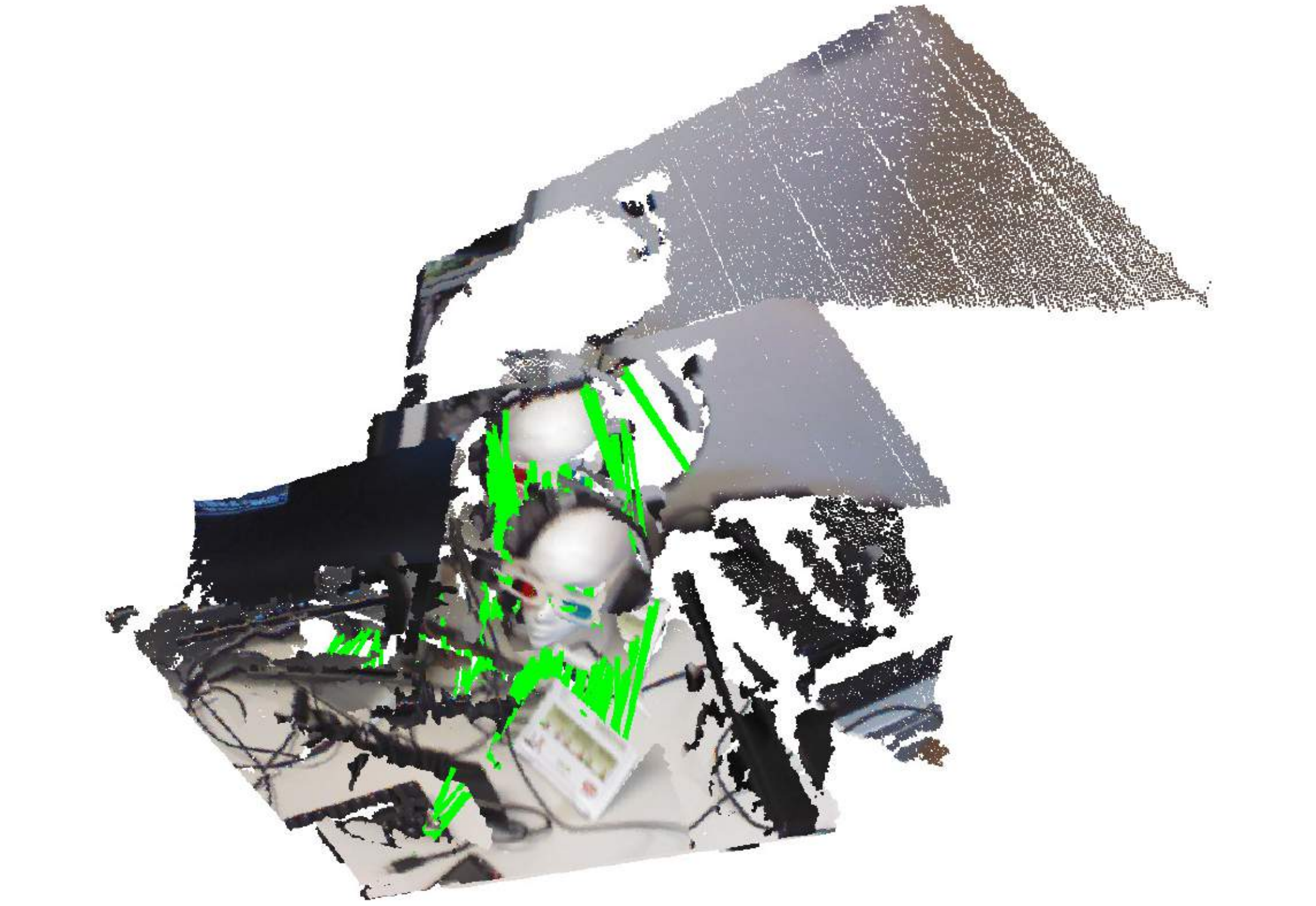}
\includegraphics[width=.48\linewidth]{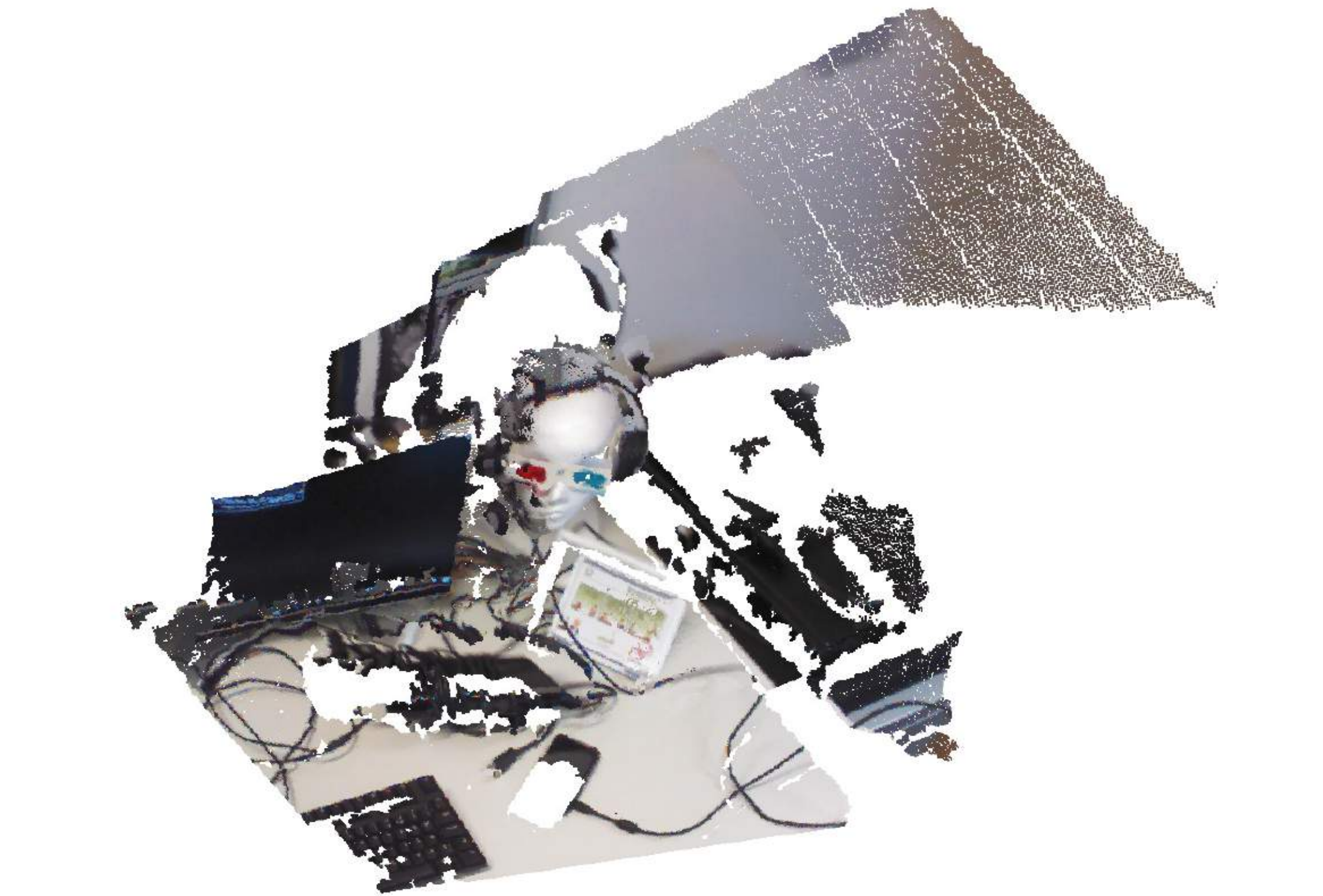}
\end{minipage}\,\,

\end{tabular}

\caption{Scan matching results on the Microsoft 7-scenes dataset~\cite{shotton2013scene}. The left-most column demonstrates the putative correspondences matched by SURF in cyan lines (already converted into the 3D space using depth and calibration information), and the rest columns show: (i) the inliers found and (ii) the qualitative scene stitching results, using FLO-RANSAC, GNC-TLS, GORE+RANSAC and DANIEL, respectively. On top of the results of each solver, from left to right, we display the stitching status (either \textcolor[rgb]{1,0,0}{Fail} or \textcolor[rgb]{0,0.5,0}{Succeed}), the RMSE and runtime (in seconds), respectively. Note that stitching with visibly distinguishable error should be regarded as \textcolor[rgb]{1,0,0}{Fail}, and when the stitching is failed, we no longer display the RMSE since it would become meaningless. Best results are shown in \textbf{bold} font. Best viewed when zoomed-in.}
\label{scan-matching}
\end{figure*}
\clearpage

\noindent we use SURF~\cite{bay2006surf} (Matlab function: \textit{detectSURFFeatures}) to match 2D keypoint correspondences across the two RGB images and then convert them into 3D correspondences by using depth data and camera intrinsic parameters. We also apply the four solvers: GNC-TLS, FLO-RANSAC, GORE+RANSAC and DANIEL, for comparative evaluation.

Both qualitative and quantitative scan matching results are displayed in Fig.~\ref{scan-matching}, where we show the raw correspondences matched by SURF, and the inliers found as well as the scan matching results by the respective robust solvers in qualitative results, and show the registration status (\textcolor[rgb]{1,0,0}{Fail} means that the scenes are wrongly stitched and \textcolor[rgb]{0,0.5,0}{Succeed} means that the stitching is successful), RMSE (Root Mean Square Error) and runtime in quantitative results. Though 2D keypoint matching is used, we can observe that the inliers are merely a small minority of the putative correspondences. The results show that GNC-TLS is the least robust solver, failing in 70\% tests, and FLO-RANSAC and GORE-RANSAC both generate a few failure cases but their runtime are too slow for practical use. Our DANIEL remains highly robust in all tests and also most often has the lowest RMSE and the fastest speed, which further validates the promising practicality of DANIEL.

\section{Conclusion}

In this paper, a novel consensus maximization method for correspondence-based robust point cloud registration with high or even extreme outliers is proposed. We design a double-layered random sampling framework with the smart application of 3D rigidity constraint, in order to tremendously reduce the computational cost for sampling pure inlier subsets even in high-outlier registration problems. We further render a stratified element-wise compatibility checking technique during the sampling process, which intends to circumvent the repeated consensus building procedure so as to accelerate the algorithm convergence. These two contributions enable both the high robustness and the high time-efficiency of the resulting solver DANIEL.  Extensive experiments over diverse datasets validate that DANIEL is superior to other existing state-of-the-art methods. To be specific, DANIEL is robust against extremely high outlier ratios as many as 99\%, is up to four orders of magnitude faster than RANSAC and also faster than other solvers in most cases, and proves to be effective in practical application problems such as object pose estimation and scan matching. The source code of DANIEL is available at: \url{https://github.com/LeiSun-98/Daniel}.

\bibliographystyle{IEEEtran}
\bibliography{egbib}

\begin{IEEEbiography}[{\includegraphics[width=1in,height=1.25in,clip,keepaspectratio]{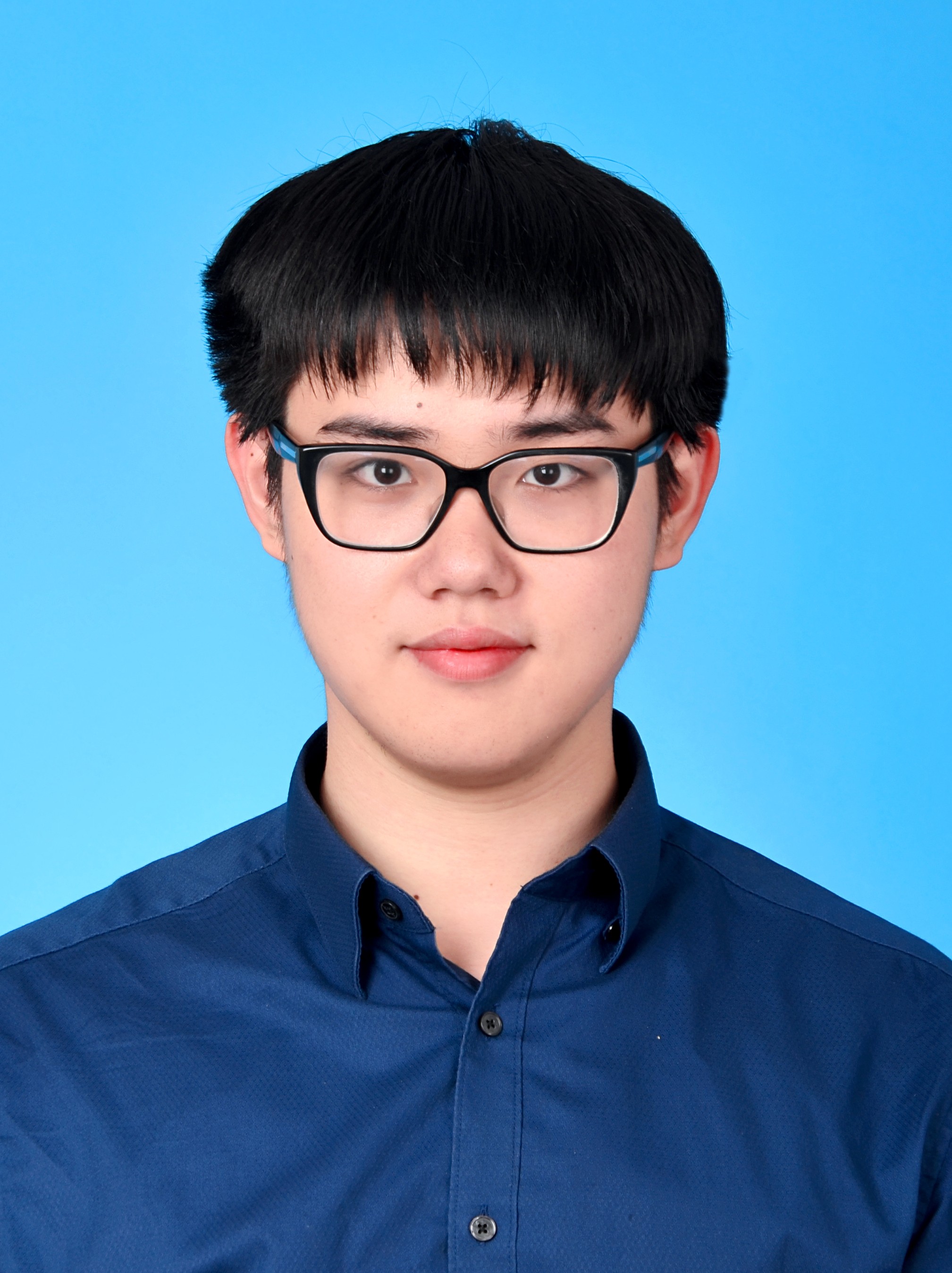}}]{Lei Sun} received the B.Eng. degree with School of Mechanical and Power Engineering at East China University of Science and Technology (ECUST), Shanghai, China. His research interests mainly include robotics perception, geometric computer vision, robust estimation and optimization.
\end{IEEEbiography}

\end{document}